%% file: hypino.tex
\documentclass{article}

\PassOptionsToPackage{square,numbers}{natbib}

\usepackage[preprint]{neurips_2025}

\usepackage[utf8]{inputenc} 
\usepackage[T1]{fontenc}    
\usepackage{hyperref}       
\usepackage{url}            
\usepackage{booktabs}       
\usepackage{amsfonts}       
\usepackage{nicefrac}       
\usepackage{microtype}      
\usepackage{xcolor}         
\usepackage{microtype}
\usepackage{graphicx}
\usepackage{wrapfig}
\usepackage{subcaption}
\usepackage{booktabs} 
\usepackage{amssymb}
\usepackage{mathtools}
\usepackage{amsthm}
\usepackage{algorithm2e}

\usepackage{booktabs}
\usepackage{siunitx}
\usepackage{float}

\title{HyPINO: Multi-Physics Neural Operators via HyperPINNs and the Method of Manufactured Solutions}

\author{%
Rafael Bischof$^{1}$ \hspace{4em} Michal Piovarči$^{1}$ \hspace{4em} Michael A. Kraus$^{2}$ \\ \\ 
\textbf{Siddhartha Mishra}$^{3}$ \hspace{4em} \textbf{Bernd Bickel}$^{1}$ \\
\\
$^{1}$Computational Design Lab, ETH Zurich, Switzerland\\
$^{2}$Institute of Structural Mechanics and Design, TU Darmstadt, Germany\\
$^{3}$Seminar for Applied Mathematics, ETH Zurich, Switzerland\\
*Correspondence to \href{mailto:rabischof@ethz.ch}{rabischof@ethz.ch}
}

\begin{document}

\maketitle

\begin{abstract}
We present HyPINO, a multi-physics neural operator designed for zero-shot generalization across a broad class of PDEs without requiring task-specific fine-tuning. Our approach combines a Swin Transformer-based hypernetwork with mixed supervision: (i) labeled data from analytical solutions generated via the Method of Manufactured Solutions (MMS), and (ii) unlabeled samples optimized using physics-informed objectives. The model maps PDE parameterizations to target Physics-Informed Neural Networks (PINNs) and can handle linear elliptic, hyperbolic, and parabolic equations in two dimensions with varying source terms, geometries, and mixed Dirichlet/Neumann boundary conditions, including interior boundaries. HyPINO achieves strong zero-shot accuracy on seven benchmark problems from PINN literature, outperforming U-Nets, Poseidon, and Physics-Informed Neural Operators (PINO). Further, we introduce an iterative refinement procedure that treats the residual of the generated PINN as "delta PDE" and performs another forward pass to generate a corrective PINN. Summing their contributions and repeating this process forms an ensemble whose combined solution progressively reduces the error on six benchmarks and achieves a >100× lower $L_2$ loss in the best case, while retaining forward-only inference. Additionally, we evaluate the fine-tuning behavior of PINNs initialized by HyPINO and show that they converge faster and to lower final error than both randomly initialized and Reptile-meta-learned PINNs on five benchmarks, performing on par on the remaining two. Our results highlight the potential of this scalable approach as a foundation for extending neural operators toward solving increasingly complex, nonlinear, and high-dimensional PDE problems. The code and model weights are publicly available at \href{https://github.com/rbischof/hypino}{\texttt{https://github.com/rbischof/hypino}}.

\end{abstract}

\section{Introduction}
\label{sec:introduction}

Neural operators have emerged as a promising paradigm for solving partial differential equations (PDEs). Their ability to generalize across families of PDEs, fast inference, and full differentiability make them appealing for a wide range of scientific computing tasks. In the longer term, such methods may serve as building blocks for general-purpose, foundational, and multi-physics simulators, sometimes referred to as "world-model predictors" \cite{ding2022pino,li2025multi,wu2024physics,yuan2025high}.

However, existing neural operators are typically sample inefficient \cite{pde_foundation_model_2024_Herde}. As a result, most prior work focuses on narrowly defined problem families \cite{deeponet}. Variations are limited to singular aspects such as specific PDE parameters (e.g., diffusion coefficients) \cite{pinn_governing_equation_coefficient_encoding}, boundary conditions \cite{duvall2021discretization}, or domain shapes \cite{ZHENG2024102243}. The support for simultaneous variations of PDE operators remains limited to subdomains, such as parametrized convection-diffusion-reaction PDEs \cite{ye2024pdeformer}.

One way to address the data requirement is by incorporating physics-informed losses. Such losses can provide self-supervision without requiring labeled simulation data \cite{pinn,physics_informed_deeponet}. While promising, existing methods often suffer from spectral bias \cite{lrannealing} and mode collapse \cite{yang2018physics}. Moreover, purely physics-based training is unstable in practice \cite{goswami2023physics_neural_operators}. Therefore, even with physics-based losses, obtaining a large, labeled dataset that spans a diverse range of PDEs remains a significant bottleneck.

To overcome these challenges, we propose HyPINO, a hybrid framework that combines physics-informed learning with synthetic supervised data. Our approach leverages a Swin Transformer-based hypernetwork \cite{hyperpinn,hypernetworks,swin} to map PDE specifications to the parameters of a target physics-informed neural network (PINN). HyPINO enables zero-shot generalization without task-specific fine-tuning. A key contribution of our work is a scalable synthetic data pipeline that generates two complementary types of training data: (i) Supervised samples generated via the Method of Manufactured Solutions (MMS) \cite{method_manufactured_solutions} by selecting target solutions and deriving the corresponding PDEs analytically. These provide direct supervision with known reference solutions. (ii) Physics-only samples constructed by randomly sampling PDE operators, source terms, and boundary / initial conditions. These are trained using physics-informed losses without requiring ground-truth solutions. This hybrid training strategy allows us to cover a broad spectrum of two-dimensional linear elliptic, hyperbolic, and parabolic PDEs with mixed Dirichlet and Neumann boundary conditions on complex domain geometries, spanning a wide range of phenomena in the natural and engineering sciences, including heat diffusion, wave propagation, acoustic scattering, and membrane deformation.

In addition, we introduce an iterative refinement procedure that builds an ensemble of corrective PINNs. At each iteration, the model evaluates the residual error and generates a "delta" PINN to improve the solution. This ensemble refinement provides a lightweight alternative to traditional fine-tuning, requiring only inference passes rather than full backward passes.

We evaluate HyPINO on seven diverse PDE benchmarks, demonstrating improved zero-shot generalization compared to baselines such as U-Nets \cite{unet}, Poseidon \cite{pde_foundation_model_2024_Herde}, and PINO \cite{pino}. We also find that PINNs initialized with our method fine-tune more efficiently than those starting from random or meta-learned initializations, achieving faster convergence and lower final errors.

In summary, our main contributions are (i) a hybrid physics-informed and supervised learning framework for multi-physics PDE solving, (ii) a scalable data generation pipeline combining random physics sampling with MMS-based supervised examples, (iii) an ensemble-based refinement mechanism that improves prediction quality without expensive retraining, and (iv) empirical results showing strong zero-shot and fine-tuning performance across multiple PDE benchmarks compared to SOTA.

We believe these contributions offer a practical step toward more general-purpose, data-efficient neural operators for multi-physics problems and world simulator foundation models.

\begin{figure}[ht]
    \centering
    \includegraphics[width=0.9\linewidth]{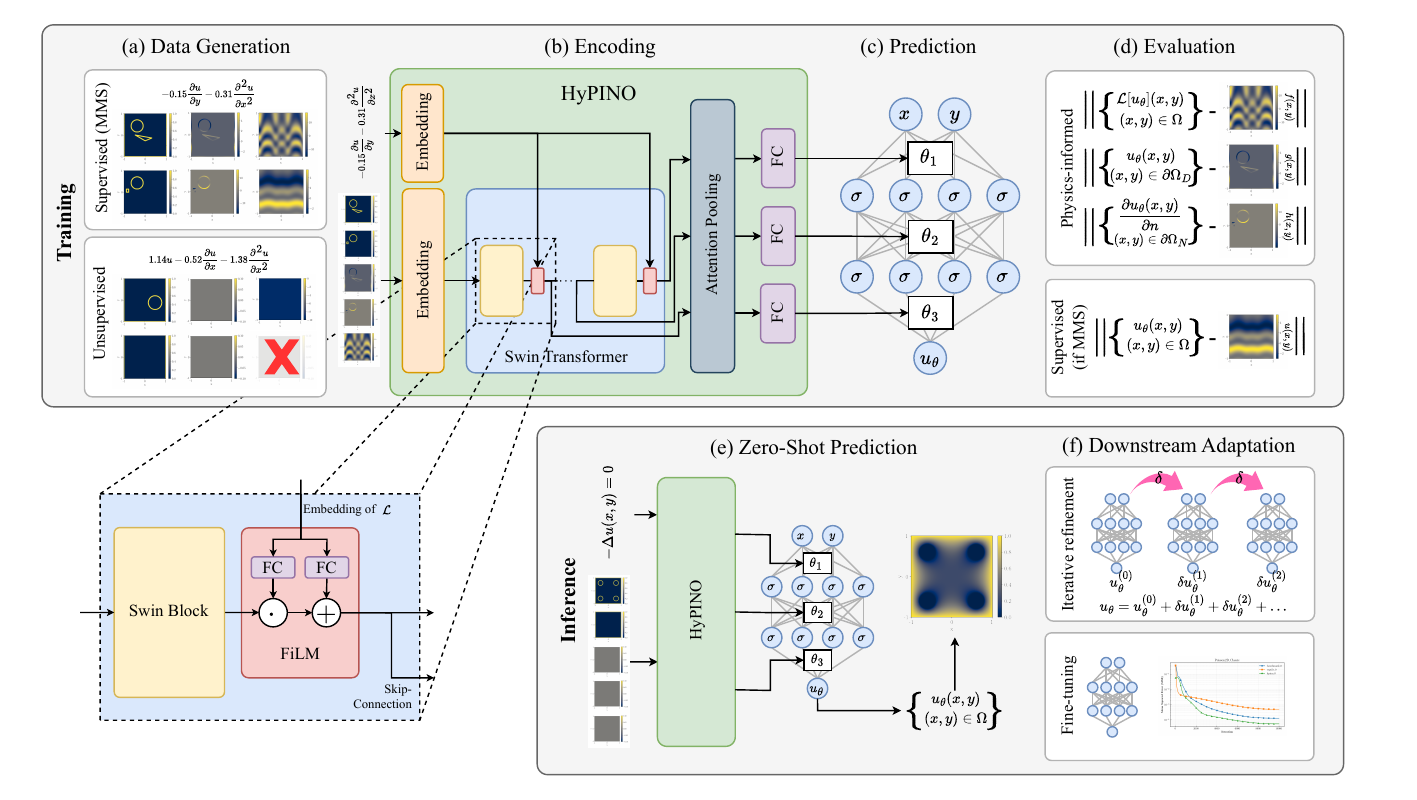}
    \caption{\textbf{Overview of the HyPINO pipeline.}
(a) Training data includes supervised samples from MMS and unsupervised physics-informed samples without ground truth.
(b) PDEs are encoded as multi-channel and vector-based inputs and processed by HyPINO to produce task-specific PINN weights.
(c) The predicted PINN maps spatial coordinates to the solution field.
(d) Training combines physics-informed residual losses as well as supervised losses for MMS data.
(e) At inference, HyPINO enables zero-shot prediction for unseen PDEs.
(f) Downstream adaptation includes iterative refinement using residual corrections or optional, task-specific fine-tuning.}
    \label{fig:pipeline}
\end{figure}

\section{Related Work}
\label{sec:related_work}

Neural operators aim to approximate solution operators that map PDE specifications to continuous solution fields, enabling fast, mesh-free inference and generalization to unseen problem instances~\cite{hao2023gnot,li2022transformer,fno,pino,deeponet,marwah2023deep}. Recent work scales these ideas toward foundation models that ingest large corpora of simulated data or equation specifications and promise broad cross-task transfer~\cite{hao2024dpot,pde_foundation_model_2024_Herde,pde_foundation_model_2023,pde_foundation_model_2024_Subramanian,ye2024pdeformer}. Despite rapid progress, most operators still target narrow PDE families (e.g.\ fixed equations with varying coefficients) and depend on expensive high-fidelity solvers for supervision~\cite{ye2024pdeformer}. Embedding the governing equations in the loss function alleviates the need for labeled data and improves physical fidelity~\cite{pinn,physics_informed_deeponet}. While the original formulation was introduced for stand-alone Physics-Informed Neural Networks (PINNs), the same residual losses have recently been integrated into operator architectures, yielding Physics-Informed Neural Operators (PINO) that train from unlabeled residual samples~\cite{boudec2024learning,goswami2023physics_neural_operators,pino}. These approaches still require careful weighting of supervision terms and often struggle with stability and spectral bias for complex PDEs.

Hypernetworks generate the parameters of a target network conditioned on an auxiliary input~\cite{hypernetworks}. In the PDE context, HyperPINNs predict PINN weights for varying coefficients~\cite{hyperpinn,lee2023hyperdeeponet}, and subsequent works extend this idea to boundary conditions, domain changes, and low-rank weight modulation~\cite{cho2023hypernetwork,duvall2021discretization,hadorn2022shift,morel2025disco}. Yet, existing models rarely support concurrent variation of multiple operators, geometries, and boundary types without task-specific fine-tuning.

The Method of Manufactured Solutions (MMS) provides analytic ground-truth pairs by choosing a target field and deriving the corresponding source term and boundary data~\cite{method_manufactured_solutions}. MMS has long served for numerical-solver verification and was recently adopted for PINN evaluation~\cite{pinns_evaluated_with_method_manufactured_solutions} and operator training~\cite{hasani2024generating}. However, prior studies focus on single equations (e.g.\ Poisson); leveraging MMS for \emph{multi-physics} operator pre-training remains largely unexplored.

Our work situates itself at the intersection of these lines: we couple a Swin Transformer hypernetwork with mixed MMS and physics-informed supervision to produce a single model that generalizes zero-shot across diverse linear, elliptic, hyperbolic, and parabolic PDEs with mixed boundary conditions and diverse 2D geometries.

\section{Methodology}
\label{sec:methodology}

We consider a family of second-order linear PDEs defined over a bounded domain $\Omega \subset \mathbb{R}^2$ with boundary $\partial \Omega = \partial \Omega_D \cup \partial \Omega_N$, where $\partial \Omega_D$ and $\partial \Omega_N$ denote the Dirichlet and Neumann boundaries, respectively. The goal is to find a function $u: \Omega \to \mathbb{R}^m$ satisfying
\begin{equation}
\mathcal{L}[u](\mathbf{x}) = f(\mathbf{x}) \quad \text{in } \Omega, \quad u(\mathbf{x}) = g(\mathbf{x}) \quad \text{on } \partial \Omega_D, \quad \frac{\partial u}{\partial n}(\mathbf{x}) = h(\mathbf{x}) \quad \text{on } \partial \Omega_N,
\label{eq:neural_operator}
\end{equation}
where $\mathcal{L}$ is a linear differential operator involving derivatives up to second order, $f: \Omega \to \mathbb{R}^m$ is a known source term, and $g$, $h$ are prescribed boundary functions. Our objective is to learn the solution operator that maps the tuple $(\mathcal{L}, f, g, h)$ to the solution $u$.

\subsection{PDE Parameterization}
\label{sec:parameterization}
\begin{figure}[ht]
    \centering
    \begin{subfigure}[b]{0.16\linewidth}
        \includegraphics[width=\linewidth]{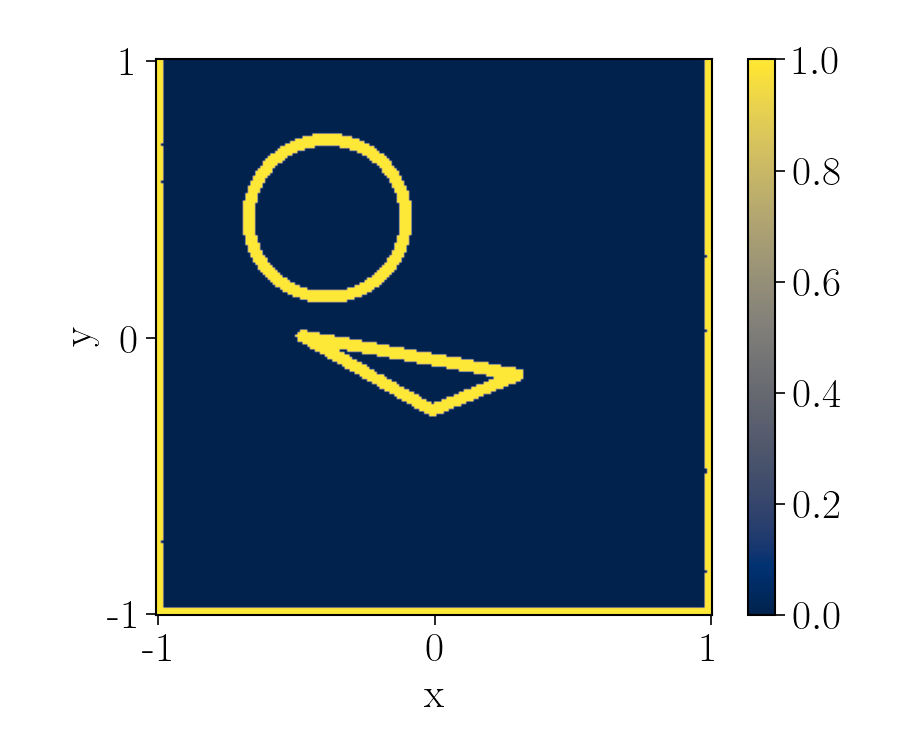}
        \caption{$\partial \Omega_D$}
        \label{fig:sampled_pde_supervised_dirichlet}
    \end{subfigure}
    \hfill
    \begin{subfigure}[b]{0.16\linewidth}
        \includegraphics[width=\linewidth]{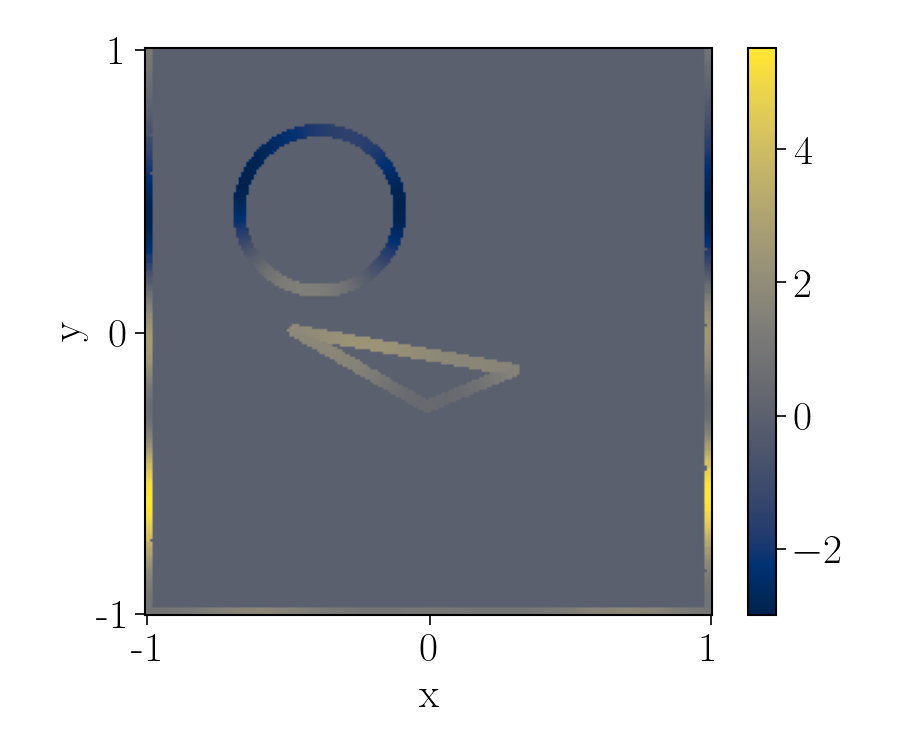}
        \caption{$g(x, y)$}
        \label{fig:sampled_pde_supervised_dirichlet_condition}
    \end{subfigure}
    \hfill
    \begin{subfigure}[b]{0.16\linewidth}
        \includegraphics[width=\linewidth]{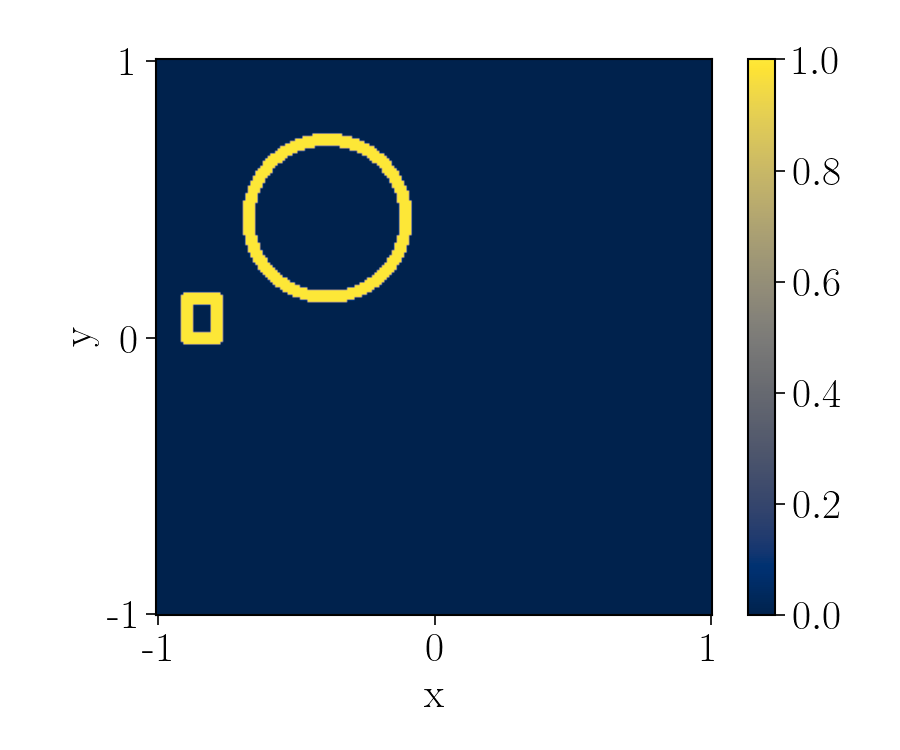}
        \caption{$\partial \Omega_N$}
        \label{fig:sampled_pde_supervised_neumann}
    \end{subfigure}
    \hfill
    \begin{subfigure}[b]{0.16\linewidth}
        \includegraphics[width=\linewidth]{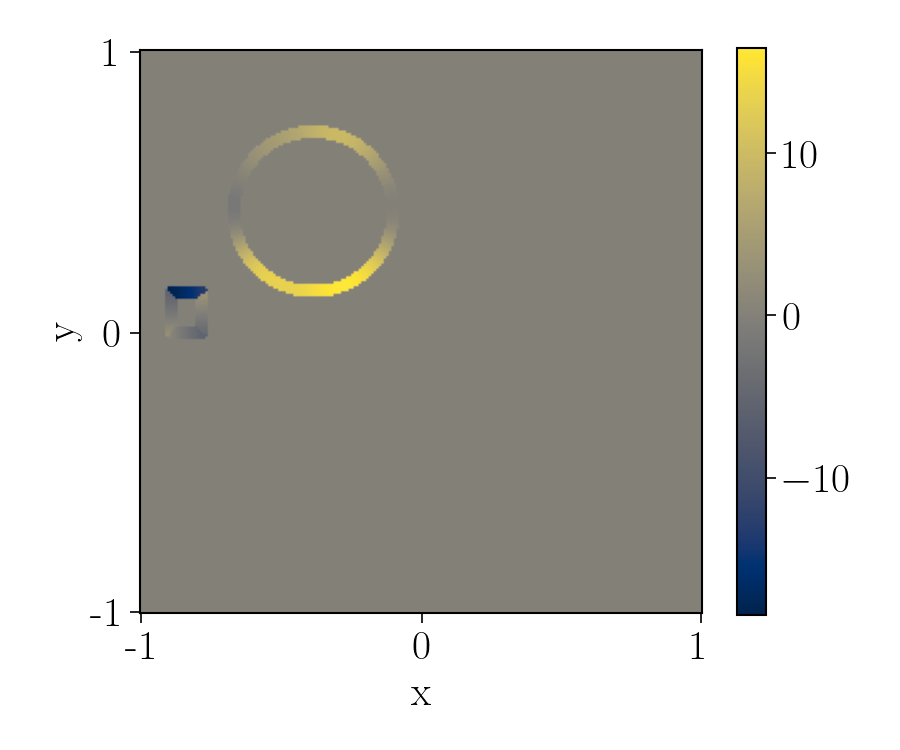}
        \caption{$h(x, y)$}
        \label{fig:sampled_pde_supervised_neumann_condition}
    \end{subfigure}
    \hfill
    \begin{subfigure}[b]{0.16\linewidth}
        \includegraphics[width=\linewidth]{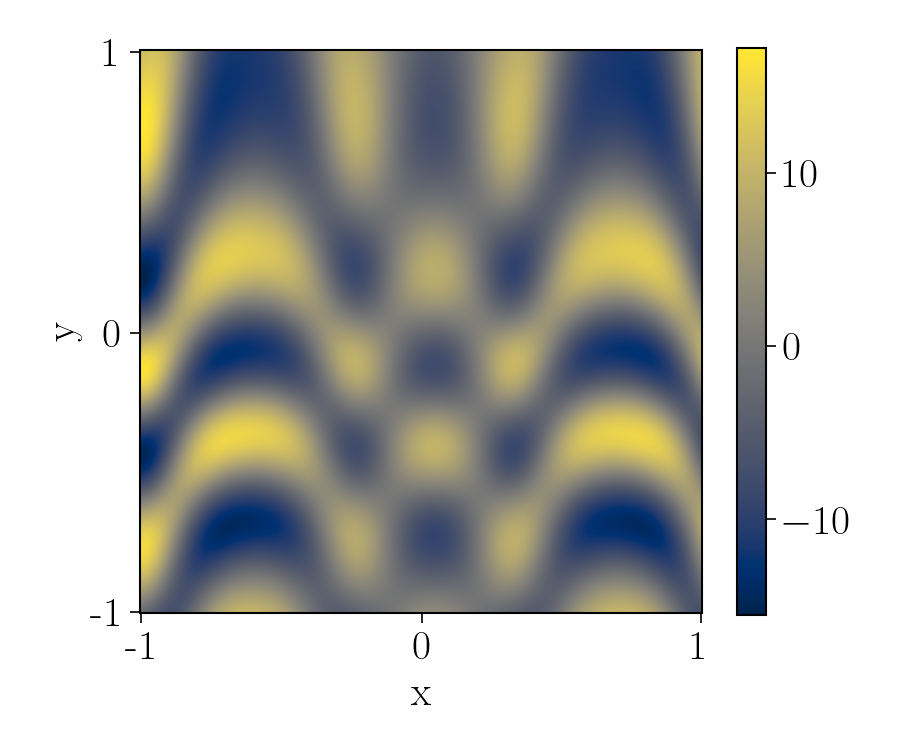}
        \caption{$f(x, y)$}
        \label{fig:sampled_pde_supervised_f}
    \end{subfigure}
    \hfill
    \begin{subfigure}[b]{0.16\linewidth}
        \includegraphics[width=\linewidth]{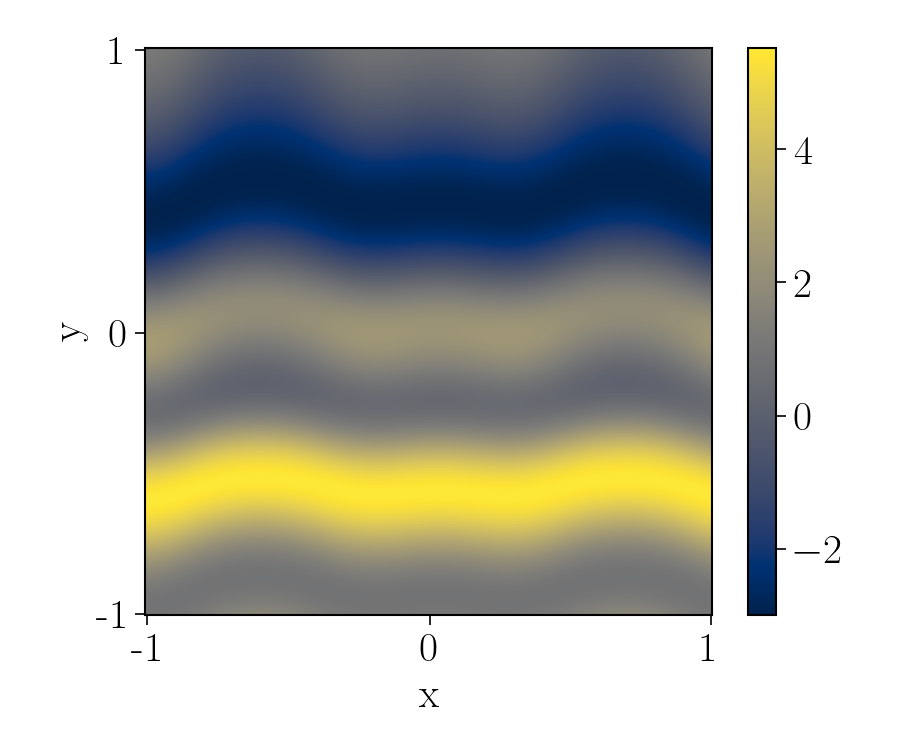}
        \caption{$u(x, y)$}
        \label{fig:sampled_pde_supervised_u}
    \end{subfigure}
    
    \caption{
    Sample generated via MMS with sampled operator $\mathcal{L}[u] = -0.31 u_{xx} - 0.15 u_y$ and sampled boundaries $\partial \Omega$: (a) Dirichlet boundary, (b) Dirichlet condition, (c) Neumann boundary, (d) Neumann condition, (e) source term, and (f) analytical solution.
    }
    \label{fig:sampled_pde_supervised}
\end{figure}

To support a wide range of linear PDEs while maintaining compatibility with modern machine learning models, we adopt a parameterization that is flexible, user-friendly, and efficiently processed by state-of-the-art architectures. The function $f$ is discretized on a uniform grid over $\Omega$, resulting in a 2D array $F$ representing its values at grid points. The boundary conditions are parametrized by creating two 2D grids per boundary type ($\partial \Omega_D$, $\partial \Omega_N$): (i) A binary mask $M$ indicating the presence of the boundary at each grid point, where we assign a value of 1 to the four grid cells closest to each boundary point and zero elsewhere; and
(ii) a value grid $V$ storing the corresponding boundary values ($g$ for Dirichlet or $h$ for Neumann conditions) at those marked cells, with zeros elsewhere. Figure~\ref{fig:sampled_pde_supervised} illustrates a sampled PDE instance with its full parameterization. Finally, following \cite{iwata2023meta}, we parameterize $\mathcal{L}$ as $\mathcal{L}[u](\mathbf{x}) = c_1 u + c_2 u_x + c_3 u_y + c_4 u_{xx} + c_5 u_{yy}$, where $\mathbf{c} = (c_1, c_2, c_3, c_4, c_5) \in \mathbb{R}^5$ encodes the operator coefficients.

\subsection{Neural Operator Architecture}
\label{sec:neural_operator}

We base our model on HyperPINN~\cite{hyperpinn}, a hypernetwork-based neural operator that maps a parametrized PDE instance to the weights $\theta$ of a PINN $u_{\theta}$ specialized to that instance. Formally, the hypernetwork realizes a mapping
\begin{equation}
\big(\mathbf{c}, \; F,\; M_g,\; M_h,\; V_g,\; V_h \big) \;\longmapsto\; \theta^\star
\quad\text{such that}\quad 
u_{\theta^\star} \approx u,
\label{eq:hyperpinn}
\end{equation}
where $\mathbf{c}$ denotes the vector of PDE coefficients, $F$ the discretized source function, $M_g$ and $M_h$ the Dirichlet and Neumann boundary condition location grids, $V_g$ and $V_h$ the Dirichlet and Neumann boundary condition value grids, and $u$ the reference solution.

The vector of operator coefficients $\mathbf{c} \in \mathbb{R}^5$ is embedded into a fixed-length representation $z_C \in \mathbb{R}^{d_C}$ using a Fourier feature mapping \cite{tancik2020fourier} which was shown to prevent spectral bias and mode collapse, in particular in physics-informed settings \cite{fourier_pinns}. The grid-based inputs $F$, $M_g$, $M_h$, $V_g$ and $V_h$ are concatenated and processed via a series of \(K\) Swin Transformer blocks \(\{\mathcal{SW}_i\}_{i=1}^K\) \cite{swin}. After each block, we introduce a FiLM layer \cite{film}, which modulates the Swin block's output conditioned on $z_C$:
\begin{equation}
z_{i+1} \;=\; \operatorname{FiLM}_i\bigl(\mathcal{SW}_i(z_i),\; z_C\bigr), \quad z_i \in \mathbb{R}^{H_i \times W_i \times C_i}
\label{eq:modulation_block}
\end{equation}
Inspired by Swin Transformer U-Net architectures~\cite{swin_unet, swin_unet_denoising}, we retain all intermediate latent representations $\{z_i\}_{i=1}^{K}$ to keep information at various semantic levels. To enable information aggregation, we flatten the spatial dimensions $H_i$ and $W_i$ and use Multi-Head Attention Pooling \cite{lee2019set,zhai2022scaling}, where a set of trainable query vectors $\{q_i\}_{i=1}^K, q_i \in \mathbb{R}^{T \times C_i}$ is defined. $T$ corresponds to the number of weight and bias tensors in the target PINN. The queries $q_i$ are then used in a multi-head attention module together with $z_i$ reshaped into $kv_i \in \mathbb{R}^{H_i \times W_i \times C_i}$:
\begin{equation}
    p_i = \operatorname{MultiHeadAttention}_i(q_i, kv_i, kv_i), \quad p_i \in \mathbb{R}^{T \times C_i}.
    \label{eq:pooling}
\end{equation}
The outputs $\{p_i\}_{i=1}^K$ are concatenated along the channel dimension to produce a unified latent matrix \(p = \left[p_1 \;\Vert\; p_2 \;\Vert\; \cdots \;\Vert\; p_K\right] \in \mathbb{R}^{T \times \left(\sum_{i=1}^K C_i\right)}\), containing an entry of aggregated information for each weight matrix and bias vector in the target PINN. Finally, dedicated MLPs project each entry into the required dimensionality for the corresponding weight matrix or bias vector.

We define the architecture of the target PINN as an MLP with Fourier feature mapping~\cite{tancik2020fourier}, which, when concatenated to the input $(x, y)$, results in a dimensionality of $2N + 2$, and multiplicative skip connections \cite{wang2021mult_skip_connections}.
Fourier encodings provide spectral expressivity for modeling high-frequency components \cite{fourier_pinns}, while the skip connections enhance gradient propagation and, in the context of hypernetworks, have the additional benefit of enabling dynamic depth modulation based on PDE complexity by allowing the hypernet to mask some layers.

For each PDE instance, the hypernetwork therefore generates the following parameter set $\theta^\star$:
\begin{equation}
\left\{W_0, U, V, b_0, b_u, b_v \right\}, \quad
\left\{W_i, b_i\right\}_{i=1}^{T-2}, \quad
W_{\text{out}}, b_{\text{out}},
\label{eq:pinn_parameter_set}
\end{equation}
where \( d \) denotes the width of the latent layers. The parameter dimensions are as follows: \( W_0, U, V \in \mathbb{R}^{d \times (2N + 2)} \) and \( b_0, b_u, b_v \in \mathbb{R}^{d} \); for \( i = 1, \dots, T-2 \), \( W_i \in \mathbb{R}^{d \times d} \) and \( b_i \in \mathbb{R}^{d} \); and finally, \( W_{\text{out}} \in \mathbb{R}^{1 \times d} \) and \( b_{\text{out}} \in \mathbb{R} \). Note that we use the $\operatorname{tanh}$ activation function due to its bounded output space, which provides stability to the hypernet's training.
\subsection{Data Sampling}\label{sec:data_sampling}
We create a synthetic dataset of PDEs by randomly drawing the differential operator \(\mathcal{L}\), domain \(\Omega\), boundary data, source term \(f\), and, when available, a reference solution \(u\). The full dataset is a mix of two classes, supervised and unsupervised samples. For supervised samples, a manufactured analytical solution \(u\) is chosen first. We then set \(f=\mathcal{L}[u]\) and derive \(g(\mathbf{x}) = u(\mathbf{x})\) and / or \(h(\mathbf{x}) = \tfrac{\partial u}{\partial n}(\mathbf{x})\) by evaluating \(u(\mathbf{x})\) and its normal derivative on \(\partial \Omega\). In addition to the physics-informed loss, samples of this class provide the analytical solution \(u(\mathbf{x})\) as well as its derivatives that can be used for additional supervised losses during training. For unsupervised samples, the reference solution \(u\) is unknown. We sample \(f\) and boundary conditions subject to constraints designed to maximize diversity and the probability of well-posedness. Training relies solely on the physics-informed loss, as reference solutions are unavailable.

Differential operators \(\mathcal{L}\) are formed by sampling \(n \sim \text{Uniform}(\{1,2,3\})\) terms from \(\mathcal{B} = \{u, u_x, u_y, u_{xx}, u_{yy}\}\) without replacement. Each selected term $T_i$ is assigned a coefficient \(c_i \sim \text{Uniform}([-2,2])\), and the operator is defined as \(\mathcal{L}[u] = \sum_{i=1}^{n} c_i T_i[u]\).

To generate supervised samples, we use MMS, an established approach for validating PDE solvers. We first construct an analytical solution \(u : \Omega \to \mathbb{R}\) on a domain \(\Omega \subset \mathbb{R}^2\) by applying \(n \sim \text{Uniform}(\{6,\dots,10\})\) iterative updates starting from \(u(x,y) \gets 0\). Each update adds a term of the form \(d \cdot \psi(a x + b y + c) + e\), where \(\psi \in \{x, \sin, \cos, \tanh, (1 + e^{-x})^{-1}, (1 + x^2)^{-1}\}\), and coefficients  \(a,b \in \{0, \text{Uniform}([-10,10])\}\), and \(c,d,e \sim \text{Uniform}([-2\pi,2\pi])\). Terms are incorporated using one of three rules chosen uniformly at random: additive, multiplicative, or compositional.

Source term generation depends on the availability of an analytical solution. For supervised samples, we compute \(f(\mathbf{x}) = \mathcal{L}[u](\mathbf{x})\) via symbolic differentiation. For unsupervised samples, where \(u\) is unknown, we set \(f(\mathbf{x}) = \mathcal{N}(0, 10^2)\), i.e., a spatially constant random source drawn from a zero-mean Gaussian.

Domains \(\Omega \subset [-1, 1]^2\) are generated via randomized Constructive Solid Geometry (CSG)~\cite{deepxde}. The outer boundary \(\partial \Omega_{\text{outer}}\) is defined as the unit square and may represent either purely spatial or spatio-temporal domains, with \(y = -1\) marking the initial time in time-dependent PDEs. Inner boundaries \(\partial \Omega_{\text{inner}, i}\) are formed by subtracting randomly sampled geometric primitives (e.g., disks, polygons, rectangles) from the outer region. These boundaries enclose regions where the source term \(f\) remains active, and their role (e.g., obstacle vs. inclusion) is encoded implicitly through the boundary type.

Each inner boundary \(\partial \Omega_{\text{inner}, i}\) is randomly assigned Dirichlet, Neumann, or both: \(u(\mathbf{x}) = g_i(\mathbf{x})\) or \(\partial u/\partial n = h_i(\mathbf{x})\). For supervised samples, where \(u(\mathbf{x})\) is known, we set \(g = u\) and \(h = \partial u/\partial n\). For unsupervised samples, boundary values are sampled to promote compatibility with the operator \(\mathcal{L}[u]\). If \(u\) appears as a standalone term, we set \(u = 0\) on \(\partial \Omega\) to avoid trivial or inconsistent configurations (e.g., \(u = f\) with nonzero \(f\)). If first-order terms (e.g., \(u_x, u_y\)) appear alone, constant Dirichlet values are used. In other cases, linear profiles are allowed, offering mild spatial variability without conflicting with the constant source term of unsupervised samples.

Despite efforts to ensure well-posedness, some unsupervised samples may still be ill-posed due to incompatible boundary and source term configurations. Nonetheless, they are essential for exposing the model to realistic complexities such as interior boundaries, inclusions, and discontinuities. These are common features in practical PDEs but are difficult to introduce through supervised data generated via MMS.
\subsection{Objective Function}
\label{sec:objective}

For each PDE instance $(\mathcal{L}, f, g, h)$ on $\Omega \subset [-1, 1]^2$ with Dirichlet ($\partial\Omega_D$) and Neumann ($\partial\Omega_N$) boundaries, HyPINO \(\Phi: (\mathcal{L}, f, g, h) \mapsto \theta^\star\) produces weights $\theta^\star$ for a target PINN $u_{\theta^\star} : \Omega \to \mathbb{R}$.
\begin{equation}
    \mathcal{J}_{\mathrm{R}} =
    \frac{1}{|\Omega|} \sum_{\mathbf{x} \in \Omega}
    \rho\left( \mathcal{L}[u_{\theta^\star}](\mathbf{x}) - f(\mathbf{x}) \right)
    \label{eq:residual_loss}
\end{equation}
is the residual loss and \(\rho(\cdot)\) the Huber function~\cite{huber1992robust}. The Dirichlet and Neumann losses are computed similarly:
\begin{equation}
    \mathcal{J}_{\mathrm{D}} =
    \frac{1}{|\partial\Omega_D|} \sum_{\mathbf{x} \in \partial \Omega_D}
    \rho\left( u_{\theta^\star}(\mathbf{x}) - g(\mathbf{x}) \right),
    \qquad
    \mathcal{J}_{\mathrm{N}} =
    \frac{1}{|\partial\Omega_N|} \sum_{\mathbf{x} \in \partial \Omega_N}
    \rho\left( \nabla u_{\theta^\star}(\mathbf{x}) \cdot \mathbf{n}(\mathbf{x}) - h(\mathbf{x}) \right).
    \label{eq:boundary_losses}
\end{equation}
For PDEs with known analytical solutions \(u\), we add a second-order Sobolev loss~\cite{sobolev} that penalizes errors in function values, gradients, and second derivatives:
\begin{equation}
    \mathcal{J}_{\mathrm{S}} =
    \frac{1}{|\Omega|} \sum_{\mathbf{x} \in \Omega}
    \sum_{k=0}^{2} \lambda_{\mathrm{S}}^{(k)} \rho\left( \nabla^k u_{\theta^\star}(\mathbf{x}) - \nabla^k u(\mathbf{x}) \right).
    \label{eq:sobolev_loss_compact}
\end{equation}
The total loss is a weighted sum of the active terms:
\begin{equation}
    \mathcal{J} =
    \lambda_{\mathrm{R}} \mathcal{J}_{\mathrm{R}} +
    \lambda_{\mathrm{D}} \mathcal{J}_{\mathrm{D}} +
    \lambda_{\mathrm{N}} \mathcal{J}_{\mathrm{N}} +
    \mathcal{J}_{\mathrm{S}},
    \label{eq:weighted_loss}
\end{equation}
where \(\mathcal{J}_{\mathrm{R}}\) is always included, \(\mathcal{J}_{\mathrm{D}}\) and \(\mathcal{J}_{\mathrm{N}}\) are applied when collocation points fall on \(\partial\Omega_D\) or \(\partial\Omega_N\), and \(\mathcal{J}_{\mathrm{S}}\) is active only when the ground-truth solution $u$ is known.

\subsection{Residual–Driven Iterative Refinement}
\label{sec:iterative-refinement}

Using a hypernetwork to generate a single PINN of fixed architecture may seem restrictive, particularly in multi-physics settings where different PDEs may demand different levels of representational complexity. However, hypernetworks offer a natural mechanism for generating ensembles of PINNs at inference time, which have proven effective in reducing prediction error \cite{bischof2022mixture,xpinn,gated_pinn}. Beyond naïvely producing multiple independent samples, our framework for linear PDEs supports an ensemble construction through an iterative refinement procedure, similar in spirit to multi-stage neural networks that progressively reduce residual error~\cite{wang2024multistage}:

Given a PDE instance $(L, f, g, h)$, the hypernetwork generates an initial solution $u^{(0)} := u_{\Phi(L, f, g, h)}$. We compute residuals $r_f^{(0)}$, $r_D^{(0)}$ and $r_N^{(0)}$ with respect to the PDE and boundary conditions, treat the residuals as a “delta PDE” and feed them back into the hypernetwork to obtain a corrective PINN:
\begin{equation}
    \delta u^{(1)} := u_{\Phi(L, r_f^{(0)}, r_D^{(0)}, r_N^{(0)})}.
    \label{eq:delta_u_1}
\end{equation}
The updated solution is \(u^{(1)} := u^{(0)} + \delta u^{(1)}\). We repeat this process for $t = 0, \dots, T{-}1$:
\begin{equation}
    u^{(t+1)} := u^{(t)} + \delta u^{(t+1)}, \quad \text{with} \quad \delta u^{(t+1)} := u_{\Phi(L, r_f^{(t)}, r_D^{(t)}, r_N^{(t)})}.
    \label{eq:refinement_step}
\end{equation}
After $T$ iterations, the final prediction is \(u^{(T)} = u^{(0)} + \sum_{t=1}^{T} \delta u^{(t)}\). We refer to this model as HyPINO$^{i}$, where $i$ defines the number of refinement rounds.

During iterative refinement, only the small PINNs are differentiated to compute residuals, whereas the hypernetwork $\Phi$ remains in inference mode. We use uniform weights for each $\delta u^{(t)}$, though adaptive weighting (e.g., scaled residual updates) remains a promising direction for future work.
\newpage
\section{Experiments}
\label{sec:experiments}

\subsection{Training}
\label{sec:training}

HyPINO generates weights for a target PINN with three hidden layers and 32 hidden units per layer. The full model has 77M trainable parameters. We train the hypernetwork for 30,000 batches using the AdamW optimizer with a cosine learning rate schedule from $10^{-4}$ to $10^{-6}$ and a batch size of 128. Training was conducted on 4 NVIDIA RTX 4090 GPUs for all experiments.

Training is divided into two phases. In the first 10,000 batches, all samples are supervised with known analytical solutions, using loss weights: $\lambda_{\mathrm{R}} = 0.01$, $\lambda_{\mathrm{S}}^{(0)} = 1$, $\lambda_{\mathrm{S}}^{(1)} = 0.1$, $\lambda_{\mathrm{S}}^{(2)} = 0.01$, $\lambda_{\mathrm{D}} = 10$, and $\lambda_{\mathrm{N}} = 1$. In the remaining 20,000 batches, each batch consists of 50\% supervised and 50\% unsupervised samples. Loss weights are updated to: $\lambda_{\mathrm{R}} = 0.1$, $\lambda_{\mathrm{S}}^{(0)} = 1$, $\lambda_{\mathrm{S}}^{(1)} = 1$, $\lambda_{\mathrm{S}}^{(2)} = 0.1$, $\lambda_{\mathrm{D}} = 10$, and $\lambda_{\mathrm{N}} = 1$.

\subsection{Baseline Models}
\label{sec:baselines}

We compare HyPINO against three baselines, each trained for 30{,}000 batches with batch size 128 and an initial learning rate of $10^{-4}$: (i) \textbf{U-Net}~\cite{unet}, which shares HyPINO's encoder but replaces the hypernetwork decoder with a convolutional decoder that directly outputs a $224 \times 224$ solution grid. It is trained solely on supervised data and has 62M trainable parameters. (ii) \textbf{Poseidon}~\cite{pde_foundation_model_2024_Herde}, a pretrained neural operator with 158M parameters. We use the Poseidon-B checkpoint and adapt it by changing the embedding and lead-time-conditioned layer normalization layers' dimensionality to match the size of our parameterization. Similarly to the U-Net, Poseidon is fine-tuned only on supervised data. (iii) \textbf{PINO}~\cite{pino}, a Fourier neural operator \cite{fno} with 33M parameters. We adapt it to accept 5-channel grid inputs and condition on the PDE operator using FiLM layers. It is trained using the same hybrid supervision and curriculum as HyPINO, including physics-informed losses.

\subsection{Evaluation}
\label{sec:evaluation}

We evaluate HyPINO and baseline models on seven standard PDE benchmarks from the PINN literature: (i) \textbf{HT} - 1D heat equation~\cite{deepxde}, (ii) \textbf{HZ} - 2D Helmholtz equation~\cite{relobralo}, (iii) \textbf{HZ-G} - Helmholtz on an irregular geometry~\cite{pinnacle}, (iv) \textbf{PS-C} - Poisson with four circular interior boundaries~\cite{pinnacle}, (v) \textbf{PS-L} - Poisson on an L-shaped domain~\cite{deepxde}, (vi) \textbf{PS-G} - Poisson with a Gaussian vorticity field~\cite{pde_foundation_model_2024_Herde}, and (vii) \textbf{WV} - 1D wave equation~\cite{pinnacle}. The exact problem statements and visualizations of the corresponding parameterizations are provided in Appendix~\ref{app:experiments}.
\begin{table}[h]
  \centering
  \caption{Model performance across seven PDE benchmarks. Each cell shows mean squared error (MSE) / symmetric mean absolute percentage error (SMAPE) \cite{smape}. Lower is better.}
  \vspace{5pt}
  \label{tab:model_performances}
  \begingroup
  \setlength{\tabcolsep}{4pt}
  \small
  \renewcommand{\arraystretch}{1.1}
  \begin{tabular}{l *{7}{c}}
    \toprule
     & \textbf{HT} & \textbf{HZ} & \textbf{HZ-G} & \textbf{PS-C} & \textbf{PS-L} & \textbf{PS-G} & \textbf{WV} \\
    \midrule
    \textbf{U-Net} & 
    3.5e-2 / 67 & 
    3.7e-2 / 68 & 
    6.9e-2 / 68 & 
    2.7e-2 / 33 & 
    3.9e-3 / 112 & 
    9.2e-1 / 159 & 
    3.7e-1 / 144 \\
    
    \textbf{Poseidon} & 
    7.1e-2 / 47 & 
    3.3e-3 / 28 & 
    1.3e-1 / 65 & 
    5.3e-2 / 93 & 
    3.5e-3 / 111 & 
    7.2e-1 / 155 & 
    8.7e-1 / 138 \\
    
    \textbf{PINO} & 
    1.4e-2 / 38 & 
    2.0e-2 / 51 & 
    6.1e-2 / 60 & 
    1.7e-1 / 65 & 
    3.3e-3 / 51 & 
    3.1e-1 / 70 & 
    3.0e-1 / 149 \\
    
    \textbf{PINO$^3$} & 
    1.3e-2 / 47 & 
    7.2e-3 / 48 & 
    4.6e-2 / 64 & 
    2.8e-2 / 63 & 
    4.6e-3 / 62 & 
    2.3e-2 / 43 & 
    3.1e-1 / 127 \\
    
    \textbf{PINO$^{10}$} & 
    3.9e-2 / 78 & 
    5.1e-3 / 39 & 
    1.4e-1 / 75 & 
    1.1e-2 / 48 & 
    1.0e-3 / 47 & 
    1.8e-2 / 38 & 
    8.5e-1 / 139 \\
    \midrule
    \textbf{HyPINO} & 
    2.3e-2 / 42 & 
    5.7e-3 / 36 & 
    1.3e-1 / 64 & 
    5.6e-2 / 86 & 
    \textbf{1.7e-4} / 39 & 
    1.8e-1 / 61 & 
    2.9e-1 / 150 \\
    
    \textbf{HyPINO$^3$} & 
    4.9e-4 / 11 & 
    2.7e-3 / 31 & 
    \textbf{1.6e-2} / \textbf{38} & 
    3.4e-3 / 18 & 
    1.9e-4 / \textbf{36} & 
    6.6e-3 / 25 & 
    2.3e-1 / 134 \\
    
    \textbf{HyPINO$^{10}$} & 
    \textbf{8.0e-5} / \textbf{7} & 
    \textbf{1.6e-3} / \textbf{22} & 
    1.9e-2 / 40 & 
    \textbf{2.3e-3} / \textbf{15} & 
    2.7e-4 / 40 & 
    \textbf{5.0e-3} / \textbf{24} & 
    \textbf{1.2e-1} / \textbf{96} \\
    \bottomrule
  \end{tabular}
  \endgroup
\end{table}

We summarize model performance across the seven PDE benchmarks in Table~\ref{tab:model_performances}. HyPINO demonstrates consistently strong results, achieving an average rank of 2.00 across all tasks, compared to 3.00 for U-Net, 2.86 for Poseidon, and 2.14 for PINO. It is important to note that neither Poseidon nor PINO was originally designed for the PDE parameterization chosen in this study. As such, some degree of performance degradation is expected. In contrast, HyPINO contains a dedicated embedding mechanism tailored to this parameterization, but faces the challenge of operating in a significantly less structured output space compared to the grid-based outputs of the baselines. Its competitive zero-shot performance under these conditions is therefore noteworthy. Across all benchmarks, models trained with physics-informed objectives generally outperform those relying solely on supervised data. This indicates that incorporating physics-based losses helps mitigate the generalization gap between the synthetic training data and the evaluation tasks. 

Table~\ref{tab:model_performances} further highlights the advantages of our proposed iterative refinement approach. After three refinement iterations (HyPINO$^3$), we observe substantial reductions in prediction error across all but one benchmark. Notably, the MSE for PS-C and PS-G decreases by more than one order of magnitude, and an even larger improvement is observed for HT (almost two orders of magnitude). With ten refinement iterations (HyPINO$^{10}$), our model achieves state-of-the-art performance on all but two evaluated benchmarks, outperforming the best baseline models by factors ranging from 2.1 (on HZ against Poseidon) to 173 (HT against PINO). Table~\ref{tab:model_predictions} shows that iterative refinement leads to a progressively more accurate prediction on the challenging WV benchmark, with the model being able to extend the undulating shape continuously further from the initial condition across the time dimension. Importantly, our results indicate that iterative refinement is not specific to HyPINO but serves as a generally effective test-time enhancement for other physics-informed neural operators, as demonstrated by the performance of PINO$^3$ and PINO$^{10}$.

We hypothesize that these improvements arise because the iterative procedure allows for correcting systematic biases introduced during training on synthetic data, which, despite its diversity and breadth, remains composed of relatively simple basis functions. As these training-induced errors tend to be consistent, artifacts produced by the initial HyPINO-generated PINNs can be systematically corrected in subsequent iterations. This residual-driven refinement yields ensembles that are significantly more effective than naive ensembles formed from independently generated target networks.

\noindent
\begin{table}[htbp]
    \centering
    \caption{Comparison of predictions and errors of HyPINO after zero, three, and 10 refinement rounds across all benchmark PDEs.}
  \vspace{5pt}
    \label{tab:model_predictions}
    \begin{tabular}{>{\centering\arraybackslash}m{0.1\textwidth}@{}m{0.125\textwidth}@{}|@{}m{0.25\textwidth}@{}|@{}m{0.25\textwidth}@{}|@{}m{0.25\textwidth}@{}}
    \textit{HT} &
    \includegraphics[width=0.94\linewidth]{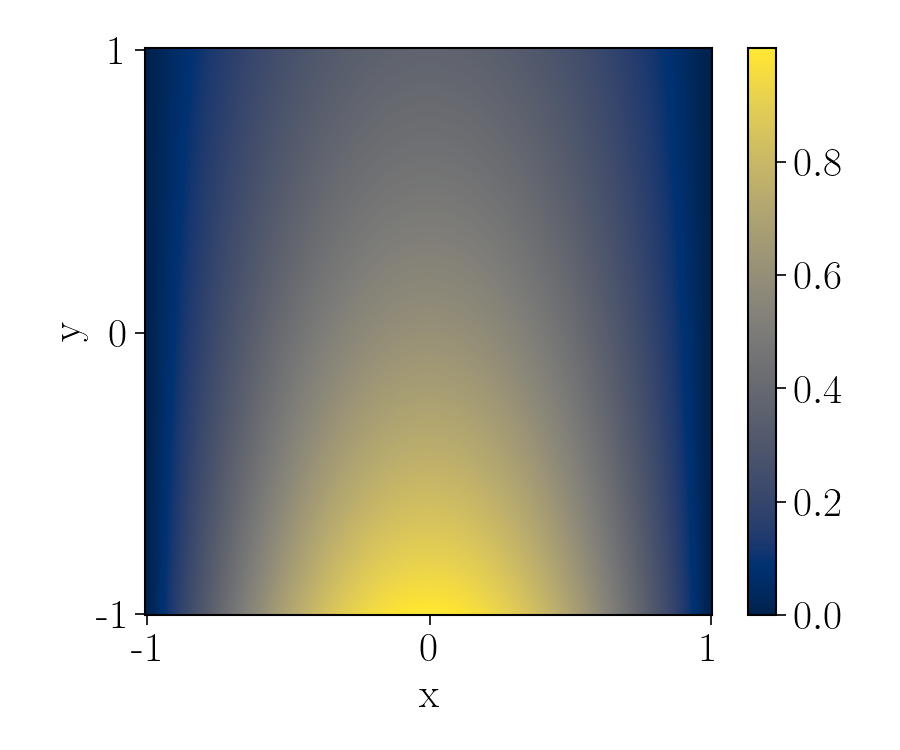} &
    \includegraphics[width=0.47\linewidth]{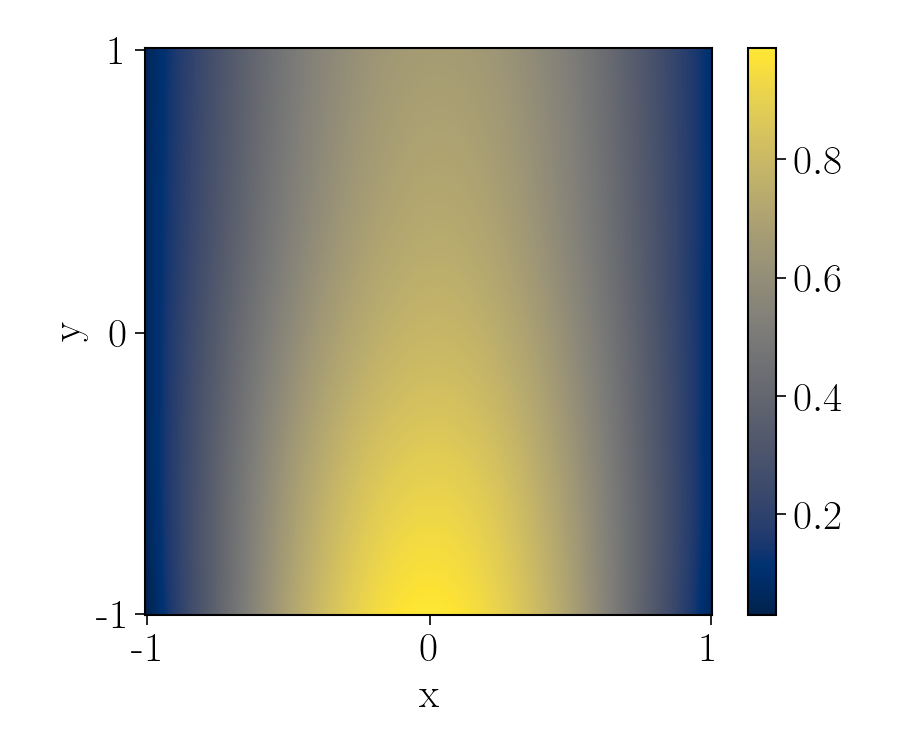}
    \includegraphics[width=0.47\linewidth]{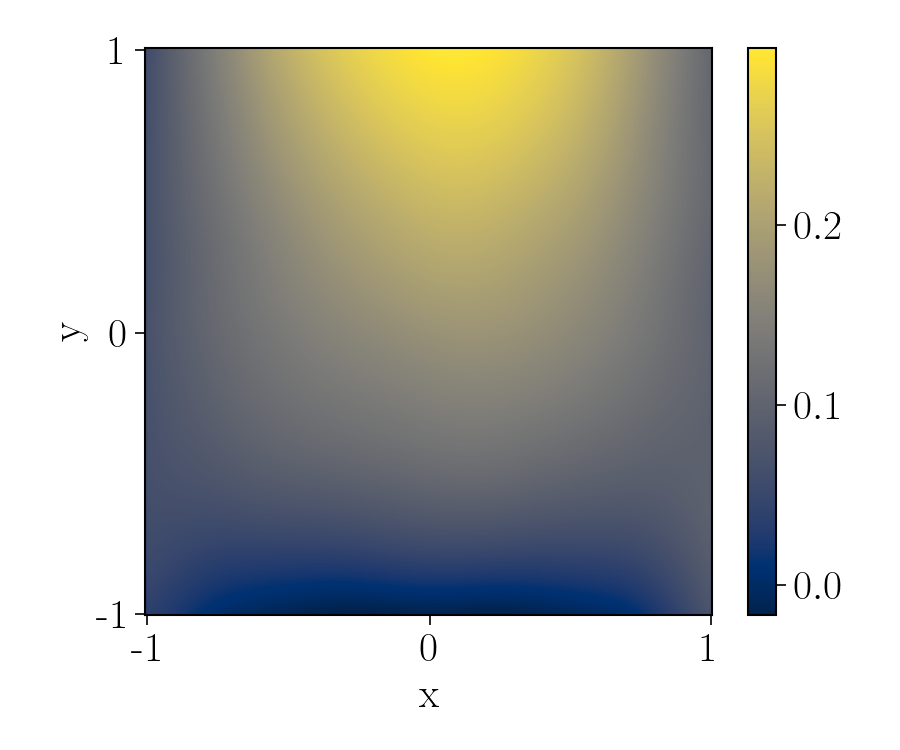} &
    \includegraphics[width=0.47\linewidth]{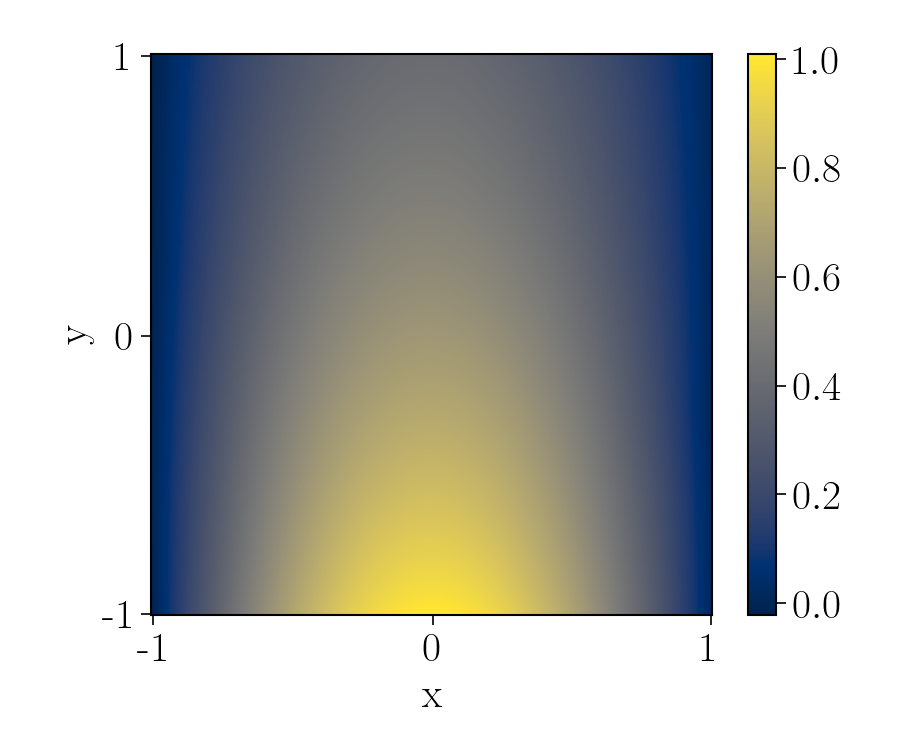}
    \includegraphics[width=0.47\linewidth]{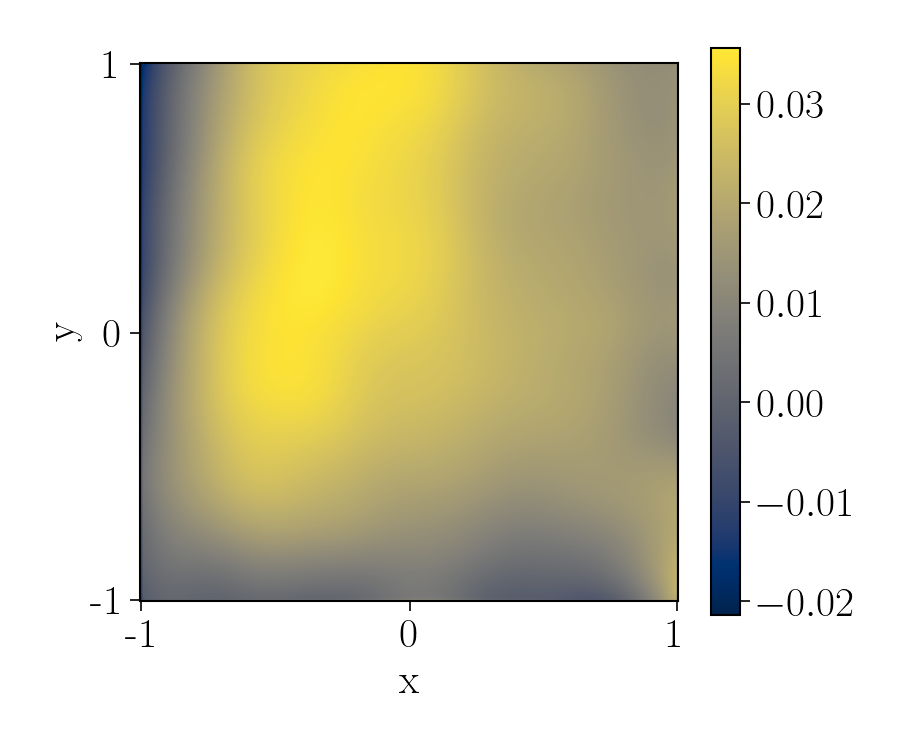} &
    \includegraphics[width=0.47\linewidth]{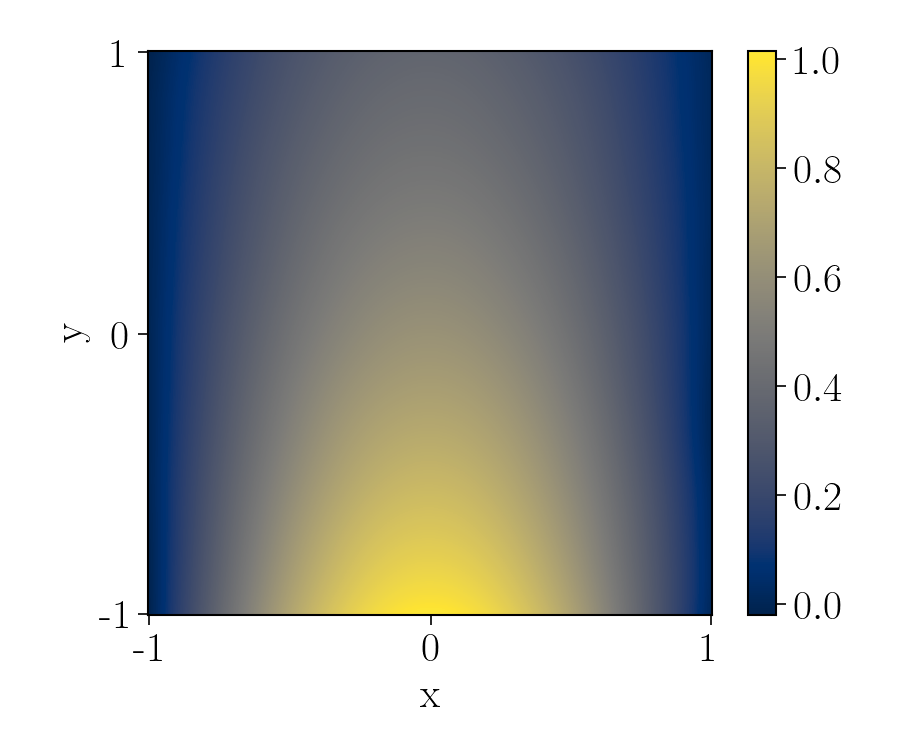}
    \includegraphics[width=0.47\linewidth]{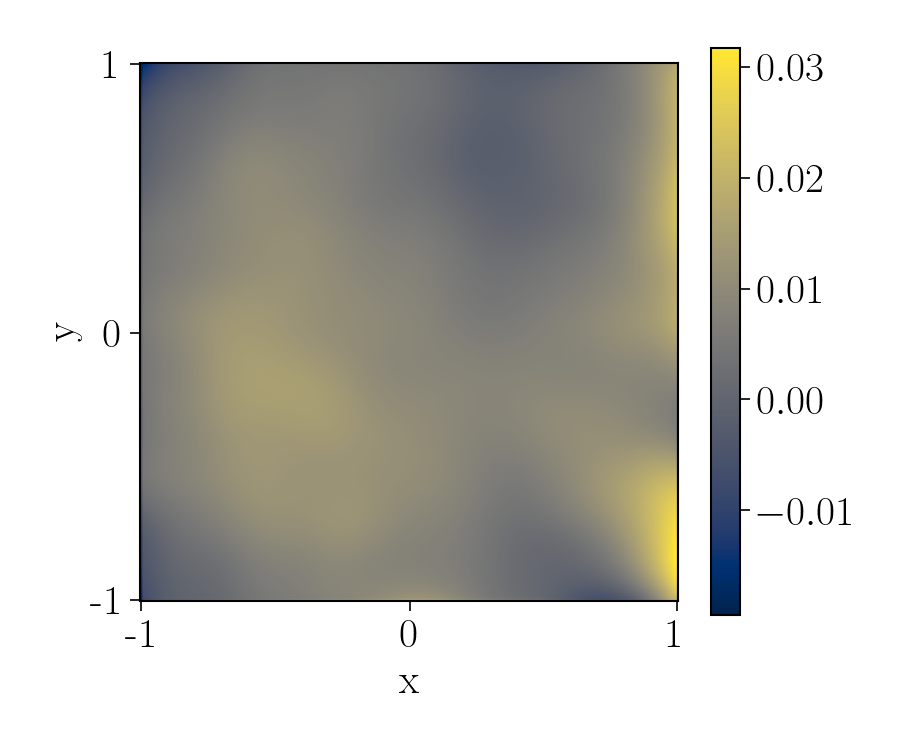} \\
    \hline
    \textit{HZ} &
    \includegraphics[width=0.94\linewidth]{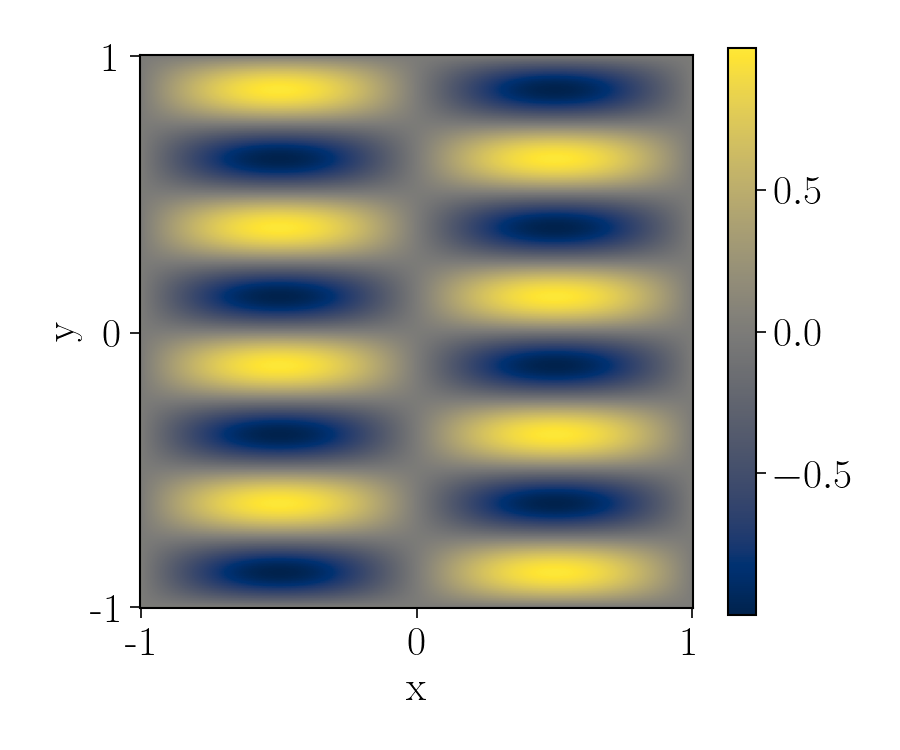} &
    \includegraphics[width=0.47\linewidth]{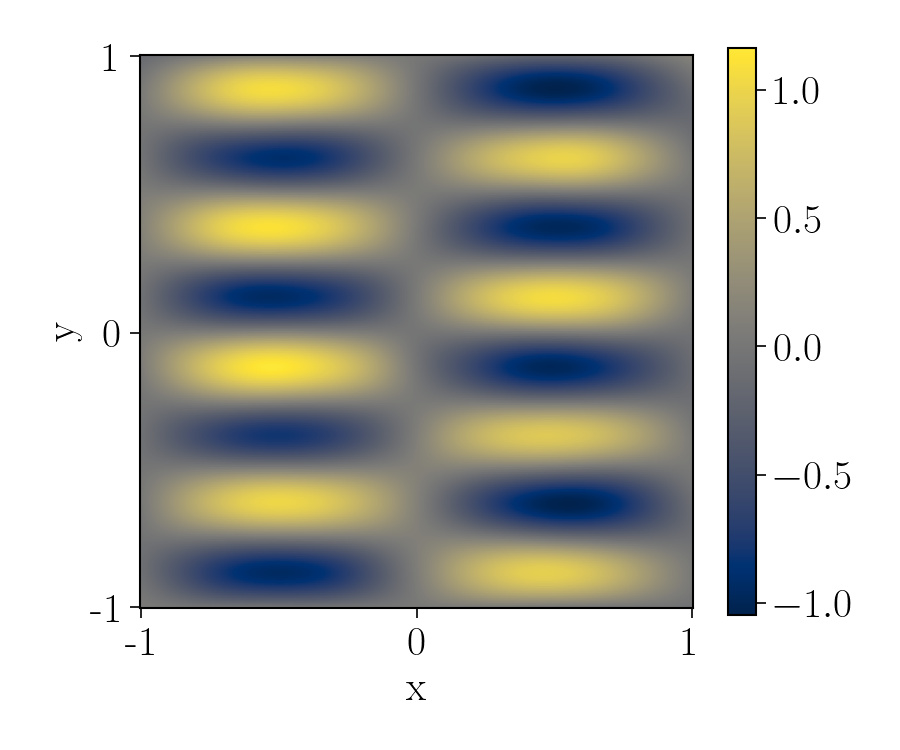}
    \includegraphics[width=0.47\linewidth]{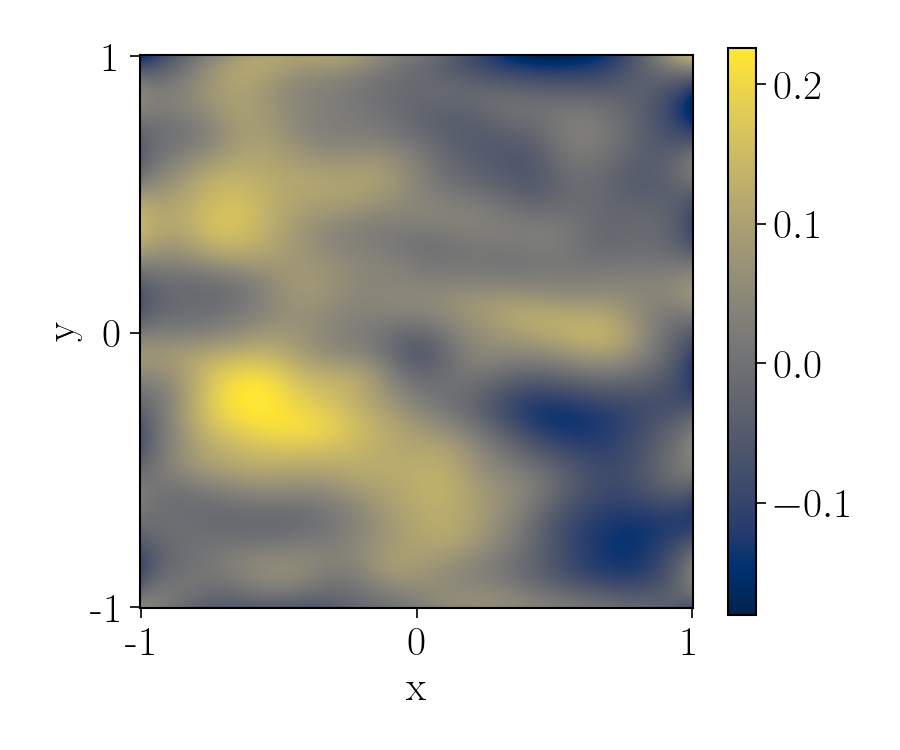} &
    \includegraphics[width=0.47\linewidth]{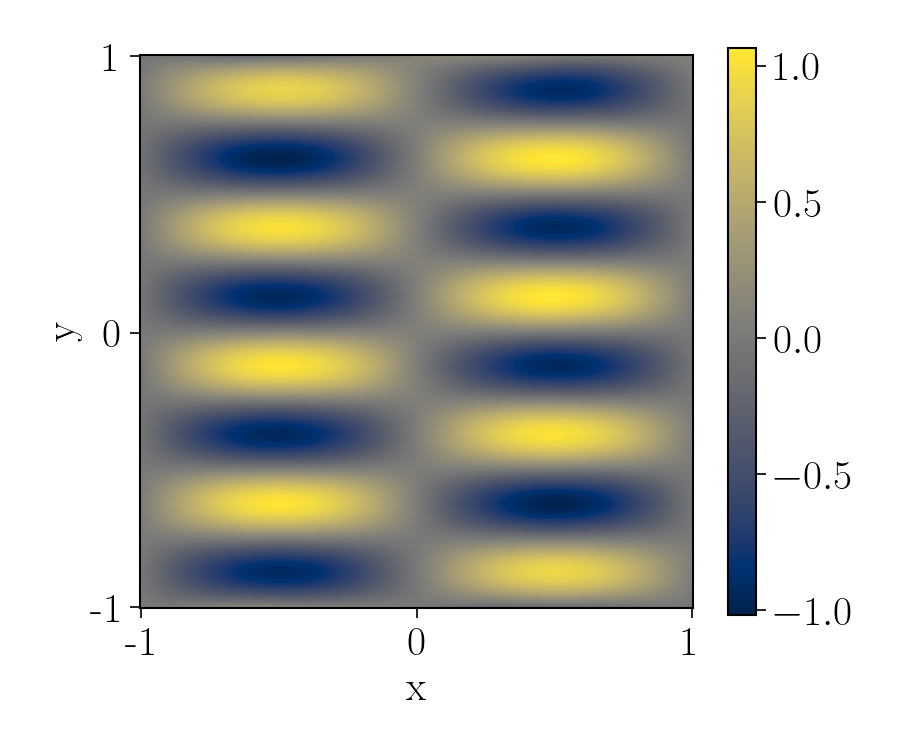}
    \includegraphics[width=0.47\linewidth]{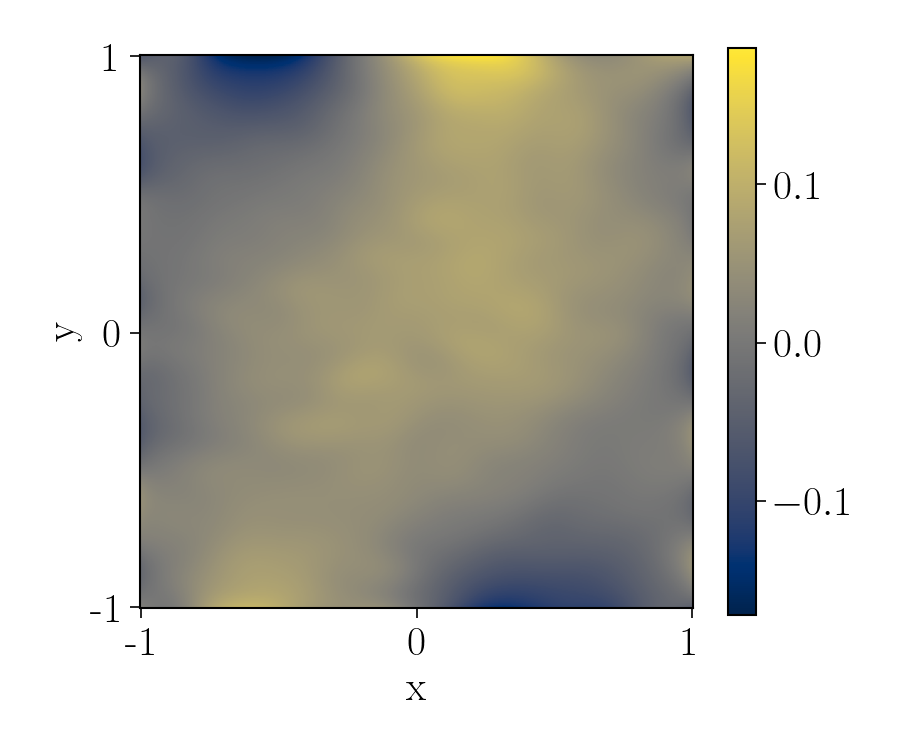} &
    \includegraphics[width=0.47\linewidth]{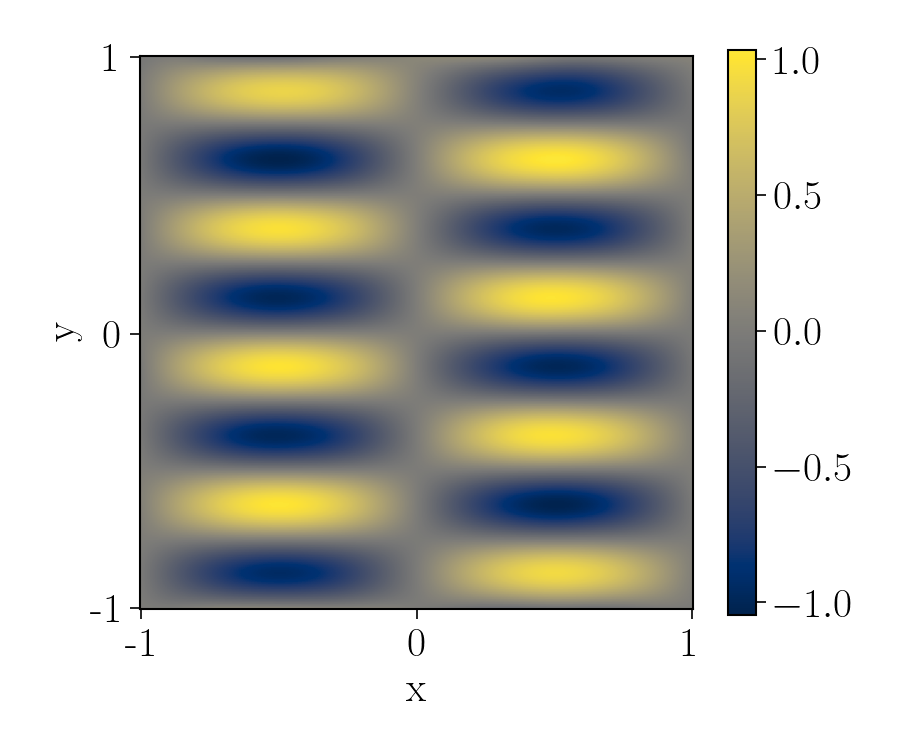}
    \includegraphics[width=0.47\linewidth]{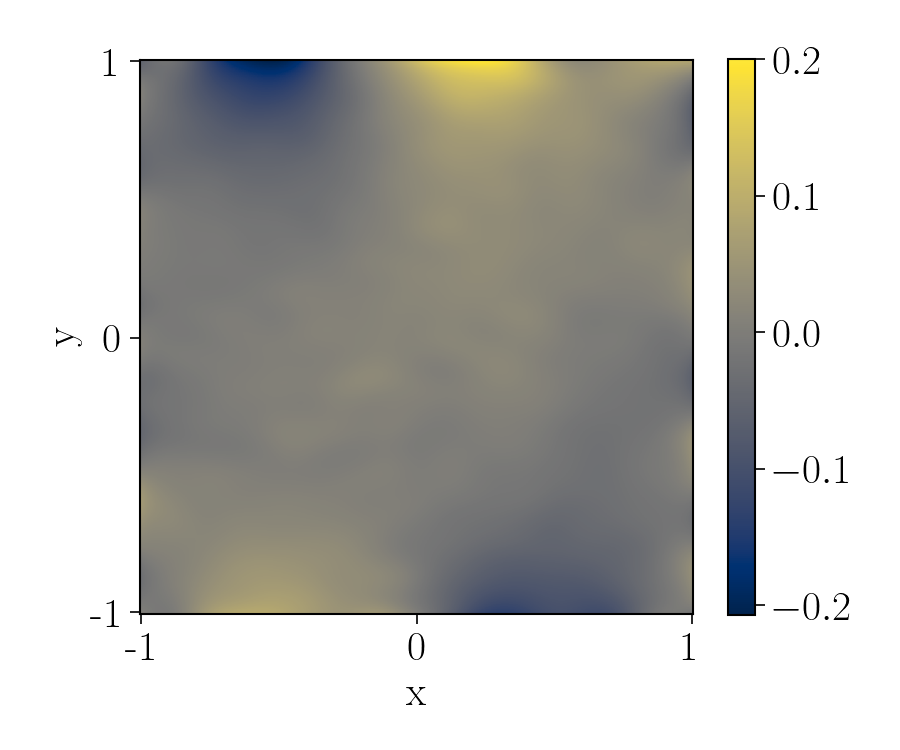} \\
    \hline
    \textit{HZ-G} &
    \includegraphics[width=0.94\linewidth]{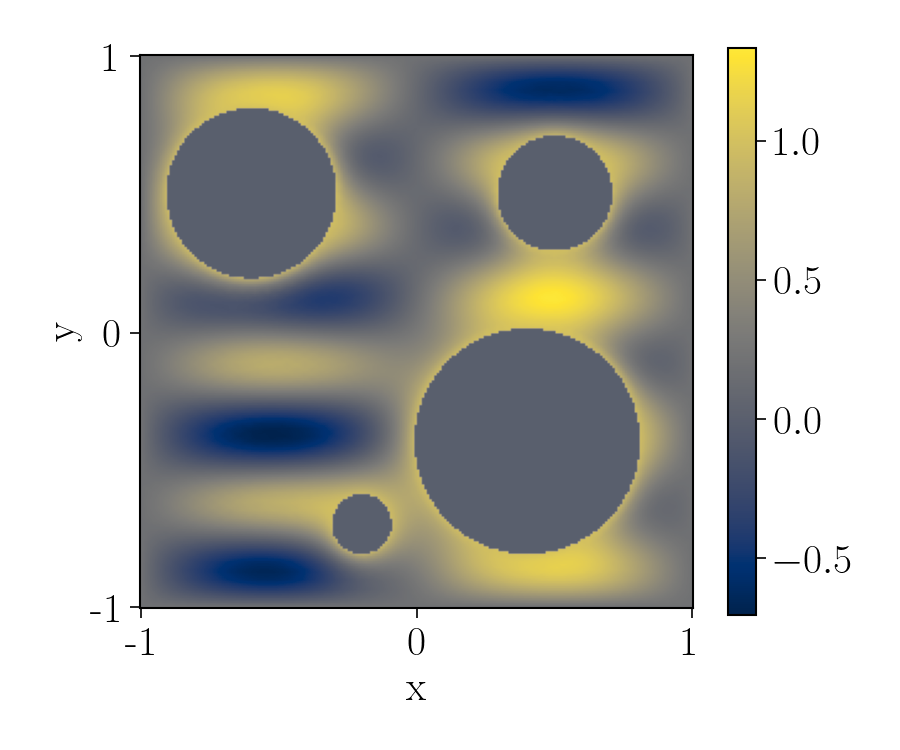} &
    \includegraphics[width=0.47\linewidth]{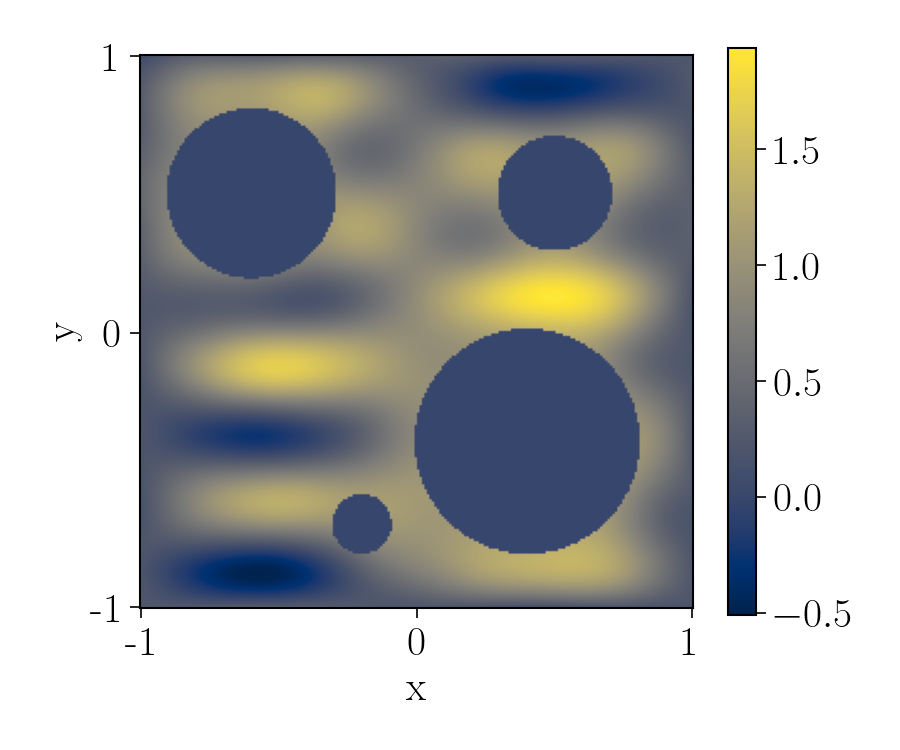}
    \includegraphics[width=0.47\linewidth]{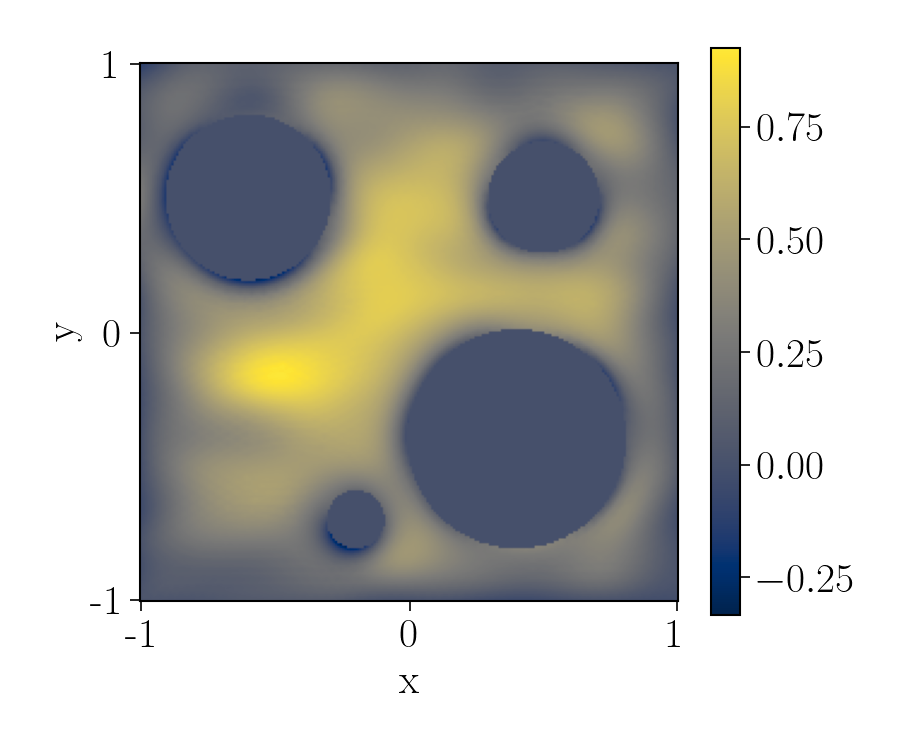} &
    \includegraphics[width=0.47\linewidth]{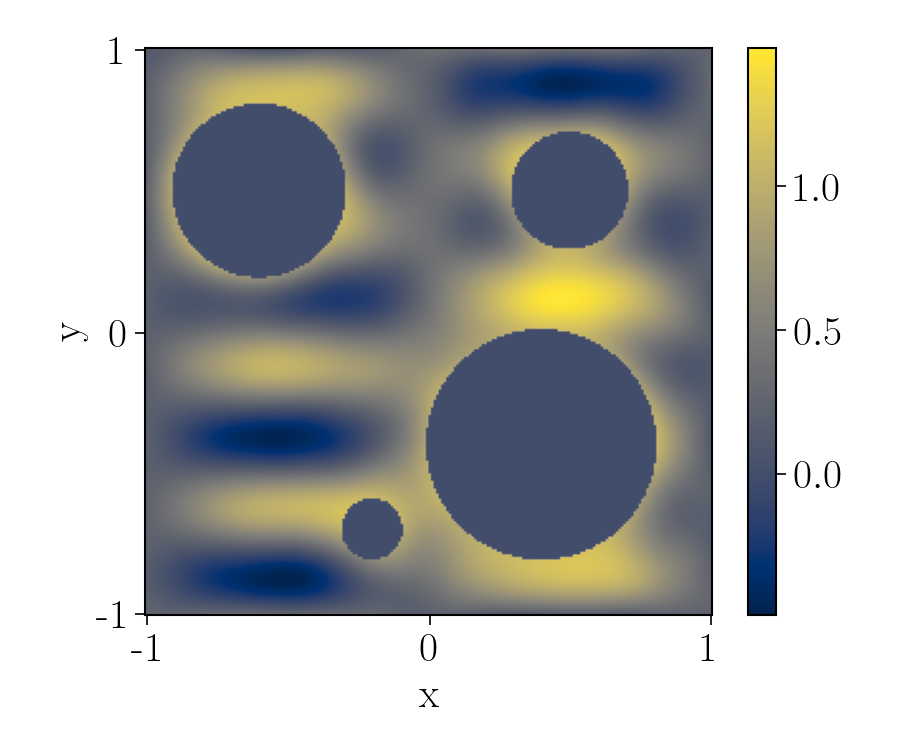}
    \includegraphics[width=0.47\linewidth]{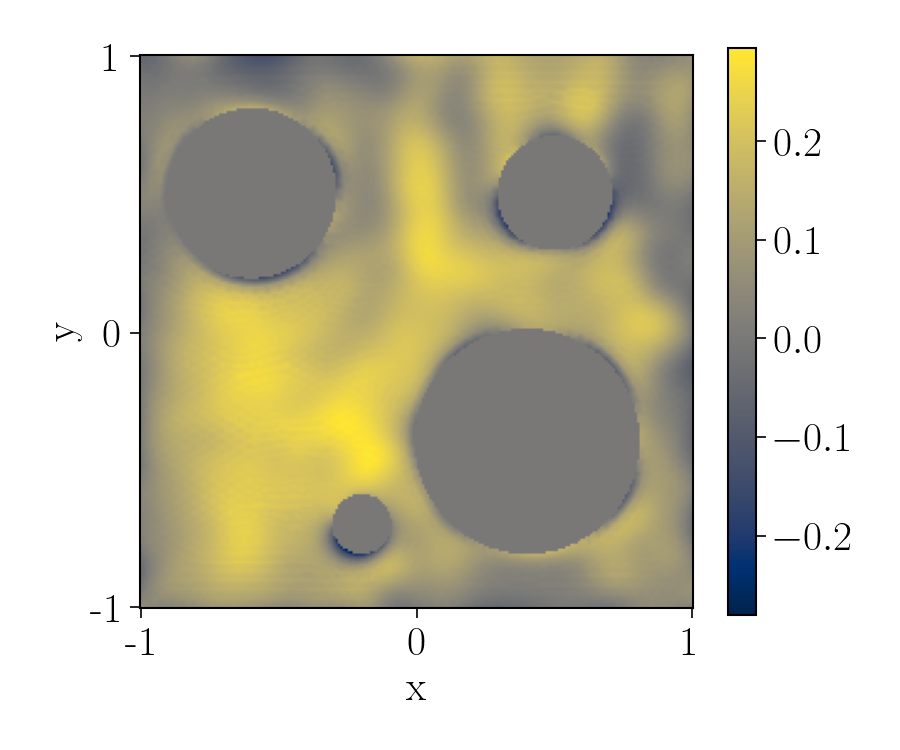} &
    \includegraphics[width=0.47\linewidth]{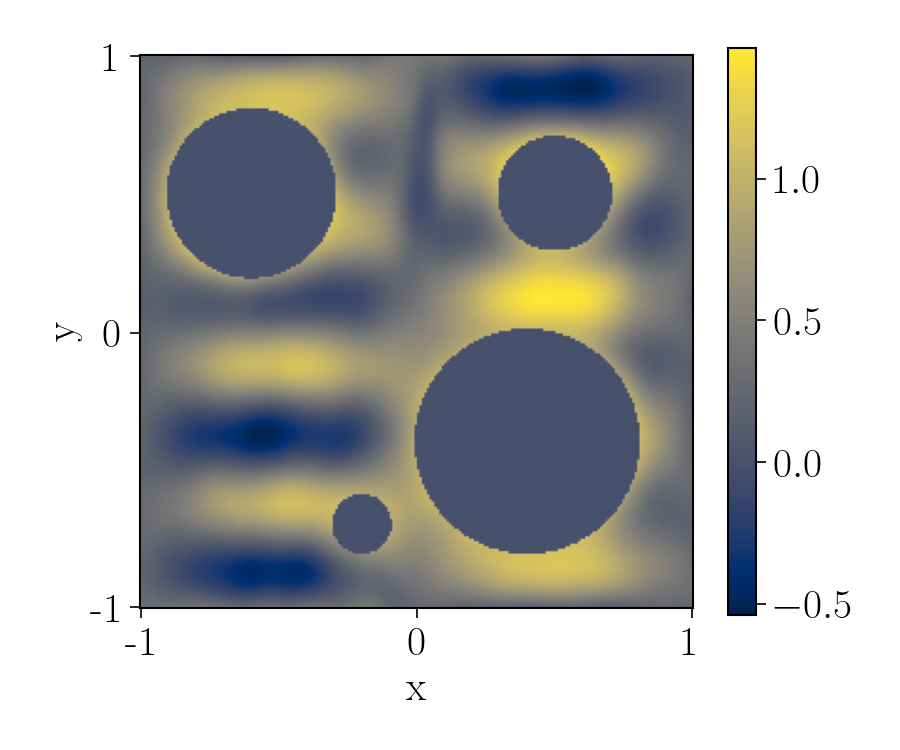}
    \includegraphics[width=0.47\linewidth]{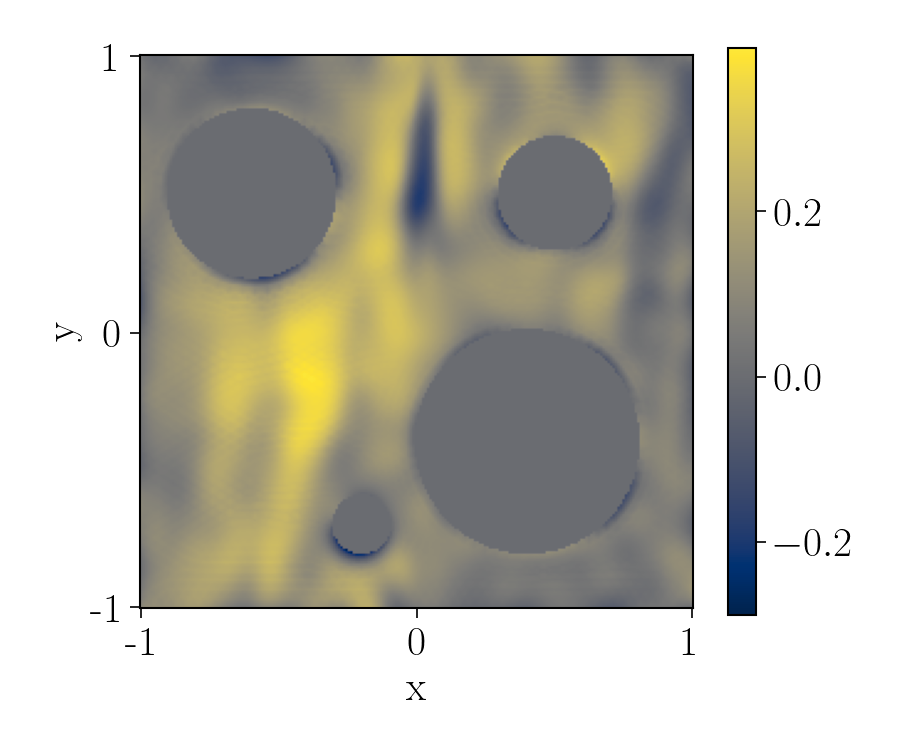} \\
    \hline
    \textit{PS-C} &
    \includegraphics[width=0.94\linewidth]{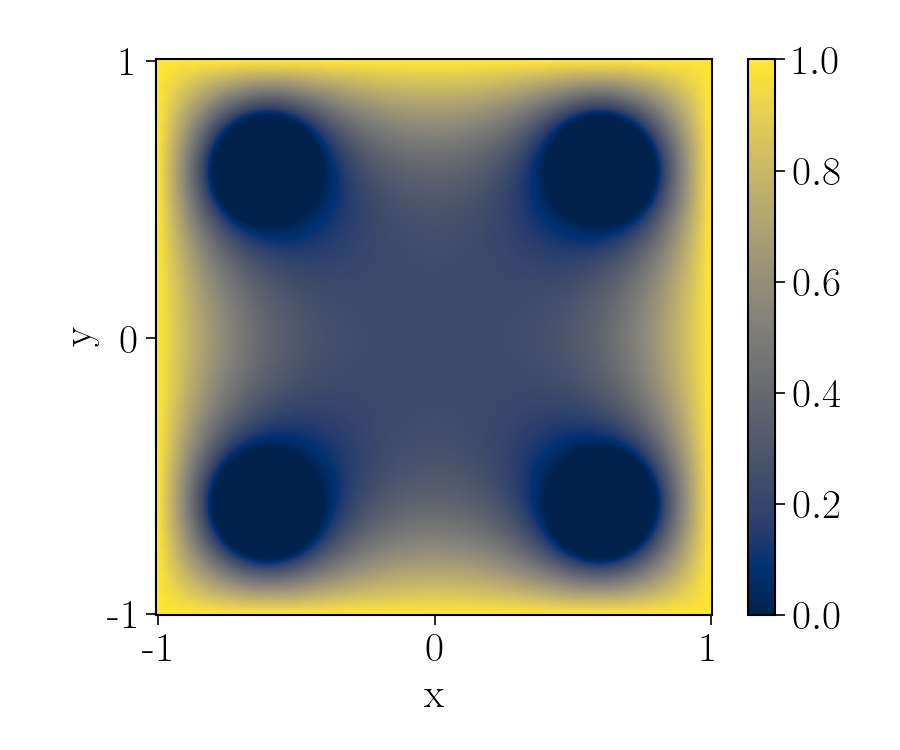} &
    \includegraphics[width=0.47\linewidth]{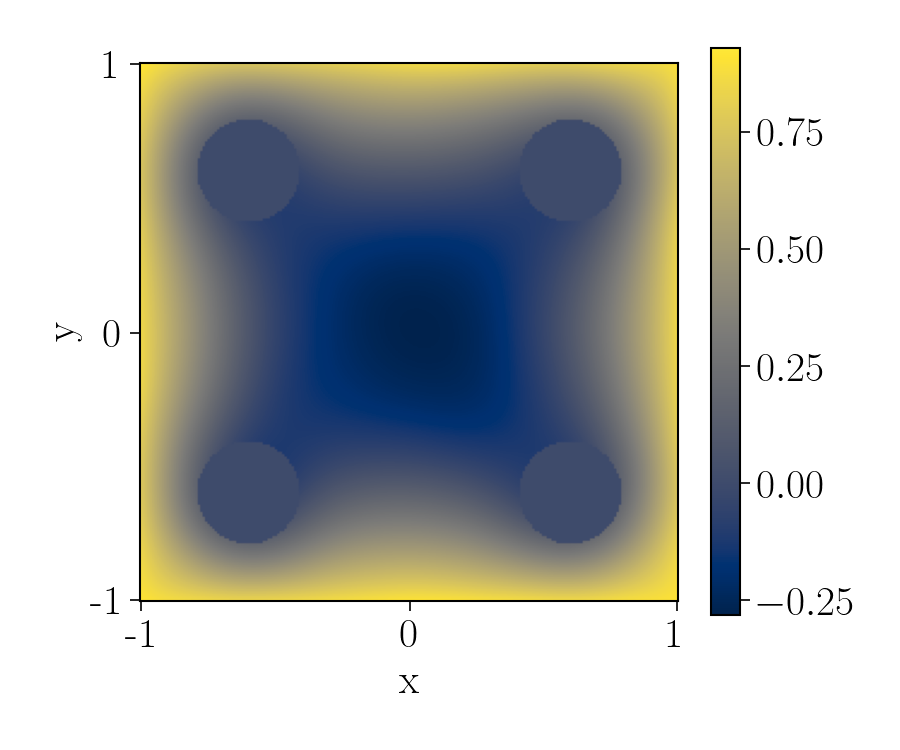}
    \includegraphics[width=0.47\linewidth]{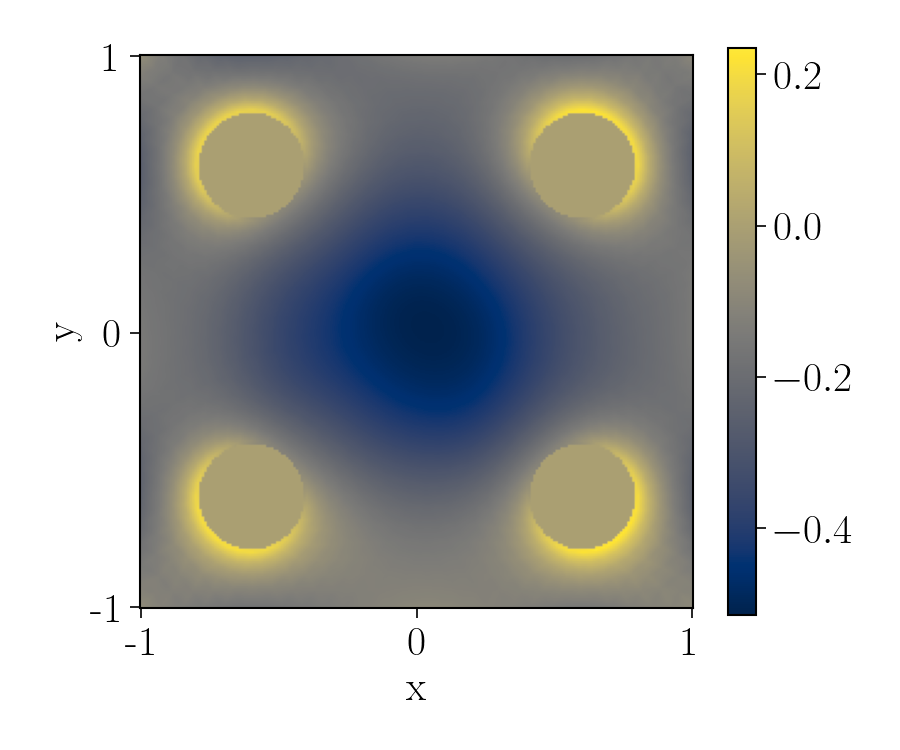} &
    \includegraphics[width=0.47\linewidth]{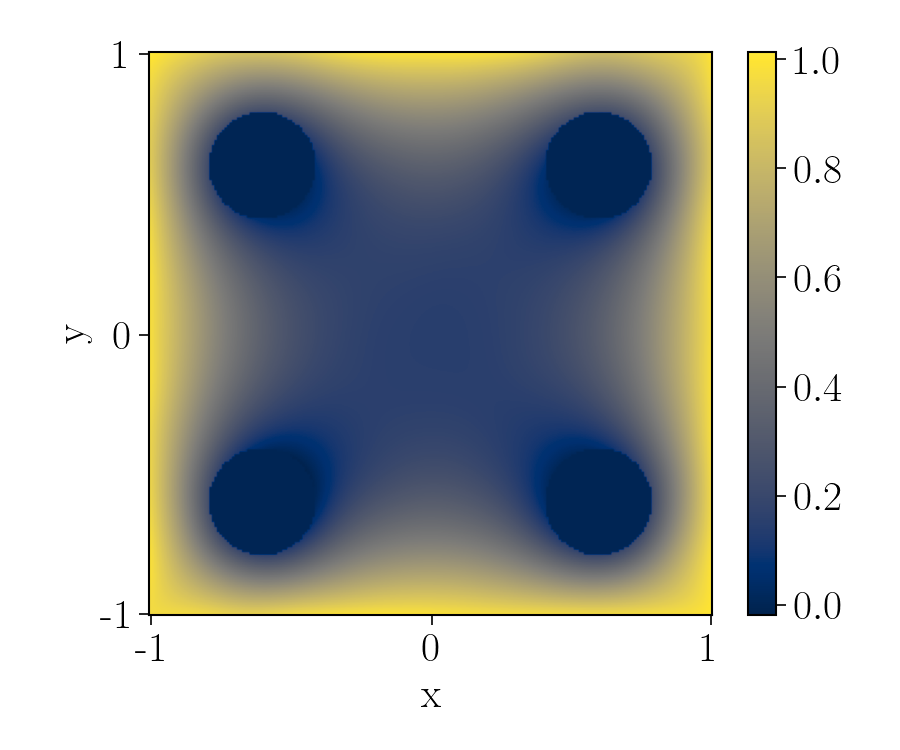}
    \includegraphics[width=0.47\linewidth]{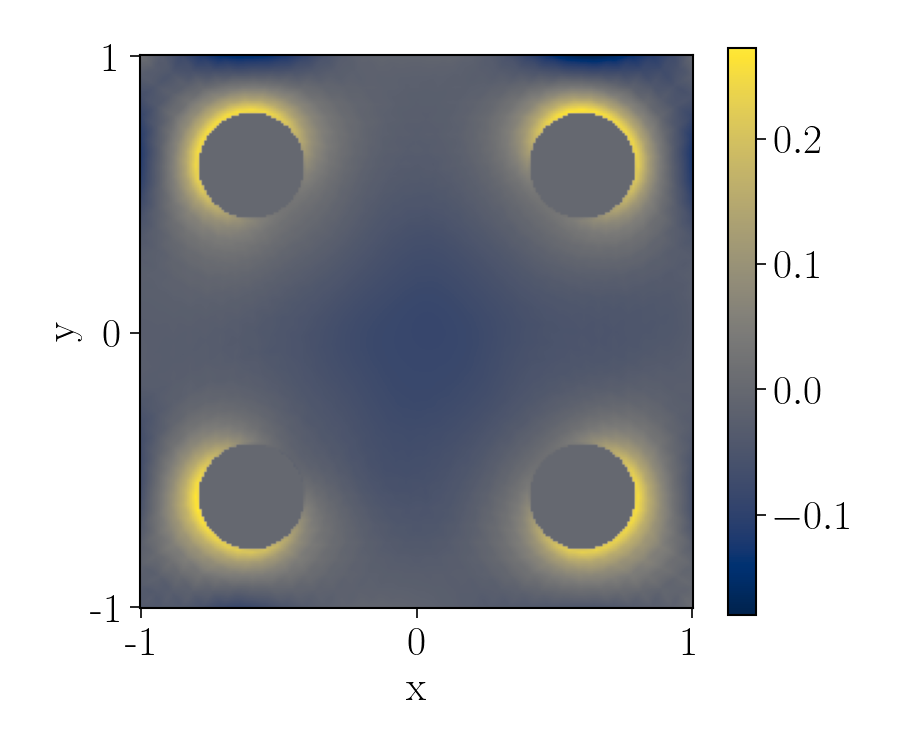} &
    \includegraphics[width=0.47\linewidth]{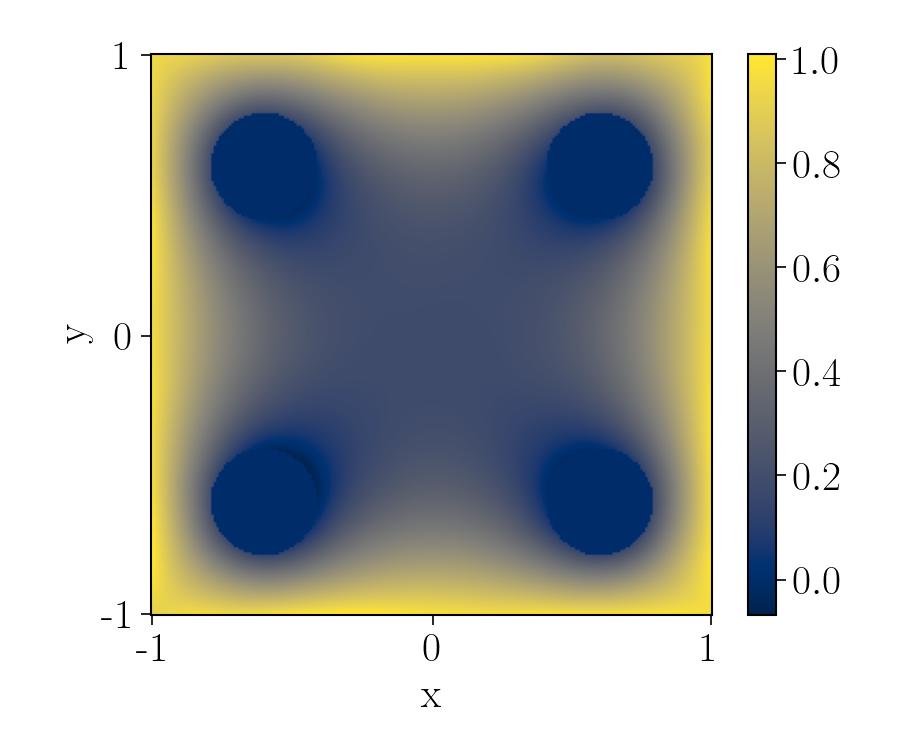}
    \includegraphics[width=0.47\linewidth]{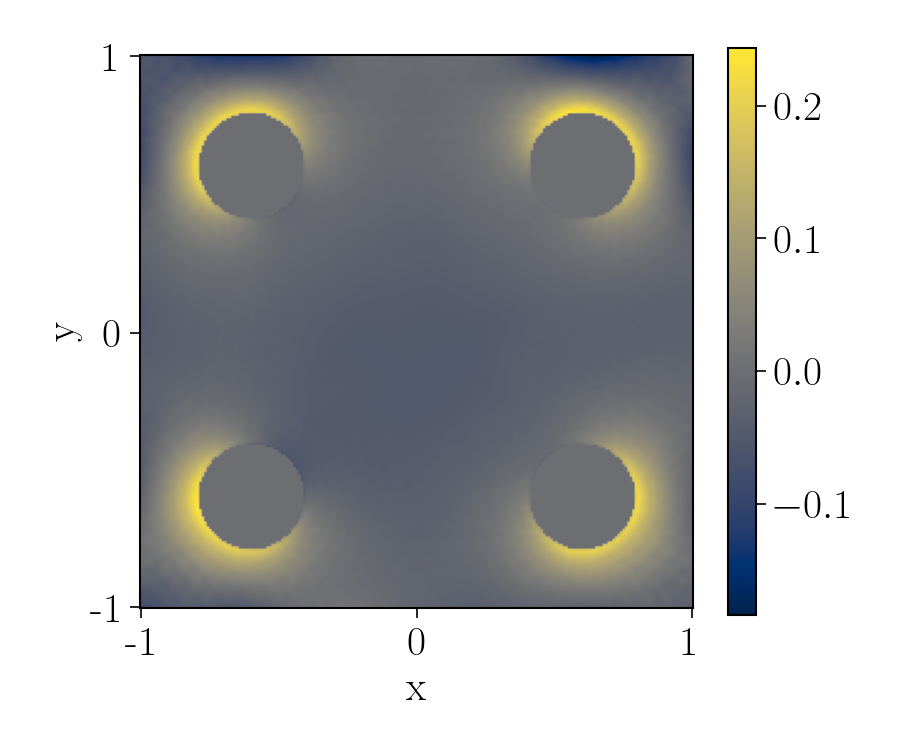} \\
    \hline
    \textit{PS-L} &
    \includegraphics[width=0.94\linewidth]{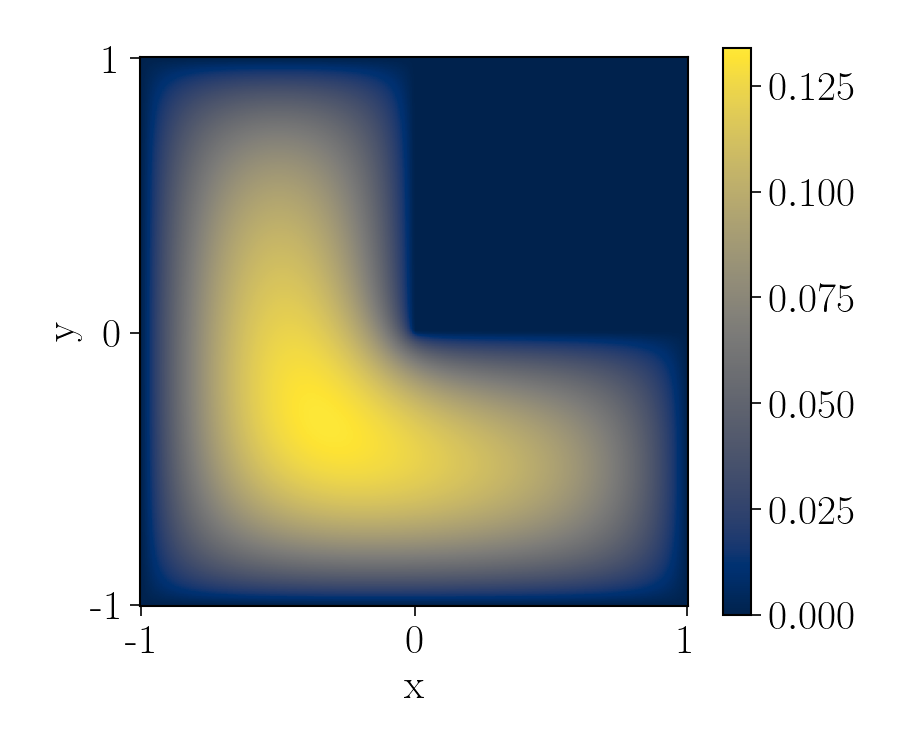} &
    \includegraphics[width=0.47\linewidth]{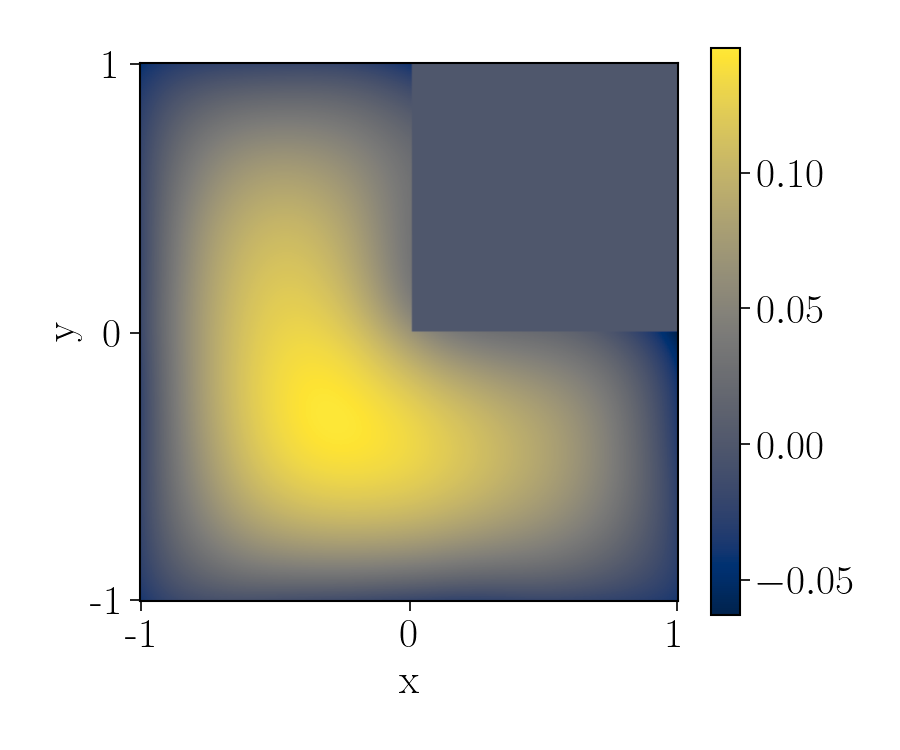}
    \includegraphics[width=0.47\linewidth]{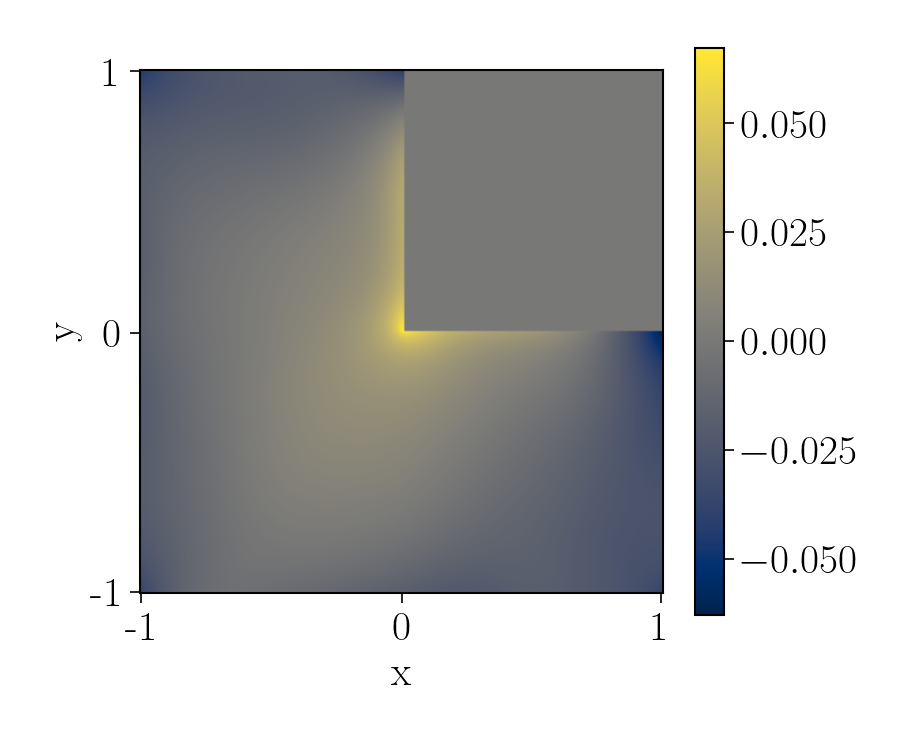} &
    \includegraphics[width=0.47\linewidth]{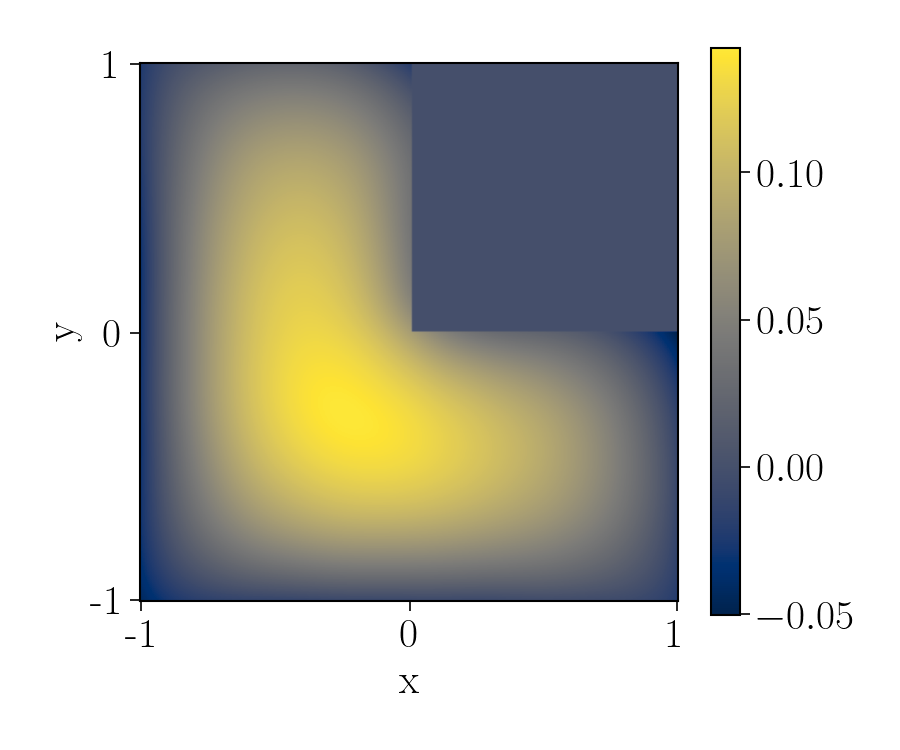}
    \includegraphics[width=0.47\linewidth]{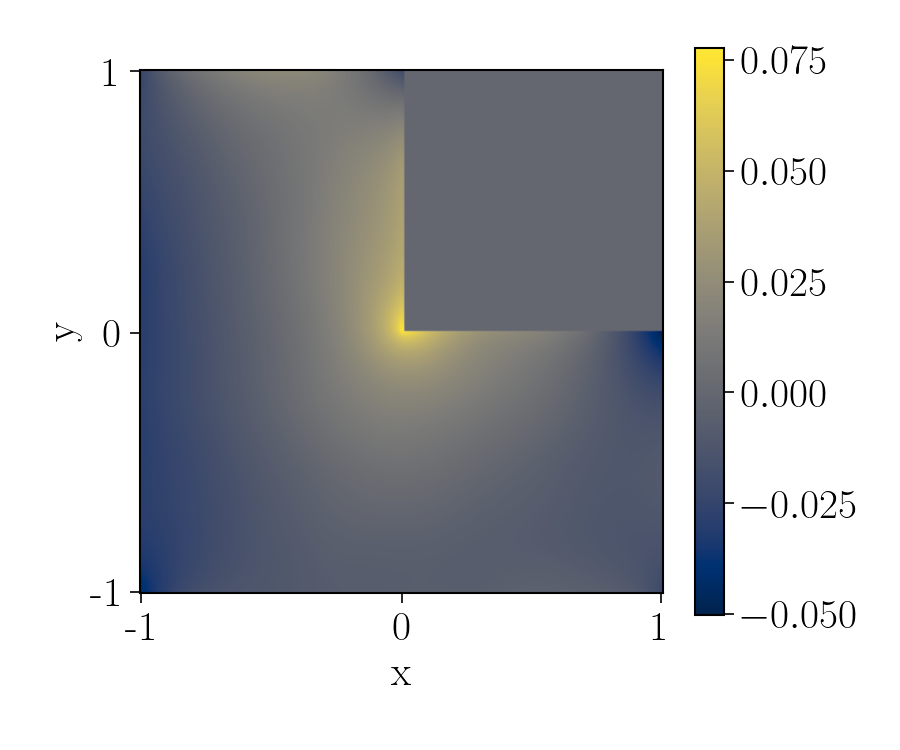} &
    \includegraphics[width=0.47\linewidth]{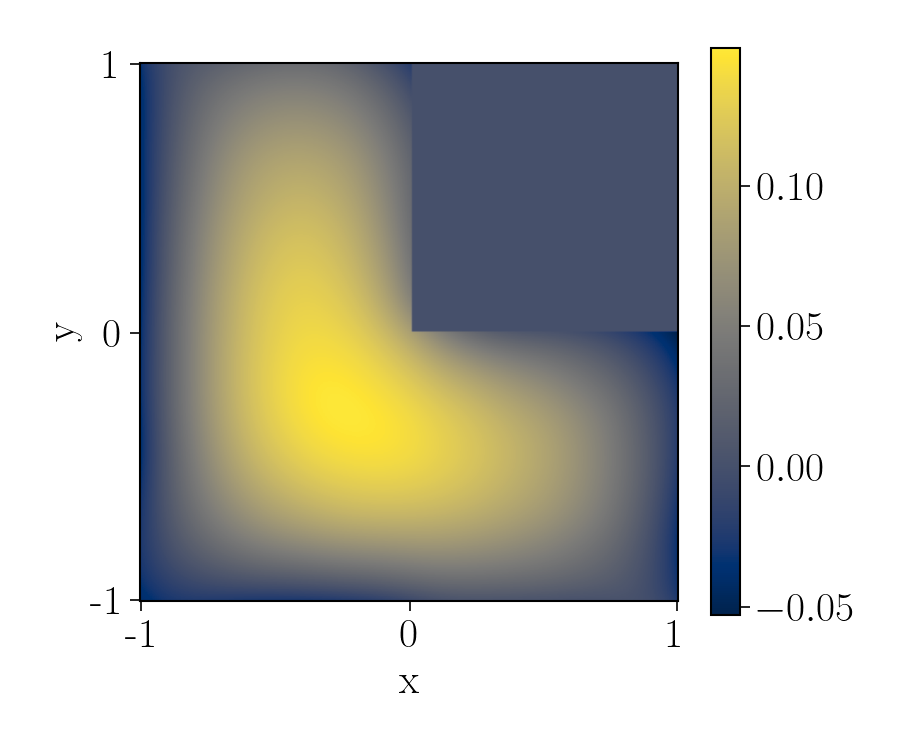}
    \includegraphics[width=0.47\linewidth]{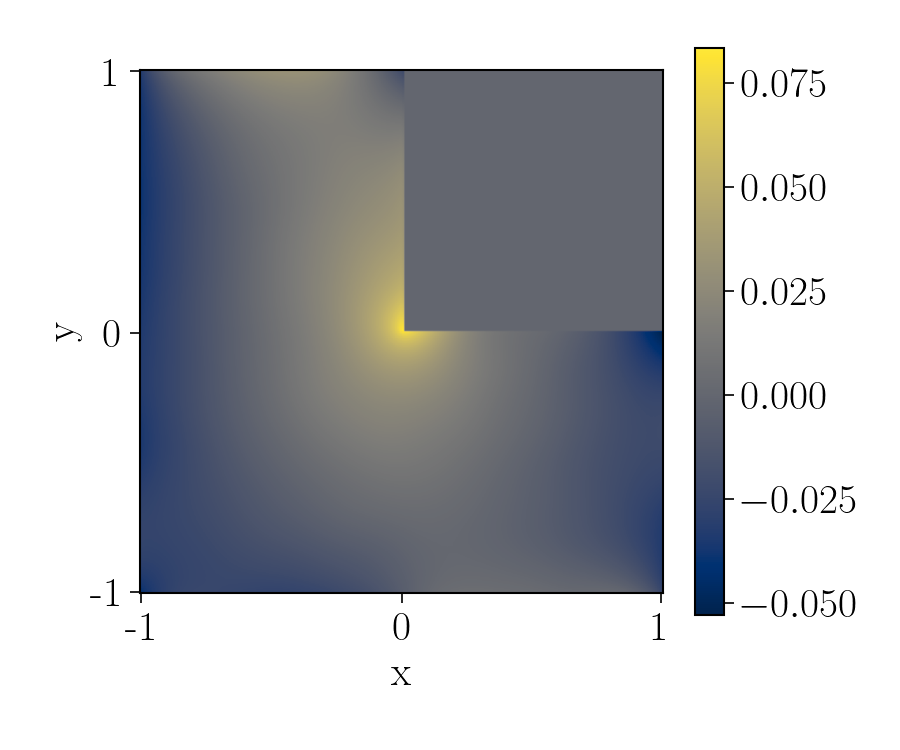} \\
    \hline
    \textit{PS-G} &
    \includegraphics[width=0.94\linewidth]{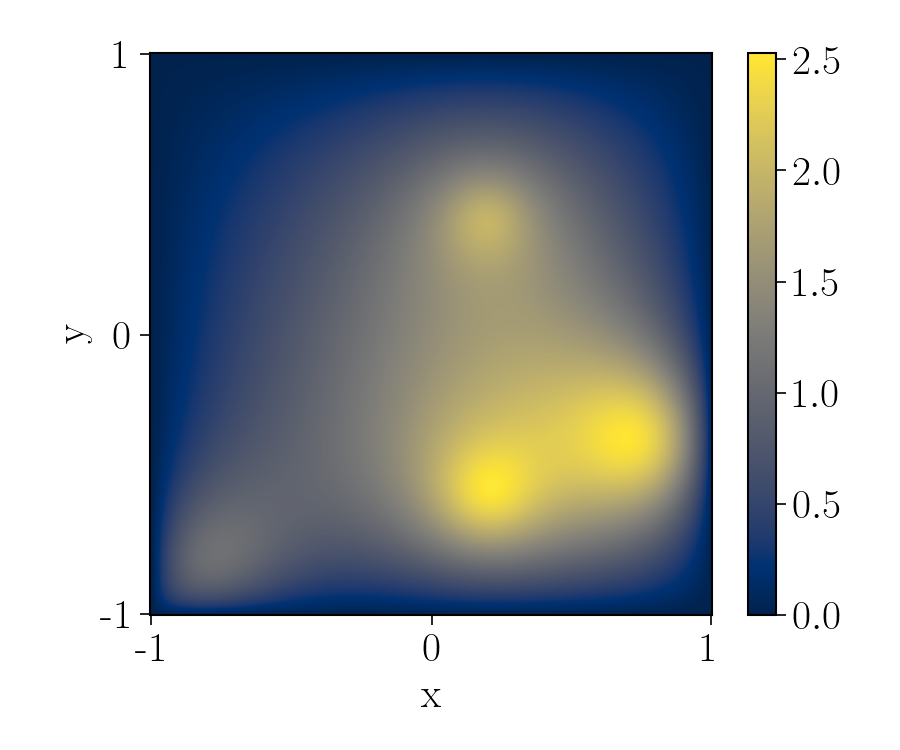} &
    \includegraphics[width=0.47\linewidth]{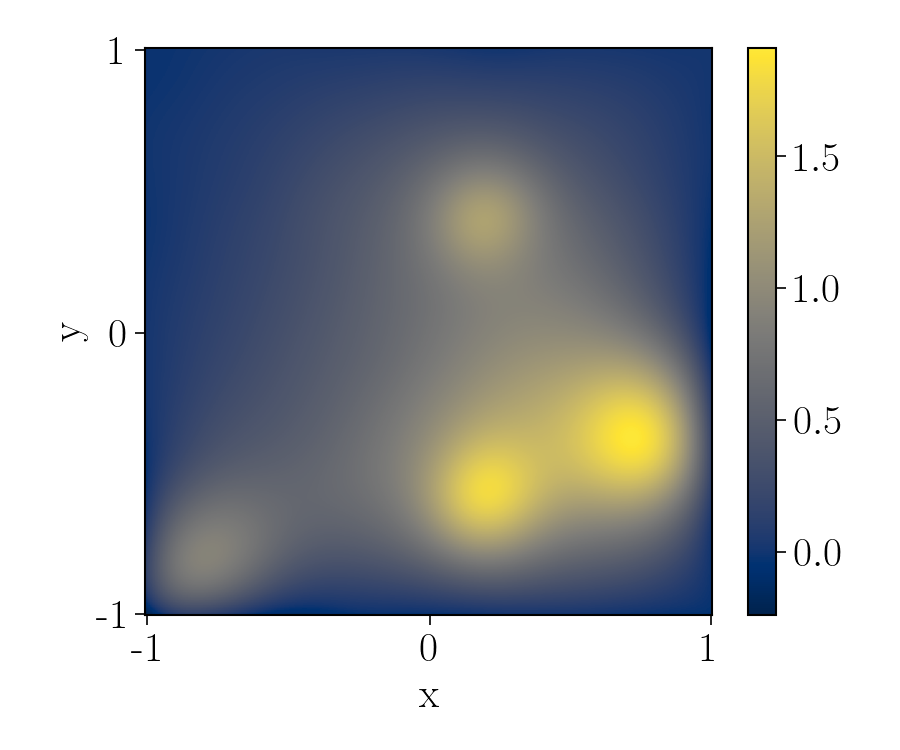}
    \includegraphics[width=0.47\linewidth]{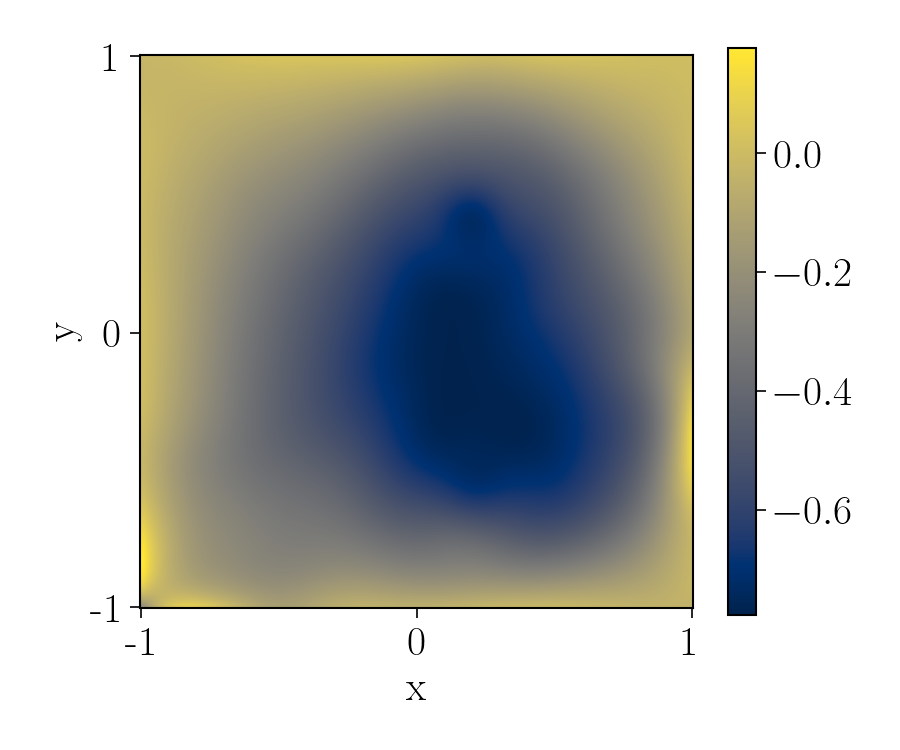} &
    \includegraphics[width=0.47\linewidth]{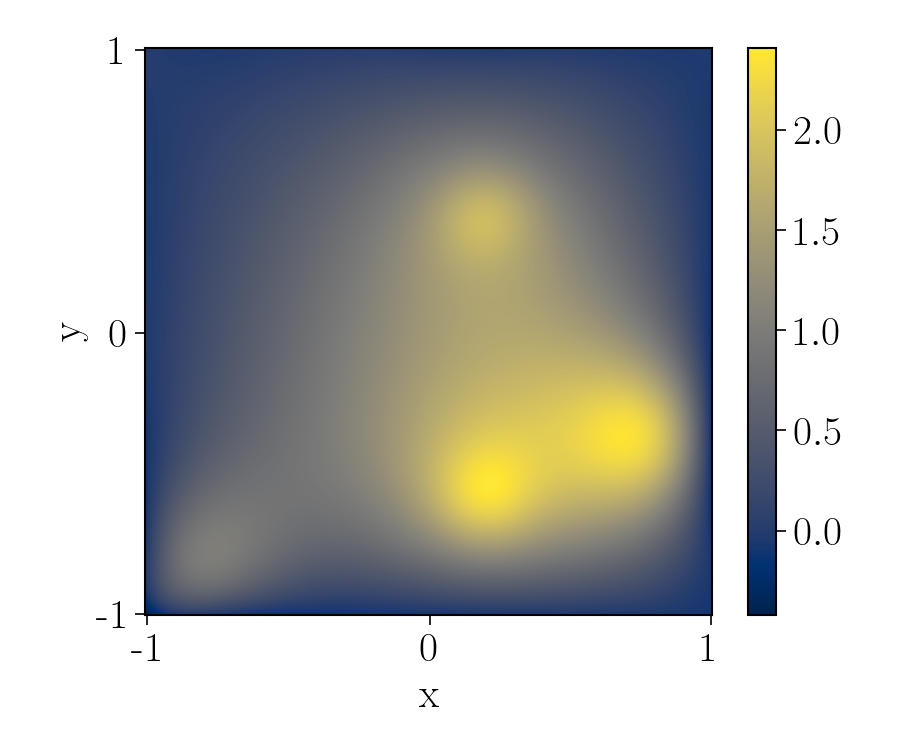}
    \includegraphics[width=0.47\linewidth]{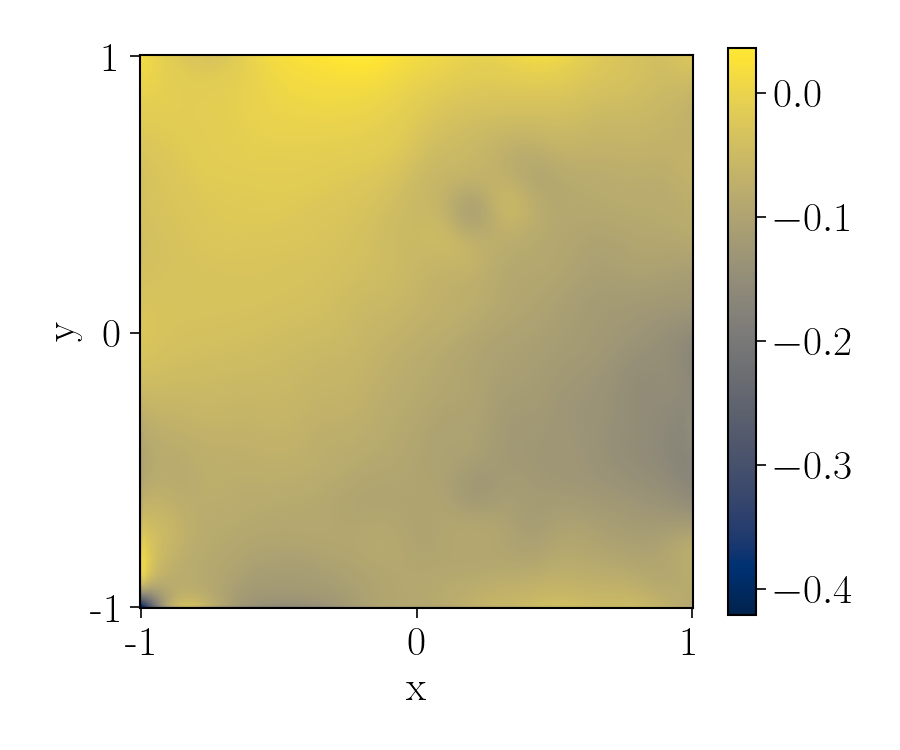} &
    \includegraphics[width=0.47\linewidth]{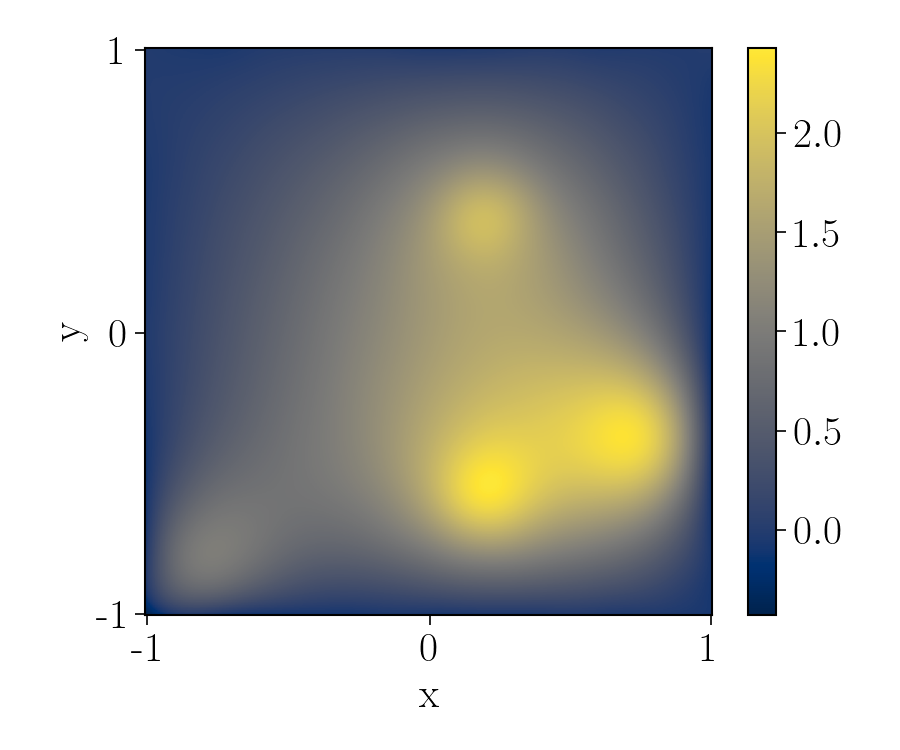}
    \includegraphics[width=0.47\linewidth]{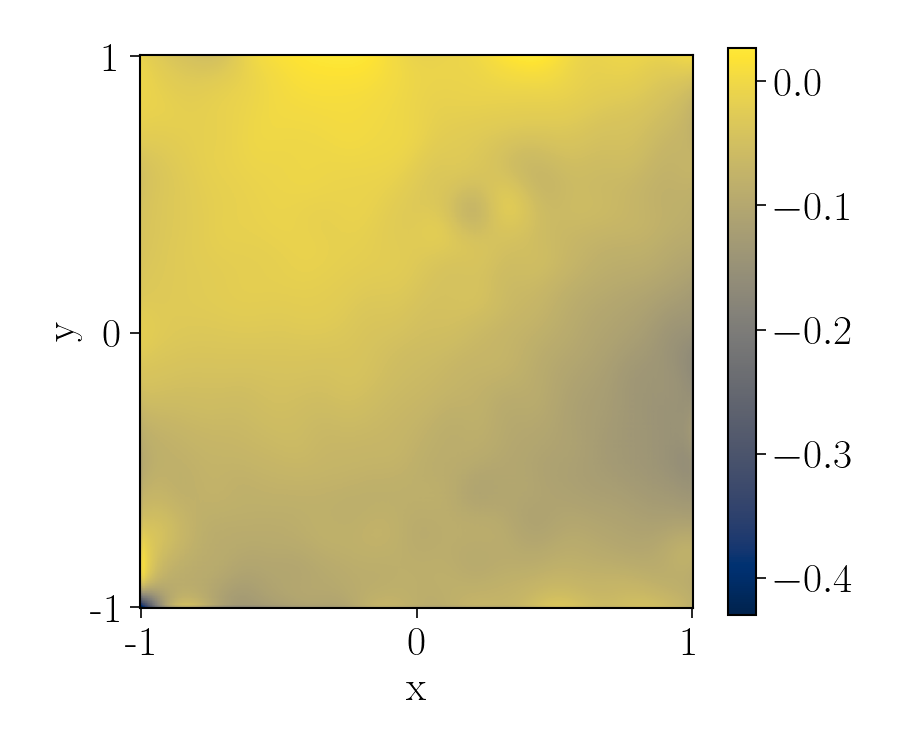} \\
    \hline
    \textit{WV} &
    \includegraphics[width=0.94\linewidth]{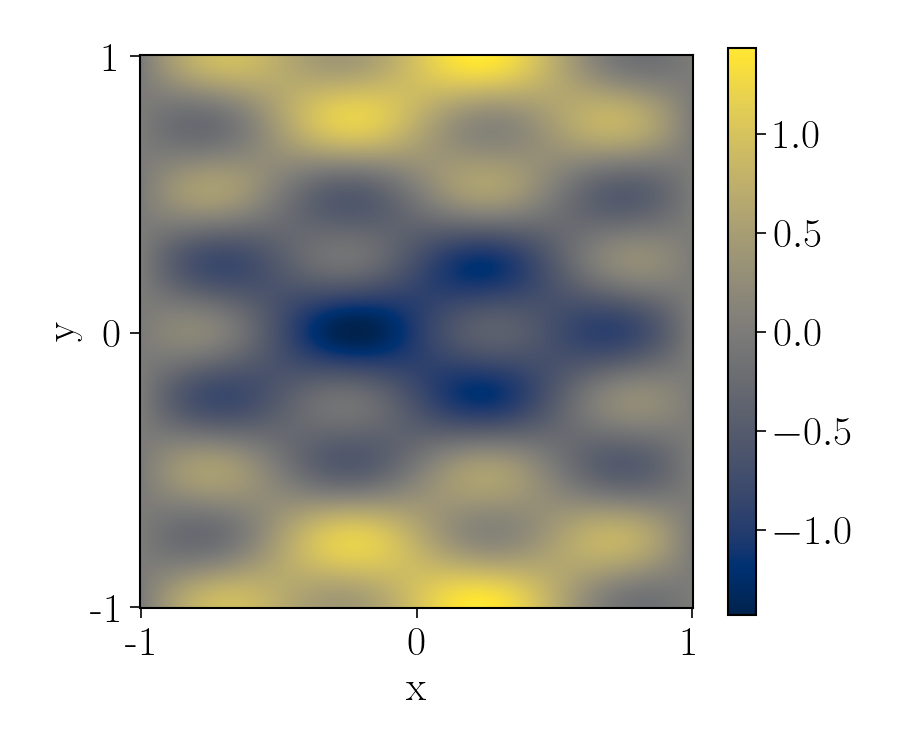} &
    \includegraphics[width=0.47\linewidth]{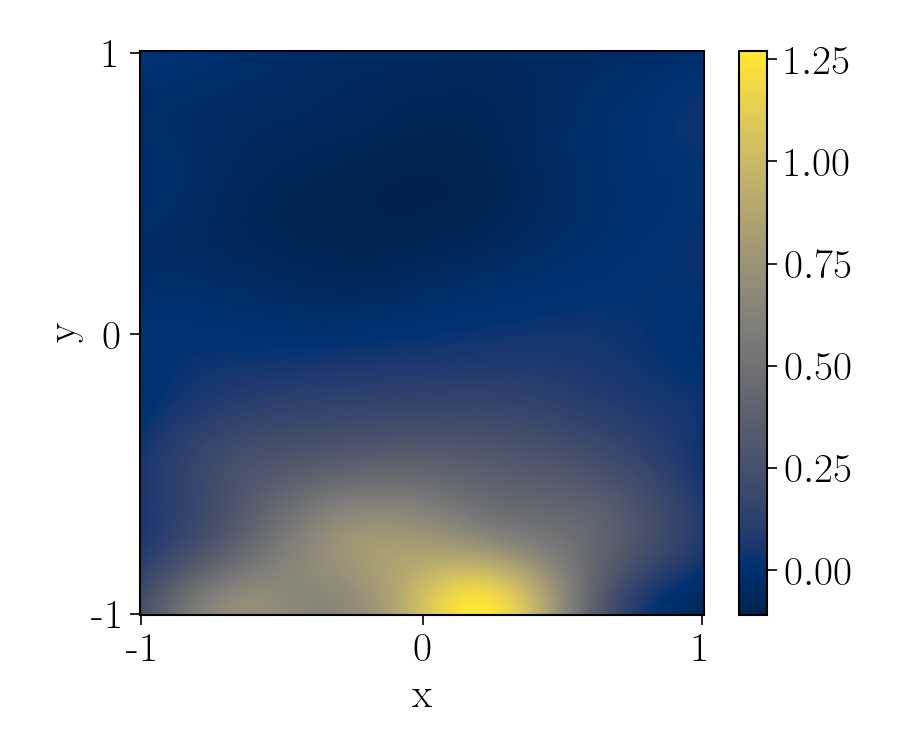}
    \includegraphics[width=0.47\linewidth]{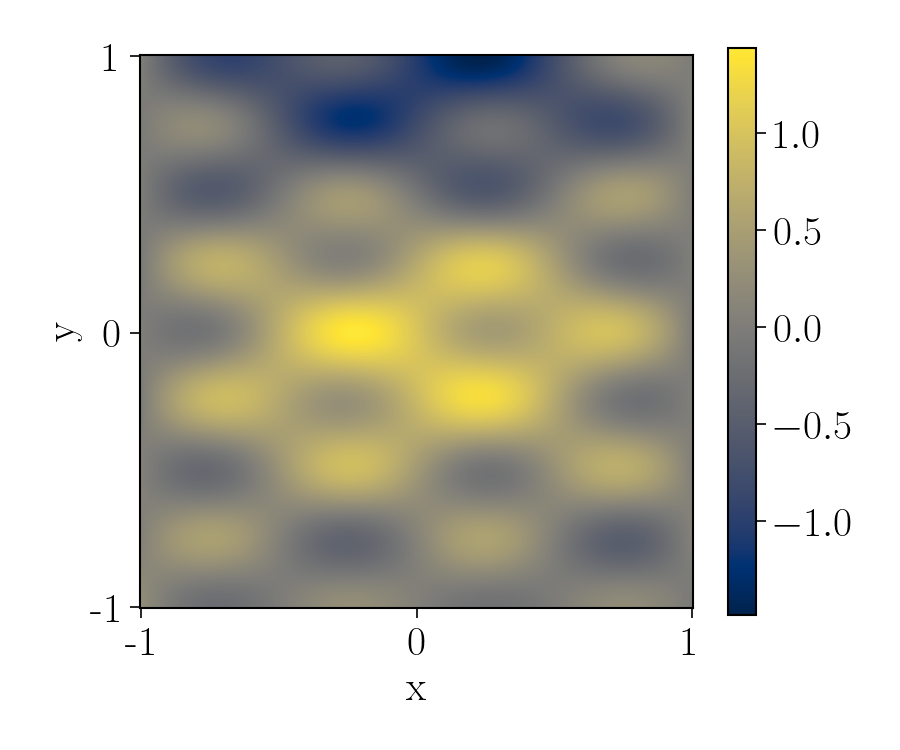} &
    \includegraphics[width=0.47\linewidth]{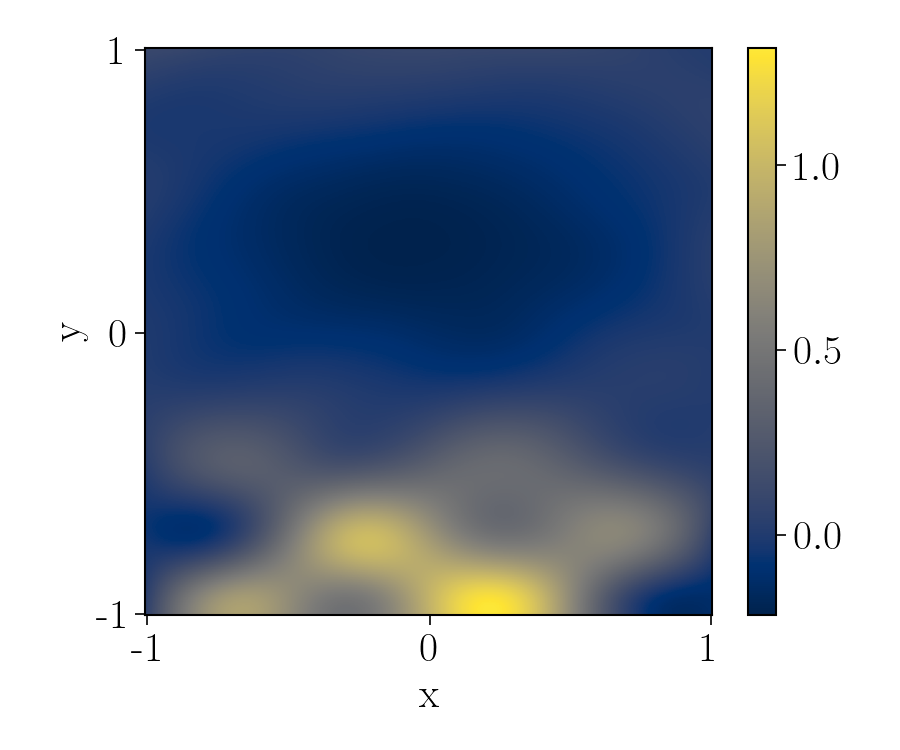}
    \includegraphics[width=0.47\linewidth]{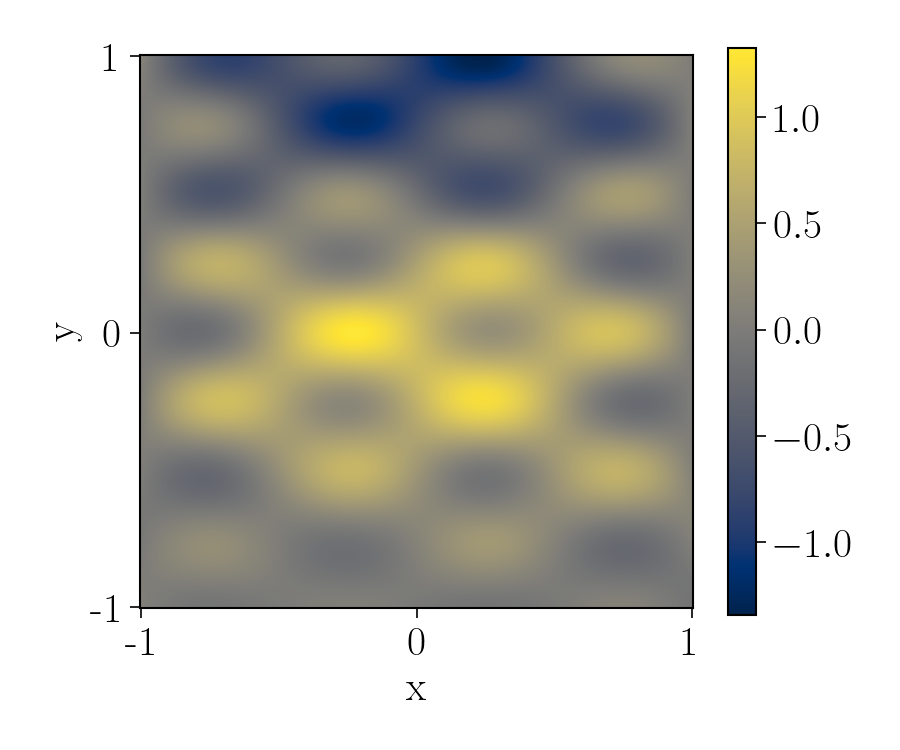} &
    \includegraphics[width=0.47\linewidth]{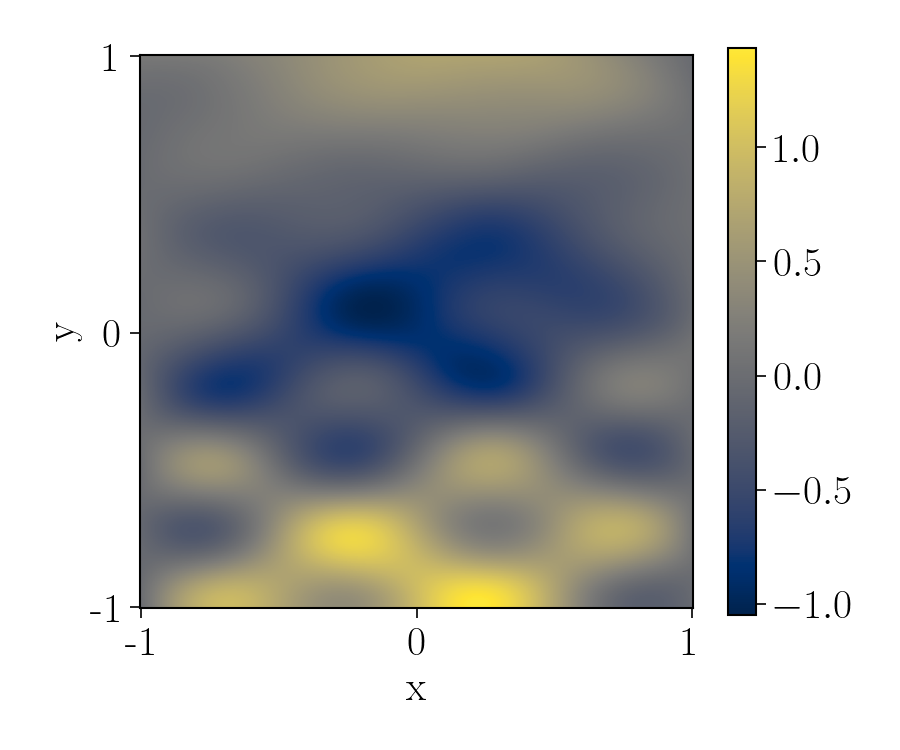}
    \includegraphics[width=0.47\linewidth]{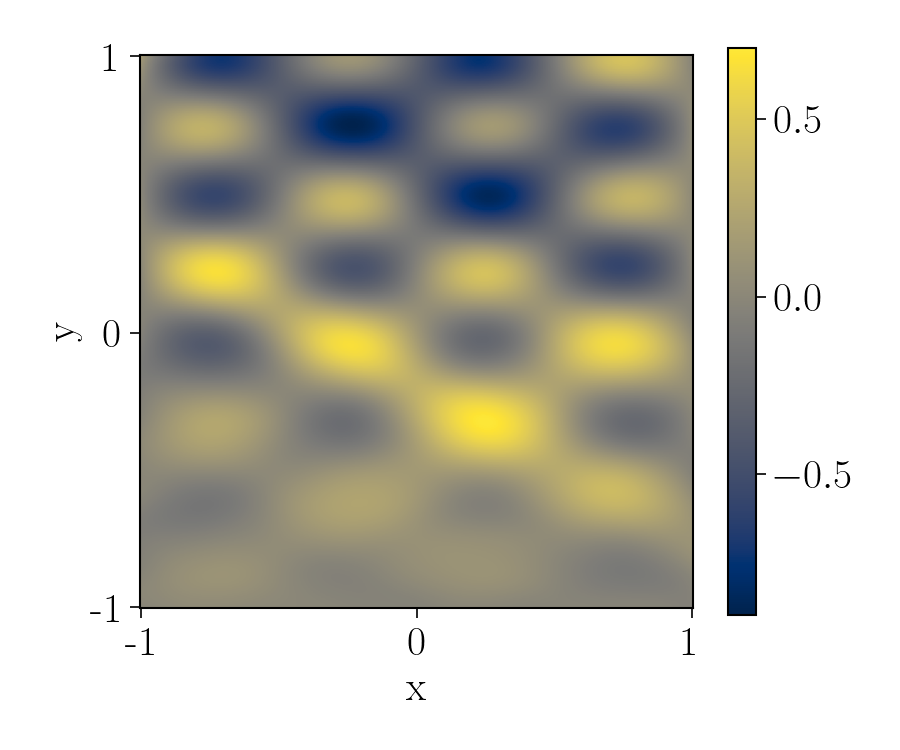} \\

    & \centering \scriptsize\textit{Reference} 
    & \scriptsize\makebox[0.12\textwidth][c]{\textit{Prediction}}%
      \makebox[0.13\textwidth][c]{\textit{Difference}} 
    & \scriptsize\makebox[0.12\textwidth][c]{\textit{Prediction}}%
      \makebox[0.13\textwidth][c]{\textit{Difference}} 
    & \scriptsize\makebox[0.12\textwidth][c]{\textit{Prediction}}%
      \makebox[0.13\textwidth][c]{\textit{Difference}} \\
    
    & & \multicolumn{1}{c|}{\textit{HyPINO}} 
      & \multicolumn{1}{c|}{\textit{HyPINO${^3}$}} 
      & \multicolumn{1}{c}{\textit{HyPINO${^{10}}$}} \\
    \end{tabular}
\end{table}

\begin{figure}[t]
    \centering
    \begin{subfigure}[b]{0.49\linewidth}
        \includegraphics[width=\linewidth]{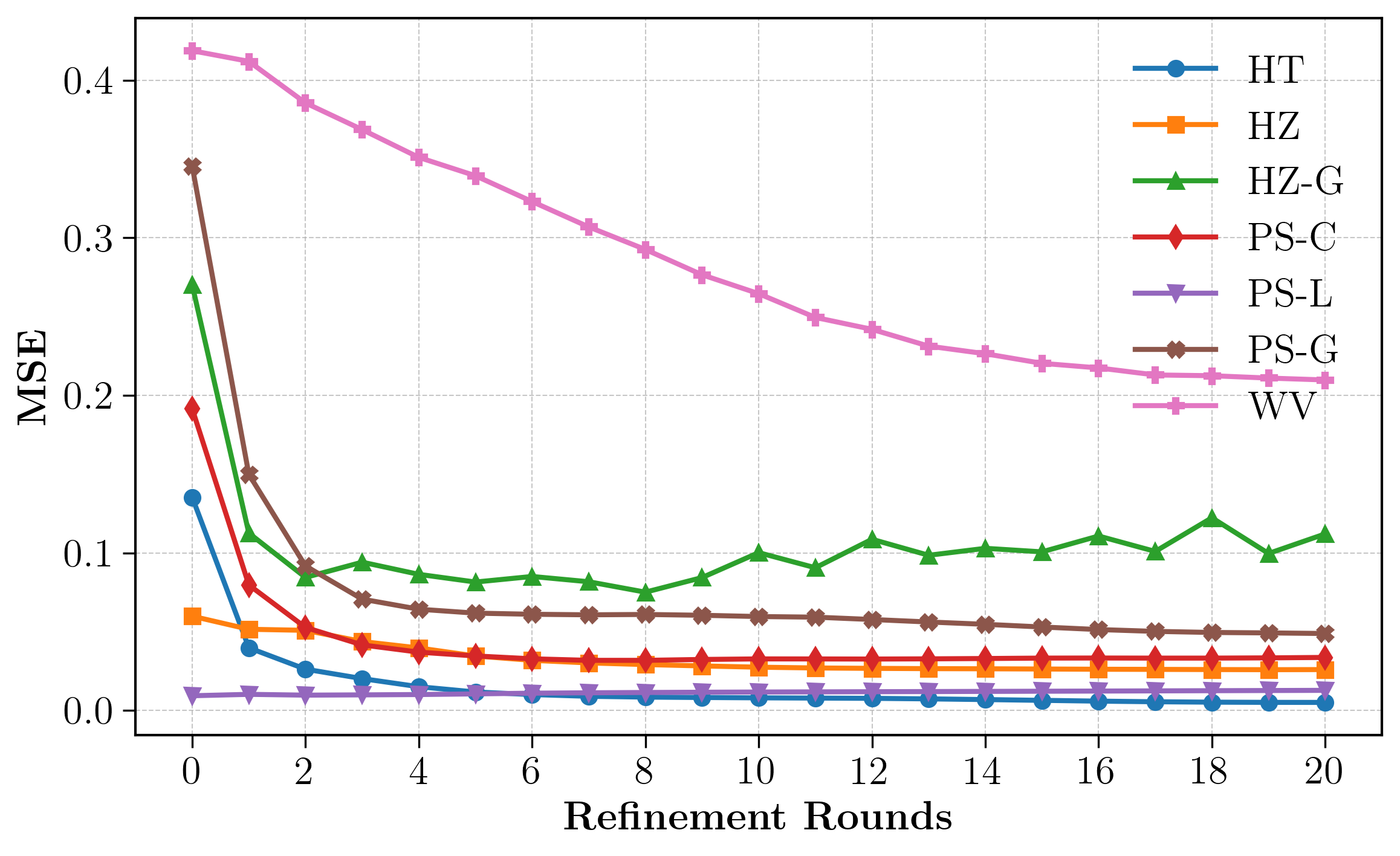}
        \label{fig:iterative_refinement_mse}
    \end{subfigure}
    \hfill
    \begin{subfigure}[b]{0.49\linewidth}
        \includegraphics[width=\linewidth]{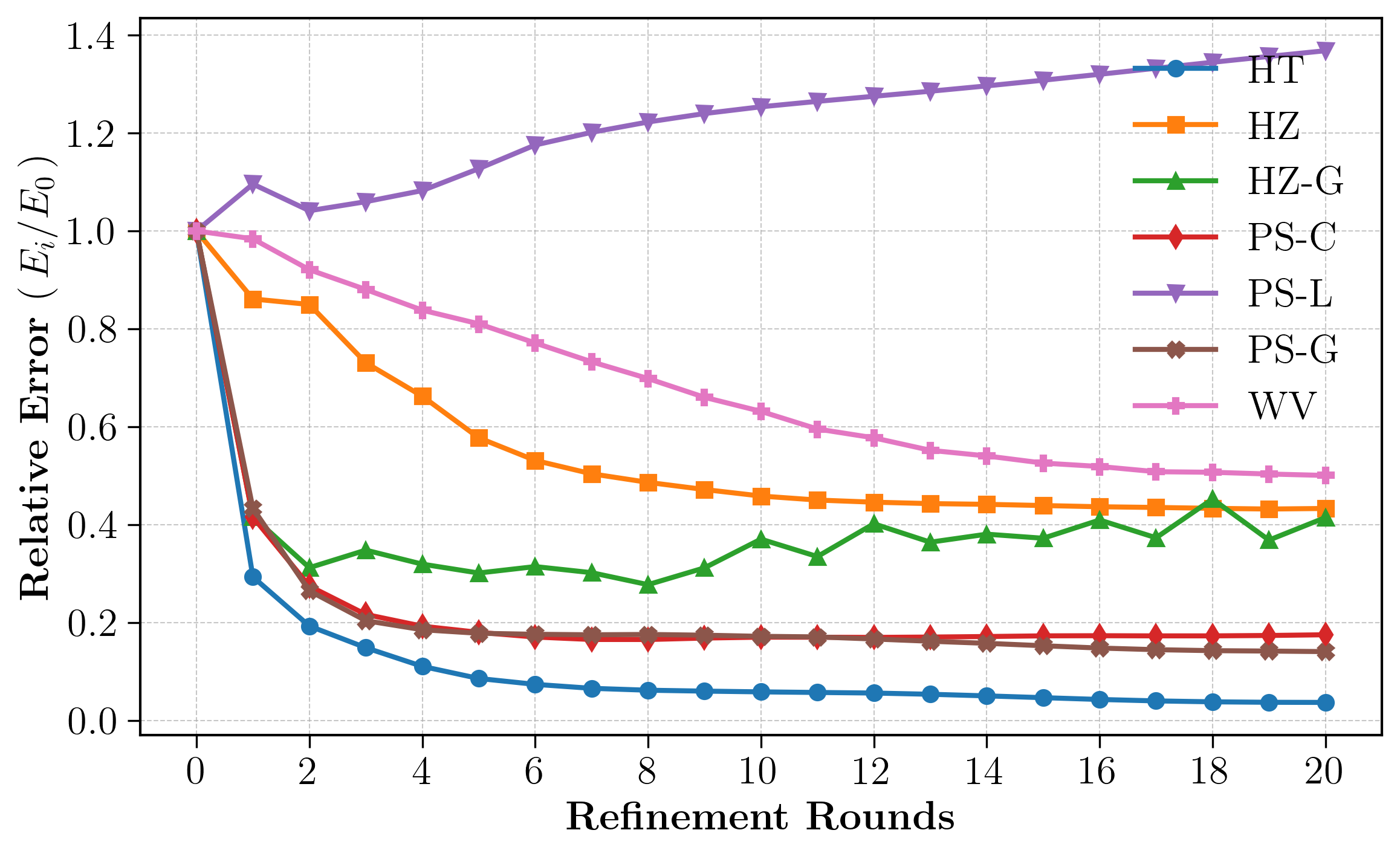}
        \label{fig:iterative_refinement_relative}
    \end{subfigure}
    
    \caption{Effect of iterative refinement on HyPINO predictions across benchmarks. MSE (left) and relative error (right) as functions of refinement iterations. Relative error at iteration~$i$ is the ratio of MSE at iteration~$i$ to that at iteration~0.}
    \label{fig:iterative_refinement_rounds}
\end{figure}

Figure~\ref{fig:iterative_refinement_rounds} illustrates these trends, showing mean squared error and relative error as functions of the number of refinement iterations. Consistent improvements are observed with additional iterations. The performance degradation on PS-L may be attributed to the already low initial error in the zeroth iteration and the small magnitudes of the solution values, resulting in correction terms that fall outside the distribution encountered during training.

\subsection{Fine-tuning}
\label{sec:finetuning}

The parameters $\theta^\star$ produced by HyPINO can be used to initialize PINNs for subsequent fine-tuning on specific PDE instances. We compare the convergence behavior of HyPINO-initialized PINNs with those initialized randomly and with Reptile meta-learning~\cite{reptile}, where Reptile was trained on our synthetic dataset using 10{,}000 outer- and 1,000 inner-loop cycles. We also evaluate ensembles generated with HyPINO$^3$ and HyPINO$^{10}$ against ensembles of equal size and architecture initialized with random weights or Reptile.

\begin{wrapfigure}{r}{0.49\textwidth}
    \centering
    \includegraphics[width=\linewidth]{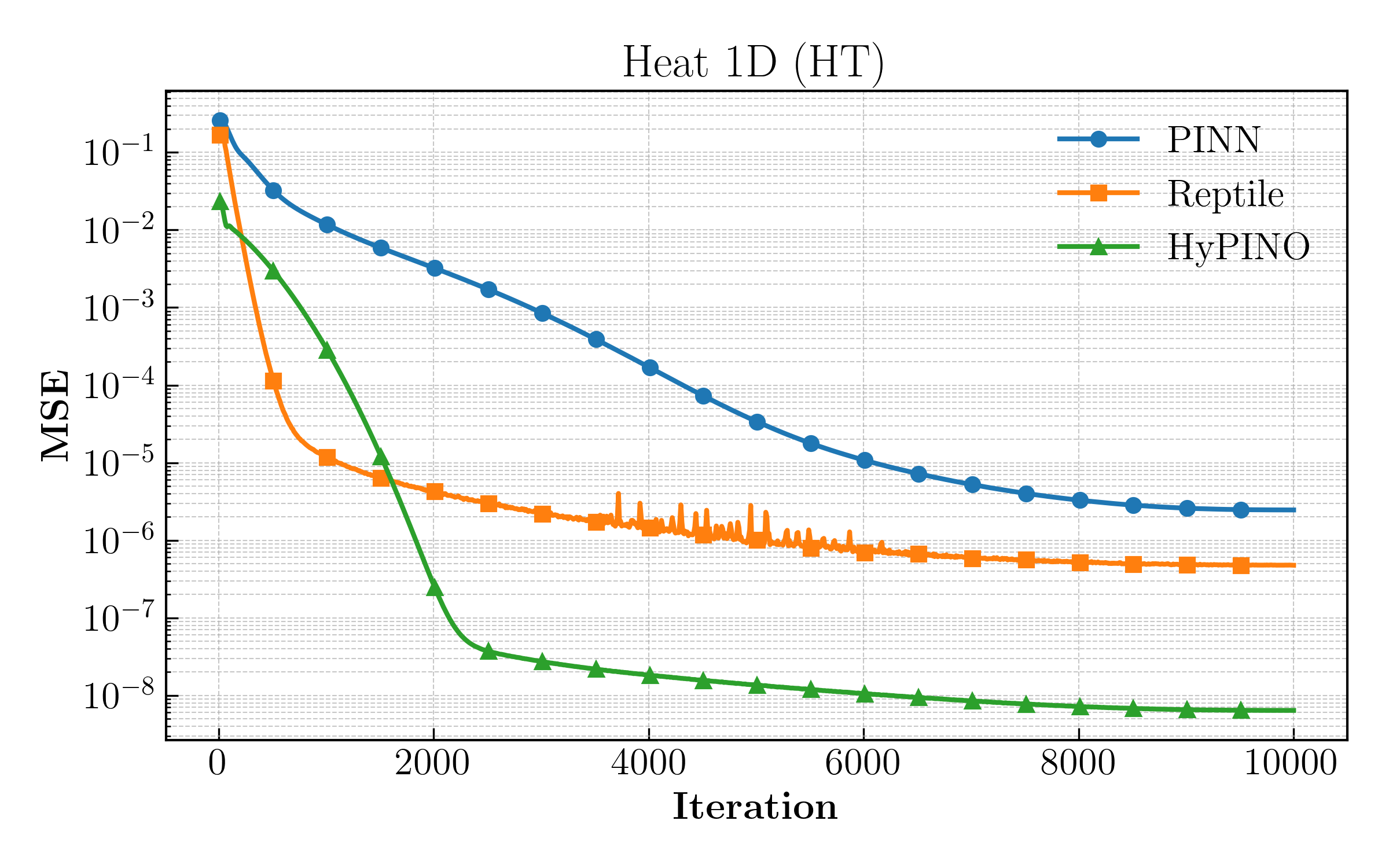}
    \caption{Convergence on the 1D Heat Equation (HT) for randomly initialized PINNs (blue), Reptile-initialized PINNs (orange), and HyPINO-initialized PINNs.}
    \label{fig:finetuning_heat_results}
\end{wrapfigure}

PINN fine-tuning is performed over 10,000 steps using the Adam optimizer, starting with a learning rate of $10^{-4}$, decayed to $10^{-7}$ via a cosine schedule. Figure~\ref{fig:finetuning_heat_results} shows convergence results on the 1D Heat Equation (HT) benchmark; results for other benchmarks and ensemble comparisons are shown in Figures~\ref{fig:finetuning_convergence} and~\ref{fig:finetuning_convergence_2}.

HyPINO-initialized PINNs consistently start with lower loss and converge to lower final error on 4 out of 7 benchmarks. On two benchmarks, they match baseline performance, and on 1, they underperform. Quantitatively, a randomly initialized PINN requires an average of 1,068 steps to reach the initial MSE of a HyPINO-initialized model. For ensembles, matching the MSE of HyPINO$^3$ and HyPINO$^{10}$ requires an average of 1,617 and 1,772 steps, respectively. Reptile-initialized PINNs converge rapidly during the first 1,000 steps, which is consistent with their meta-training configuration. However, they tend to plateau earlier and converge to higher final errors than HyPINO initializations. These findings suggest that, in addition to strong zero-shot performance, HyPINO offers a robust initialization strategy for training PINNs.

\section{Conclusions \& Outlook}
\label{sec:conclusion}
We introduce a multi-physics neural operator based on hypernetworks (HyPINO), trained on synthetic data comprising both supervised samples, constructed using the Method of Manufactured Solutions, and purely physics-informed samples without ground-truth labels. To the best of our knowledge, our framework provides the highest degree of flexibility in the input space among existing neural operators: it accommodates variations in the differential operator, source term, domain geometry (including interior boundaries), and boundary / initial conditions. Our experiments demonstrate that training on this synthetic dataset enables strong zero-shot generalization across a diverse set of benchmark PDEs. This suggests that multi-physics neural operators can be learned with significantly reduced reliance on high-fidelity, labeled training data by leveraging synthetic datasets and self-supervised objectives. In addition, we propose a lightweight and effective iterative refinement strategy that significantly improves prediction accuracy. Notably, this refinement mechanism is generic and can be applied to other physics-informed neural operator frameworks as well. We also show that HyPINO-generated parameters provide excellent initialization for fine-tuning PINNs on specific PDE instances, yielding faster convergence and lower final errors compared to both randomly initialized and Reptile-initialized baselines.

Nonetheless, several limitations remain. Our current implementation is restricted to linear 2D PDEs with spatially uniform coefficients, which narrows the class of PDEs that HyPINO can currently address. However, the framework is inherently extensible. Future work will explore increasing the input dimensionality, incorporating spatially varying coefficients, supporting nonlinear PDEs, and modeling coupled systems. Some of these extensions may be achievable through modest modifications to the data generation process, the model’s input encoding architecture, or extended training. Others may necessitate increased model capacity, either through scaling the architecture or improving the target networks’ parameter generation process.

\bibliography{references}
\bibliographystyle{plainnat}


\newpage
\appendix
\input{appendix}

\end{document}

%% file: appendix.tex
\section{Methodology}
\label{app:methodology}

\subsection{Neural Operator Architecture}
\label{app:neural_operator}

We build on the HyperPINN framework~\cite{hyperpinn} and design a neural operator based on a hypernetwork that generates the weights \(\theta^\star\) of a Physics-Informed Neural Network (PINN) \(u_{\theta}\), conditioned on a given PDE instance. Specifically, the hypernetwork learns a mapping
\begin{equation}
\big(\mathbf{c}, \; f,\; g,\; h \big) \;\longmapsto\; \theta^\star
\quad\text{such that}\quad 
u_{\theta^\star} \approx u,
\label{eq_app:hyperpinn}
\end{equation}
where $\mathbf{c}$ denotes the vector of PDE coefficients, $f$ the interior source term, $g$ and $h$ the Dirichlet and Neumann boundary conditions, respectively, and $u$ the true solution.

\paragraph{Grid embeddings.}
Each grid-valued input is first passed through a Fourier feature mapping~\cite{tancik2020fourier}, which augments the input with sinusoidal encodings using five exponentially increasing bands $\text{frequencies} = 0.1 \cdot 2^i, \quad i \in \{0, 1, 2, 3, 4\}$. This enhances the network’s ability to represent high-frequency content and reduces spectral bias. The Fourier mapping layer is followed by two convolutional layers with kernel size three and strides of two. For the boundary location grids $M$ (cf. Section~\ref{sec:parameterization}), we compute four embeddings: $z_D^{1},\; z_D^{2},\; z_N^{1},\; z_N^{2}$. For the boundary value grids $V$, we compute $z_g$ (Dirichlet values) and $z_h$ (Neumann values). The source term yields the embedding $z_f$.

We define the final spatial embedding $z_G$ by
\begin{equation}
\label{eq_app:zG}
z_G
= \left[
z_D^{1} \odot z_g + z_D^{2} \;\Vert\;
z_N^{1} \odot z_h + z_N^{2} \;\Vert\;
z_f
\right],
\end{equation}
where $\odot$ denotes element-wise multiplication and $[\cdot \Vert \cdot \Vert \cdot]$ denotes concatenation along the channel dimension. This composition naturally applies spatial masking to the boundary value embeddings using the boundary location masks, ensuring that information is injected only at semantically meaningful locations.

\paragraph{Coefficient embedding.}
The vector of operator coefficients $\mathbf{c} \in \mathbb{R}^5$ is embedded into a fixed-length representation $z_C \in \mathbb{R}^{d_C}$ using a Fourier feature encoder followed by a fully connected layer.

\paragraph{Encoding.}
The grid embedding \(z_G\) is processed by a sequence of \(K\) Swin Transformer blocks \(\{\mathcal{SW}_i\}_{i=1}^K\).  Denoting by \( z^{(i)} \in \mathbb R^{H_i \times W_i \times C_i} \)
the output of block \(\mathcal{SW}_i\) and \( z^{(0)} = z_G \), we interleave each block with a FiLM modulation \cite{film} conditioned on the coefficient embeddings $z_C$. Concretely, we define
\begin{equation}
\label{eq_app:film_mlps}
\gamma_i(z),\,\beta_i(z)\;\colon\;\mathbb R^{d_C}\;\to\;\mathbb R^{C_i}
\end{equation}
via small MLPs, and write
\begin{equation}
\label{eq_app:film}
z^{(i+1)} \;=\; \gamma_i(z_C) \;\odot\; \mathcal{SW}_i\bigl(z_G^{(i)}\bigr)\;+\;\beta_i(z_C),
\end{equation}
where “\(\odot\)” denotes channel-wise scaling broadcast across spatial dimensions. This design ensures that at each stage, the latent grid features are adaptively modulated by both the global operator coefficients \(z_C\). 

Inspired by Swin Transformer U-Net architectures~\cite{swin_unet, swin_unet_denoising}, we retain all intermediate latent representations from the Swin blocks $\{z^{(i)}\}_{i=1}^{K}$ to keep information at various semantic levels.

\paragraph{Pooling.}
To aggregate spatial information into a compact latent representation suitable for parameterizing the target PINN, we perform Multi-Head Attention Pooling~\cite{lee2019set,zhai2022scaling} across the flattened outputs of each Swin Transformer block. Specifically, let $z_i \in \mathbb{R}^{H_i \times W_i \times C_i}$ denote the output of the $i$-th FiLM-modulated Swin block. We reshape it into a sequence of tokens \( kv_i \in \mathbb{R}^{H_i W_i \times C_i} \), which serve as both keys and values in the attention mechanism.

For each layer \( i \in \{1, \dots, K\} \), we define a set of \( T \) trainable query vectors \( q_i \in \mathbb{R}^{T \times C_i} \), where \( T \) corresponds to the number of weight and bias tensors in the target PINN. We then compute the pooled representation via multi-head attention:
\begin{equation}
    \label{eq_app:map}
    p_i = \operatorname{MultiHeadAttention}_i(q_i, kv_i, kv_i), \quad p_i \in \mathbb{R}^{T \times C_i}.
\end{equation}

The pooled outputs \( \{p_i\}_{i=1}^K \) are concatenated along the channel dimension to produce a unified latent matrix,
\begin{equation}
    \label{eq_app:concat_latent}
    p = \left[p_1 \;\Vert\; p_2 \;\Vert\; \cdots \;\Vert\; p_K\right] \in \mathbb{R}^{T \times \left(\sum_{i=1}^K C_i\right)}.
\end{equation}

This matrix \( p \) contains one latent vector per target weight or bias tensor, each embedding multi-scale information aggregated across the Swin hierarchy. To obtain the actual PINN parameters, we apply a dedicated MLP to each row of \( p \), mapping it to the appropriate shape and dimensionality required by the corresponding weight matrix or bias vector.

\paragraph{Target PINN.}
We define the architecture of the target PINN as an MLP with Fourier feature mapping \cite{tancik2020fourier} and multiplicative skip connections \cite{wang2021mult_skip_connections}.
Fourier encodings provide spectral expressivity for modeling high-frequency components \cite{fourier_pinns}, while the skip connections enhance gradient propagation and, in the context of hypernetworks, have the additional benefit of enabling dynamic depth modulation based on PDE complexity by masking some layers.

Given a spatial input $\mathbf{x} \in \mathbb{R}^2$, the (non-trainable) encoding is defined as:

\begin{equation}
    \label{eq_app:fourier_mapping}
    \xi(\mathbf{x}) = \left[
    \sin\left(2\pi \mathbf{B} \mathbf{x}\right),\;
    \cos\left(2\pi \mathbf{B} \mathbf{x}\right),\;
    \mathbf{x}
    \right] \in \mathbb{R}^{2N + 2},
\end{equation}

where $\mathbf{B} \in \mathbb{R}^{N \times 2}$ contains exponentially spaced frequency bands.

Following \citet{wang2021mult_skip_connections}, the encoded input is projected through three parallel transformations:

\begin{equation}
    \label{eq_app:pinn_input_layer}
    z_0 = \tanh(W_{\text{in}} \xi + b_0), \quad
    z_u = \tanh(U \xi + b_u), \quad
    z_v = \tanh(V \xi + b_v),
\end{equation}

where $W_{\text{in}}, U, V \in \mathbb{R}^{d \times (2N + 2)}$ and $b_0, b_u, b_v \in \mathbb{R}^d$. The hidden layers of the PINN are computed via:

\begin{equation}
    \label{eq_app:pinn_hidden_layer}
    z_{i+1} = z_u \odot \tanh(W_i z_i + b_i) + z_v \odot \left(1 - \tanh(W_i z_i + b_i)\right), \quad i = 0, \dots, T-2,
\end{equation}

with weight matrices $W_i \in \mathbb{R}^{d \times d}$ and biases $b_i \in \mathbb{R}^d$. Note that we use the $\tanh$ activation due to its bounded output range, which prevents exploding values during the hypernetwork training. 

The final prediction is obtained by a linear transformation:

\begin{equation}
    \label{eq_app:pinn_output_layer}
    u_\theta(\mathbf{x}) = W_{\text{out}} z_{T-1} + b_{\text{out}}, \quad
    W_{\text{out}} \in \mathbb{R}^{1 \times d}, \;
    b_{\text{out}} \in \mathbb{R}.
\end{equation}

For each PDE instance, the hypernetwork therefore generates the following parameter set $\theta^\star$:
\begin{equation}
\left\{W_0, U, V, b_0, b_u, b_v \right\}, \quad
\left\{W_i, b_i\right\}_{i=1}^{T-2}, \quad
W_{\text{out}}, b_{\text{out}},
\label{eq_app:pinn_parameter_set}
\end{equation}
\subsection{Data Sampling}
\label{app:data_sampling}

\subsubsection{Classes of PDEs}
\label{app:pde_classes}
We construct a synthetic dataset of PDE instances by systematically sampling the governing equations, the domain, boundary conditions, source terms, and (optionally) known solutions. Two classes of samples are considered:

\paragraph{Class I: Supervised PDEs} 
We generate a set of PDEs with analytical solutions via MMS. Specifically, we sample:
\begin{enumerate}
    \item The differential operator \(\mathcal{L}\).
    \item The domain \(\Omega\) (along with \(\partial \Omega\)).
    \item An analytical solution \(u(\mathbf{x})\).
\end{enumerate}
From the chosen solution \(u(\mathbf{x})\), we then compute:
\begin{itemize}
    \item The source term \(f(\mathbf{x})\) by applying \(\mathcal{L}\) to \(u\).
    \item The boundary conditions \(g(\mathbf{x}) = u(\mathbf{x})\) and / or \(h(\mathbf{x}) = \tfrac{\partial u}{\partial n}(\mathbf{x})\) by evaluating \(u(\mathbf{x})\) and its normal derivative on \(\partial \Omega\).
\end{itemize}
In addition to the self-supervised physics-informed loss, samples of this class provide \(u(\mathbf{x})\) as well as its derivatives that can be used as additional supervised losses during training.

\paragraph{Class II: Unsupervised PDEs}
In this class, the analytical solution \(u(\mathbf{x})\) is not known \emph{a priori}. We create samples by choosing:
\begin{enumerate}
    \item The differential operator \(\mathcal{L}\).
    \item The domain \(\Omega\) (and \(\partial \Omega\)).
    \item The source term \(f(\mathbf{x})\).
    \item Boundary conditions, subject to constraints designed to maximize the probability of well-posedness.
\end{enumerate}
Since the ground-truth solution is not available, samples from this class rely on the self-supervised physics-informed loss to train the model. The full dataset consists of a mix of samples from both types, with the loss containing a switch to ignore the supervised loss if no analytical solution is available.

\subsubsection{Sampling Differential Operators}
\label{app:sampling_governing_equations}

Considering \(\mathcal{B} = \{u, u_{x}, u_{y}, u_{xx}, u_{yy}\}\) to be the set of all terms that can appear in our differential operators, we sample the number of terms \(n \sim \text{Uniform}({1, 2, 3})\). We then randomly select \(n\) terms from \(\mathcal{B}\) without repetition and obtain their coefficients (cf. Section~\ref{sec:parameterization}) \(c_i \sim \text{Uniform}([-2, 2])\). The sum of the selected terms multiplied by their respective coefficients constitutes the final differential operator.

\subsubsection{Sampling or Deriving the Source Terms} 
\label{app:sampling_source_terms}

The source term \(f(\mathbf{x})\) is handled differently based on whether the sample has a known analytical solution. For cases with an analytical solution, \(u(\mathbf{x})\) is sampled (see Section~\ref{app:sampling_analytical_solutions}), and the source is computed by inserting \(u\) into the differential operator. For samples without analytical solution, we set the source function to a constant \(f(\mathbf{x}) = \mathcal{N}(0,10^2)\).

\subsubsection{Sampling Analytical Solutions via MMS}
\label{app:sampling_analytical_solutions}

We generate analytical solutions \( u : \Omega \to \mathbb{R} \), with \( \Omega \subset \mathbb{R}^2 \) and \( \mathbf{x} = (x, y) \), by iteratively combining \( n \) randomly constructed terms, as detailed in Algorithm~\ref{alg:sample_analytical_solutions}. The number of terms is drawn from a discrete uniform distribution, \( n \sim \text{Uniform}(\{6, 7, \ldots, 10\}) \). The initial solution is set to zero: \( u(x, y) \gets 0 \).

Each term is constructed by selecting a nonlinear function \( \psi \) from a predefined library:
\[
\{x, \sin, \cos, \tanh, \frac{1}{1 + e^{-x}}, \frac{1}{1 + x^2}\}
\]

The coefficients \( a \) and \( b \) are sampled from the set \( \{0, \text{Uniform}([-10, 10])\} \). The remaining coefficients \( c, d, e \) are sampled as \(c, d, e \sim \text{Uniform}([-2\pi, 2\pi])\). A term is then computed as \( d \cdot \psi(a x + b y + c) + e \), and integrated into the current state of \( u(x, y) \) using one of three randomly chosen rules:

\begin{tabbing}
\hspace{3cm} \= \kill
\textbf{Addition:} \> \( u(x, y) \gets u(x, y) + d \cdot \psi(a x + b y + c) + e \) \\
\textbf{Multiplication:} \> \( u(x, y) \gets u(x, y) \cdot d \cdot \psi(a x + b y + c) + e \) \\
\textbf{Composition:} \> \( u(x, y) \gets d \cdot \psi(a \cdot u(x, y) + c) + e \)
\end{tabbing}

\begin{algorithm}
\SetAlgoLined
\DontPrintSemicolon
Initialize \( u(x, y) \gets 0 \)\; 
Sample \( n \sim \text{Uniform}(\{6, 7, \ldots, 10\}) \)\; 
\For{\( i = 1 \) to \( n \)}{
    Sample \( a \sim \{0, \text{Uniform}([-10, 10])\} \)\; 
    Sample \( b \sim \{0, \text{Uniform}([-10, 10])\} \)\; 
    Sample \( c, d, e \sim \text{Uniform}([-2\pi, 2\pi]) \)\; 
    Randomly select \( \psi(x) \in \{\sin, \cos, \tanh, \sigma, x, \phi(x)\} \)\; 
    Compute \( \text{term} \gets d \cdot \psi(a x + b y + c) + e \)\; 
    Randomly choose combination rule:\;
    \uIf{add}{
        \( u(x, y) \gets u(x, y) + \text{term} \)\;
    }
    \uElseIf{multiply}{
        \( u(x, y) \gets u(x, y) \cdot \text{term} \)\;
    }
    \ElseIf{compose}{
        \( u(x, y) \gets d \cdot \psi(a \cdot u(x, y) + c) + e \)\;
    }
}
\Return \( u(x, y) \)\;
\caption{Sampling procedure for random, differentiable functions that can be used as analytical solutions with MMS.}
\label{alg:sample_analytical_solutions}
\end{algorithm}

\subsubsection{Sampling Physical Domains}
\label{app:sampling_domains}

We employ a randomized sampling procedure based on Constructive Solid Geometry (CSG)~\cite{deepxde} to generate complex and diverse domains.

To begin, we define the domain \(\Omega\) as the bounding box \([-1, 1]^2\), representing the outer boundary \(\partial \Omega_{\text{outer}}\). This initial outer region can describe a purely spatial or a spatiotemporal domain. Although we continue to use \((x, y)\) to denote the coordinate variables, in certain PDE classes (e.g., parabolic or hyperbolic), the variable \(y\) may represent the temporal dimension, with \(y = -1\) corresponding to the initial time. We then create inner boundaries \(\partial \Omega_{\text{inner}, i}\) (\(i = 1, 2, \ldots, n\)) by randomly generating geometric shapes (e.g., triangles, polygons, disks, rectangles) and subtracting them from the outer region using CSG operations. An example of a sampled domain is shown in Figure~\ref{fig:sampled_pde_supervised}.

\subsubsection{Sampling Boundary Conditions}
\label{app:sampling_boundary_conditions}

We consider two types of boundary conditions on \(\partial \Omega\): Dirichlet and Neumann. 
Note that in our setting, the computational domain is \(\Omega \subset [-1, 1]^2\). 

To maximize the likelihood of obtaining well-posed PDEs, we first categorize the PDE as elliptic, parabolic, or hyperbolic. Based on this classification, the following boundary conditions are imposed on the outer boundary \(\partial \Omega_{\text{outer}}\):

\begin{itemize}
    \item \textbf{Elliptic PDEs}:
    Dirichlet conditions on \(\partial \Omega_{\text{outer}}\):
    \begin{equation}
        u(\mathbf{x}) = g(\mathbf{x}),
        \quad \mathbf{x} \in \partial \Omega_{\text{outer}}.
    \end{equation}

    \item \textbf{Parabolic PDEs}:
    Interpreting \(y\) as time, the \emph{initial condition} is enforced at \(y = -1\). In addition, we impose Dirichlet conditions on the spatial boundaries:
    \begin{equation}
         u(\mathbf{x}) = g(\mathbf{x}), \; \mathbf{x} \in \partial \Omega_{\text{outer}} \setminus \{y=1\}.
    \end{equation}

    \item \textbf{Hyperbolic PDEs}:
    Similar to the parabolic setup, we set \(y=-1\) as the initial time and enforce
    \begin{equation}
         u(\mathbf{x}) = g(\mathbf{x}), \; \mathbf{x} \in \partial \Omega_{\text{outer}} \setminus \{y=1\},
    \end{equation}
    \begin{equation}
        \frac{\partial u}{\partial n}(\mathbf{x}) \;=\; h(\mathbf{x}), \; \mathbf{x} \in \partial \Omega_{\text{outer}} \cap \{y = -1\}.
    \end{equation}
\end{itemize}

For inner boundaries (created by subtracting geometric shapes via CSG, cf.~Section~\ref{app:sampling_domains}), each component \(\partial \Omega_{\text{inner}, i}\), with \(i \in \{0, 1, \dots, n\}\), is independently assigned either a Dirichlet or Neumann condition, or both:

\begin{equation}
    u(\mathbf{x}) = g_i(\mathbf{x}) 
    \quad\text{or}\quad 
    \frac{\partial u}{\partial n}(\mathbf{x}) = h_i(\mathbf{x}),
    \quad \mathbf{x} \in \partial \Omega_{\text{inner}, i}.
\end{equation}

For samples from Class~I (Section~\ref{app:pde_classes}), where an analytical solution \(u(\mathbf{x})\) is known, boundary conditions follow directly from:

\begin{equation}
    g(\mathbf{x}) = u(\mathbf{x}),\quad 
    \mathbf{x} \in \partial \Omega_D
\end{equation}
\begin{equation}
    h(\mathbf{x}) = \frac{\partial u}{\partial n}(\mathbf{x}),\quad 
    \mathbf{x} \in \partial \Omega_N
    .
\end{equation}

Here, \(g\) and \(h\) are computed by evaluating \(u\) and its normal derivative at the relevant boundary segments (outer or inner). For samples from Class~II, we set the source term \(f(\mathbf{x})\) to zero on the boundary (see Section~\ref{app:sampling_source_terms}) and sample boundary values in a manner consistent with \(\mathcal{L}[u]\) to ensure that the governing equation and boundary conditions remain compatible. Specifically:

\begin{itemize}
    \item If \(u\) appears as a standalone term in \(\mathcal{L}[u]\), we set \(u(\mathbf{x}) = 0\) on \(\partial \Omega\).
    \item If first-order terms such as \(u_{x}\) or \(u_{y}\) appear stand-alone, we allow the corresponding Dirichlet boundary value \(u(\mathbf{x})\) to be a random constant.
    \item Otherwise, we also include linear functions as possible boundary values.
\end{itemize}

Despite efforts to ensure well-posedness in the generation of synthetic, unsupervised training samples, some configurations may still result in ill-posed problems due to conflicting boundary constraints and source functions. Nevertheless, these unsupervised samples are essential for enabling the model to learn from domains with interior boundaries. While supervised samples can also include inner boundaries, they are, by construction, overconstrained: the combination of the source term and outer boundary conditions suffices to determine the analytical solution. As a result, the model primarily learns to represent and adapt to interior boundary effects through unsupervised data, where such features introduce structural variability without the aid of explicit targets. Given the importance of accurately modeling interior boundaries in practical applications, we consider this trade-off acceptable.

\newpage
\section{Experiments}
\label{app:experiments}
All problems are reformulated over the canonical domain \([{-1}, 1]^2\). In particular, problems originally defined over domains \([a_x, b_x] \times [a_y, b_y]\) are mapped to \([{-1}, 1]^2\) through affine transformations of the form:
\(
\tilde{x} = \frac{2(x - a_x)}{b_x - a_x} - 1, \quad
\tilde{y} = \frac{2(y - a_y)}{b_y - a_y} - 1,
\)
where \((x, y)\) are original spatial coordinates and \((\tilde{x}, \tilde{y}) \in [-1, 1]^2\) are the normalized coordinates.

\subsection{Heat 1D (HT)}
\label{app:heat}
Consider the one-dimensional heat equation:
\begin{equation}
\label{eq:heat_equation}
\frac{\partial u}{\partial t} = \alpha \frac{\partial^2 u}{\partial x^2}, 
\quad x \in [0, 1], \quad t \in [0, 1],
\end{equation}
where \(\alpha = 0.1\) denotes the thermal diffusivity constant.

Dirichlet boundary conditions are imposed as:
\begin{equation}
\label{eq:heat_bc}
u(0, t) = u(1, t) = 0.
\end{equation}

The initial condition is given by a periodic (sinusoidal) function:
\begin{equation}
\label{eq:heat_ic}
u(x, 0) = \sin\left(\frac{n \pi x}{L}\right),
\quad 0 < x < L, \quad n = 1, 2, \dots,
\end{equation}
where \(L = 1\) is the length of the domain, and \(n\) is the frequency parameter.

The corresponding exact solution is:
\begin{equation}
\label{eq:heat_exact}
u(x, t) = \exp\left(-\frac{n^2 \pi^2 \alpha t}{L^2}\right)
\sin\left(\frac{n \pi x}{L}\right).
\end{equation}

\begin{figure}[H]
    \centering
    \begin{subfigure}[b]{0.16\linewidth}
        \includegraphics[width=\linewidth]{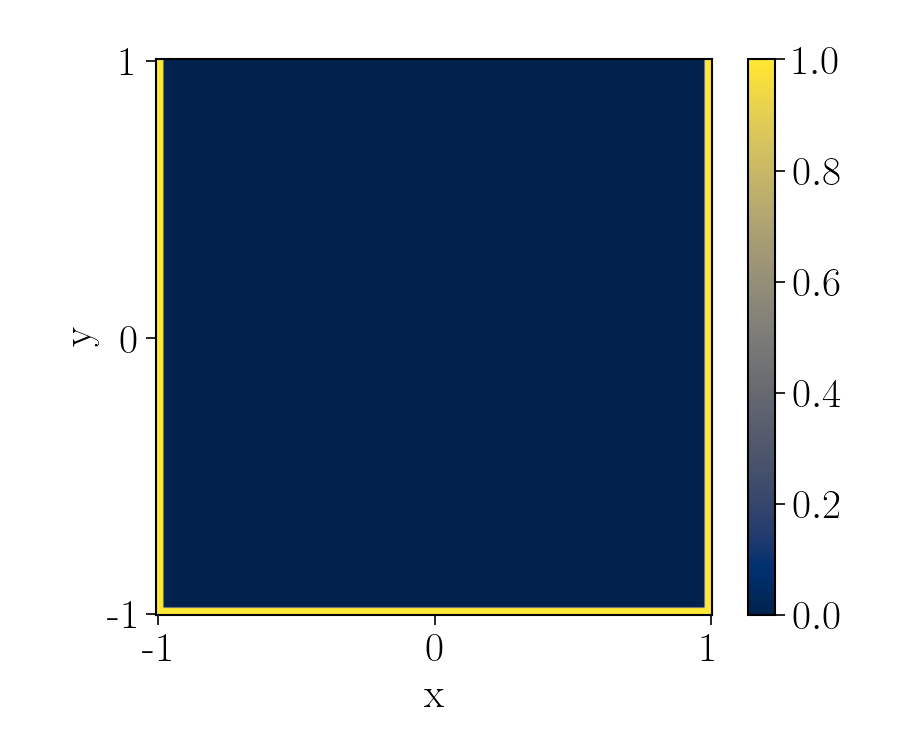}
        \caption{$\partial \Omega_D$}
    \end{subfigure}
    \begin{subfigure}[b]{0.16\linewidth}
        \includegraphics[width=\linewidth]{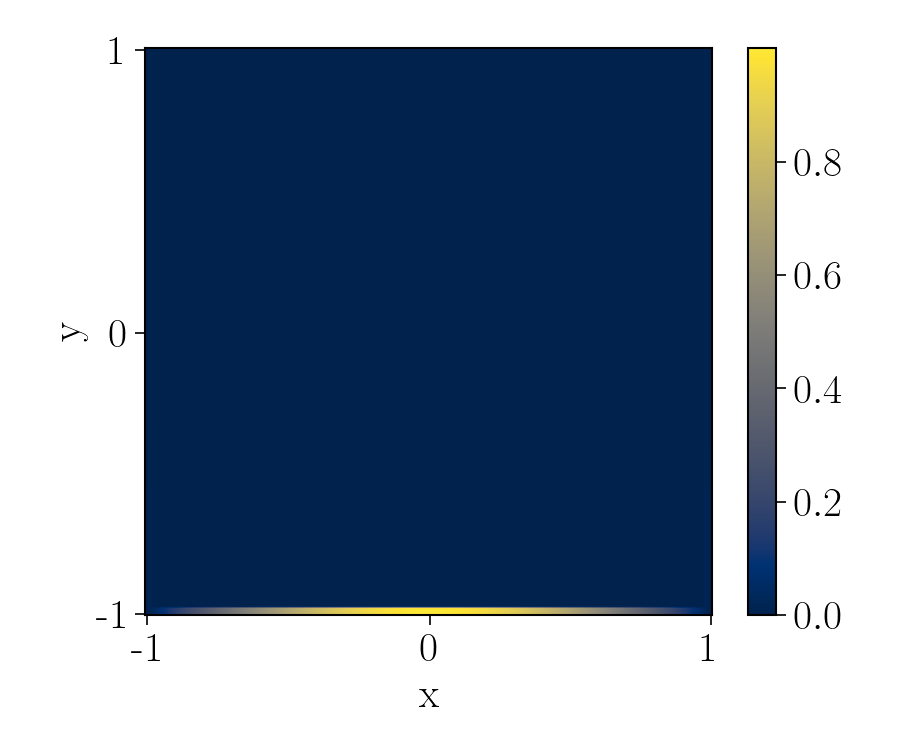}
        \caption{$g(x)$}
    \end{subfigure}
    \begin{subfigure}[b]{0.16\linewidth}
        \includegraphics[width=\linewidth]{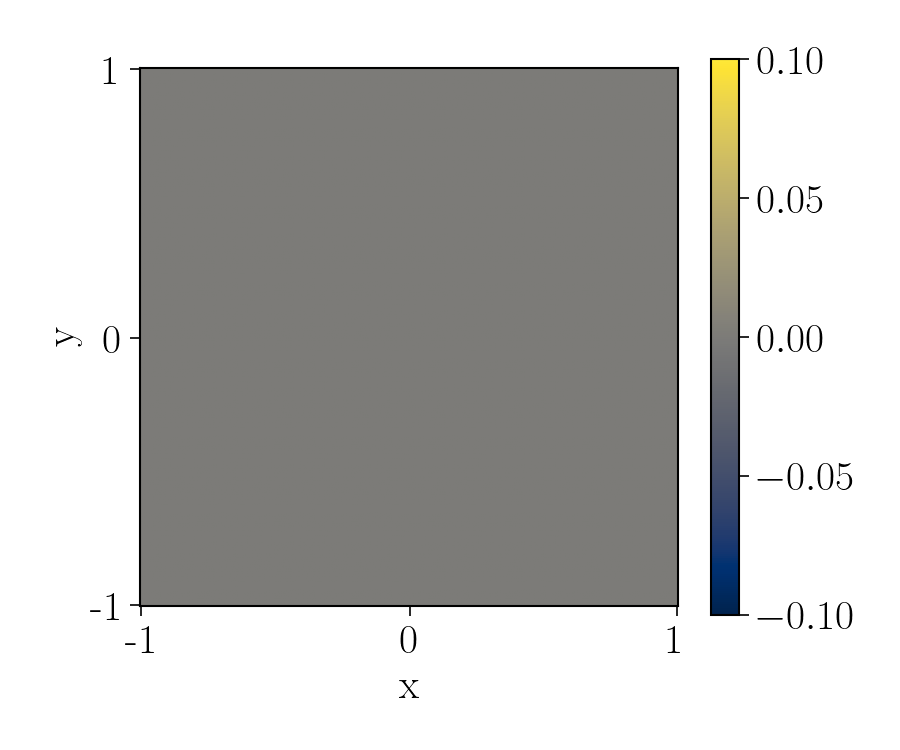}
        \caption{$\partial \Omega_N$}
    \end{subfigure}
    \begin{subfigure}[b]{0.16\linewidth}
        \includegraphics[width=\linewidth]{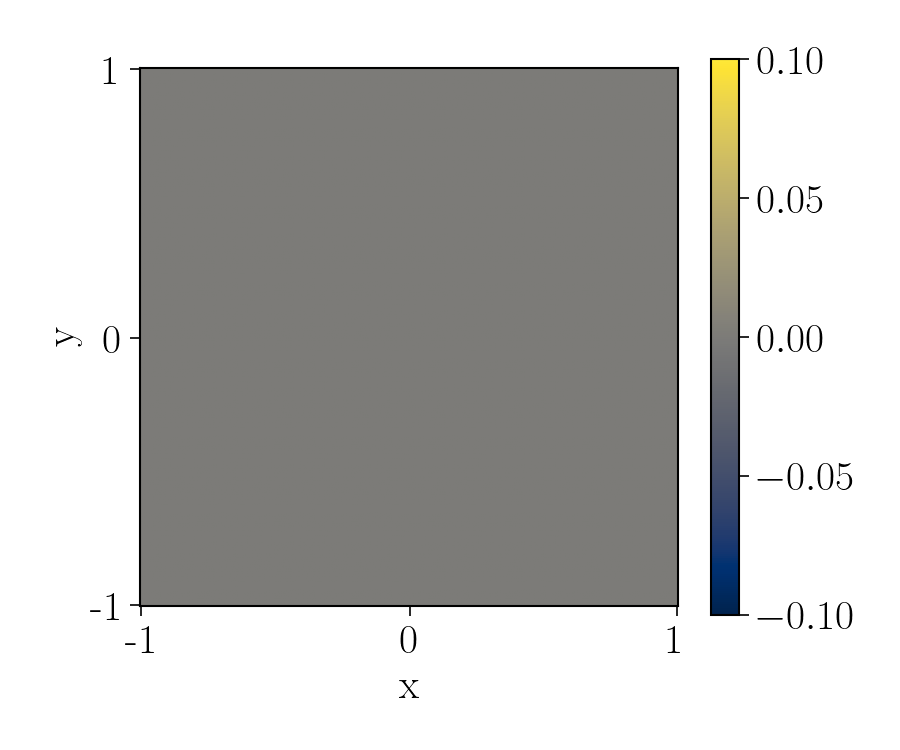}
        \caption{$h(x)$}
    \end{subfigure}
    \begin{subfigure}[b]{0.16\linewidth}
        \includegraphics[width=\linewidth]{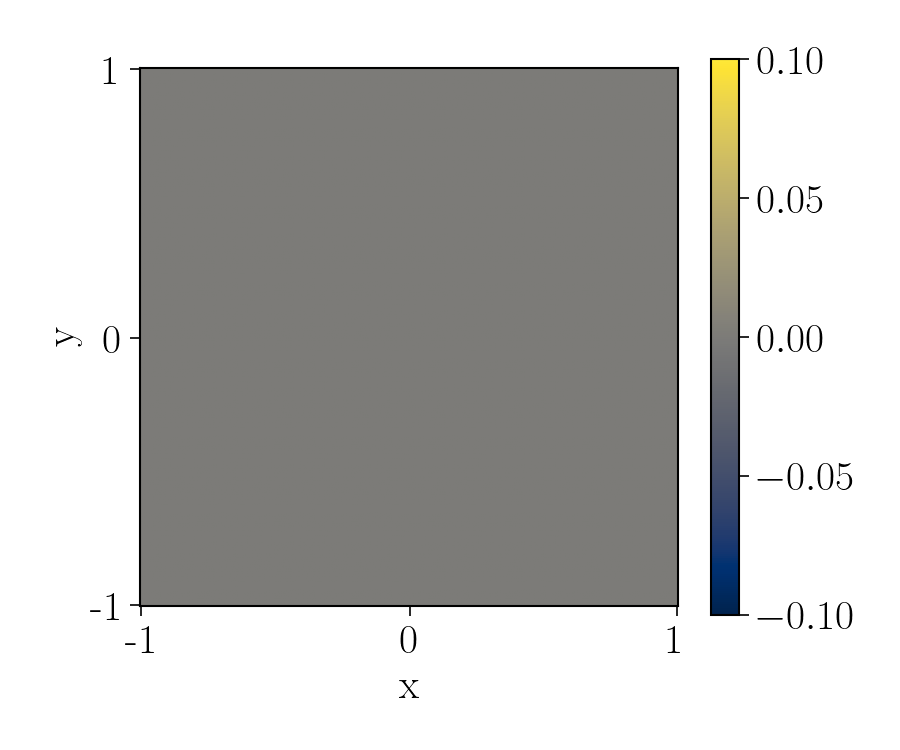}
        \caption{$f(x)$}
    \end{subfigure}
    \begin{subfigure}[b]{0.16\linewidth}
        \includegraphics[width=\linewidth]{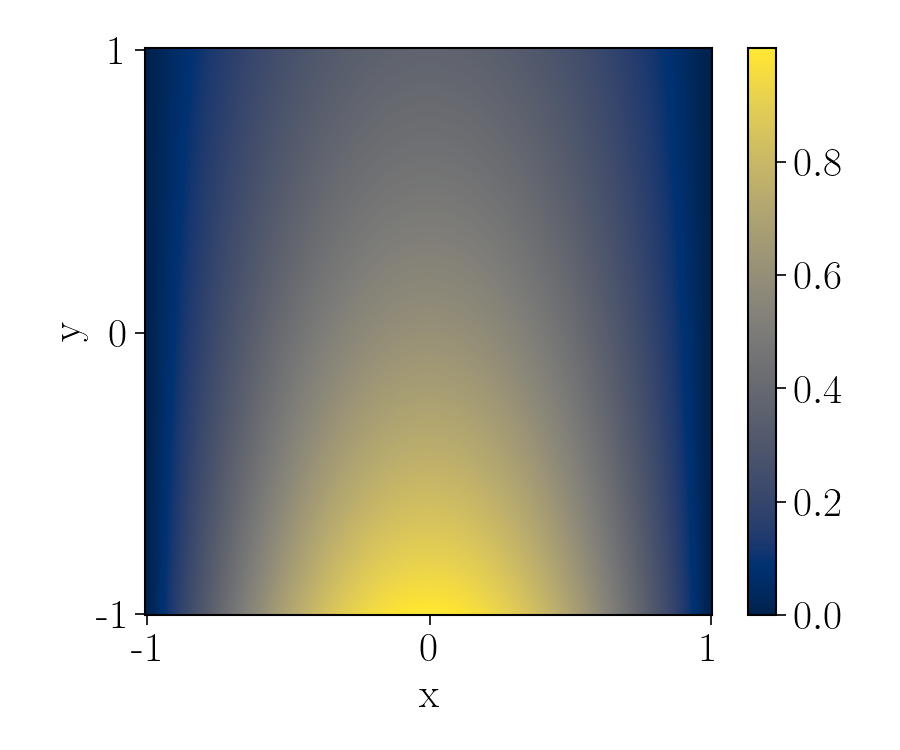}
        \caption{$u(x)$}
    \end{subfigure}

    \caption{Parameterization of the 1D Heat PDE.}
    \label{fig:heat_parametrization}
\end{figure}

This benchmark problem is adapted from DeepXDE~\cite{deepxde}.
Figure~\ref{fig:heat_parametrization} shows the parameterization of the different PDE components.

\subsection{Helmholtz 2D (HZ)}
\label{app:helmholtz}
Consider the two-dimensional Helmholtz equation:
\begin{equation}
\label{eq:helmholtz}
\Delta u(x, y) + k^2 u(x, y) = f(x, y), \quad (x, y) \in [-1, 1]^2,
\end{equation}
where \( k \) is the wave number.

Dirichlet boundary conditions are imposed as:
\[
u(-1, y) = u(1, y) = u(x, -1) = u(x, 1) = 0.
\]

A commonly used instance with an analytical solution is:
\begin{equation}
\label{eq:helmholtz_analytical}
\begin{aligned}
f(x, y) &= \left( -\pi^2 - (4\pi)^2 + k^2 \right) \sin(\pi x) \sin(4\pi y), \\
u(x, y) &= \sin(\pi x) \sin(4\pi y).
\end{aligned}
\end{equation}

\begin{figure}[H]
    \centering
    \begin{subfigure}[b]{0.16\linewidth}
        \includegraphics[width=\linewidth]{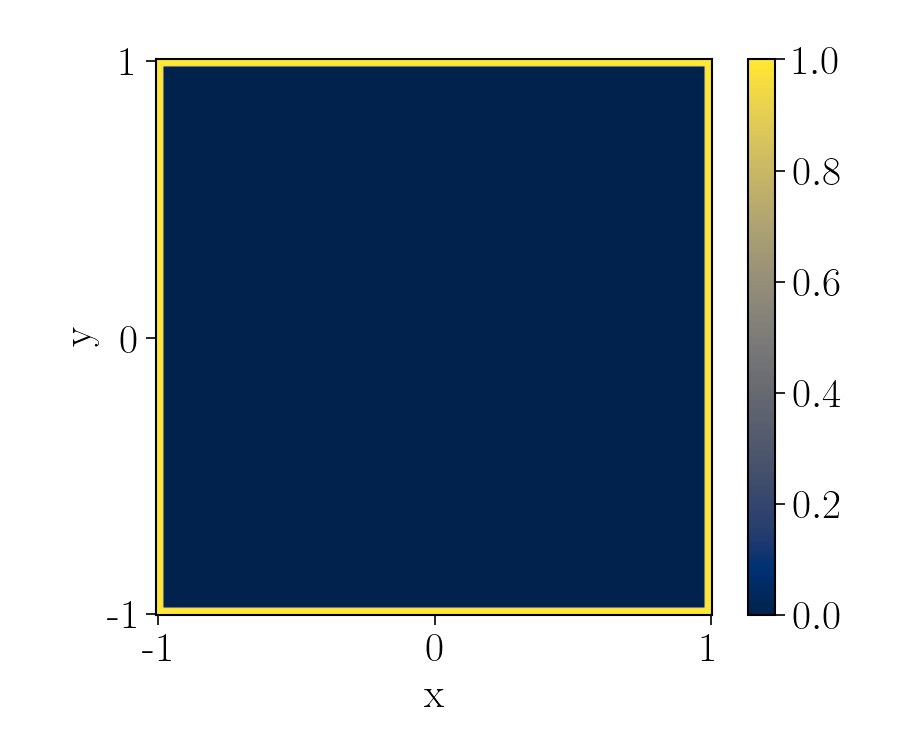}
        \caption{$\partial \Omega_D$}
    \end{subfigure}
    \begin{subfigure}[b]{0.16\linewidth}
        \includegraphics[width=\linewidth]{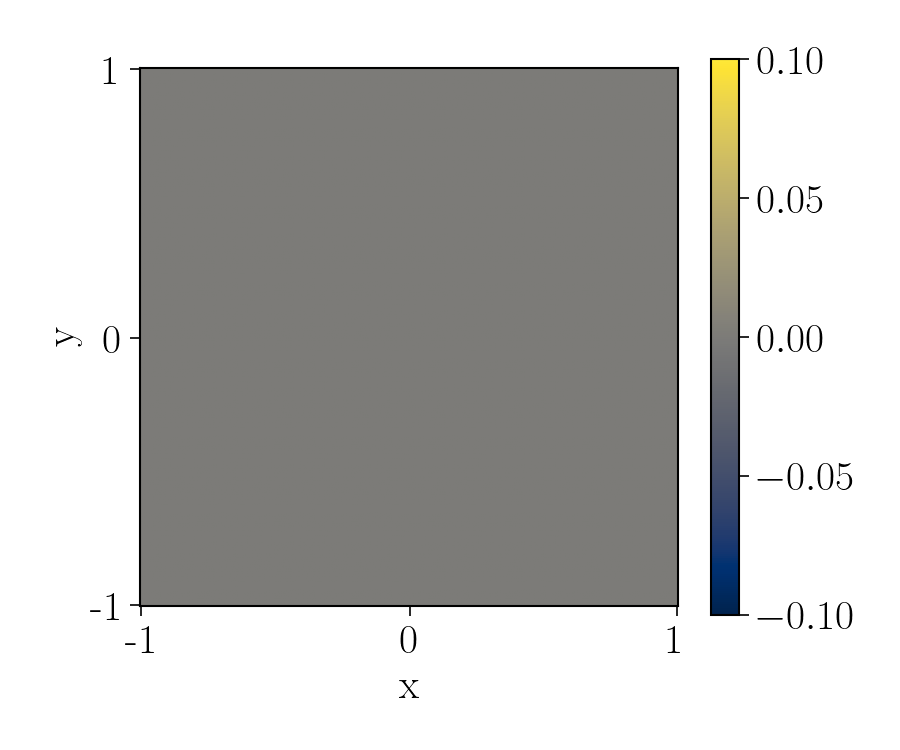}
        \caption{$g(x)$}
    \end{subfigure}
    \begin{subfigure}[b]{0.16\linewidth}
        \includegraphics[width=\linewidth]{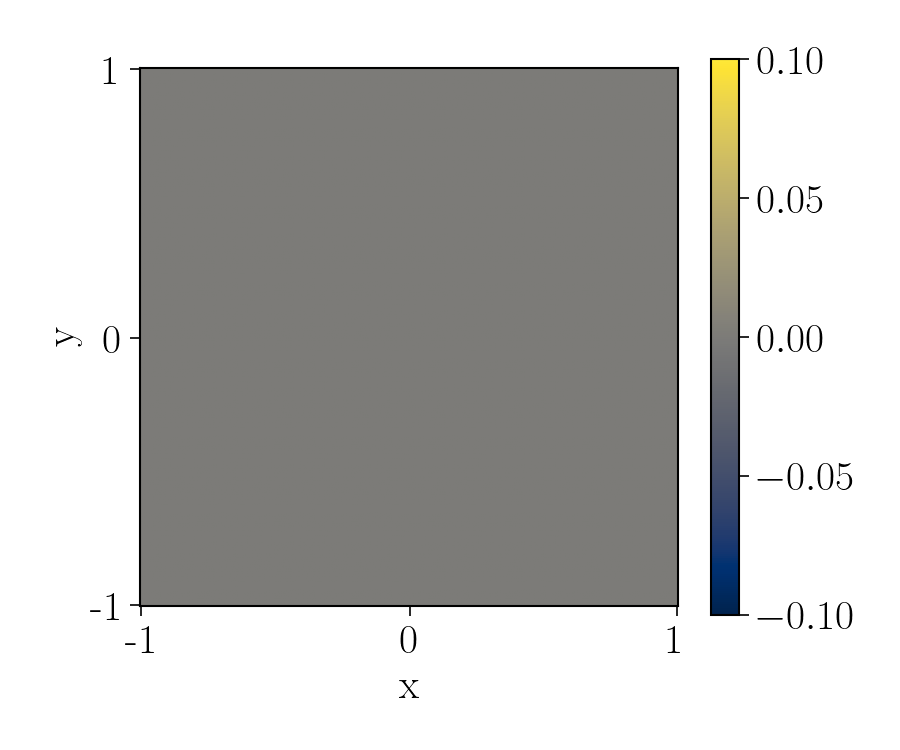}
        \caption{$\partial \Omega_N$}
    \end{subfigure}
    \begin{subfigure}[b]{0.16\linewidth}
        \includegraphics[width=\linewidth]{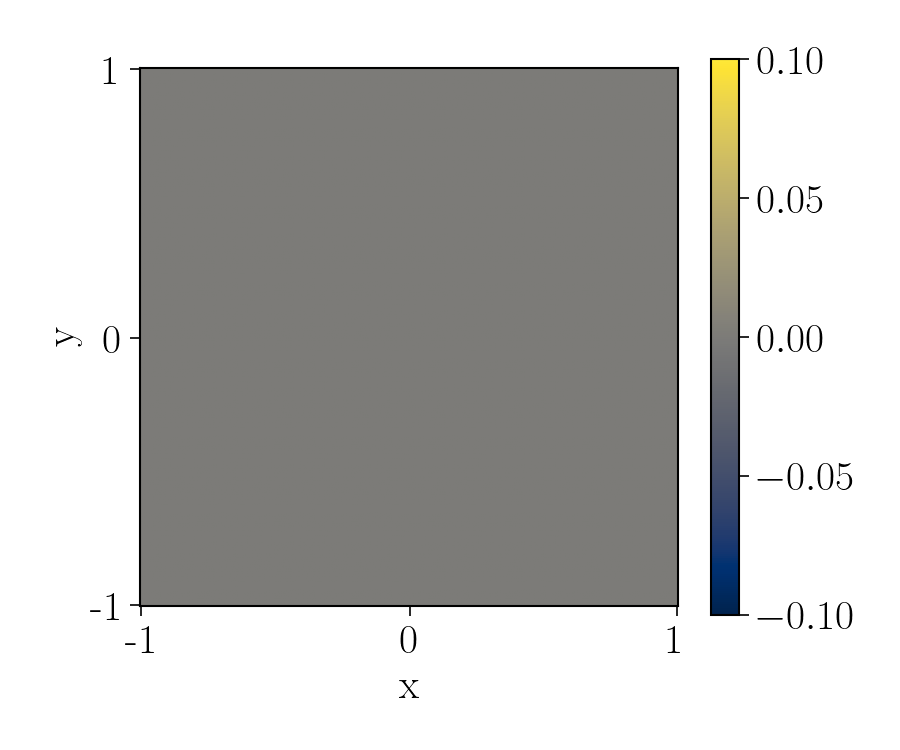}
        \caption{$h(x)$}
    \end{subfigure}
    \begin{subfigure}[b]{0.16\linewidth}
        \includegraphics[width=\linewidth]{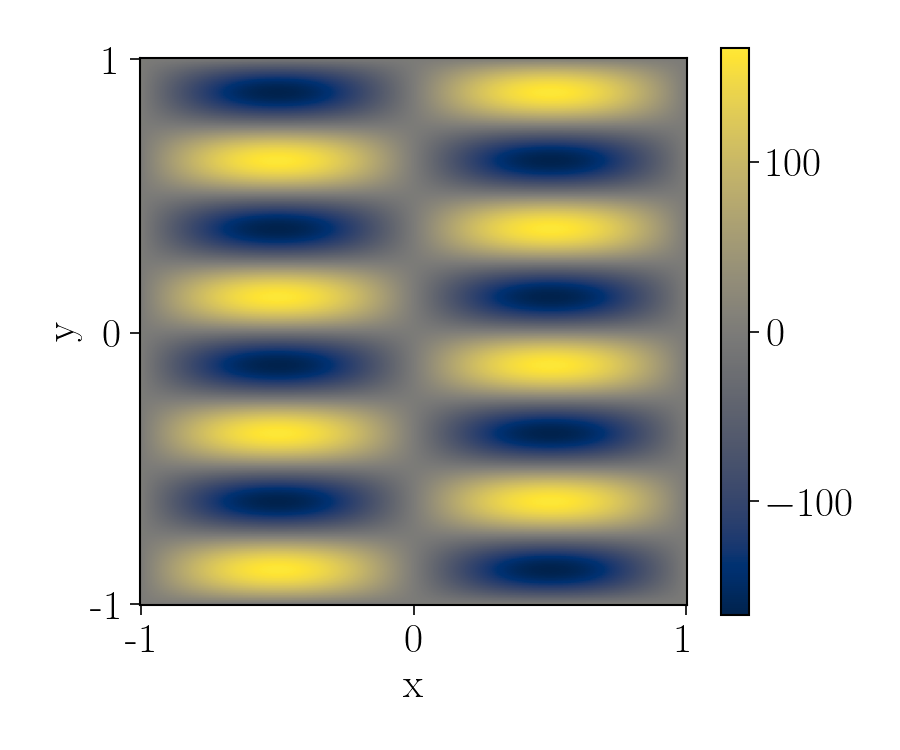}
        \caption{$f(x)$}
    \end{subfigure}
    \begin{subfigure}[b]{0.16\linewidth}
        \includegraphics[width=\linewidth]{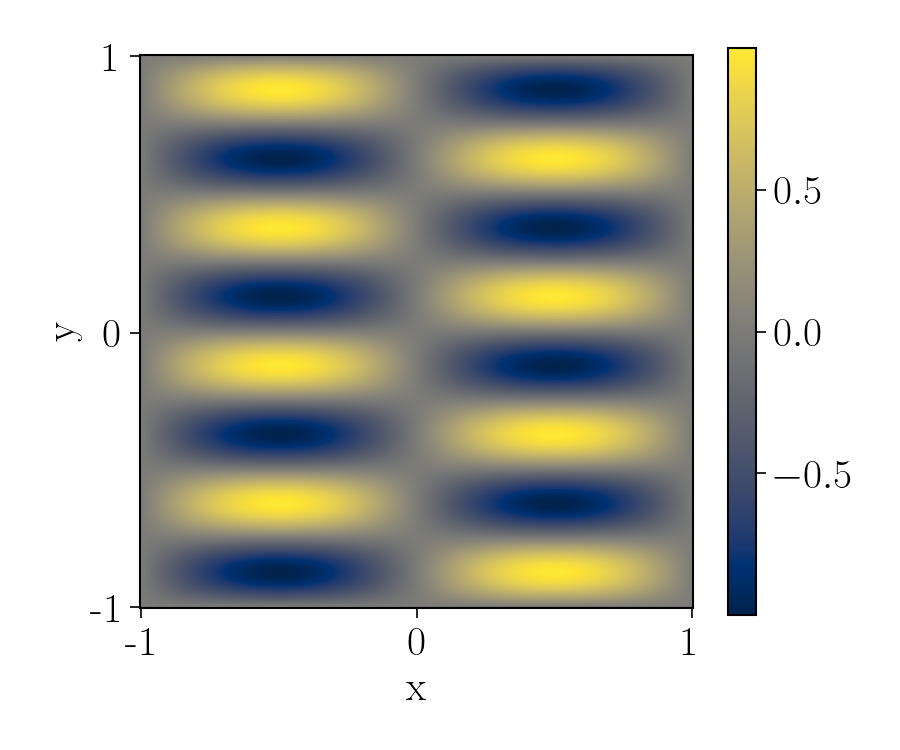}
        \caption{$u(x)$}
    \end{subfigure}

    \caption{Parameterization of the 2D Helmholtz PDE.}
    \label{fig:helmholtz_parametrization}
\end{figure}

This benchmark problem is adapted from DeepXDE~\cite{deepxde}.
Figure~\ref{fig:helmholtz_parametrization} shows the parameterization of the different PDE components.

\subsection{Helmholtz 2D - Complex Geometry (HZ-G)}
\label{app:poisson_b}
Consider the two-dimensional Poisson–Boltzmann (Helmholtz-type) equation:
\begin{equation}
\label{eq:pb_equation}
-\Delta u(x, y) + k^2 u(x, y) = f(x, y), \quad (x, y) \in \Omega,
\end{equation}
where the domain \(\Omega = [-1, 1]^2 \setminus \Omega_{\text{circle}}\) consists of the square \([-1,1]^2\) with four circular regions removed.

The source term is defined as:
\begin{equation}
\label{eq:pb_source}
f(x_1, x_2) = A \cdot \mu_2 x_2 \sin(\mu_1 \pi x_1) \sin(\mu_2 \pi x_2),
\end{equation}
with parameters \(\mu_1 = 1\), \(\mu_2 = 4\), \(k = 8\), and \(A = 10\).

Dirichlet boundary conditions are imposed as:
\begin{equation}
\label{eq:pb_bc}
u(x, y) = 
\begin{cases}
0.2, & (x, y) \in \partial \Omega_{\text{rec}}, \\
1.0, & (x, y) \in \partial \Omega_{\text{circle}},
\end{cases}
\end{equation}
where \(\partial \Omega_{\text{rec}}\) denotes the outer rectangular boundary, and \(\partial \Omega_{\text{circle}} = \cup_{i=1}^{4} \partial R_i\) are the boundaries of the interior circles.

The circles defining the removed interior regions are given by:
\begin{align*}
R_1 &= \{(x, y) : (x - 0.5)^2 + (y - 0.5)^2 \leq 0.2^2\}, \\
R_2 &= \{(x, y) : (x - 0.4)^2 + (y + 0.4)^2 \leq 0.4^2\}, \\
R_3 &= \{(x, y) : (x + 0.2)^2 + (y + 0.7)^2 \leq 0.1^2\}, \\
R_4 &= \{(x, y) : (x + 0.6)^2 + (y - 0.5)^2 \leq 0.3^2\}.
\end{align*}

\begin{figure}[H]
    \centering
    \begin{subfigure}[b]{0.16\linewidth}
        \includegraphics[width=\linewidth]{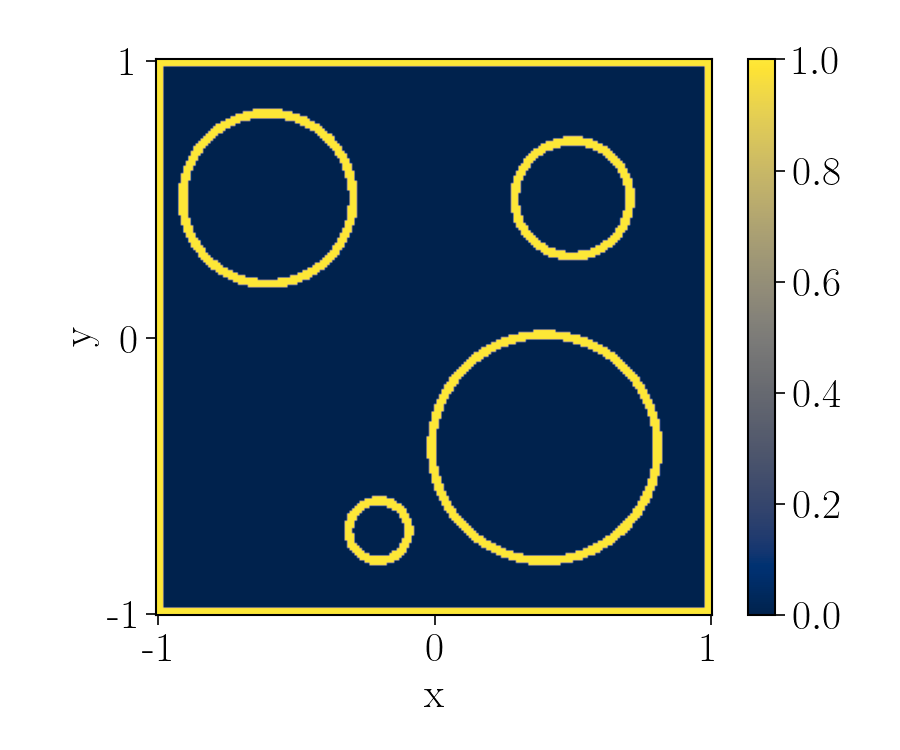}
        \caption{$\partial \Omega_D$}
    \end{subfigure}
    \begin{subfigure}[b]{0.16\linewidth}
        \includegraphics[width=\linewidth]{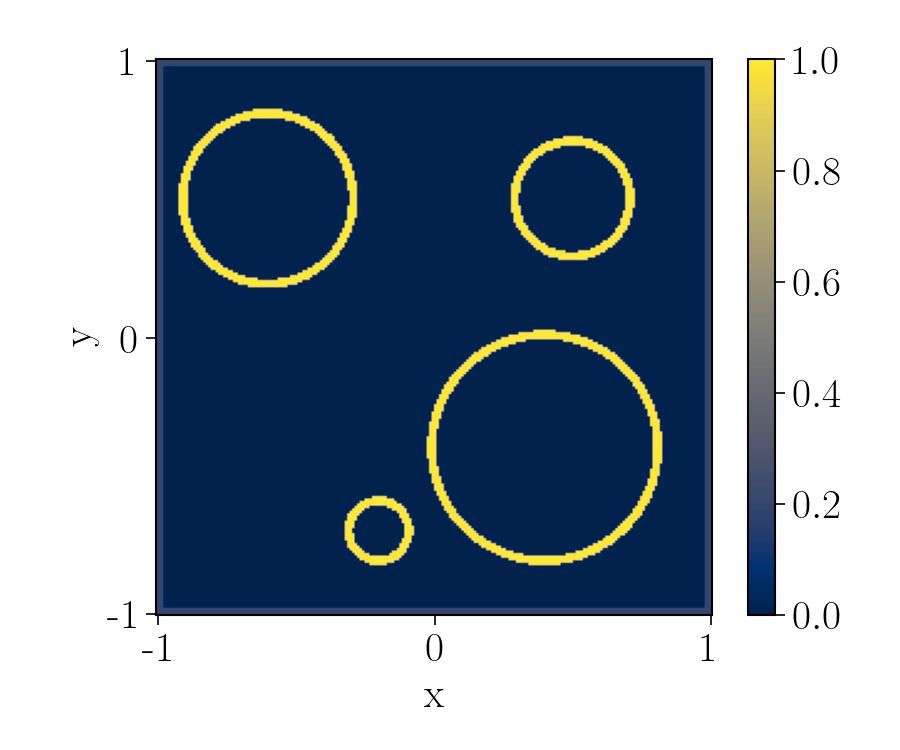}
        \caption{$g(x)$}
    \end{subfigure}
    \begin{subfigure}[b]{0.16\linewidth}
        \includegraphics[width=\linewidth]{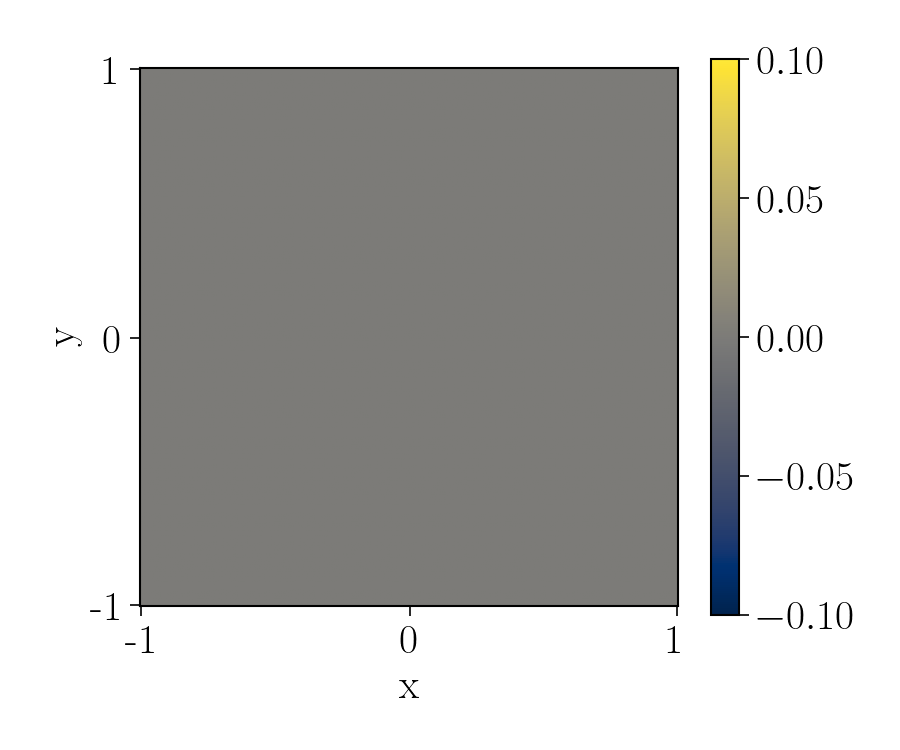}
        \caption{$\partial \Omega_N$}
    \end{subfigure}
    \begin{subfigure}[b]{0.16\linewidth}
        \includegraphics[width=\linewidth]{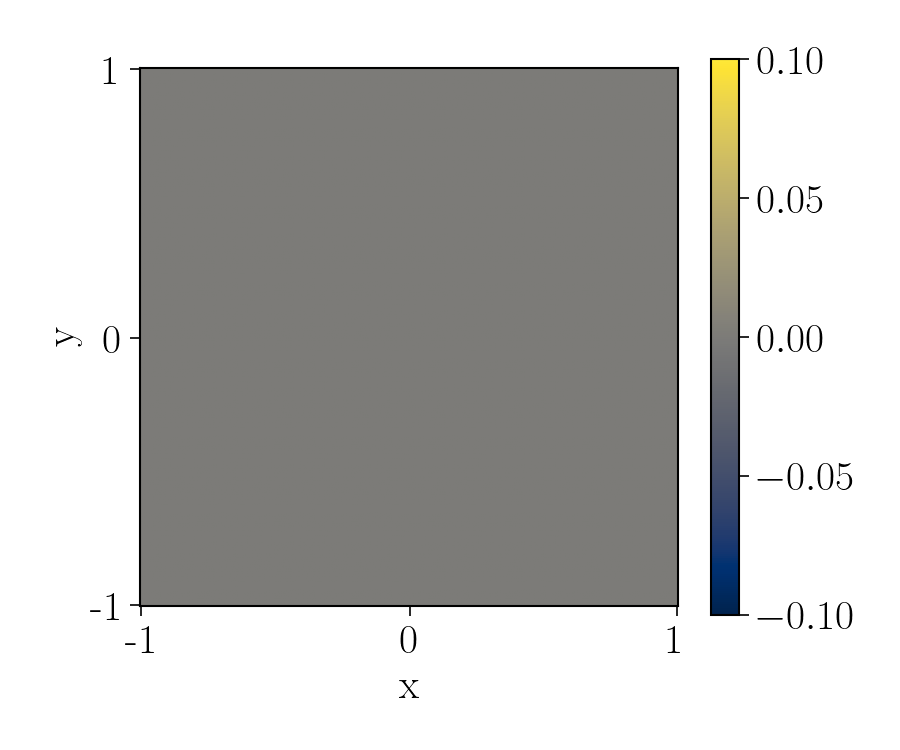}
        \caption{$h(x)$}
    \end{subfigure}
    \begin{subfigure}[b]{0.16\linewidth}
        \includegraphics[width=\linewidth]{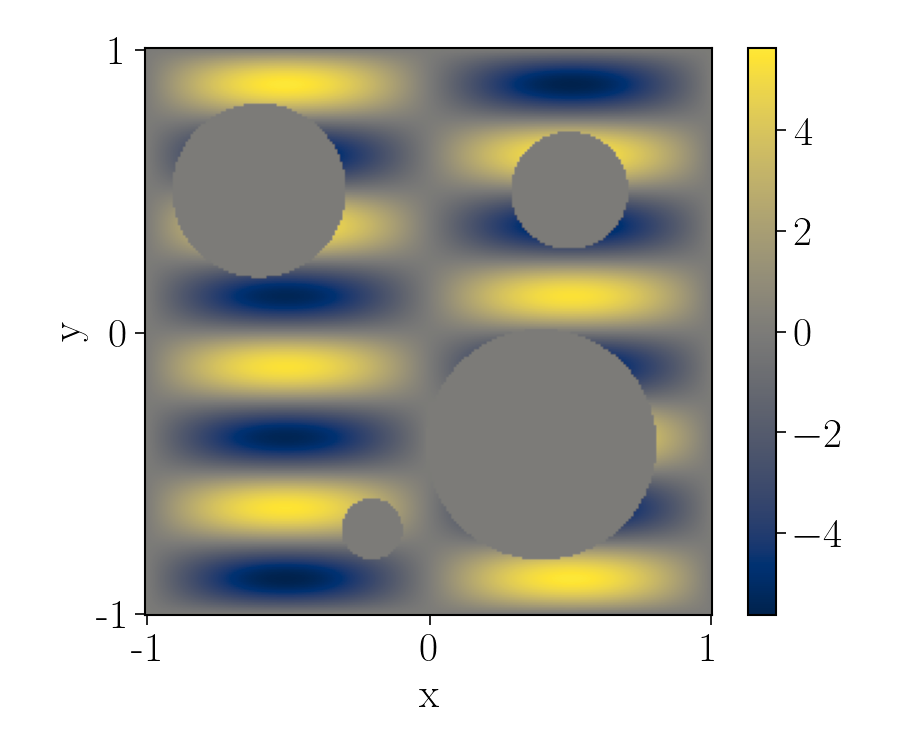}
        \caption{$f(x)$}
    \end{subfigure}
    \begin{subfigure}[b]{0.16\linewidth}
        \includegraphics[width=\linewidth]{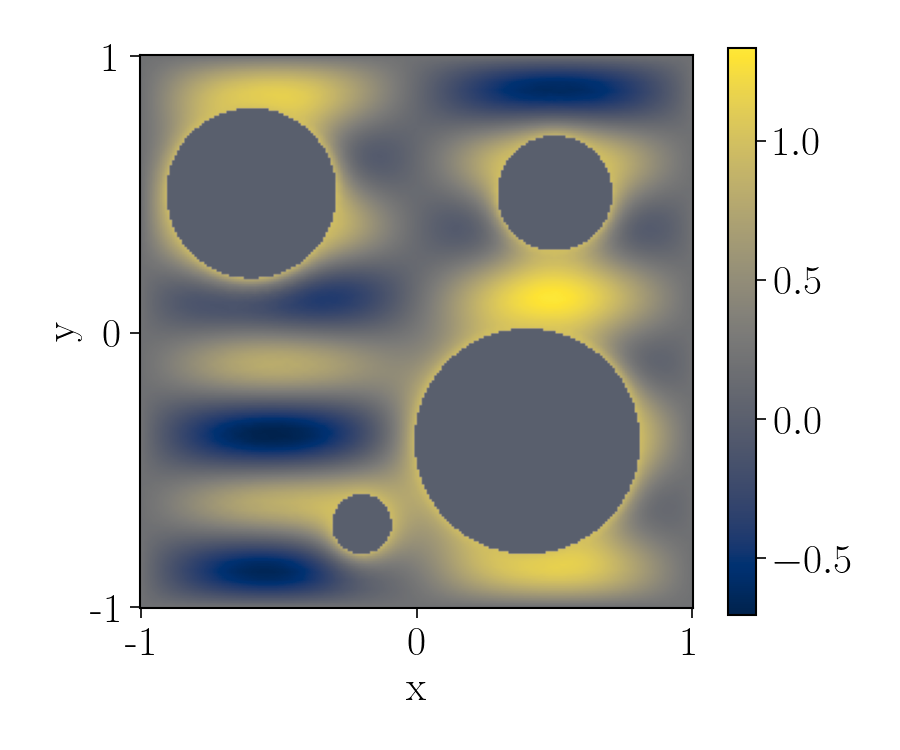}
        \caption{$u(x)$}
    \end{subfigure}

    \caption{Parameterization of the 2D Helmholtz-type (Poisson–Boltzmann) PDE with complex geometry.}
    \label{fig:poisson_B_parametrization}
\end{figure}

This benchmark problem is adapted from PINNacle~\cite{pinnacle}. Figure~\ref{fig:poisson_B_parametrization} shows the parameterization of the different PDE components.

\subsection{Poisson 2D - Circles (PS-C)}
\label{app:poisson_c}

Consider the two-dimensional Poisson equation:
\begin{equation}
\label{eq:poisson_eq}
-\Delta u(x, y) = 0, \quad (x, y) \in \Omega,
\end{equation}
where the domain is defined as a rectangle with four interior circular exclusions:
\[
\Omega = \Omega_{\text{rec}} \setminus \bigcup_{i=1}^4 R_i,
\quad \text{with} \quad \Omega_{\text{rec}} = [-0.5, 0.5]^2,
\]
and the circular regions \( R_i \) given by:
\begin{equation}
\label{eq:poisson_circles}
\begin{aligned}
R_1 &= \left\{(x, y) \;\middle|\; (x - 0.3)^2 + (y - 0.3)^2 \leq 0.1^2 \right\}, \\
R_2 &= \left\{(x, y) \;\middle|\; (x + 0.3)^2 + (y - 0.3)^2 \leq 0.1^2 \right\}, \\
R_3 &= \left\{(x, y) \;\middle|\; (x - 0.3)^2 + (y + 0.3)^2 \leq 0.1^2 \right\}, \\
R_4 &= \left\{(x, y) \;\middle|\; (x + 0.3)^2 + (y + 0.3)^2 \leq 0.1^2 \right\}.
\end{aligned}
\end{equation}

Dirichlet boundary conditions are applied as follows:
\begin{equation}
\label{eq:poisson_bc}
u(x, y) = 
\begin{cases}
0, & (x, y) \in \partial R_i, \\
1, & (x, y) \in \partial \Omega_{\text{rec}}.
\end{cases}
\end{equation}

\begin{figure}[H]
    \centering
    \begin{subfigure}[b]{0.16\linewidth}
        \includegraphics[width=\linewidth]{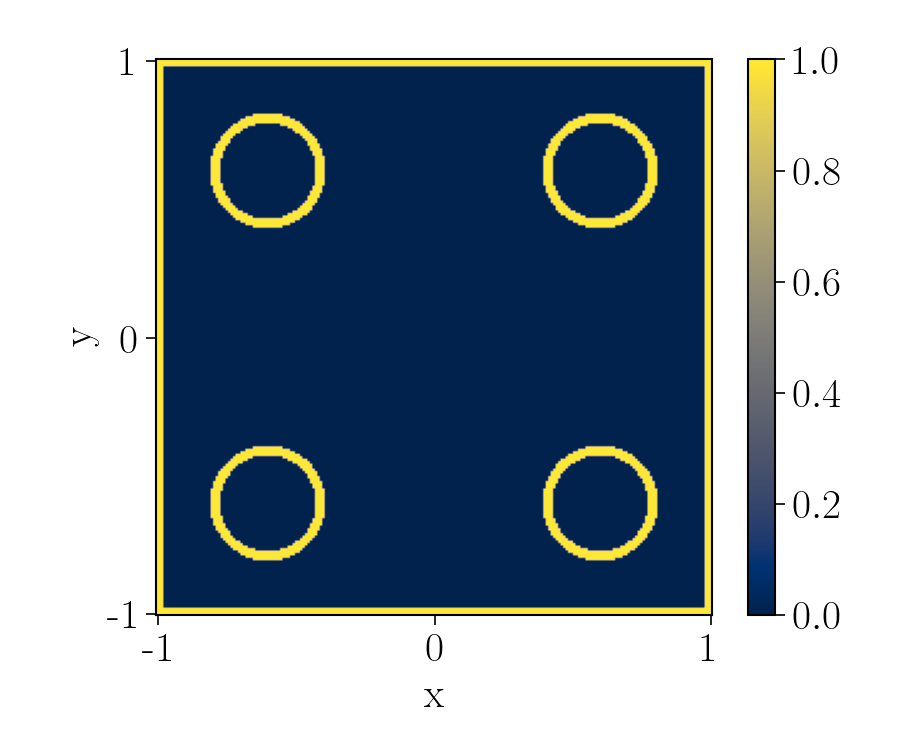}
        \caption{$\partial \Omega_D$}
    \end{subfigure}
    \begin{subfigure}[b]{0.16\linewidth}
        \includegraphics[width=\linewidth]{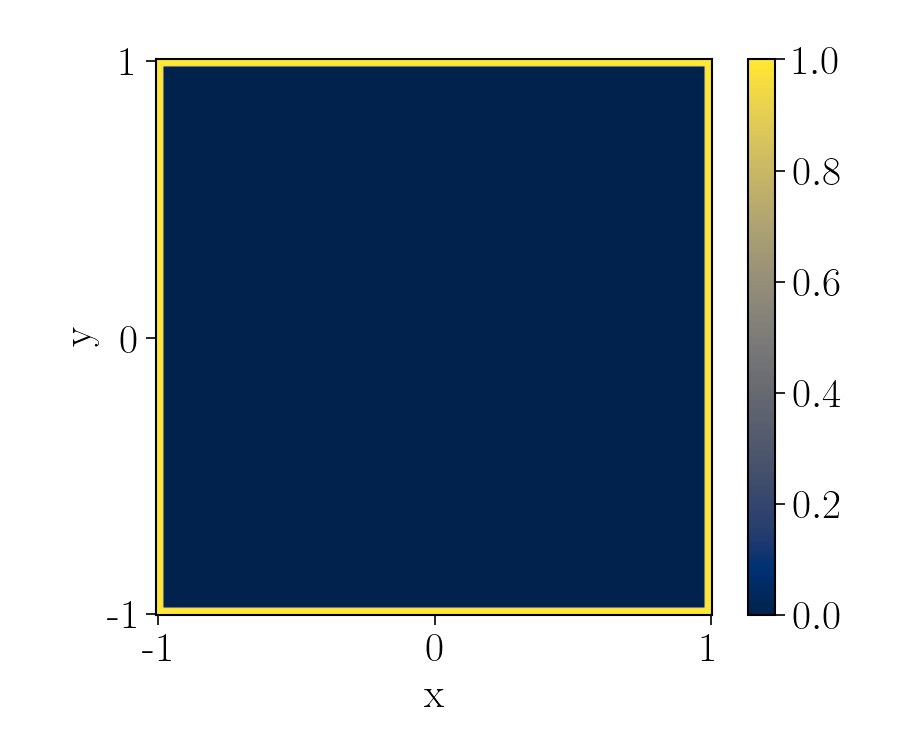}
        \caption{$g(x)$}
    \end{subfigure}
    \begin{subfigure}[b]{0.16\linewidth}
        \includegraphics[width=\linewidth]{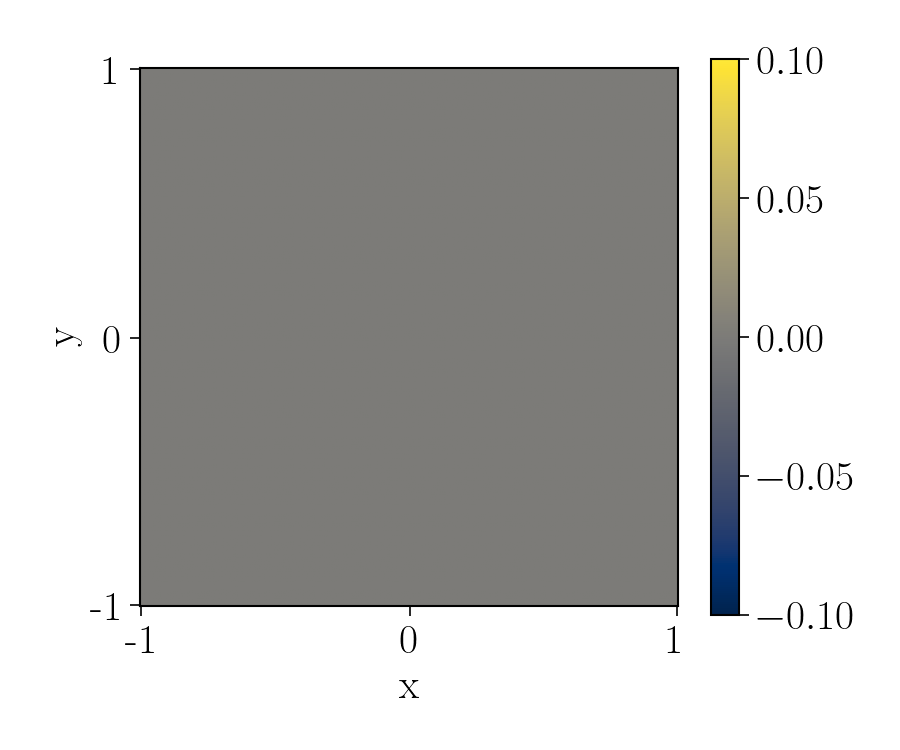}
        \caption{$\partial \Omega_N$}
    \end{subfigure}
    \begin{subfigure}[b]{0.16\linewidth}
        \includegraphics[width=\linewidth]{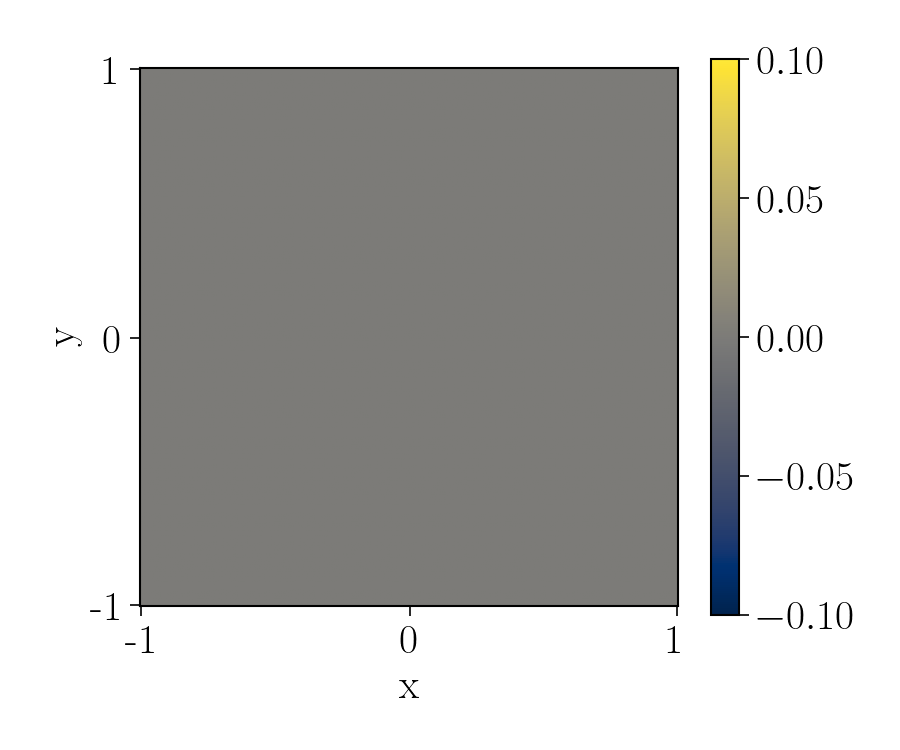}
        \caption{$h(x)$}
    \end{subfigure}
    \begin{subfigure}[b]{0.16\linewidth}
        \includegraphics[width=\linewidth]{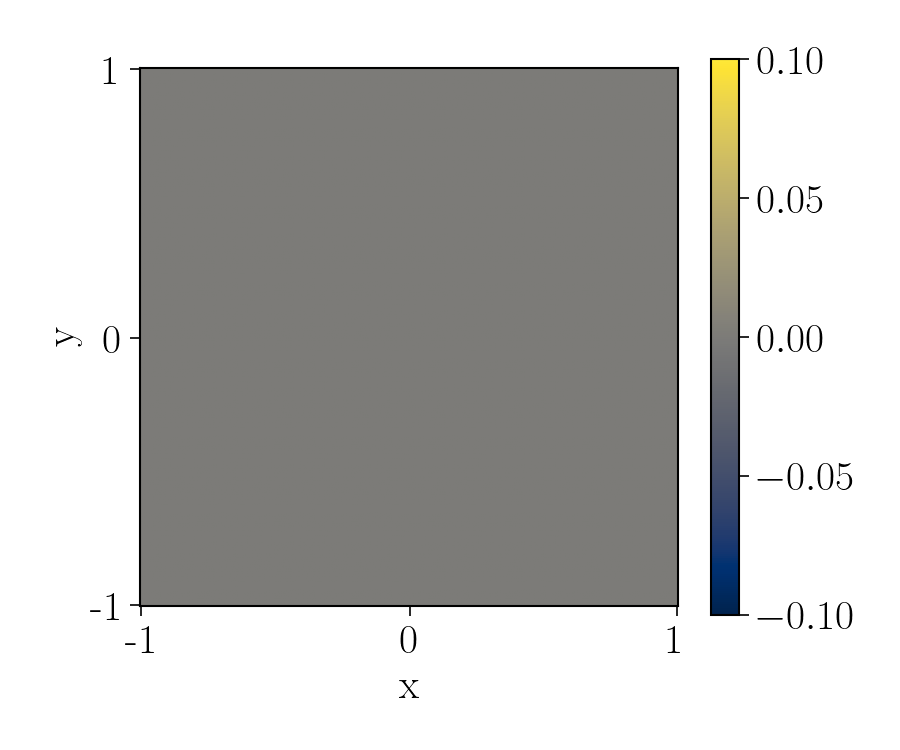}
        \caption{$f(x)$}
    \end{subfigure}
    \begin{subfigure}[b]{0.16\linewidth}
        \includegraphics[width=\linewidth]{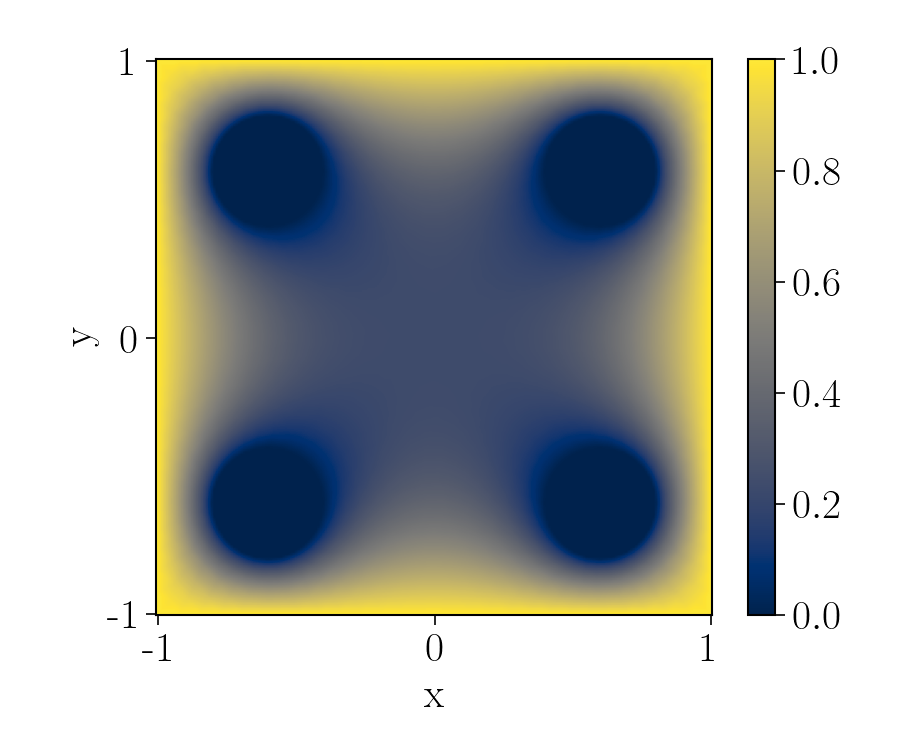}
        \caption{$u(x)$}
    \end{subfigure}

    \caption{Parameterization of the 2D Poisson PDE with circular inner boundaries.}
    \label{fig:poisson_C_parametrization}
\end{figure}

This benchmark problem is adapted from PINNacle~\cite{pinnacle}. Figure~\ref{fig:poisson_C_parametrization} shows the parameterization of the different PDE components.

\subsection{Poisson 2D - L-Domain (PS-L)}
\label{app:poisson-l}

Consider the two-dimensional Poisson equation:
\begin{equation}
\label{eq:poisson_ld}
- u_{xx} - u_{yy} = 1, \quad (x, y) \in \Omega,
\end{equation}
where the domain is an L-shaped region:
\[
\Omega = [-1, 1]^2 \setminus [0, 1]^2.
\]

Dirichlet boundary conditions are applied as:
\begin{equation}
\label{eq:poisson_ld_bc}
u(x, y) = 0, \quad (x, y) \in \partial \Omega.
\end{equation}

\begin{figure}[H]
    \centering
    \begin{subfigure}[b]{0.16\linewidth}
        \includegraphics[width=\linewidth]{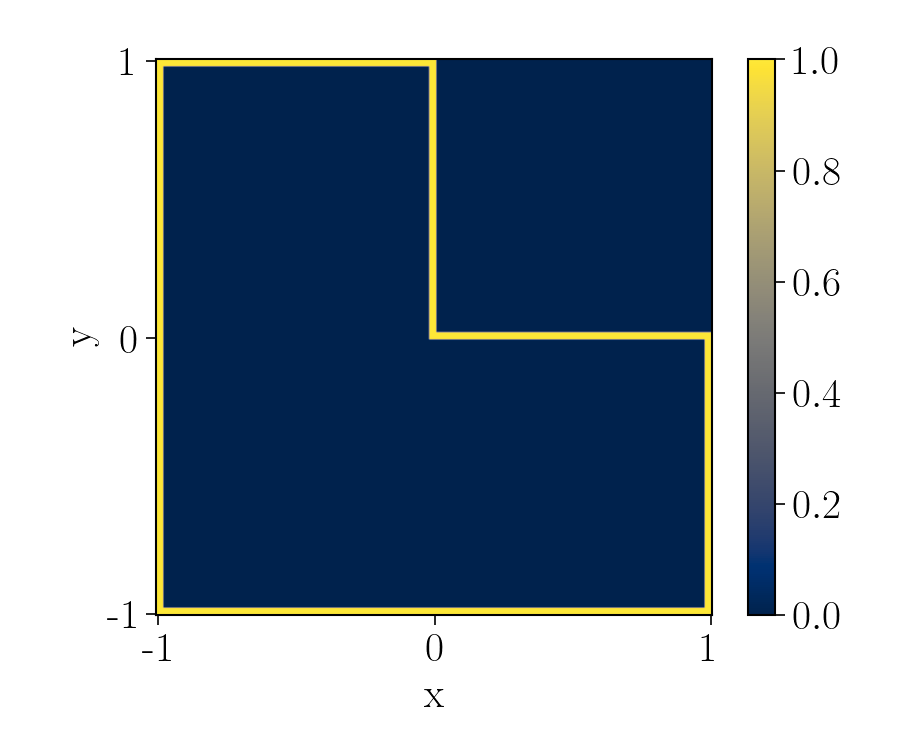}
        \caption{$\partial \Omega_D$}
    \end{subfigure}
    \begin{subfigure}[b]{0.16\linewidth}
        \includegraphics[width=\linewidth]{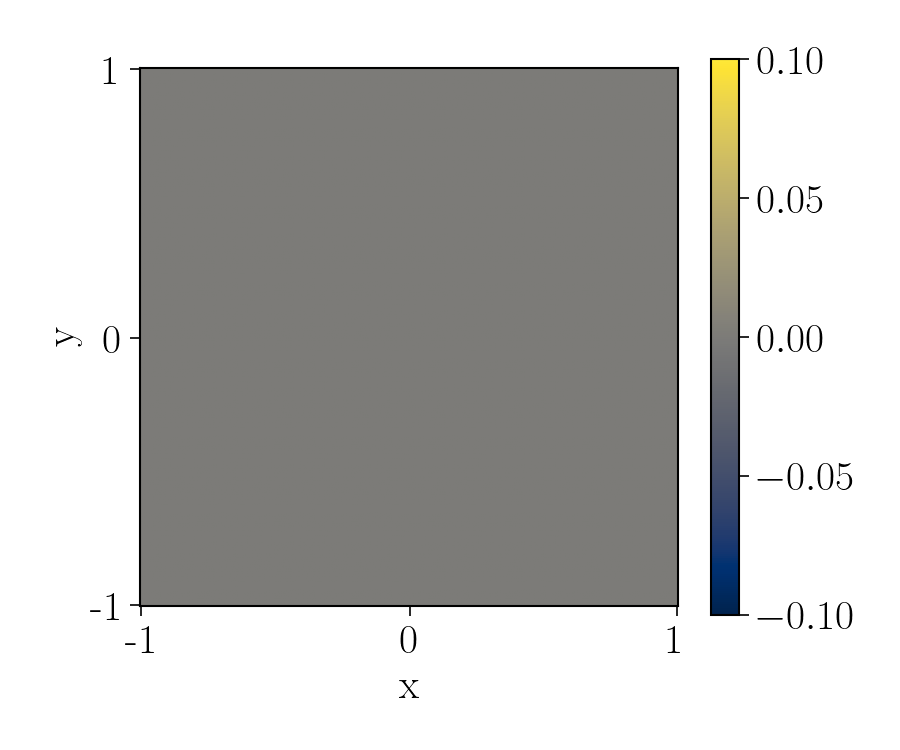}
        \caption{$g(x)$}
    \end{subfigure}
    \begin{subfigure}[b]{0.16\linewidth}
        \includegraphics[width=\linewidth]{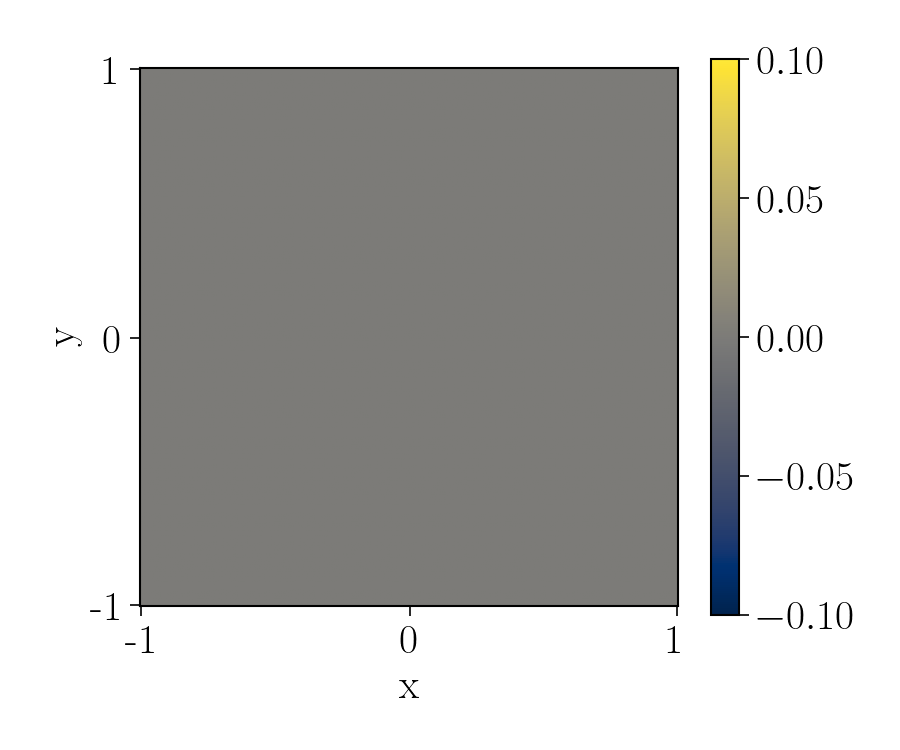}
        \caption{$\partial \Omega_N$}
    \end{subfigure}
    \begin{subfigure}[b]{0.16\linewidth}
        \includegraphics[width=\linewidth]{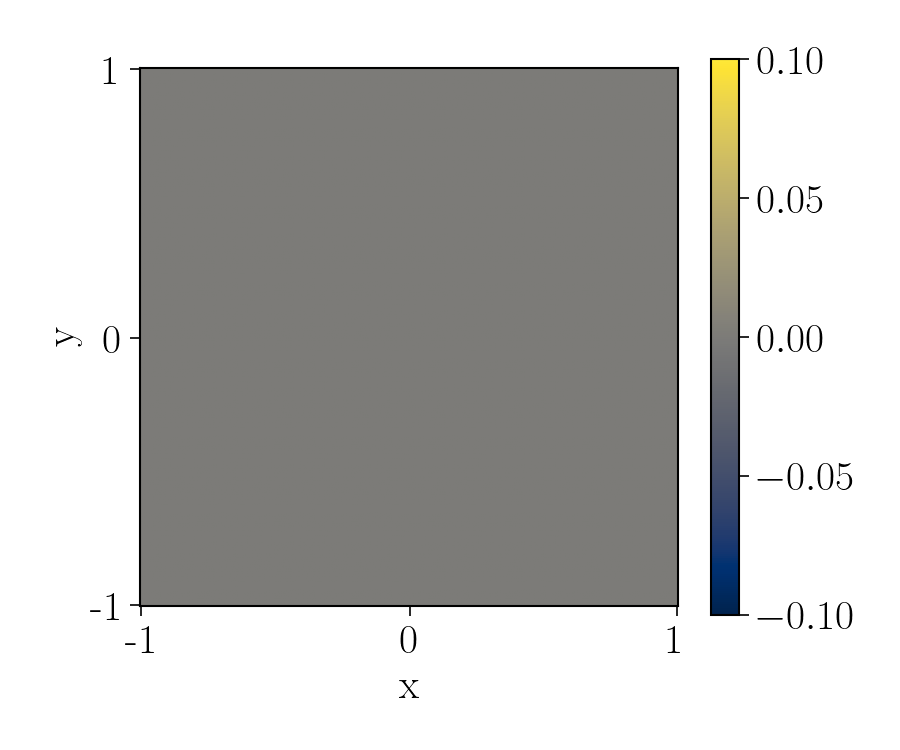}
        \caption{$h(x)$}
    \end{subfigure}
    \begin{subfigure}[b]{0.16\linewidth}
        \includegraphics[width=\linewidth]{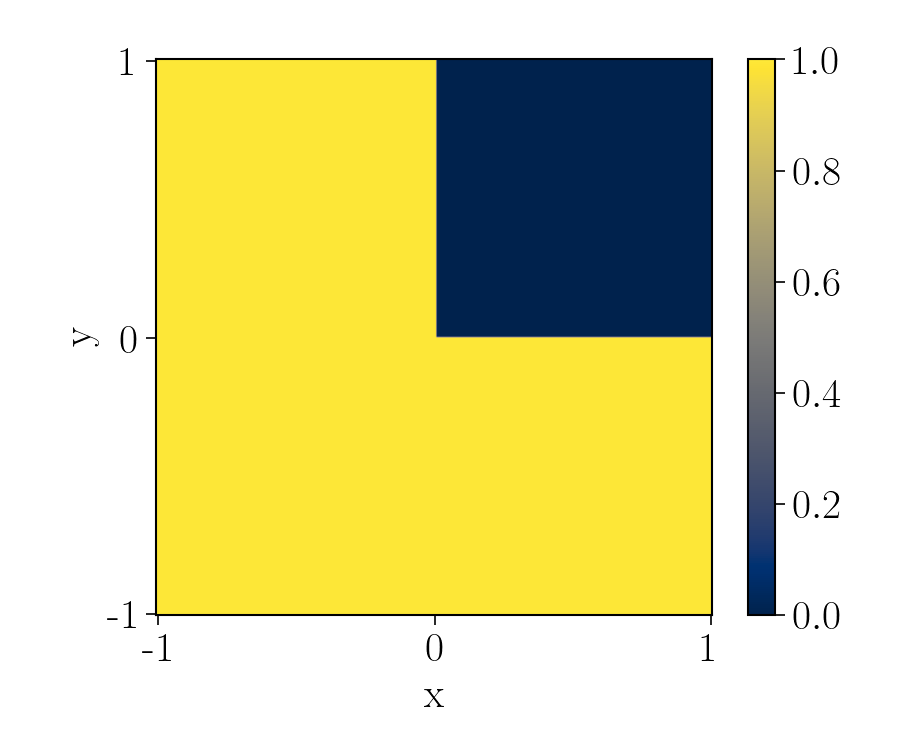}
        \caption{$f(x)$}
    \end{subfigure}
    \begin{subfigure}[b]{0.16\linewidth}
        \includegraphics[width=\linewidth]{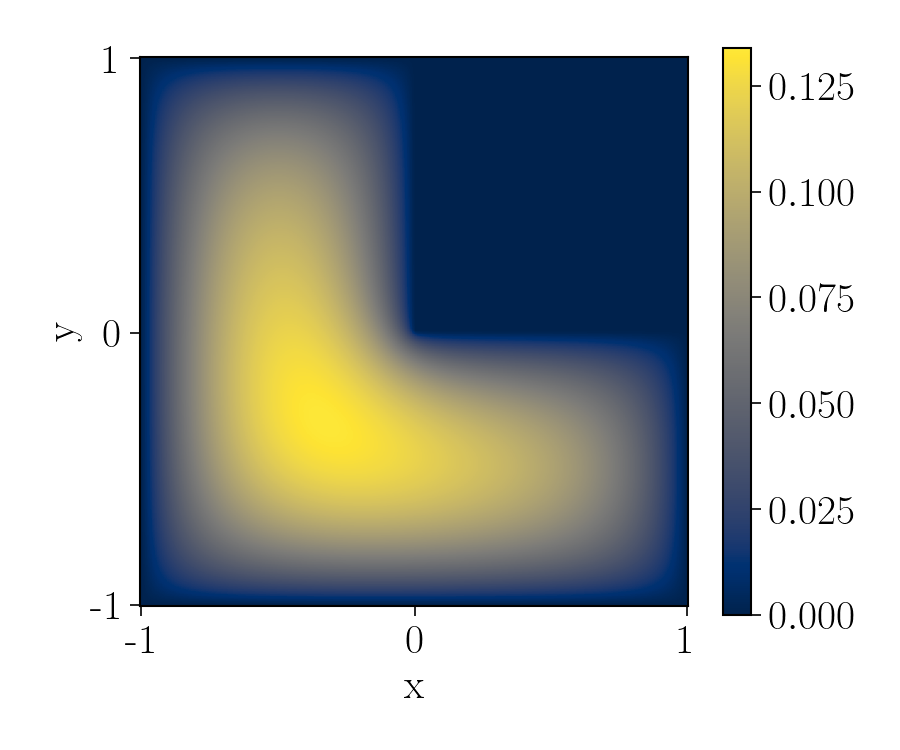}
        \caption{$u(x)$}
    \end{subfigure}

    \caption{Parameterization of the 2D Poisson PDE on an L-shaped domain.}
    \label{fig:poisson_L_parametrization}
\end{figure}

This benchmark problem is adapted from DeepXDE~\cite{deepxde}.
Figure~\ref{fig:poisson_L_parametrization} shows the parameterization of the different PDE components.

\subsection{Poisson 2D - Gauss (PS-G)}
\label{app:poisson-g}

Consider the two-dimensional Poisson equation:
\begin{equation}
\label{eq:poisson}
- \Delta u(x, y) = f(x, y), \quad (x, y) \in (0, 1)^2,
\end{equation}
with homogeneous Dirichlet boundary conditions:
\begin{equation}
u(x, y) = 0, \quad (x, y) \in \partial \Omega.
\end{equation}

The source term \( f \) is defined as a superposition of a random number \( N \) of Gaussian functions:
\begin{equation}
\label{eq:poisson_source}
f(x, y) = \sum_{i=1}^{N} \exp\left( -\frac{(x - \mu_{x,i})^2 + (y - \mu_{y,i})^2}{2\sigma_i^2} \right),
\end{equation}
where \( N \sim \text{Geom}(0.4) \), \( \mu_{x,i}, \mu_{y,i} \sim \mathcal{U}[0, 1] \), and \( \sigma_i \sim \mathcal{U}[0.025, 0.1] \).

\begin{figure}[H]
    \centering
    \begin{subfigure}[b]{0.16\linewidth}
        \includegraphics[width=\linewidth]{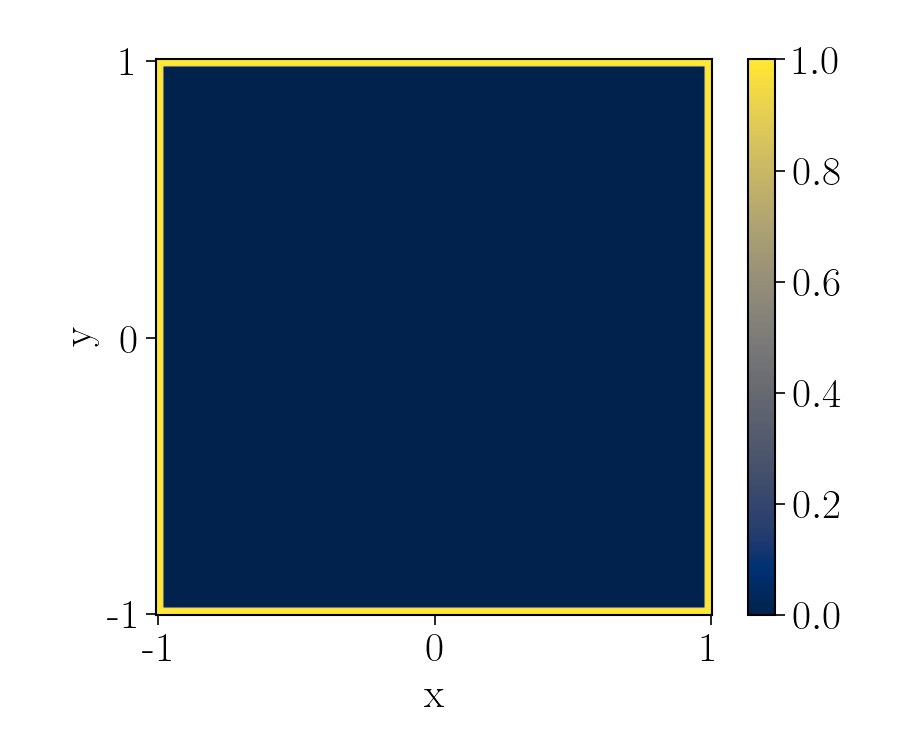}
        \caption{$\partial \Omega_D$}
    \end{subfigure}
    \begin{subfigure}[b]{0.16\linewidth}
        \includegraphics[width=\linewidth]{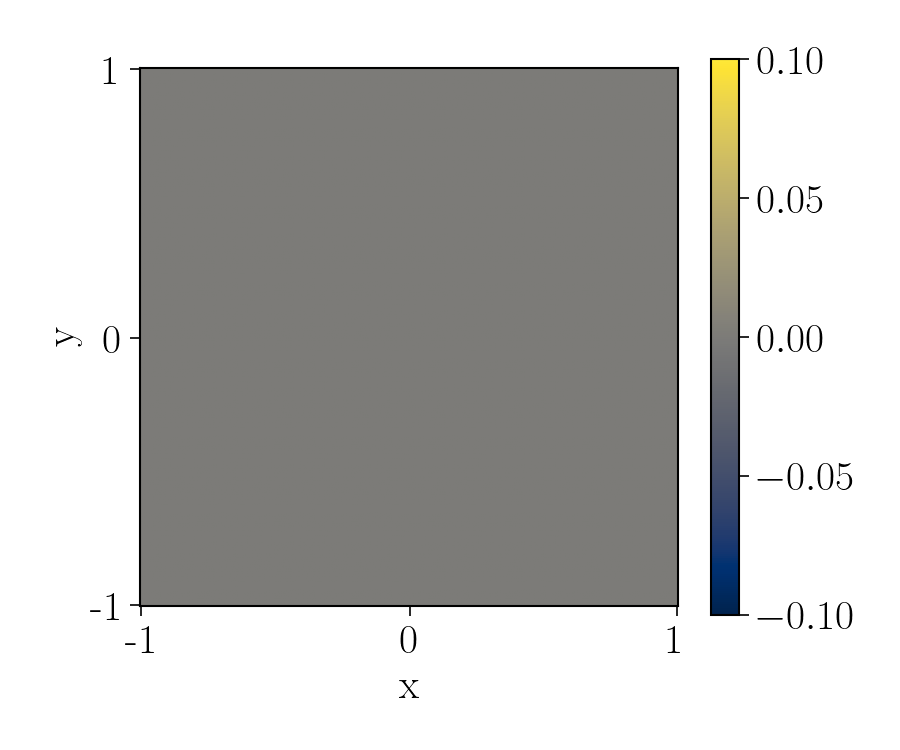}
        \caption{$g(x)$}
    \end{subfigure}
    \begin{subfigure}[b]{0.16\linewidth}
        \includegraphics[width=\linewidth]{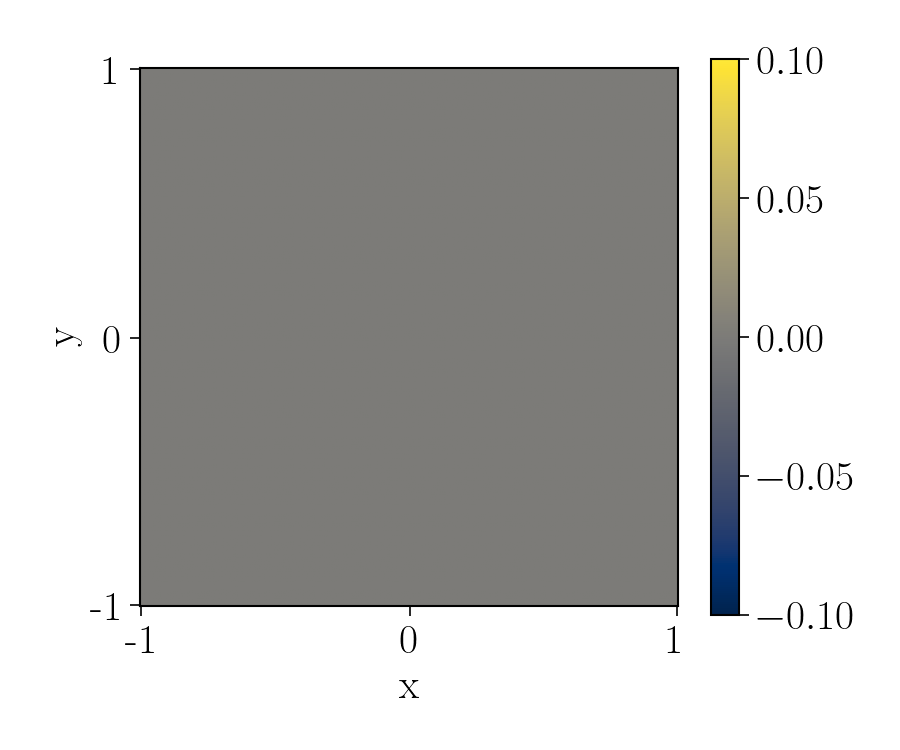}
        \caption{$\partial \Omega_N$}
    \end{subfigure}
    \begin{subfigure}[b]{0.16\linewidth}
        \includegraphics[width=\linewidth]{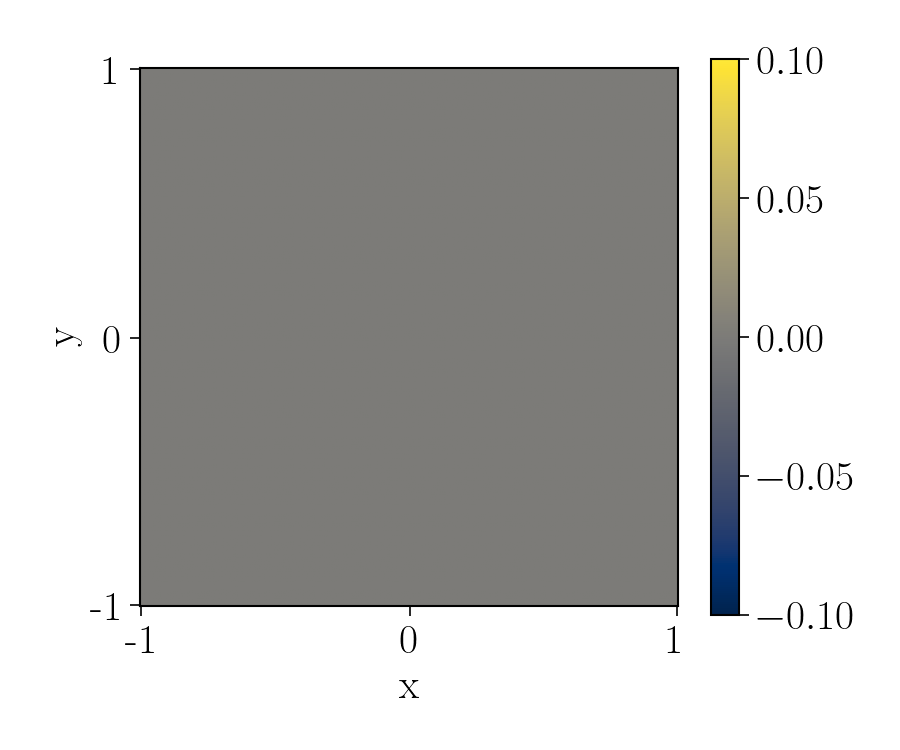}
        \caption{$h(x)$}
    \end{subfigure}
    \begin{subfigure}[b]{0.16\linewidth}
        \includegraphics[width=\linewidth]{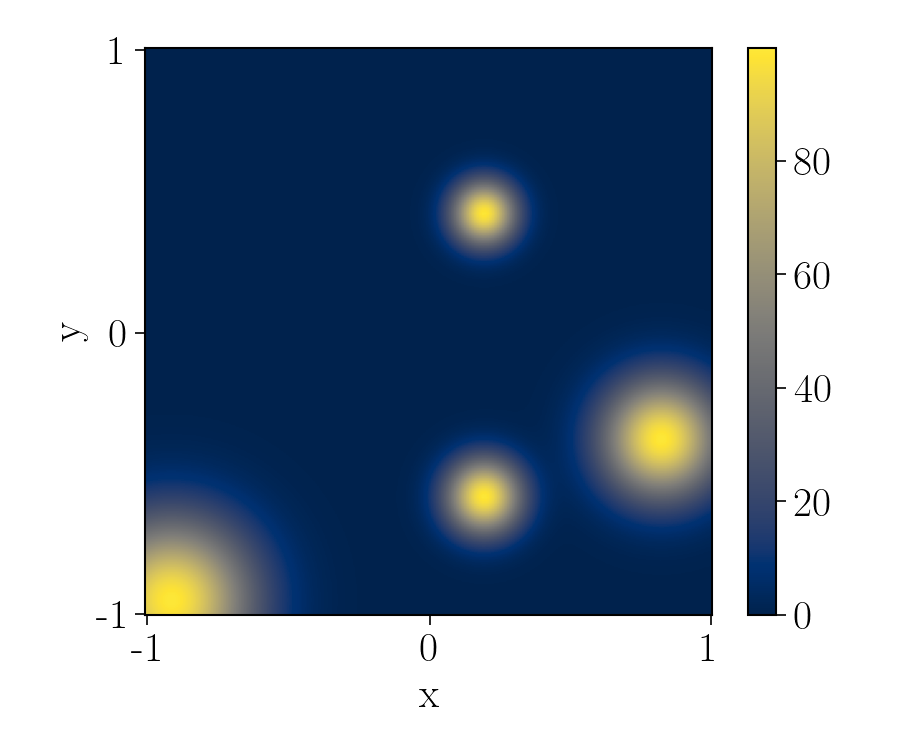}
        \caption{$f(x)$}
    \end{subfigure}
    \begin{subfigure}[b]{0.16\linewidth}
        \includegraphics[width=\linewidth]{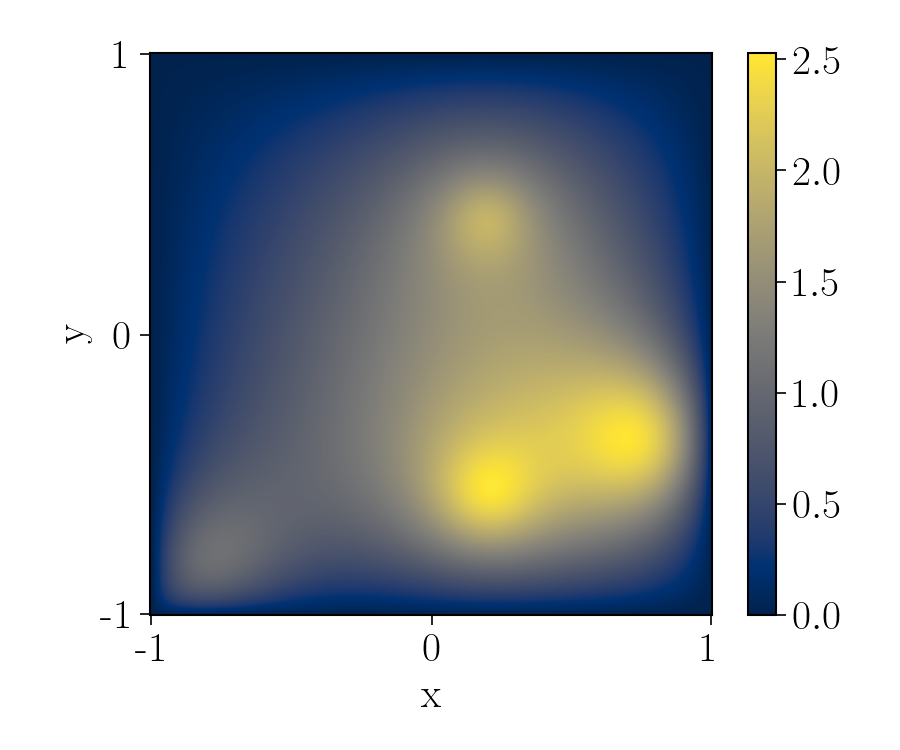}
        \caption{$u(x)$}
    \end{subfigure}

    \caption{Parameterization of the 2D Poisson PDE with Gaussian superposition vorticity field.}
    \label{fig:poisson_G_parametrization}
\end{figure}

We select a sample from the dataset introduced in~\cite{pde_foundation_model_2024_Herde}. Figure~\ref{fig:poisson_G_parametrization} shows the parameterization of the different PDE components.

\subsection{Wave 1D (WV)}
\label{app:wave}

Consider the one-dimensional wave equation:
\begin{equation}
\label{eq:wave_equation}
\frac{\partial^2 u}{\partial t^2} - 4 \frac{\partial^2 u}{\partial x^2} = 0,
\quad (x, t) \in [0, 1] \times [0, 1].
\end{equation}

Dirichlet boundary conditions are imposed as:
\begin{equation}
\label{eq:wave_bc}
u(0, t) = u(1, t) = 0.
\end{equation}

The initial conditions are given by:
\begin{equation}
\label{eq:wave_ic1}
u(x, 0) = \sin(\pi x) + \frac{1}{2} \sin(4\pi x),
\end{equation}
\begin{equation}
\label{eq:wave_ic2}
\frac{\partial u}{\partial t}(x, 0) = 0.
\end{equation}

The corresponding exact solution is:
\begin{equation}
\label{eq:wave_exact}
u(x, t) = \sin(\pi x) \cos(2\pi t) + \frac{1}{2} \sin(4\pi x) \cos(8\pi t).
\end{equation}

\begin{figure}[H]
    \centering
    \begin{subfigure}[b]{0.16\linewidth}
        \includegraphics[width=\linewidth]{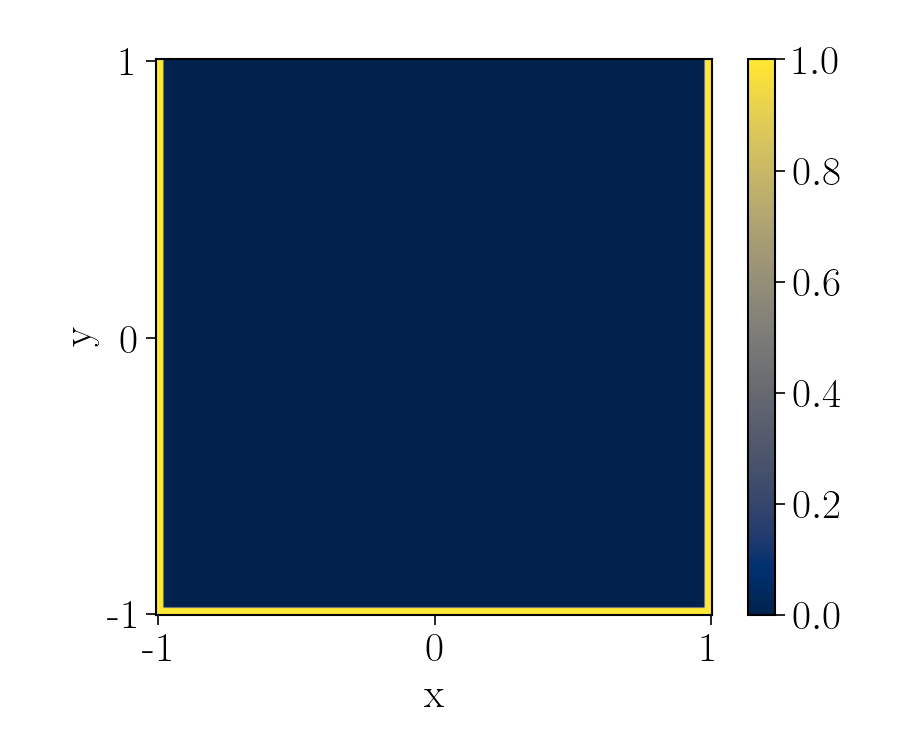}
        \caption{$\partial \Omega_D$}
    \end{subfigure}
    \begin{subfigure}[b]{0.16\linewidth}
        \includegraphics[width=\linewidth]{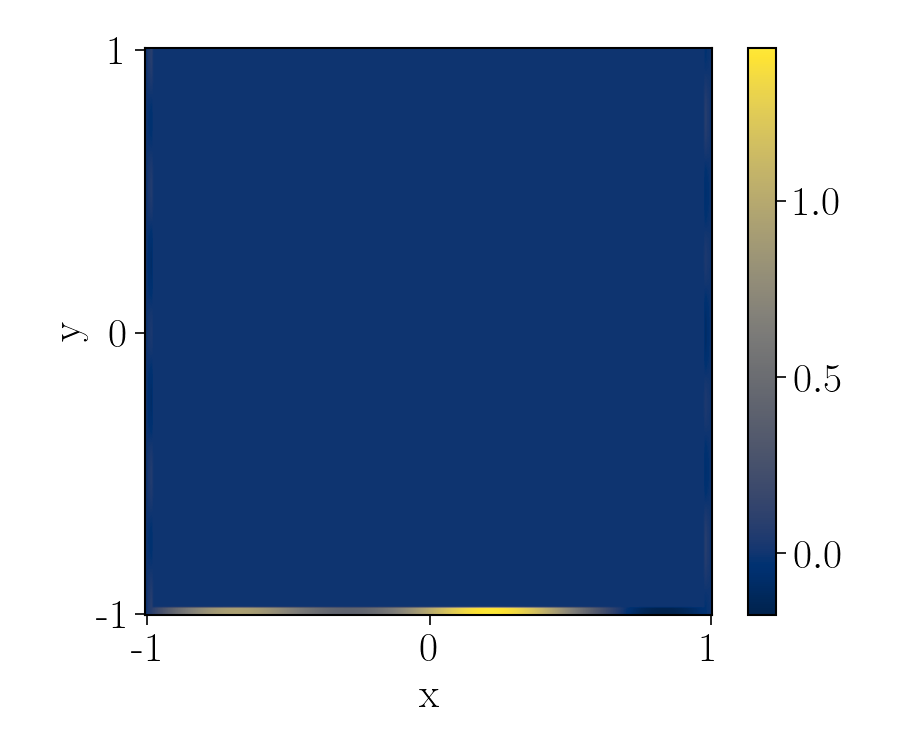}
        \caption{$g(x)$}
    \end{subfigure}
    \begin{subfigure}[b]{0.16\linewidth}
        \includegraphics[width=\linewidth]{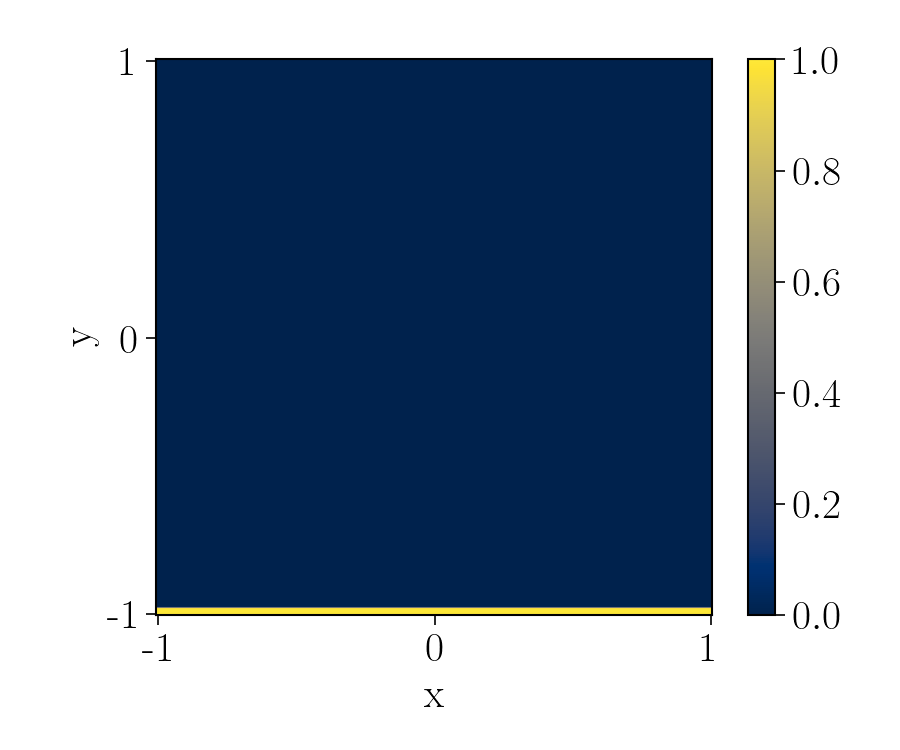}
        \caption{$\partial \Omega_N$}
    \end{subfigure}
    \begin{subfigure}[b]{0.16\linewidth}
        \includegraphics[width=\linewidth]{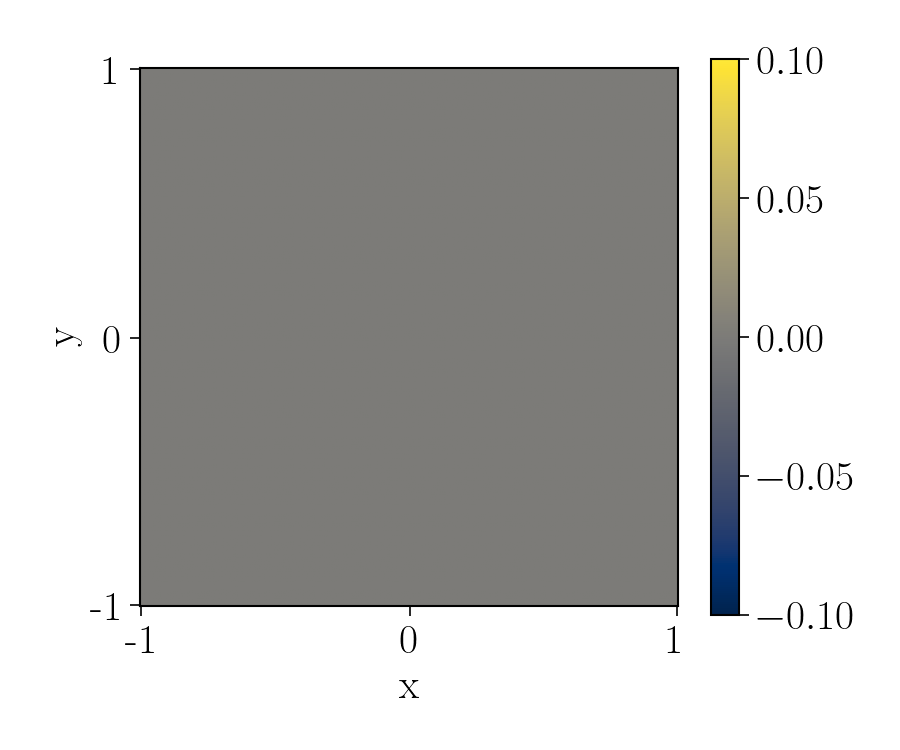}
        \caption{$h(x)$}
    \end{subfigure}
    \begin{subfigure}[b]{0.16\linewidth}
        \includegraphics[width=\linewidth]{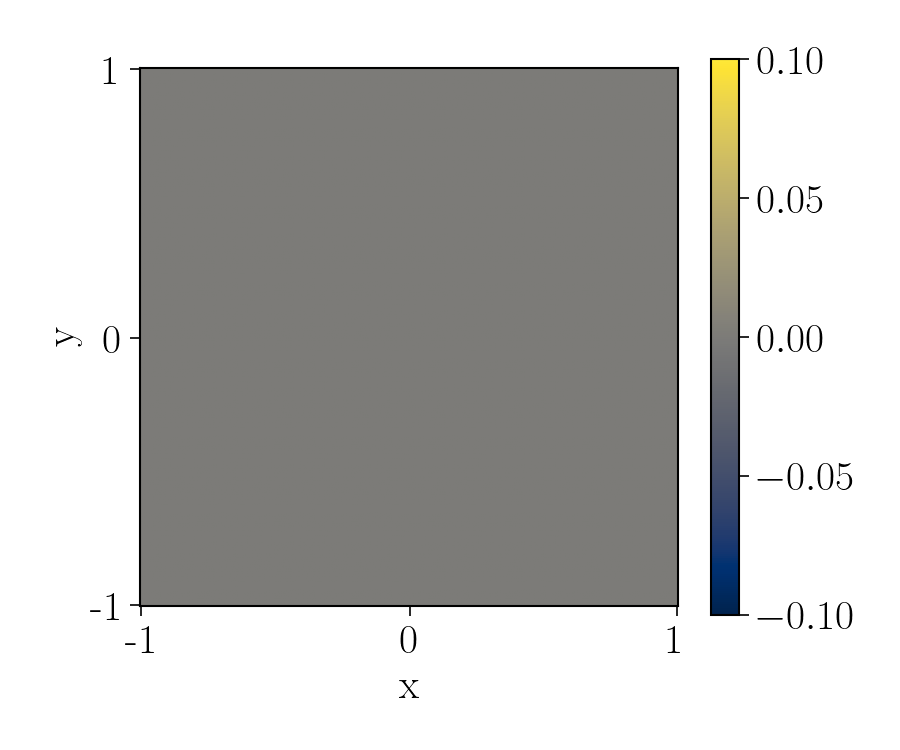}
        \caption{$f(x)$}
    \end{subfigure}
    \begin{subfigure}[b]{0.16\linewidth}
        \includegraphics[width=\linewidth]{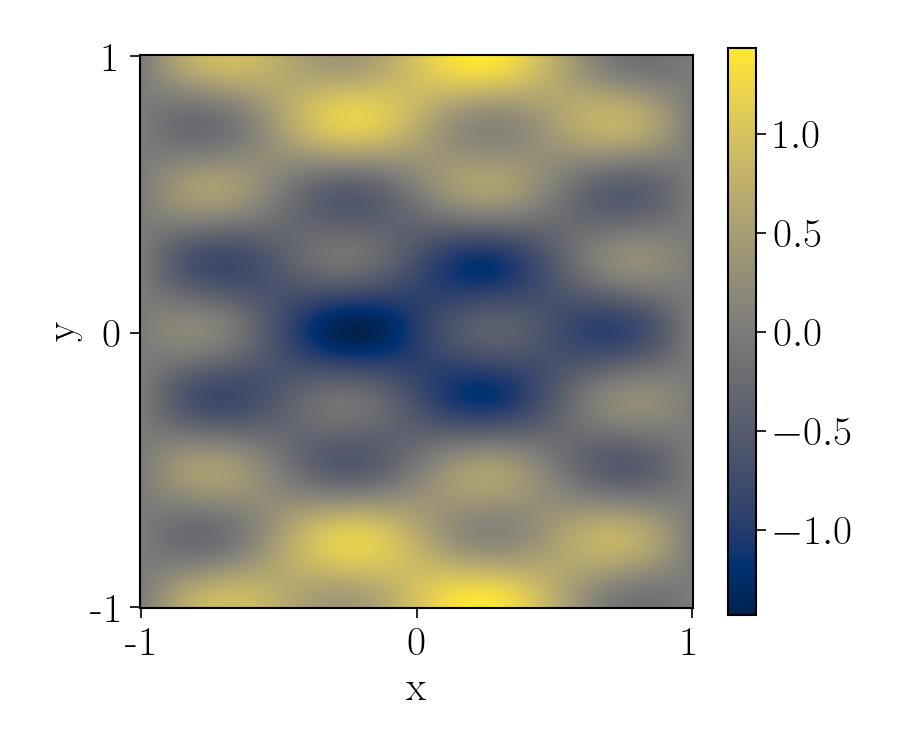}
        \caption{$u(x)$}
    \end{subfigure}

    \caption{Parameterization of the 1D Wave PDE.}
    \label{fig:wave_parametrization}
\end{figure}

This benchmark problem is adapted from PINNacle~\cite{pinnacle}. 
Figure~\ref{fig:wave_parametrization} shows the parameterization of the different PDE components.

\subsection{Baseline Models}
\label{app:baselines}

We compare our method against three baselines. All models are trained for 30{,}000 batches with a batch size of 128 and an initial learning rate of $10^{-4}$.

\textbf{U-Net}~\cite{unet}. A convolutional encoder–decoder network that shares the same input encoding architecture as HyPINO, but replaces the transformer-based decoder with a purely convolutional upsampling stack that directly outputs a solution grid of shape $(224 \times 224)$, matching the resolution of the input tensors. It is trained exclusively on supervised PDEs with analytical solutions, using a batch size of 128, an initial learning rate of $10^{-4}$, and for 30{,}000 training batches. The U-Net has a total parameter count of 62M.

\textbf{Poseidon}~\cite{pde_foundation_model_2024_Herde}. A large pretrained operator network with approximately 158M parameters. We use the Poseidon-B checkpoint and adapt it by changing the input dimensionality to 5 to accept all grid-based inputs. Additionally, the lead-time-conditioned layer normalization layers—originally designed to condition on a 1D time input—are modified to condition on the 5D vector of differential operator coefficients. Poseidon is fine-tuned exclusively on supervised samples, using the same training setup as the U-Net (30{,}000 batches, batch size 128, initial learning rate $10^{-4}$).

\textbf{PINO}~\cite{pino}. A Fourier neural operator~\cite{fno} architecture with 33M parameters, trained with joint physics-informed and supervised losses computed in Fourier space. We adapt the model to accept 5-channel grid inputs and condition on the PDE operator using FiLM layers. It follows the same hybrid supervision and training curriculum as HyPINO, including physics-informed losses, and is trained for 30{,}000 batches with a batch size of 128 and an initial learning rate of $10^{-4}$.

\subsection{Evaluation}
\label{app:evaluation}

\begin{figure}[H]
    \centering

    \begin{subfigure}[b]{0.19\linewidth}
        \includegraphics[width=\linewidth]{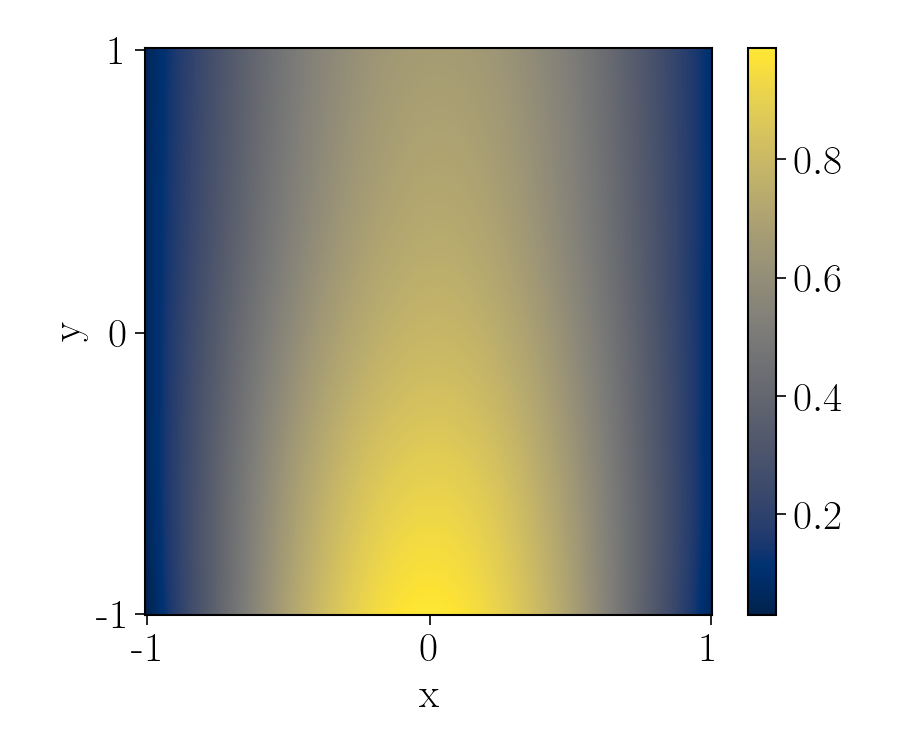}
    \end{subfigure}
    \begin{subfigure}[b]{0.19\linewidth}
        \includegraphics[width=\linewidth]{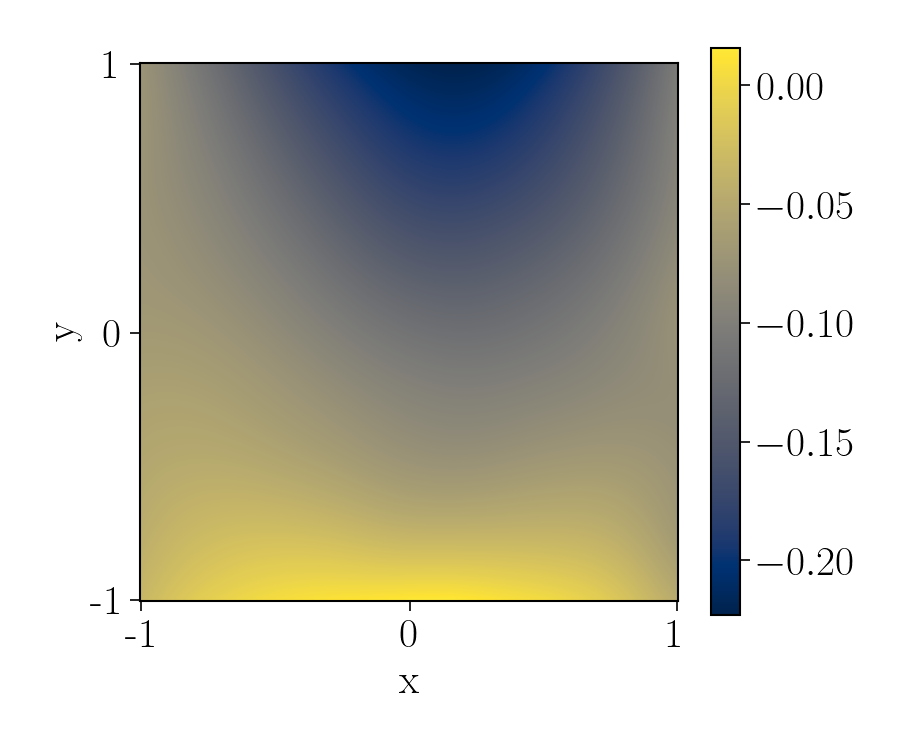}
    \end{subfigure}
    \begin{subfigure}[b]{0.19\linewidth}
        \includegraphics[width=\linewidth]{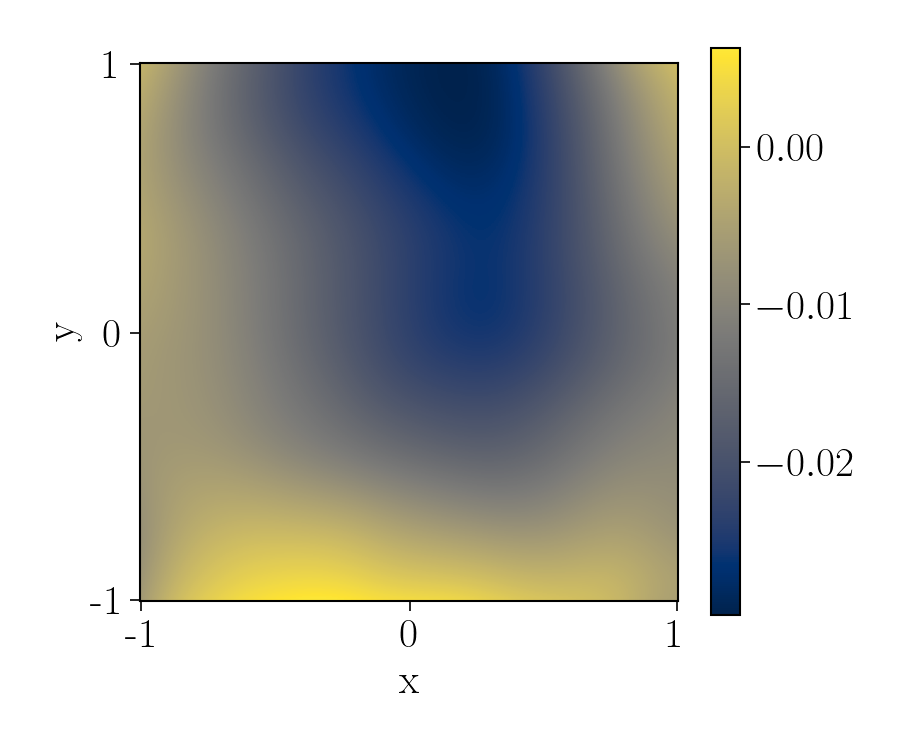}
    \end{subfigure}
    \begin{subfigure}[b]{0.19\linewidth}
        \includegraphics[width=\linewidth]{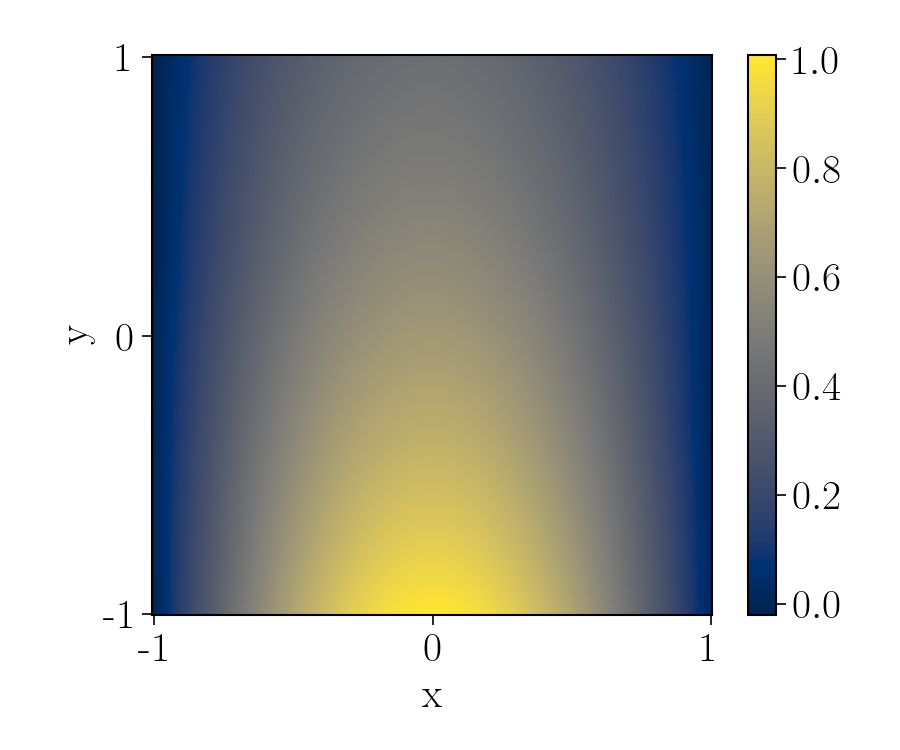}
    \end{subfigure}
    \begin{subfigure}[b]{0.19\linewidth}
        \includegraphics[width=\linewidth]{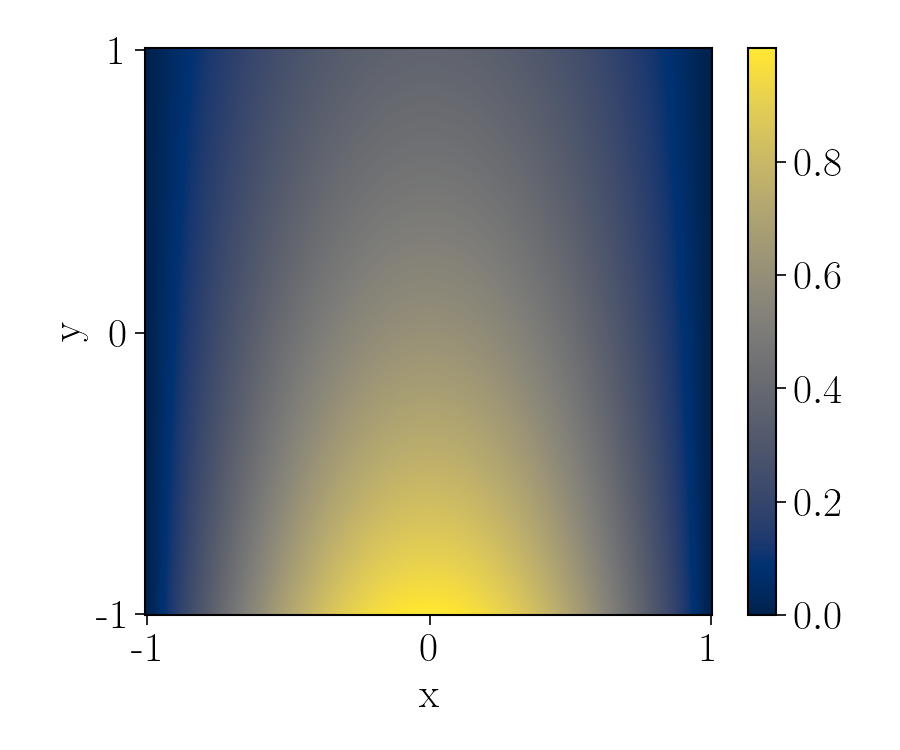}
    \end{subfigure}

    \vspace{1em}

    \begin{subfigure}[b]{0.19\linewidth}
        \includegraphics[width=\linewidth]{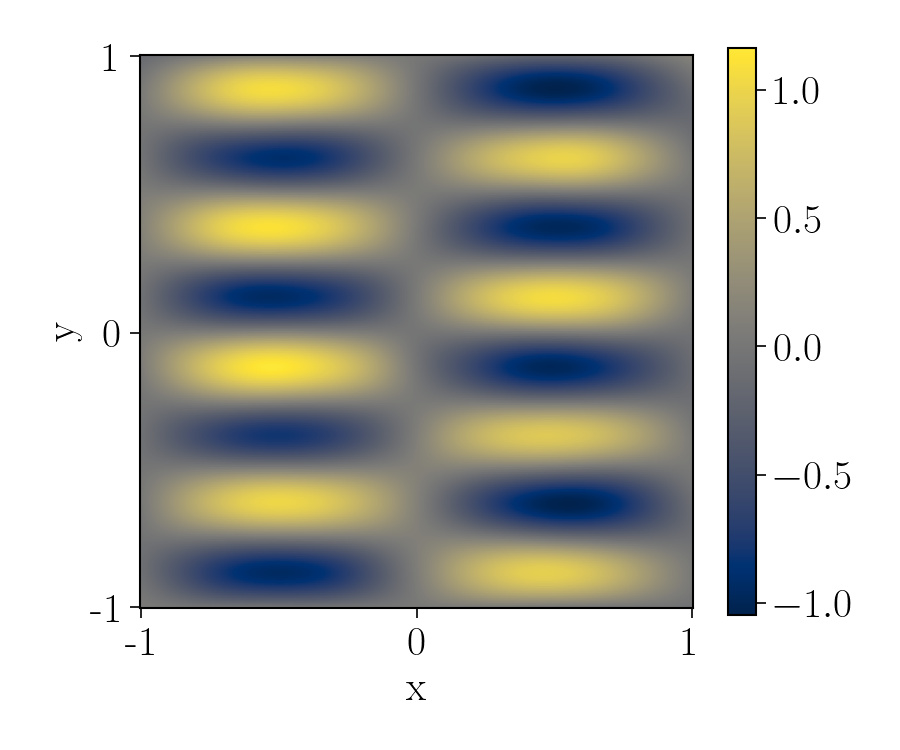}
    \end{subfigure}
    \begin{subfigure}[b]{0.19\linewidth}
        \includegraphics[width=\linewidth]{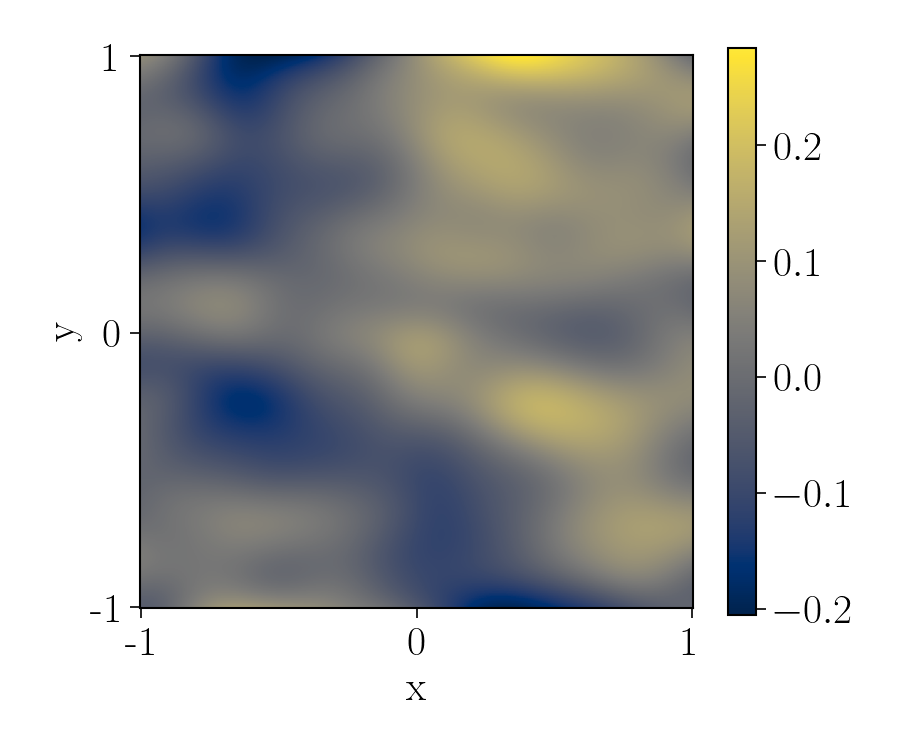}
    \end{subfigure}
    \begin{subfigure}[b]{0.19\linewidth}
        \includegraphics[width=\linewidth]{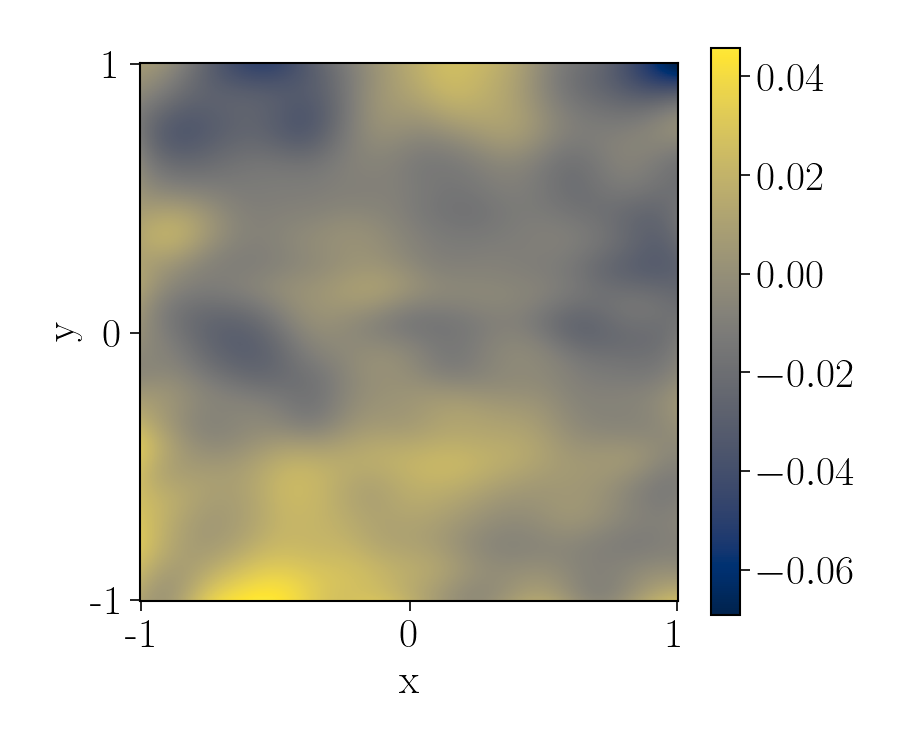}
    \end{subfigure}
    \begin{subfigure}[b]{0.19\linewidth}
        \includegraphics[width=\linewidth]{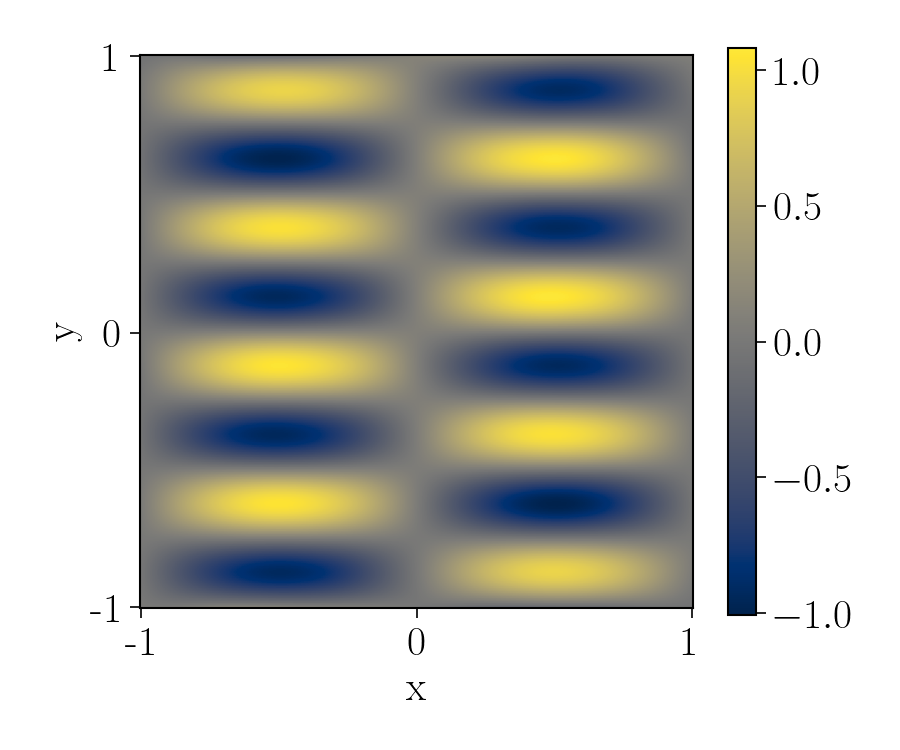}
    \end{subfigure}
    \begin{subfigure}[b]{0.19\linewidth}
        \includegraphics[width=\linewidth]{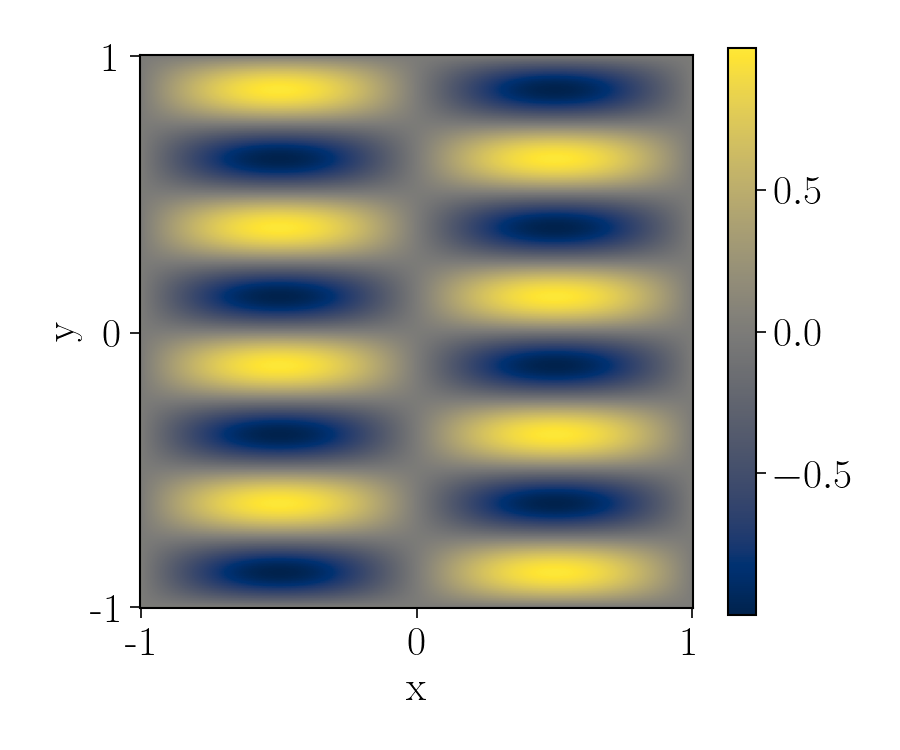}
    \end{subfigure}

    \vspace{1em}

    \begin{subfigure}[b]{0.19\linewidth}
        \includegraphics[width=\linewidth]{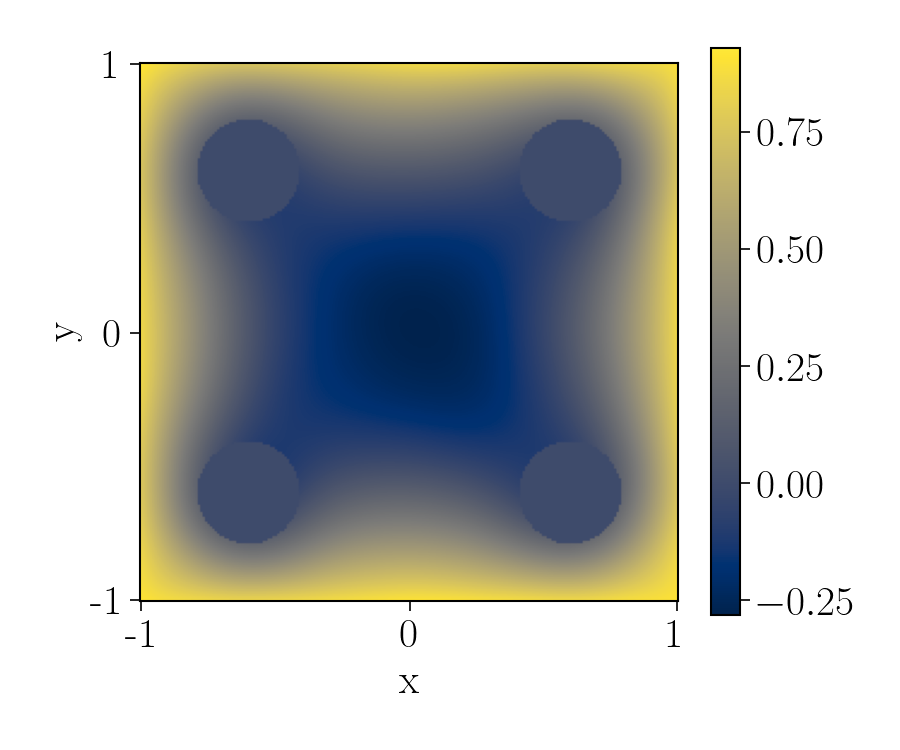}
    \end{subfigure}
    \begin{subfigure}[b]{0.19\linewidth}
        \includegraphics[width=\linewidth]{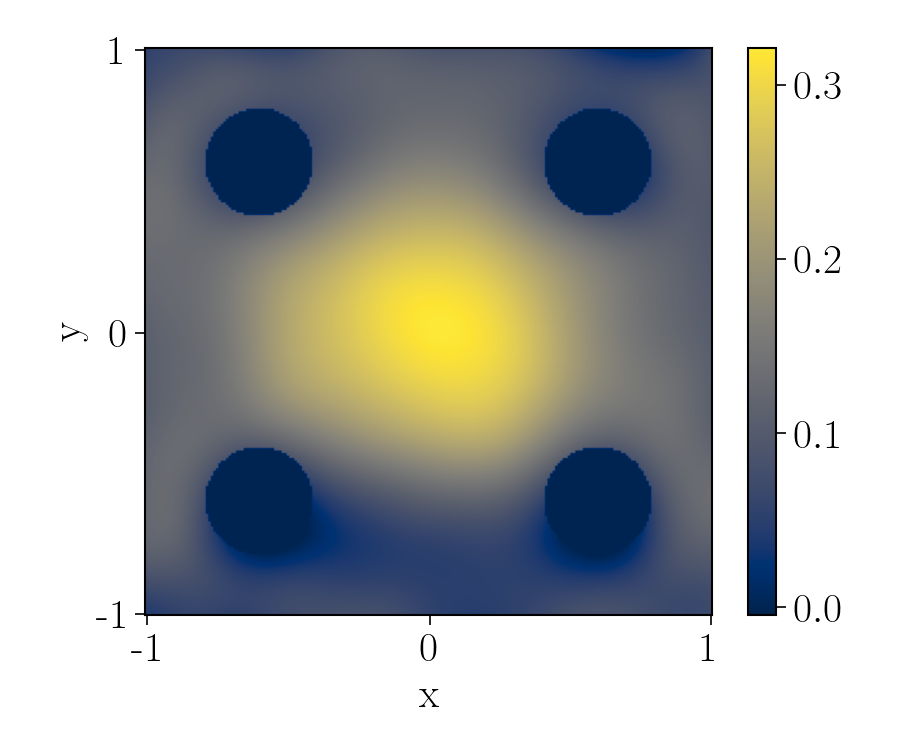}
    \end{subfigure}
    \begin{subfigure}[b]{0.19\linewidth}
        \includegraphics[width=\linewidth]{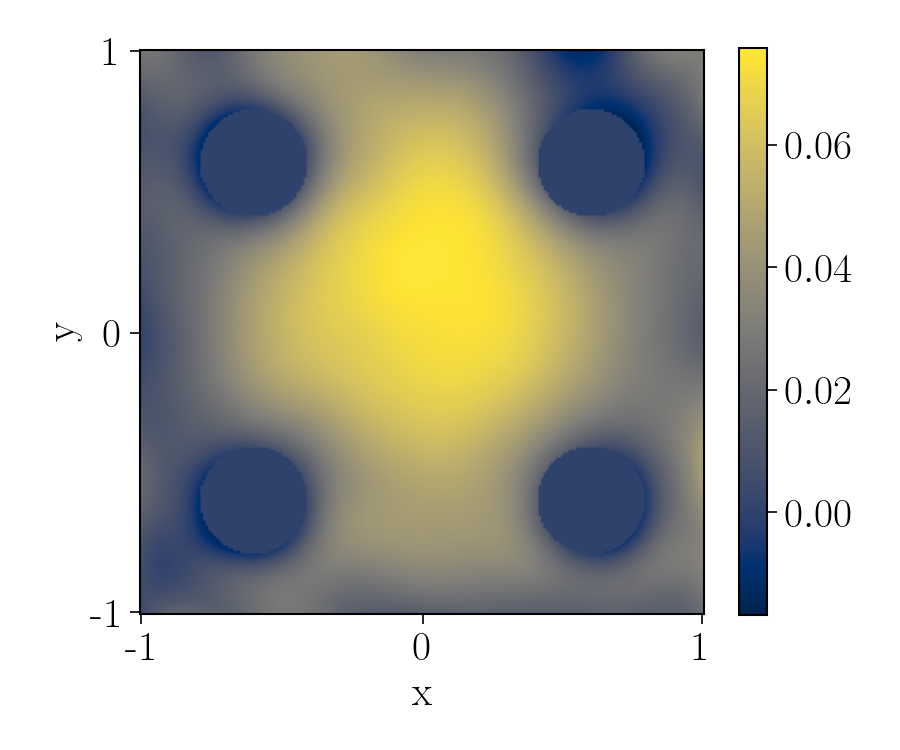}
    \end{subfigure}
    \begin{subfigure}[b]{0.19\linewidth}
        \includegraphics[width=\linewidth]{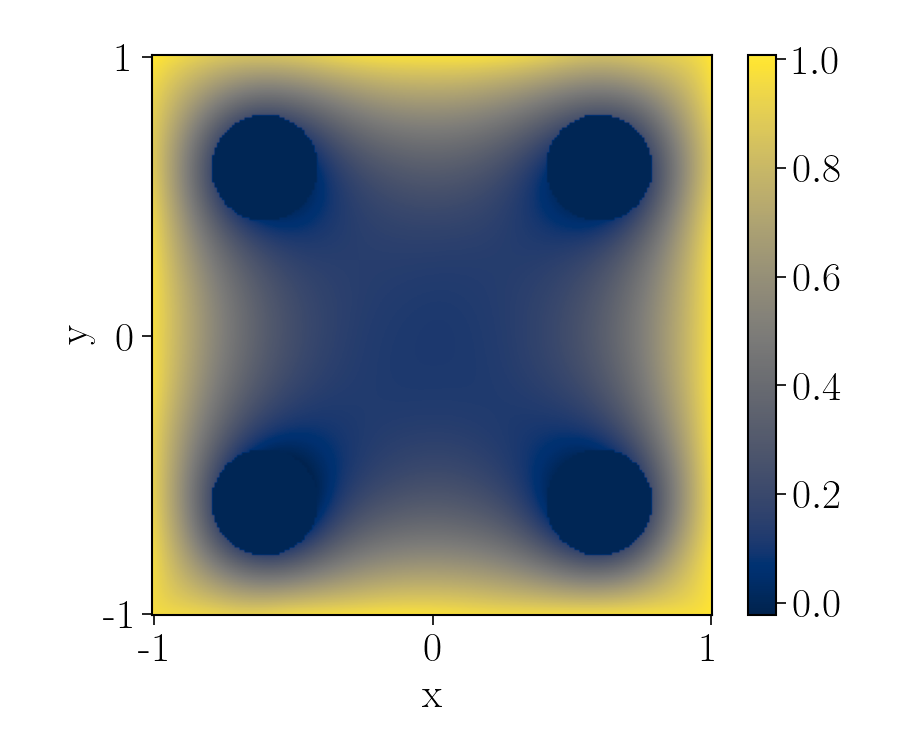}
    \end{subfigure}
    \begin{subfigure}[b]{0.19\linewidth}
        \includegraphics[width=\linewidth]{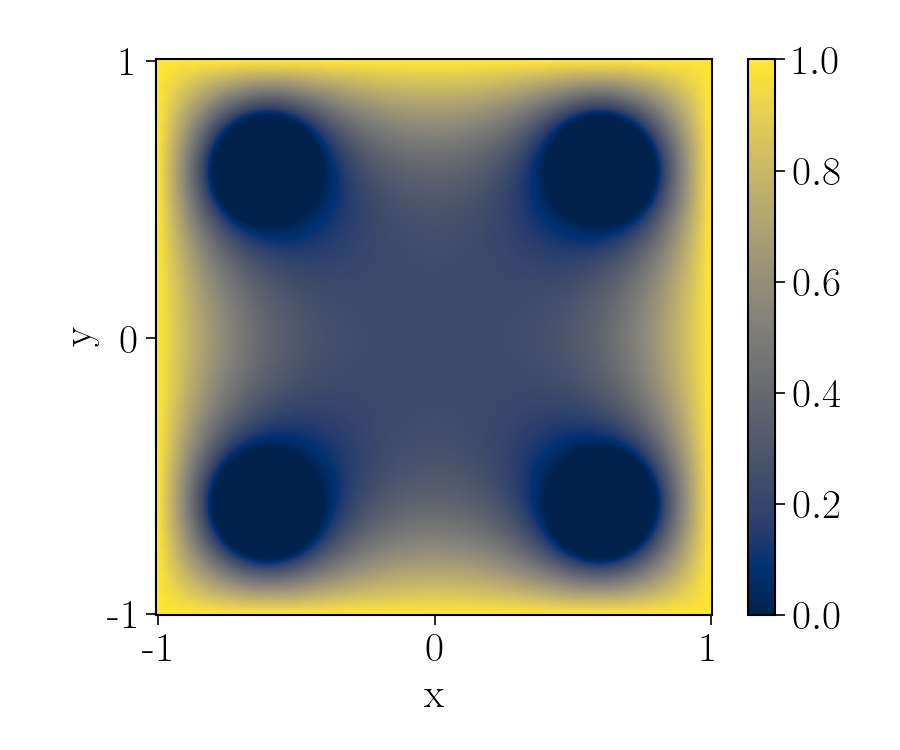}
    \end{subfigure}
    
    \vspace{1em}

    \begin{subfigure}[b]{0.19\linewidth}
        \includegraphics[width=\linewidth]{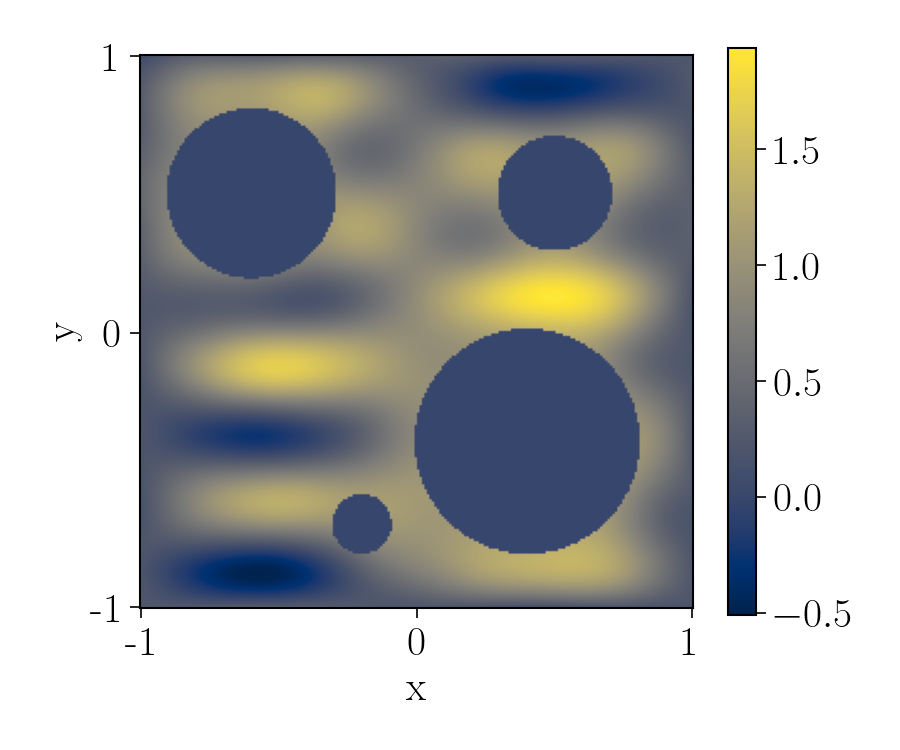}
    \end{subfigure}
    \begin{subfigure}[b]{0.19\linewidth}
        \includegraphics[width=\linewidth]{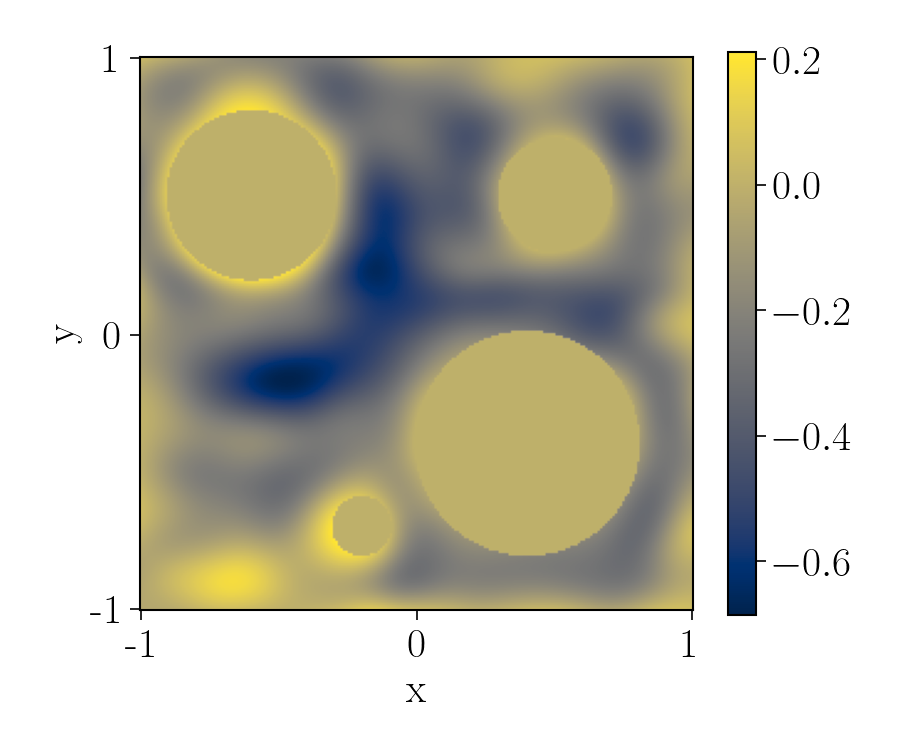}
    \end{subfigure}
    \begin{subfigure}[b]{0.19\linewidth}
        \includegraphics[width=\linewidth]{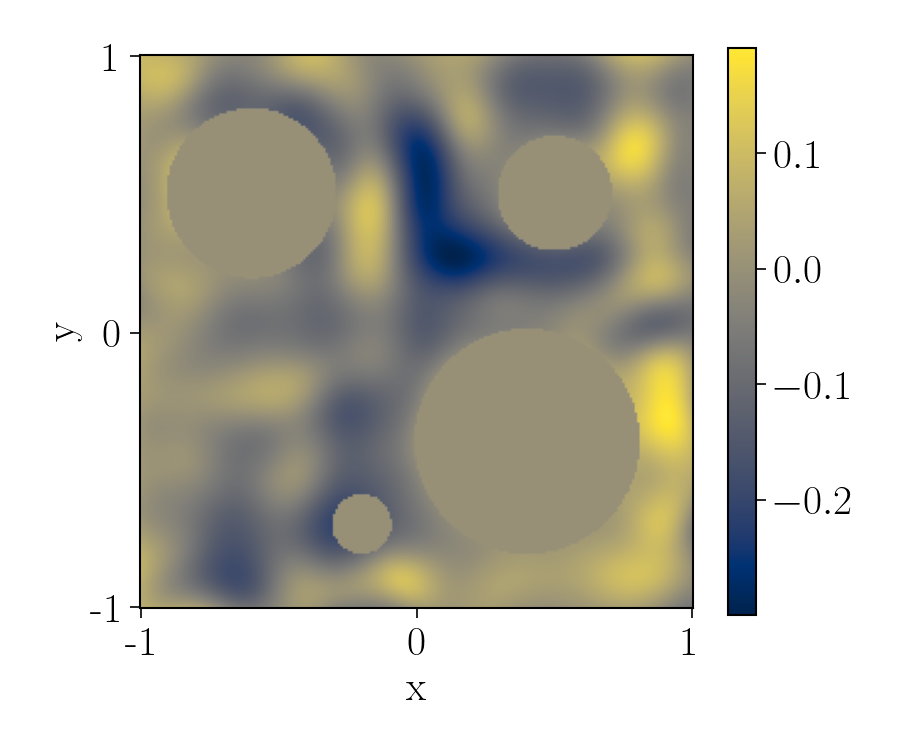}
    \end{subfigure}
    \begin{subfigure}[b]{0.19\linewidth}
        \includegraphics[width=\linewidth]{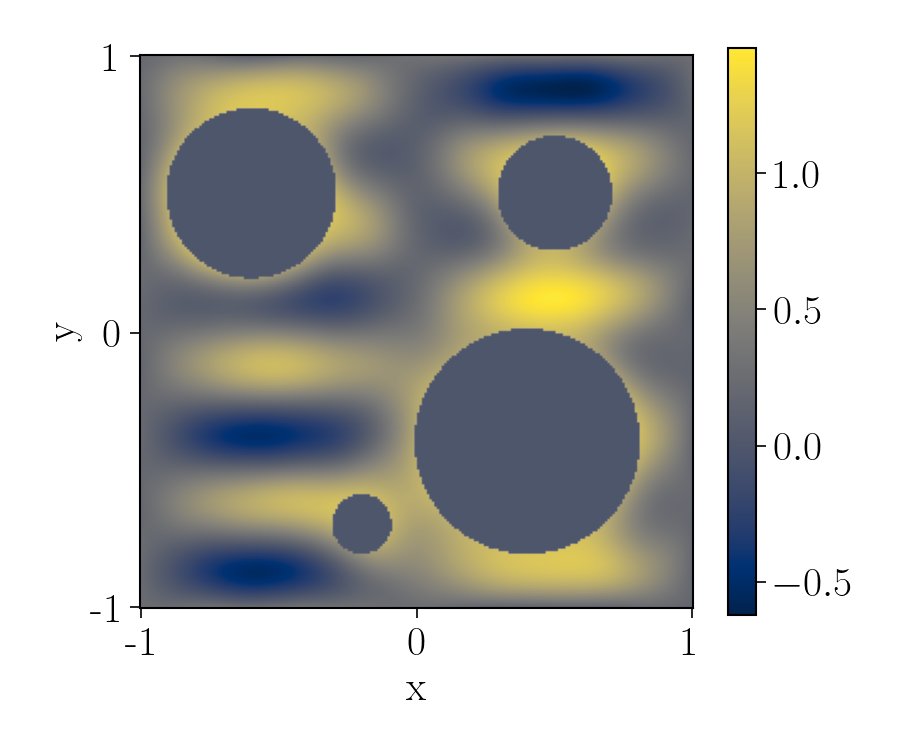}
    \end{subfigure}
    \begin{subfigure}[b]{0.19\linewidth}
        \includegraphics[width=\linewidth]{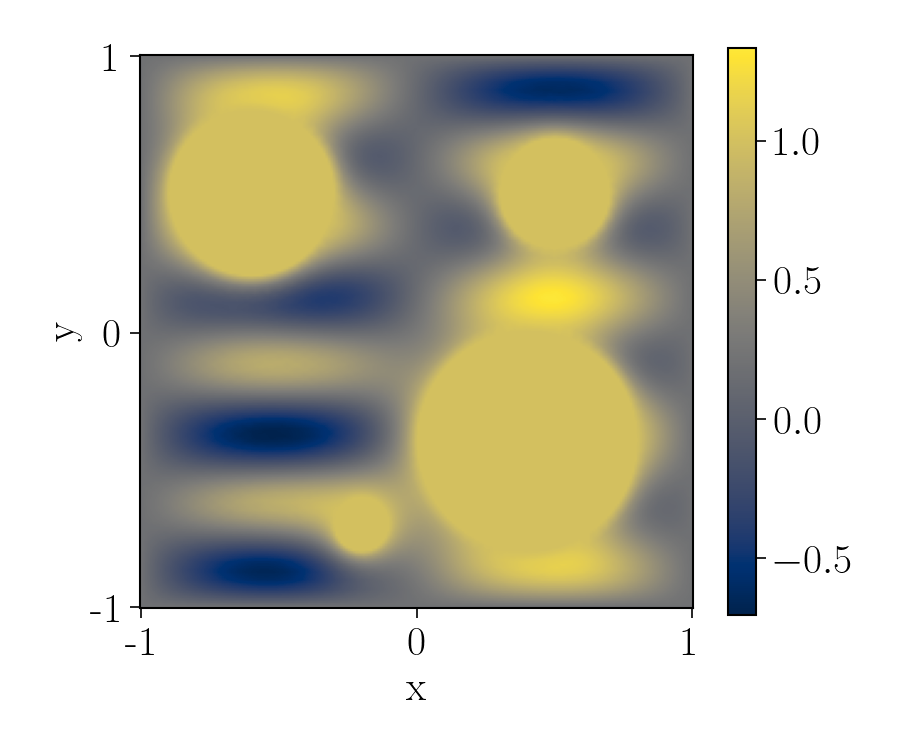}
    \end{subfigure}
    
    \vspace{1em}

    \begin{subfigure}[b]{0.19\linewidth}
        \includegraphics[width=\linewidth]{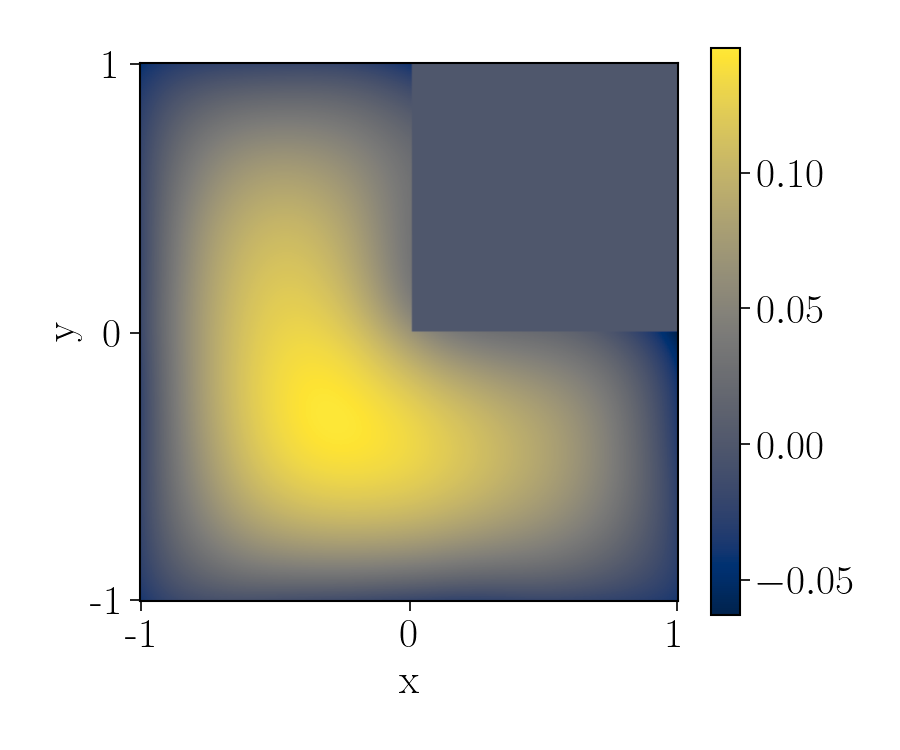}
    \end{subfigure}
    \begin{subfigure}[b]{0.19\linewidth}
        \includegraphics[width=\linewidth]{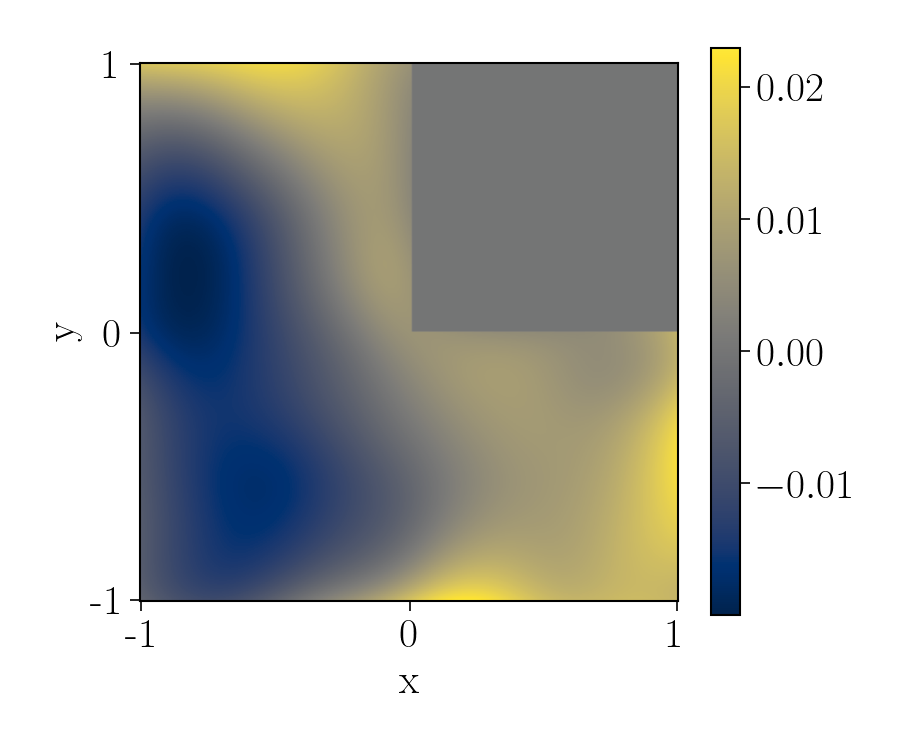}
    \end{subfigure}
    \begin{subfigure}[b]{0.19\linewidth}
        \includegraphics[width=\linewidth]{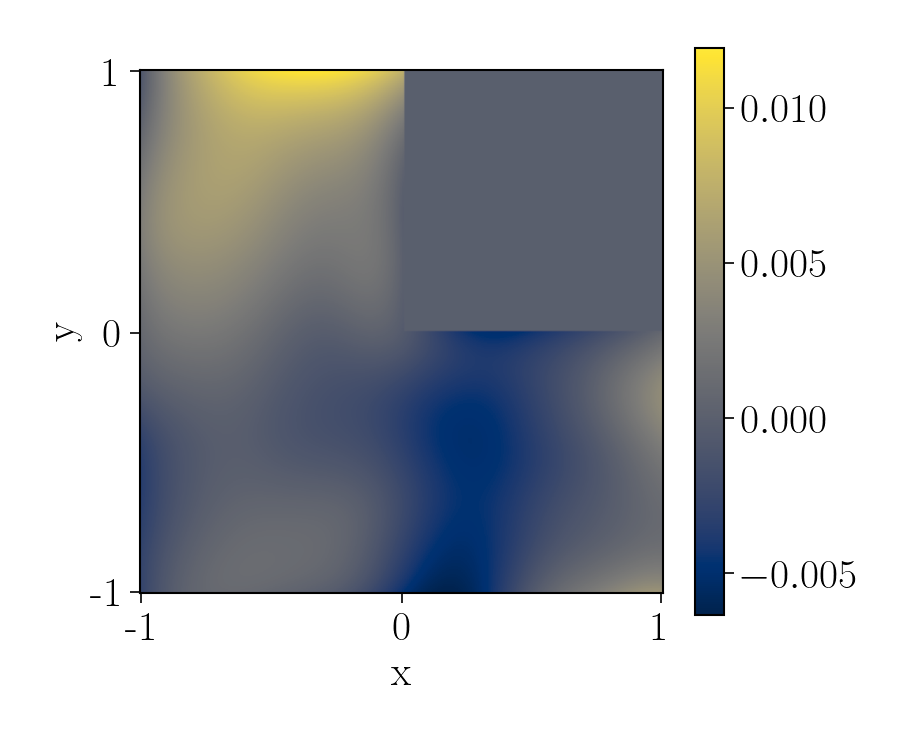}
    \end{subfigure}
    \begin{subfigure}[b]{0.19\linewidth}
        \includegraphics[width=\linewidth]{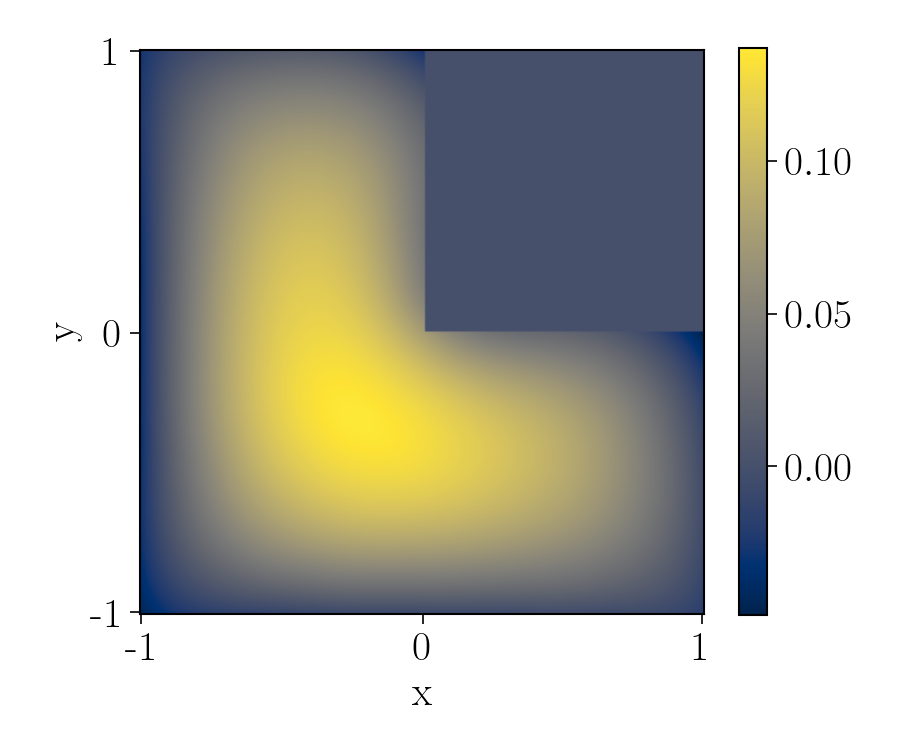}
    \end{subfigure}
    \begin{subfigure}[b]{0.19\linewidth}
        \includegraphics[width=\linewidth]{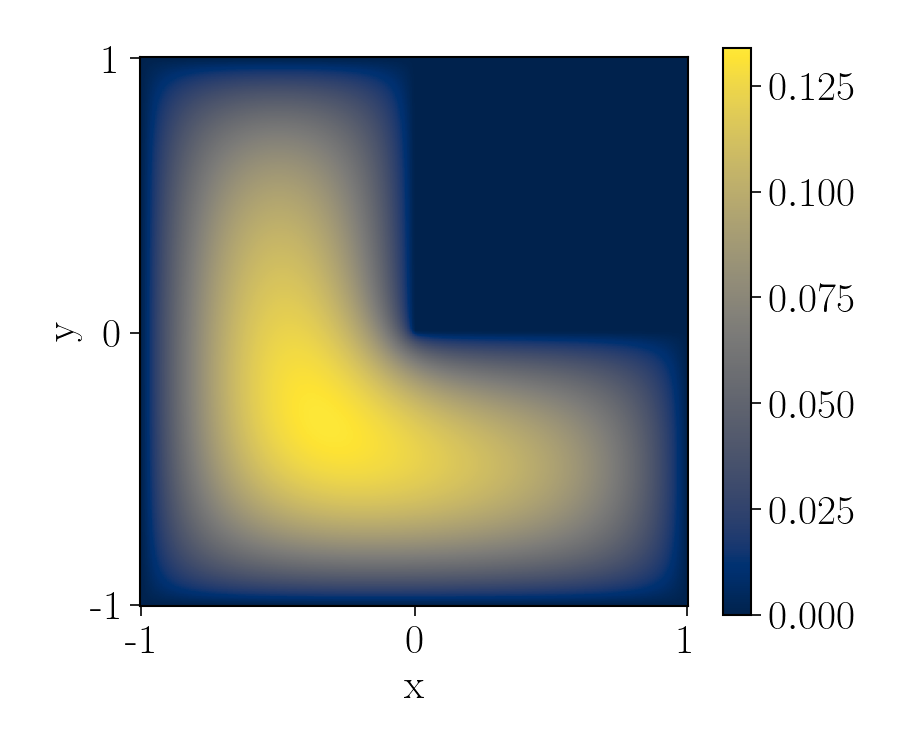}
    \end{subfigure}
    
    \vspace{1em}

    \begin{subfigure}[b]{0.19\linewidth}
        \includegraphics[width=\linewidth]{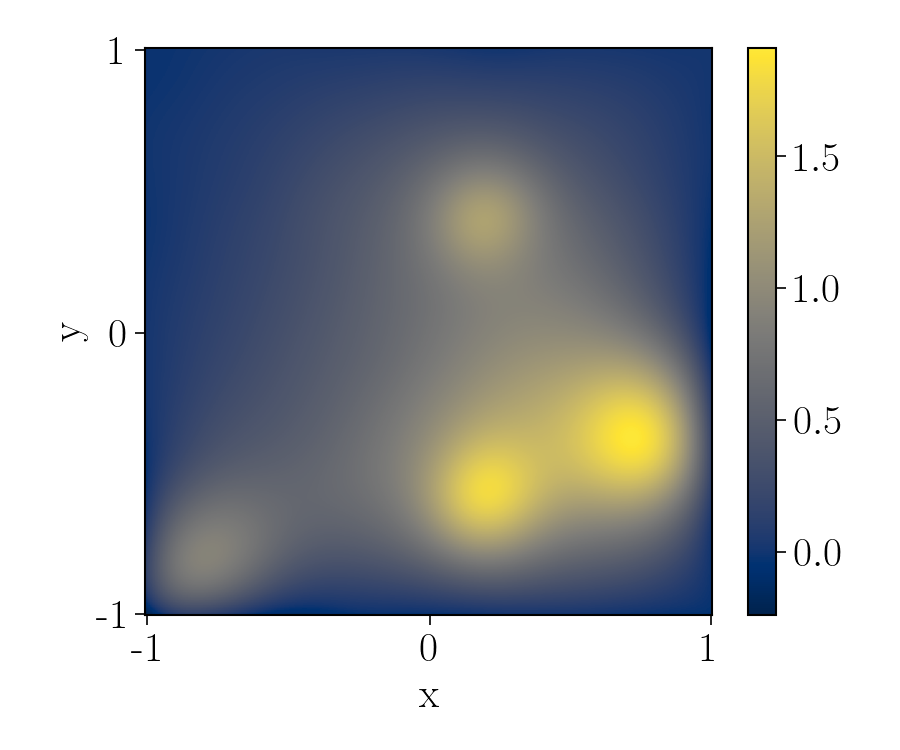}
    \end{subfigure}
    \begin{subfigure}[b]{0.19\linewidth}
        \includegraphics[width=\linewidth]{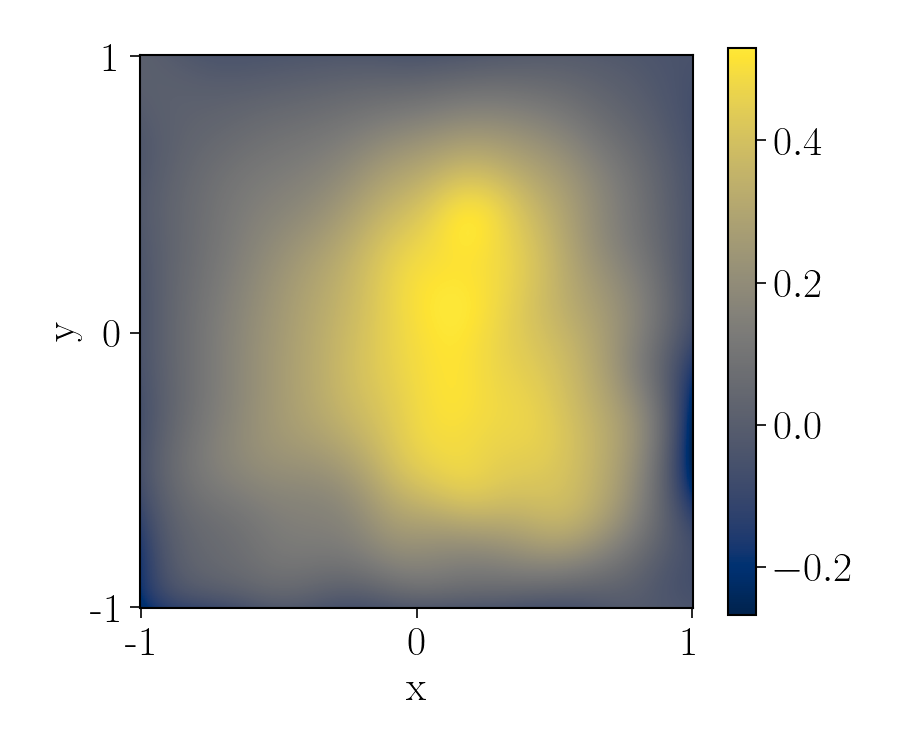}
    \end{subfigure}
    \begin{subfigure}[b]{0.19\linewidth}
        \includegraphics[width=\linewidth]{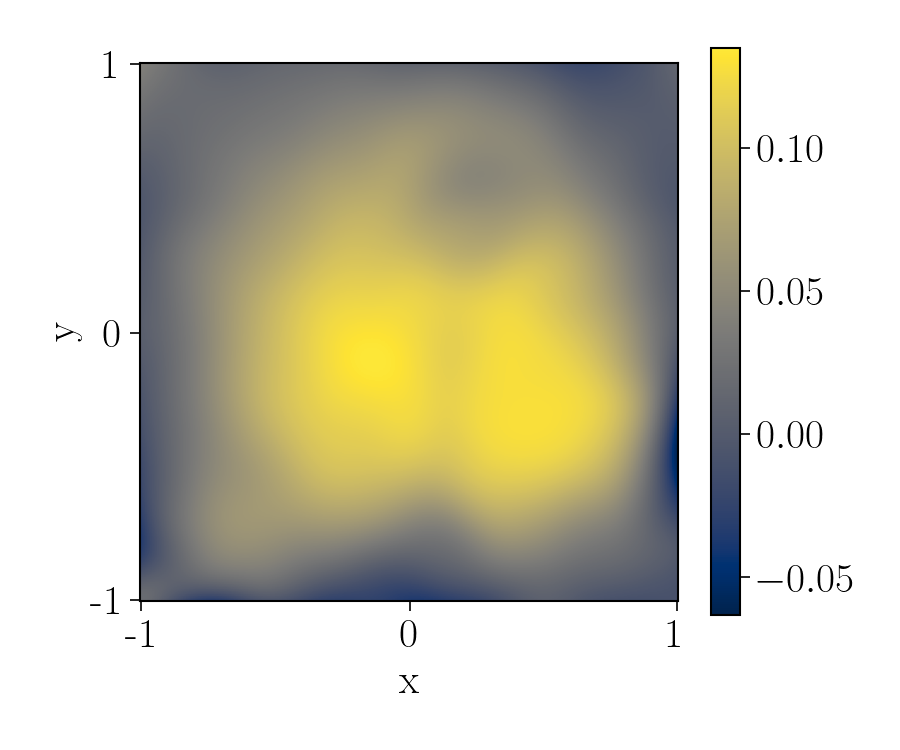}
    \end{subfigure}
    \begin{subfigure}[b]{0.19\linewidth}
        \includegraphics[width=\linewidth]{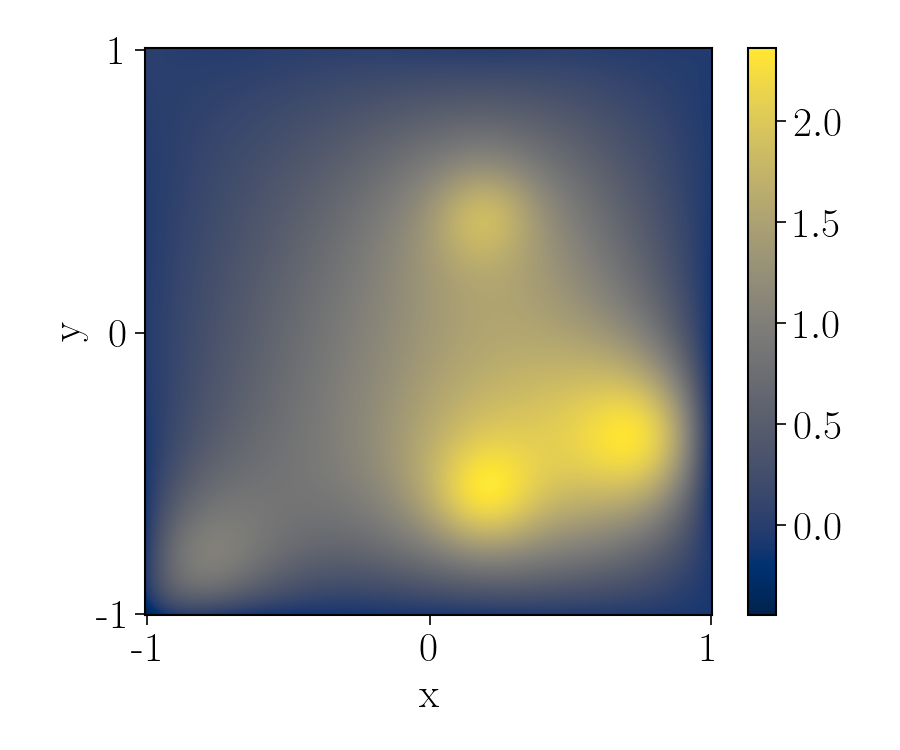}
    \end{subfigure}
    \begin{subfigure}[b]{0.19\linewidth}
        \includegraphics[width=\linewidth]{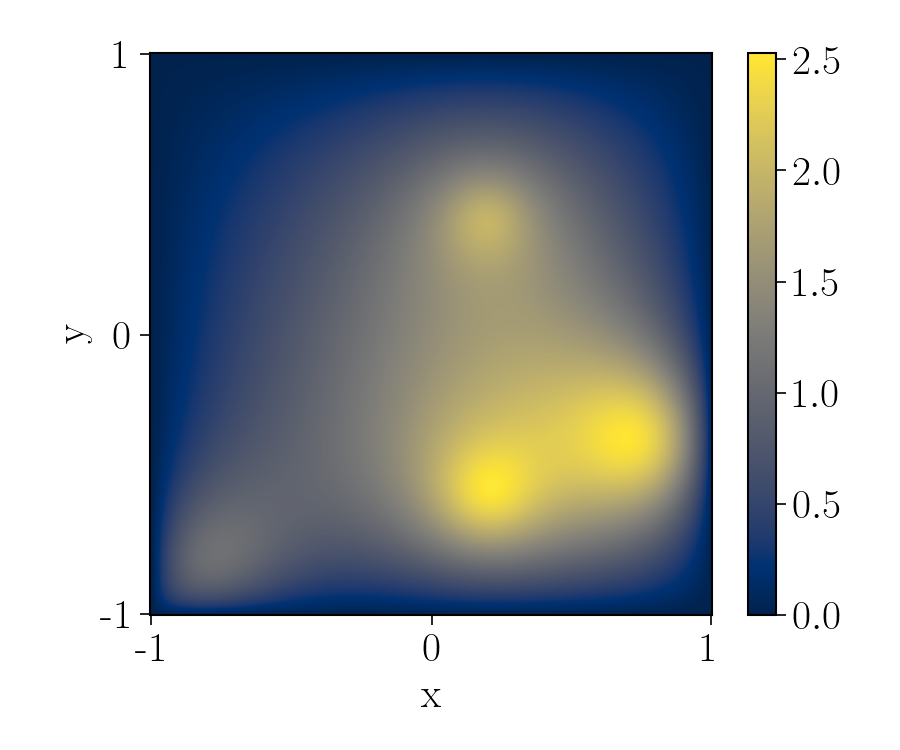}
    \end{subfigure}
    
    \vspace{1em}

    \begin{subfigure}[b]{0.19\linewidth}
        \includegraphics[width=\linewidth]{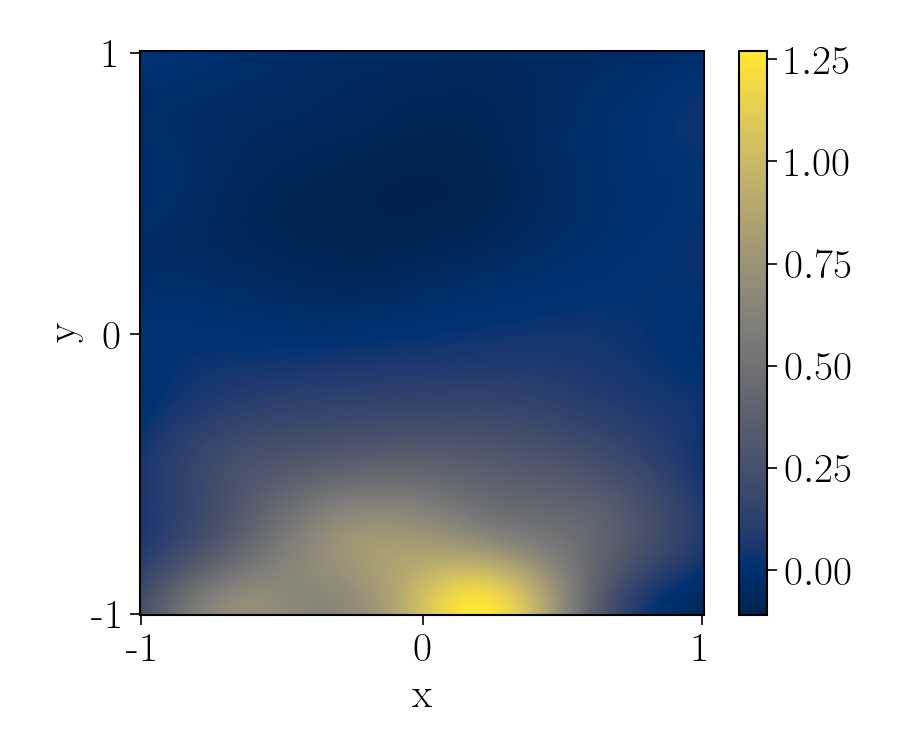}
        \caption{}
    \end{subfigure}
    \begin{subfigure}[b]{0.19\linewidth}
        \includegraphics[width=\linewidth]{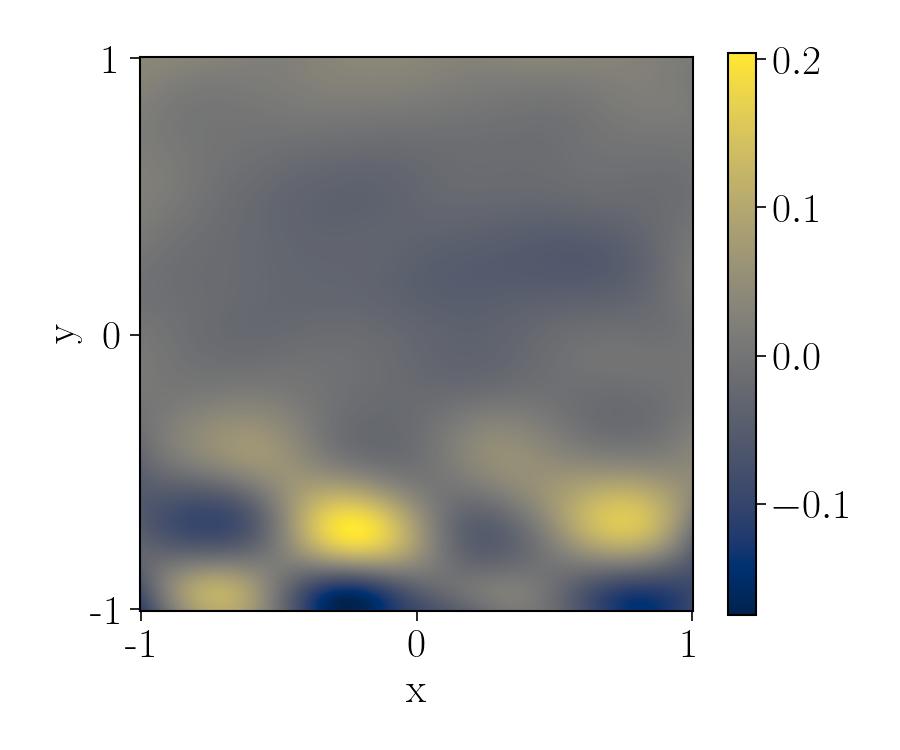}
        \caption{}
    \end{subfigure}
    \begin{subfigure}[b]{0.19\linewidth}
        \includegraphics[width=\linewidth]{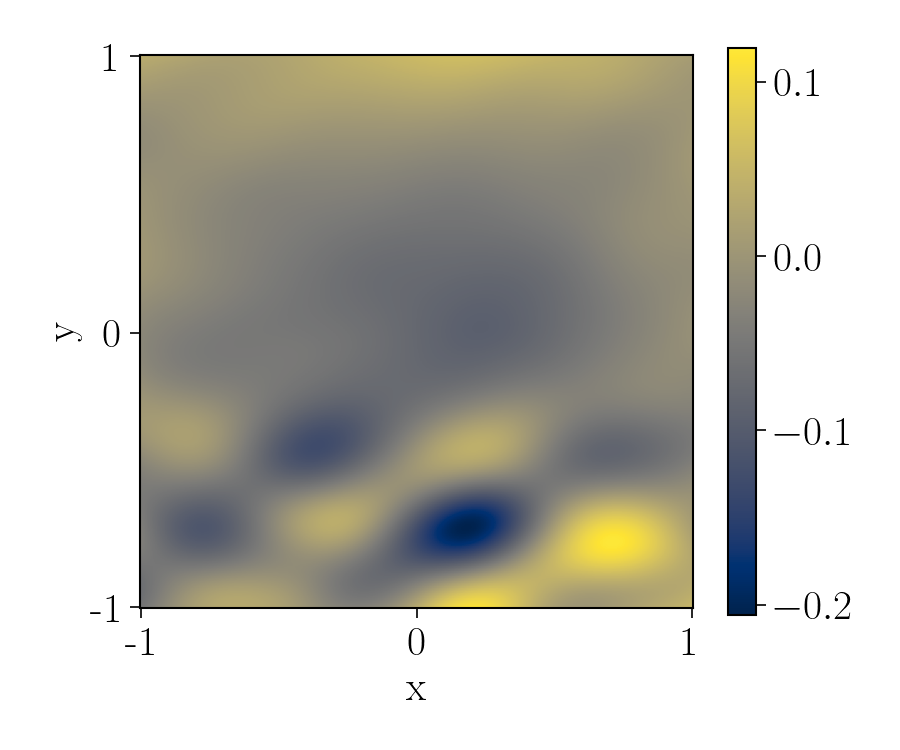}
        \caption{}
    \end{subfigure}
    \begin{subfigure}[b]{0.19\linewidth}
        \includegraphics[width=\linewidth]{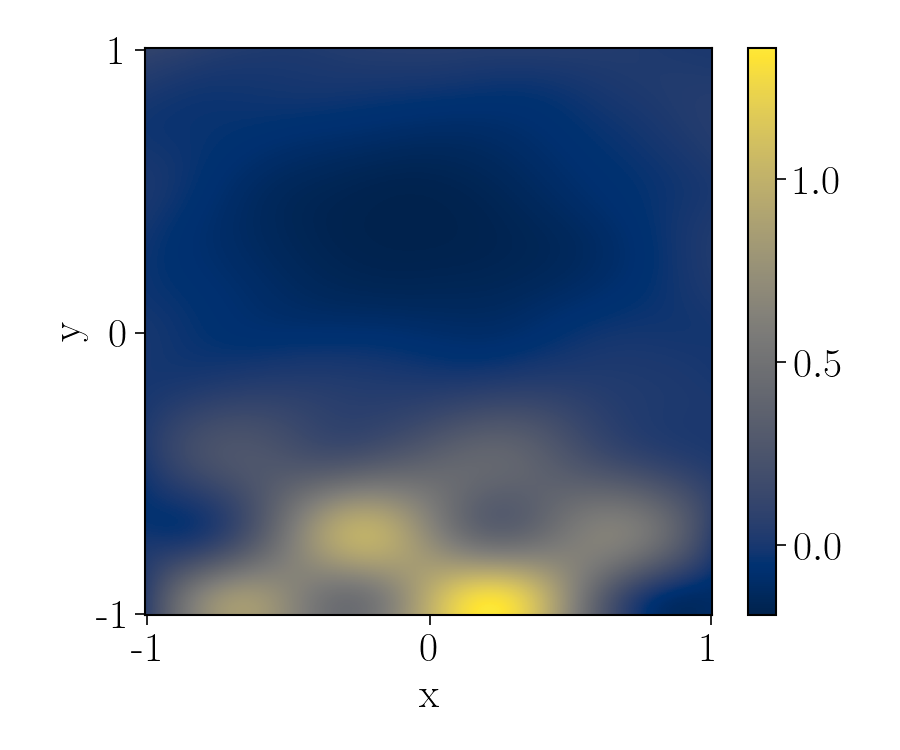}
        \caption{}
    \end{subfigure}
    \begin{subfigure}[b]{0.19\linewidth}
        \includegraphics[width=\linewidth]{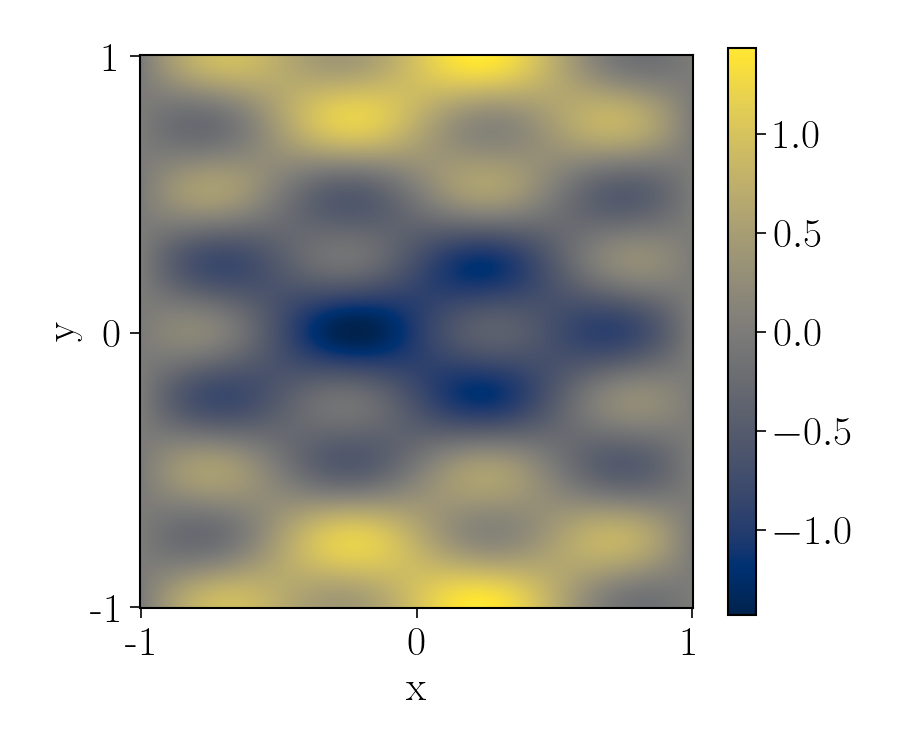}
        \caption{}
    \end{subfigure}

    \caption{Visual progression of iterative refinement across different samples. Each row shows: (a) HyPINO prediction $u^{(0)}$, (b) 1st Refinement $\delta u^{(1)}$, (c) 2nd Refinement $\delta u^{(2)}$, (d) Final prediction $u^{(0)} + \delta u^{(1)} + \delta u^{(2)}$, and ground truth (e).}
    \label{fig:iterative_refinement_grid}
\end{figure}

\newpage
\subsection{Fine-tuning}
\label{app:finetuning}

We investigate the utility of HyPINO-generated PINN parameters $\theta^\star$ as a prior for rapid adaptation to specific PDE instances. We compare three initialization strategies: (i) HyPINO-initialized PINNs, (ii) randomly initialized PINNs, and (iii) PINNs initialized via Reptile meta-learning~\cite{reptile}. For Reptile, we use 10{,}000 outer-loop steps on our synthetic dataset and 1,000 inner-loop updates per sample with an inner learning rate of 0.01.

In addition to single-network performance, we evaluate the effect of initialization on ensemble-based methods. We compare ensembles of size 3 and 10 generated using HyPINO (denoted HyPINO$^3$ and HyPINO$^{10}$) against ensembles of the same size and architecture initialized either randomly or via Reptile. For the latter, we replicate the Reptile-initialized weights across all ensemble members. Note that a HyPINO$^i$ ensemble consists of the base PINN as well as $i$ refinement (or delta) PINNs: $u^{(0)} + \sum_{t=1}^{T} \delta u^{(t)}$, thus creating an ensemble with $i+1$ experts.

\begin{figure}[ht]
    \centering

    \begin{subfigure}[b]{0.32\linewidth}
        \includegraphics[width=\linewidth]{figures/finetuning_convergence_plots/heat_0.png}
    \end{subfigure}
    \begin{subfigure}[b]{0.32\linewidth}
        \includegraphics[width=\linewidth]{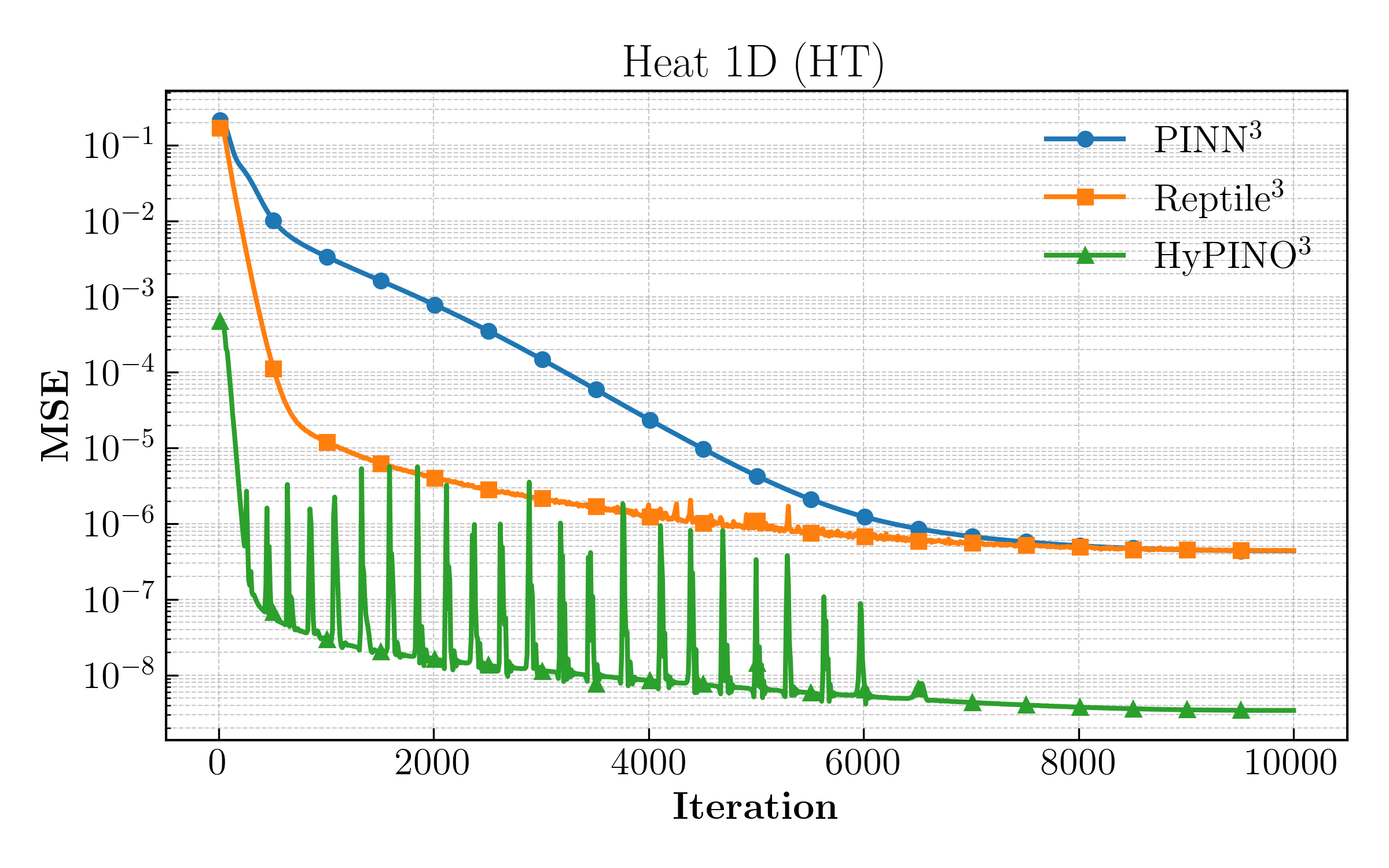}
    \end{subfigure}
    \begin{subfigure}[b]{0.32\linewidth}
        \includegraphics[width=\linewidth]{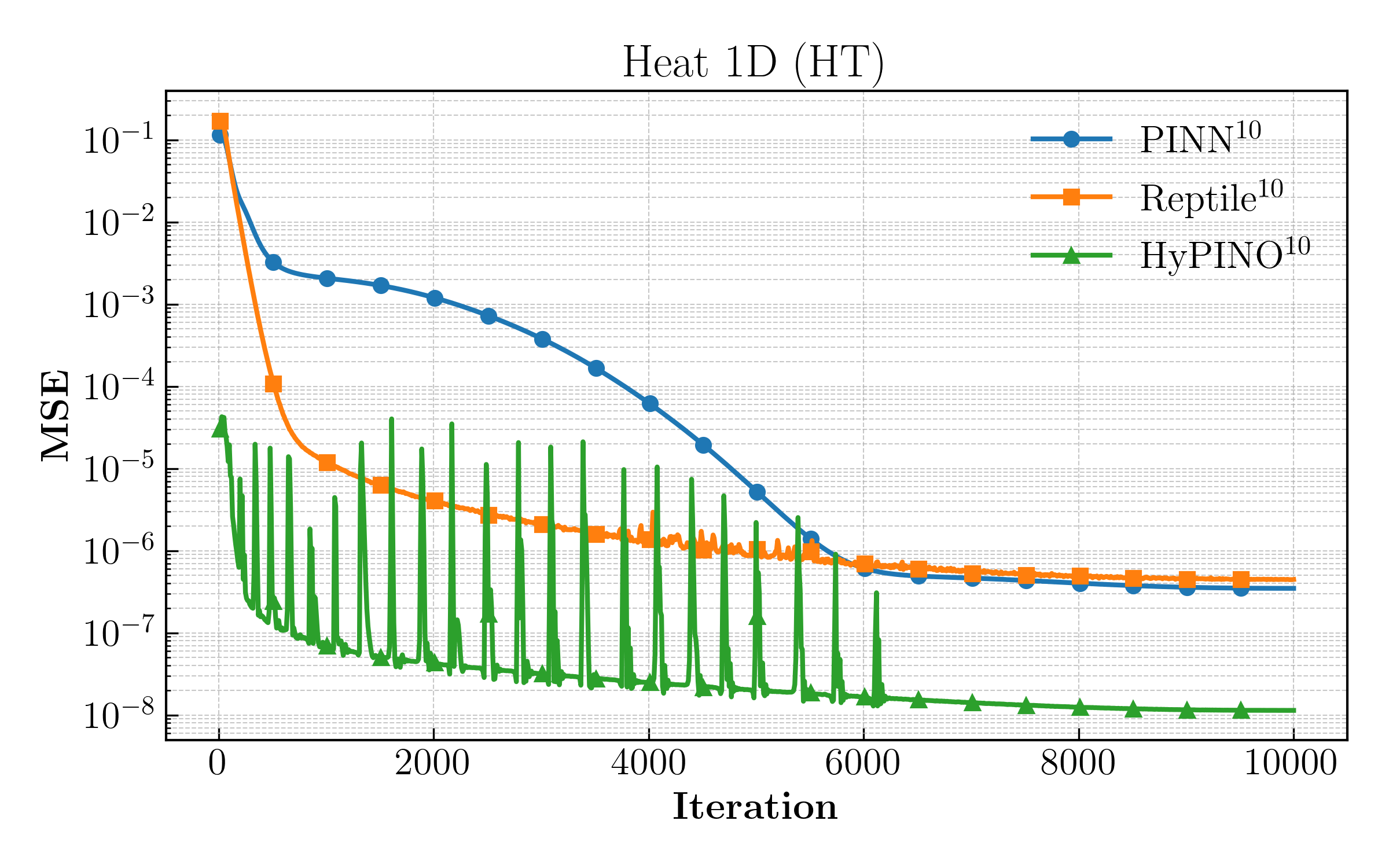}
    \end{subfigure}

    \vspace{1em}

    \begin{subfigure}[b]{0.32\linewidth}
        \includegraphics[width=\linewidth]{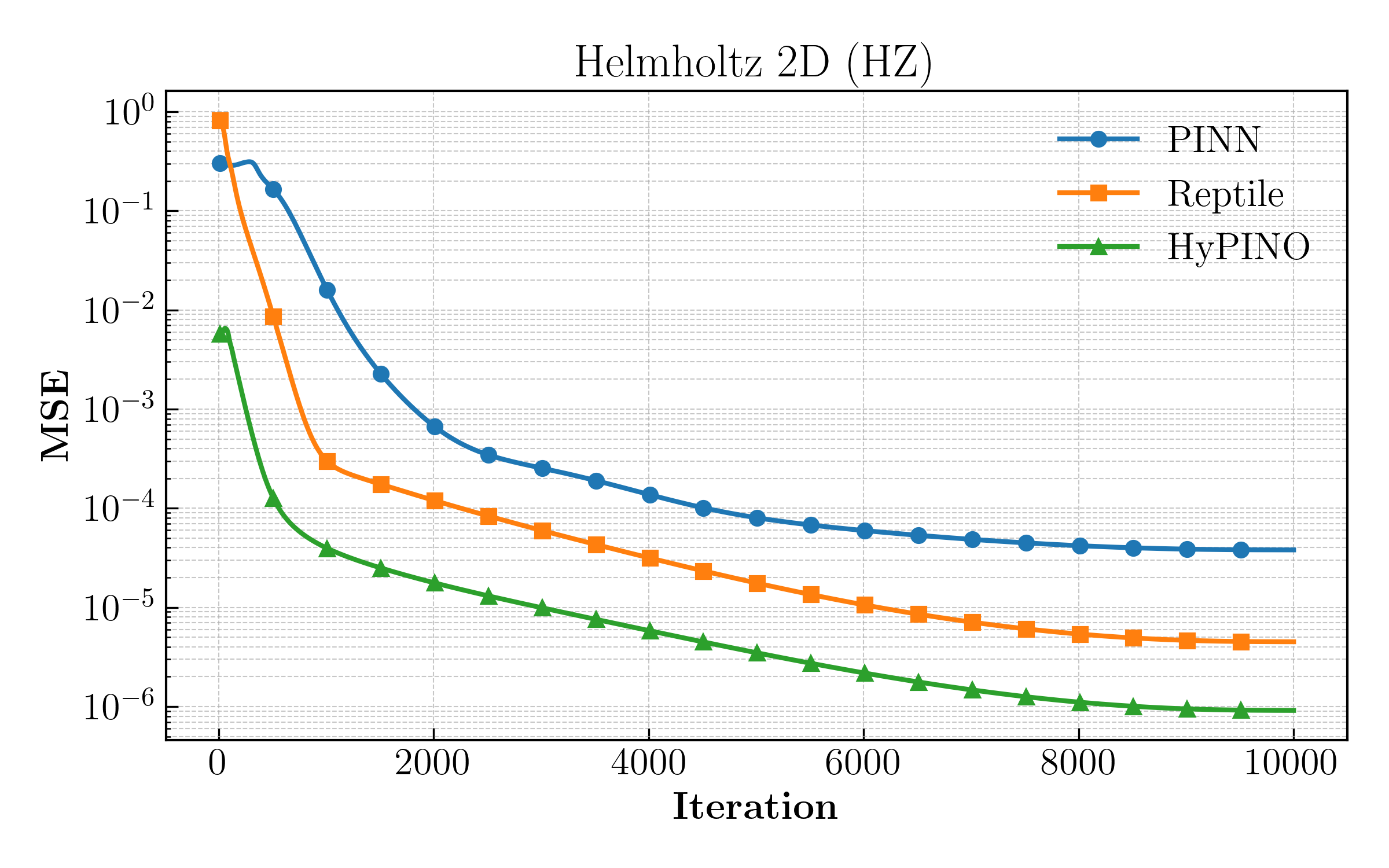}
    \end{subfigure}
    \begin{subfigure}[b]{0.32\linewidth}
        \includegraphics[width=\linewidth]{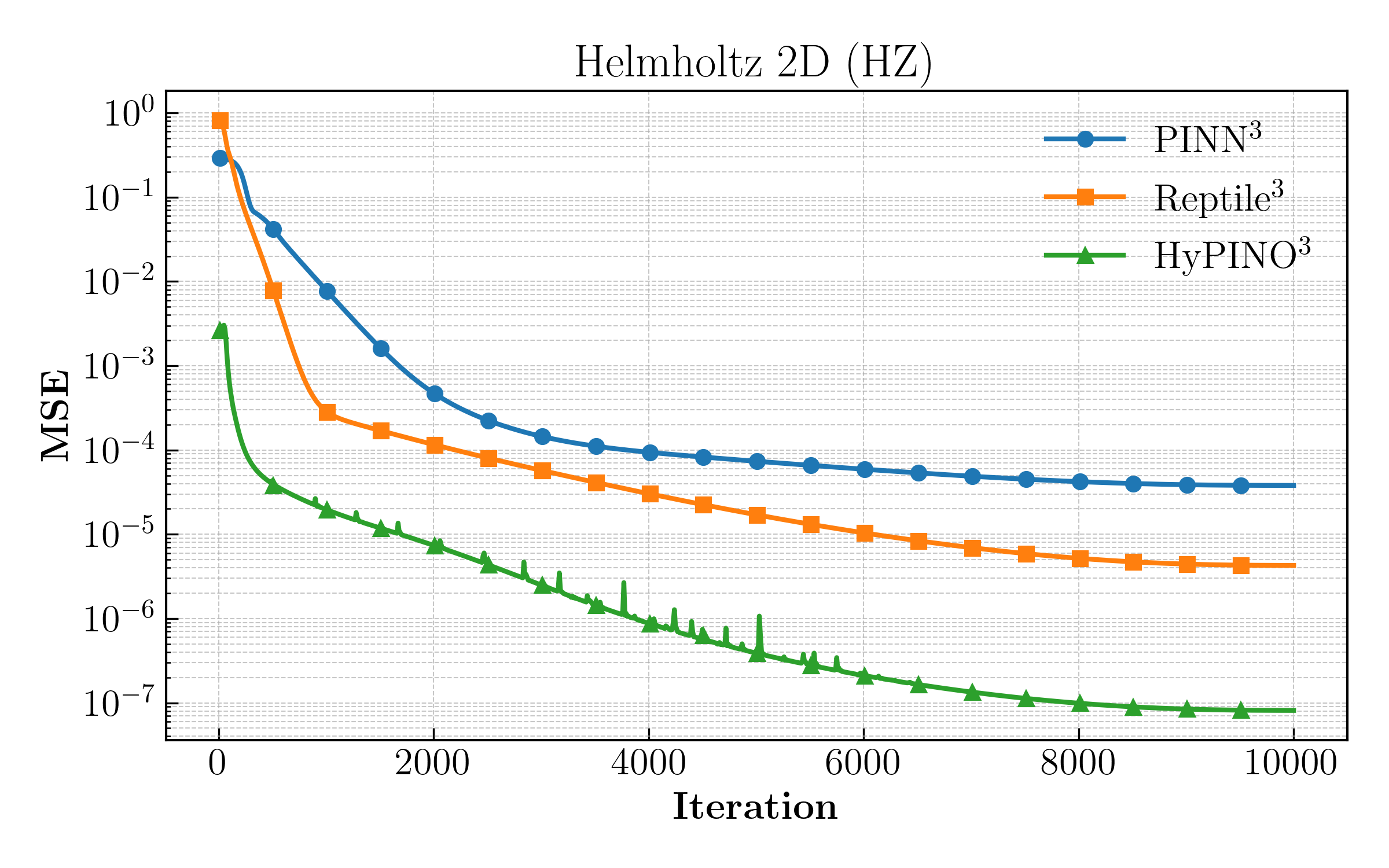}
    \end{subfigure}
    \begin{subfigure}[b]{0.32\linewidth}
        \includegraphics[width=\linewidth]{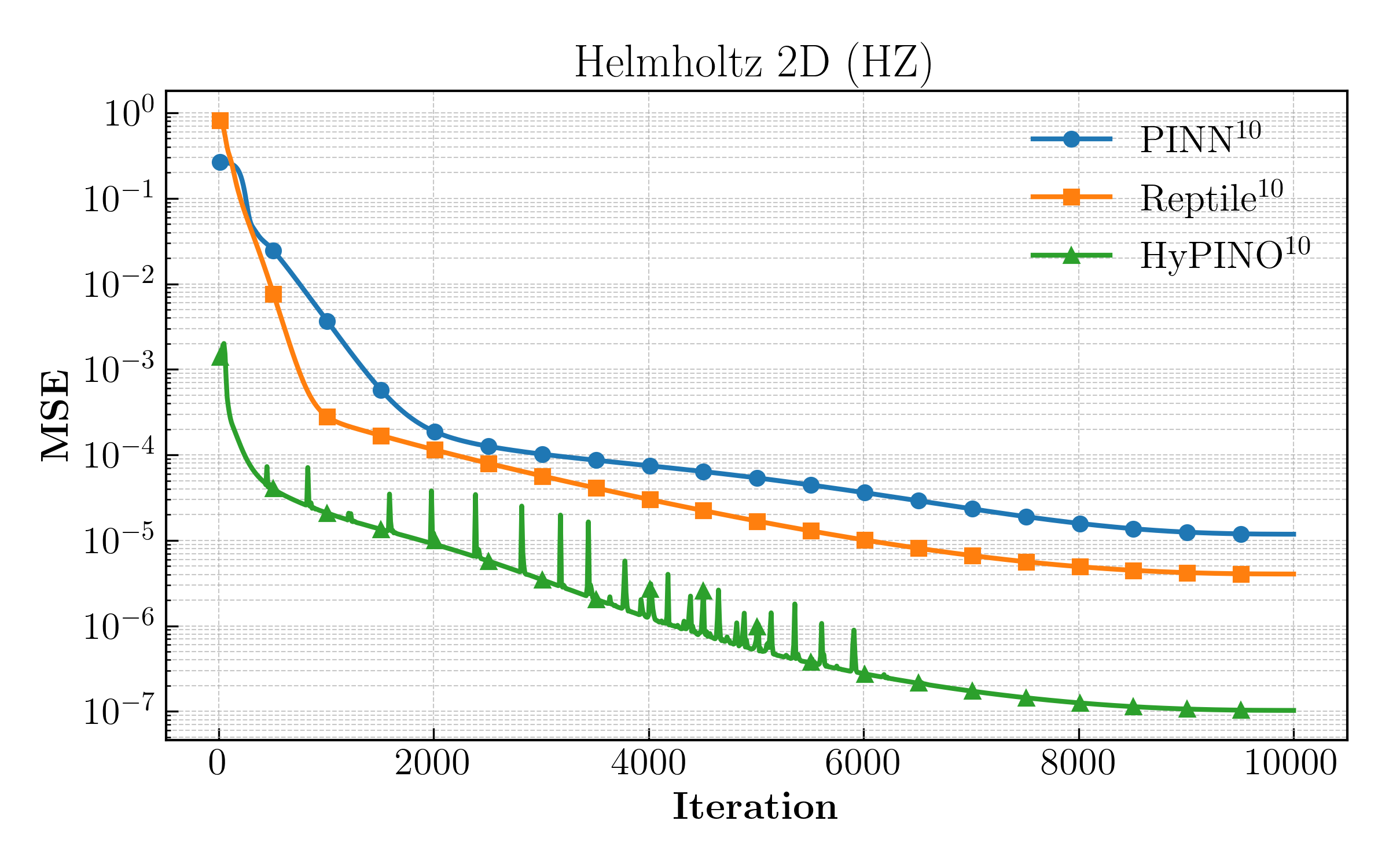}
    \end{subfigure}

    \vspace{1em}

    \begin{subfigure}[b]{0.32\linewidth}
        \includegraphics[width=\linewidth]{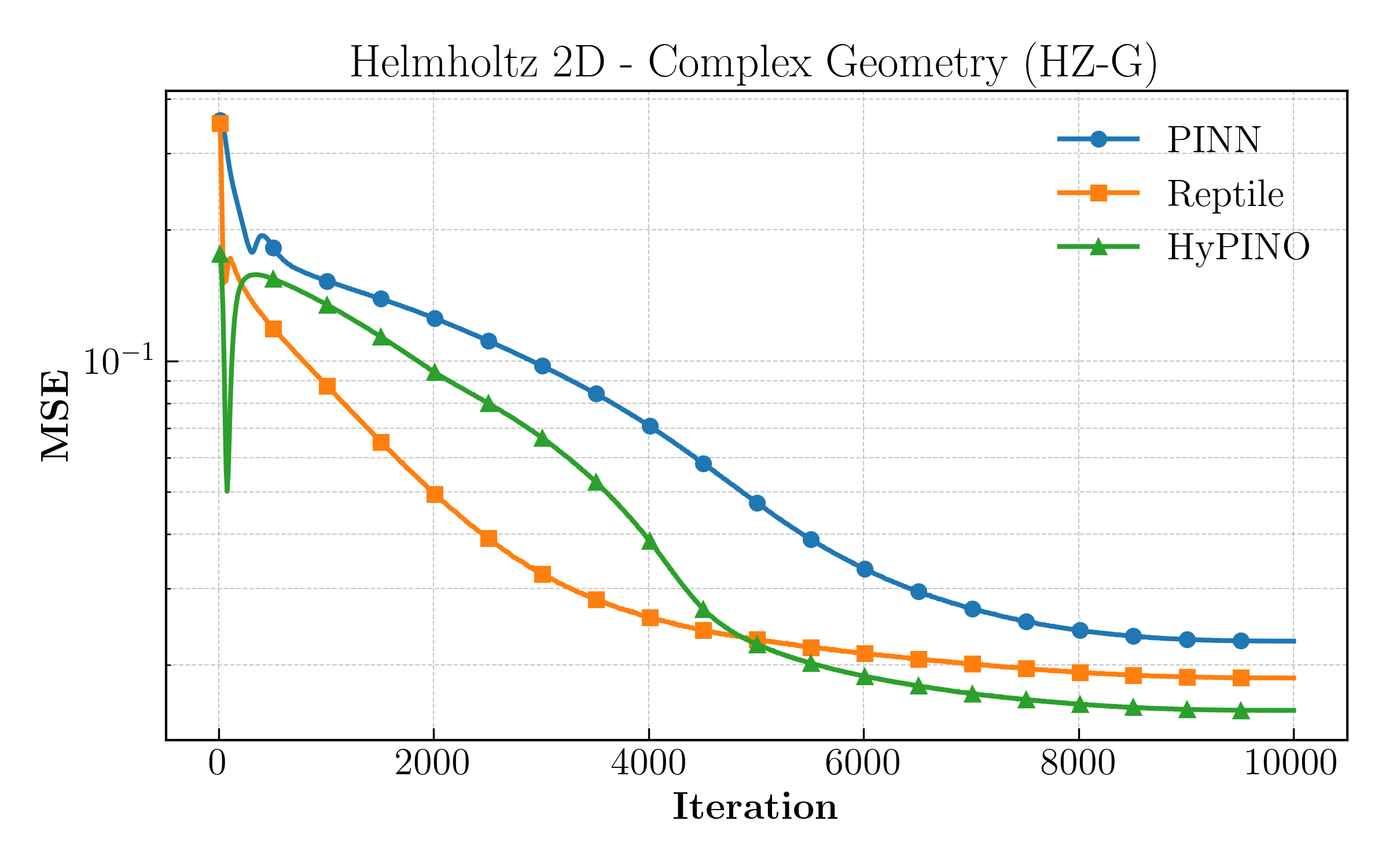}
    \end{subfigure}
    \begin{subfigure}[b]{0.32\linewidth}
        \includegraphics[width=\linewidth]{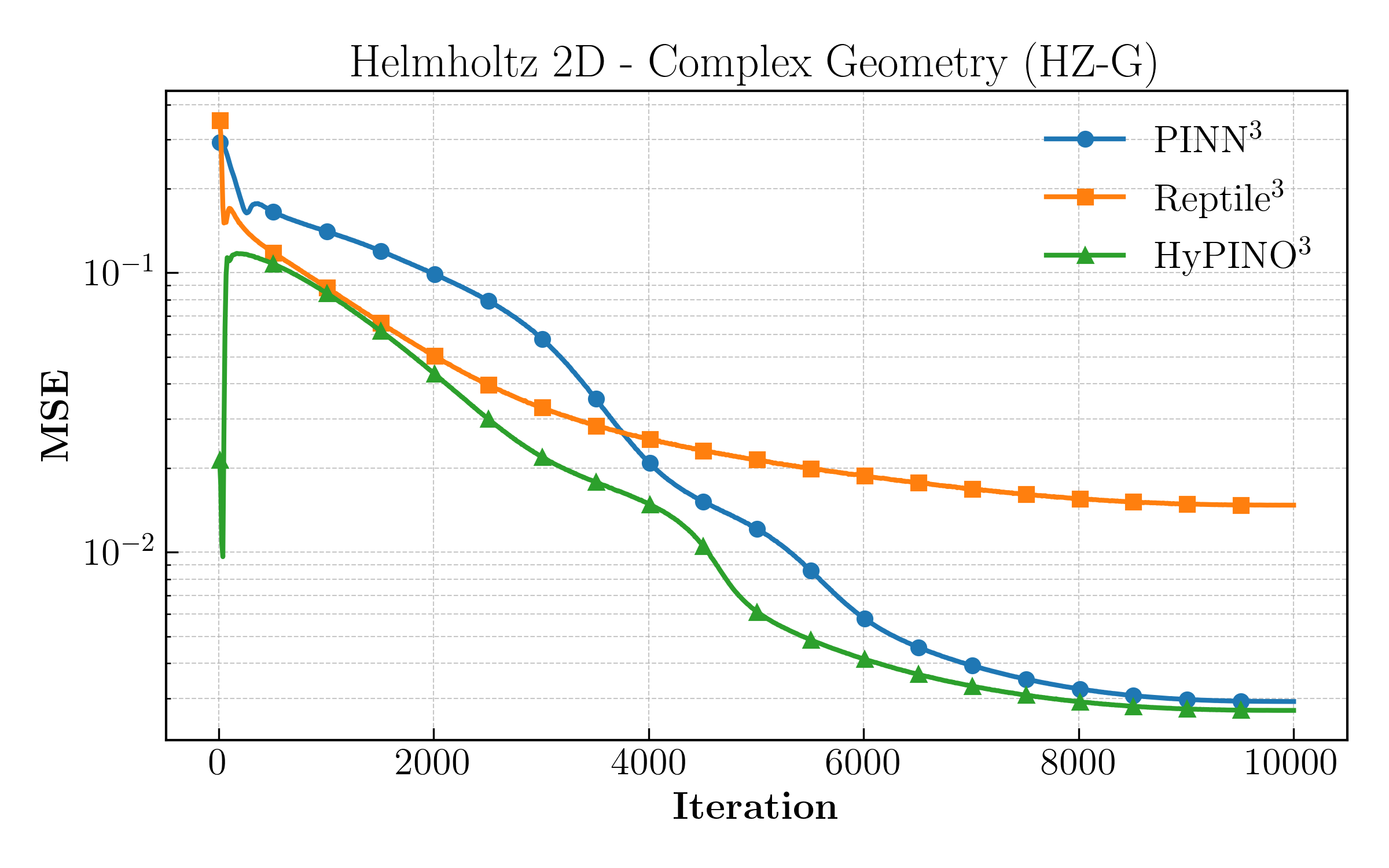}
    \end{subfigure}
    \begin{subfigure}[b]{0.32\linewidth}
        \includegraphics[width=\linewidth]{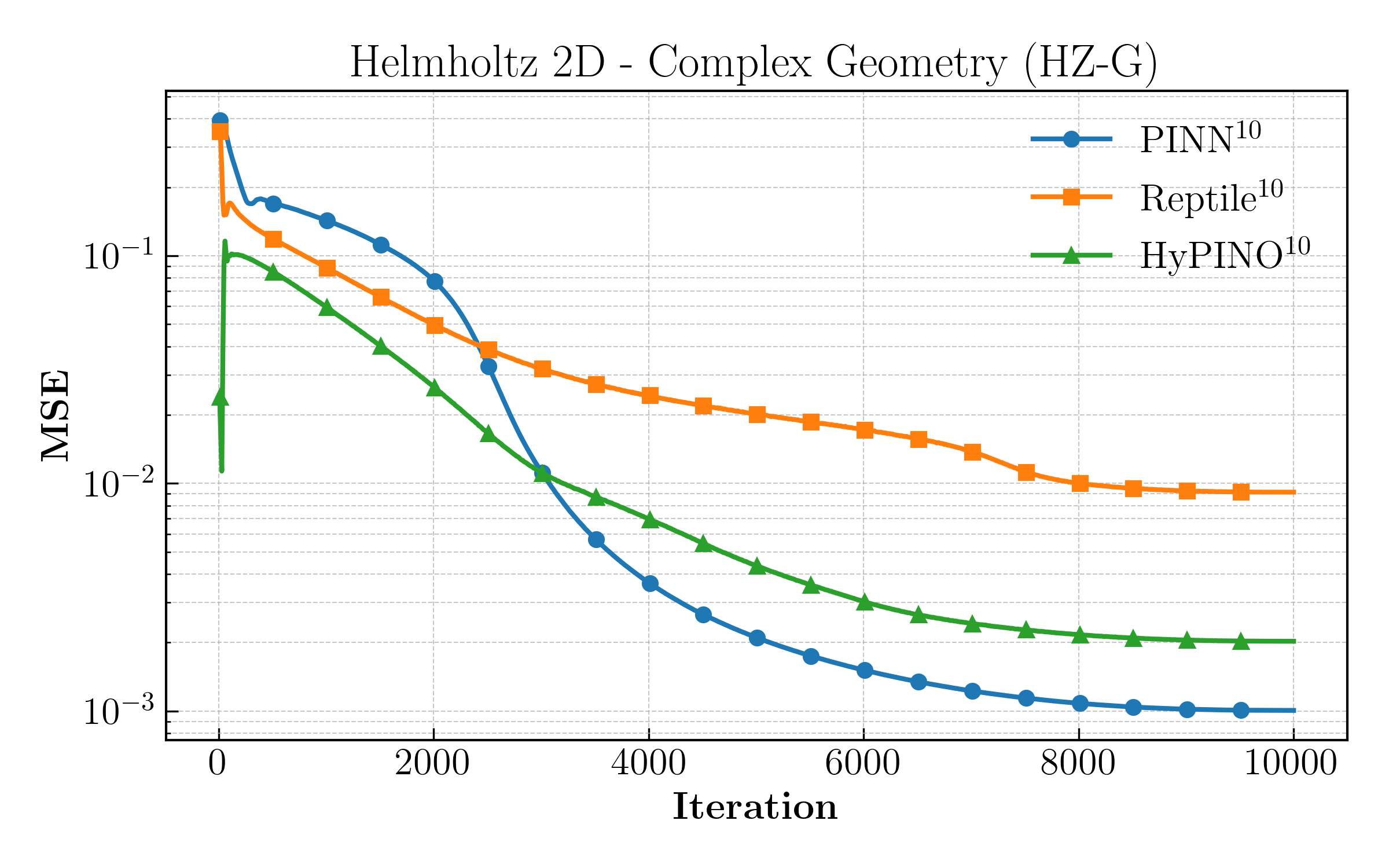}
    \end{subfigure}
    
    \vspace{1em}

    \begin{subfigure}[b]{0.32\linewidth}
        \includegraphics[width=\linewidth]{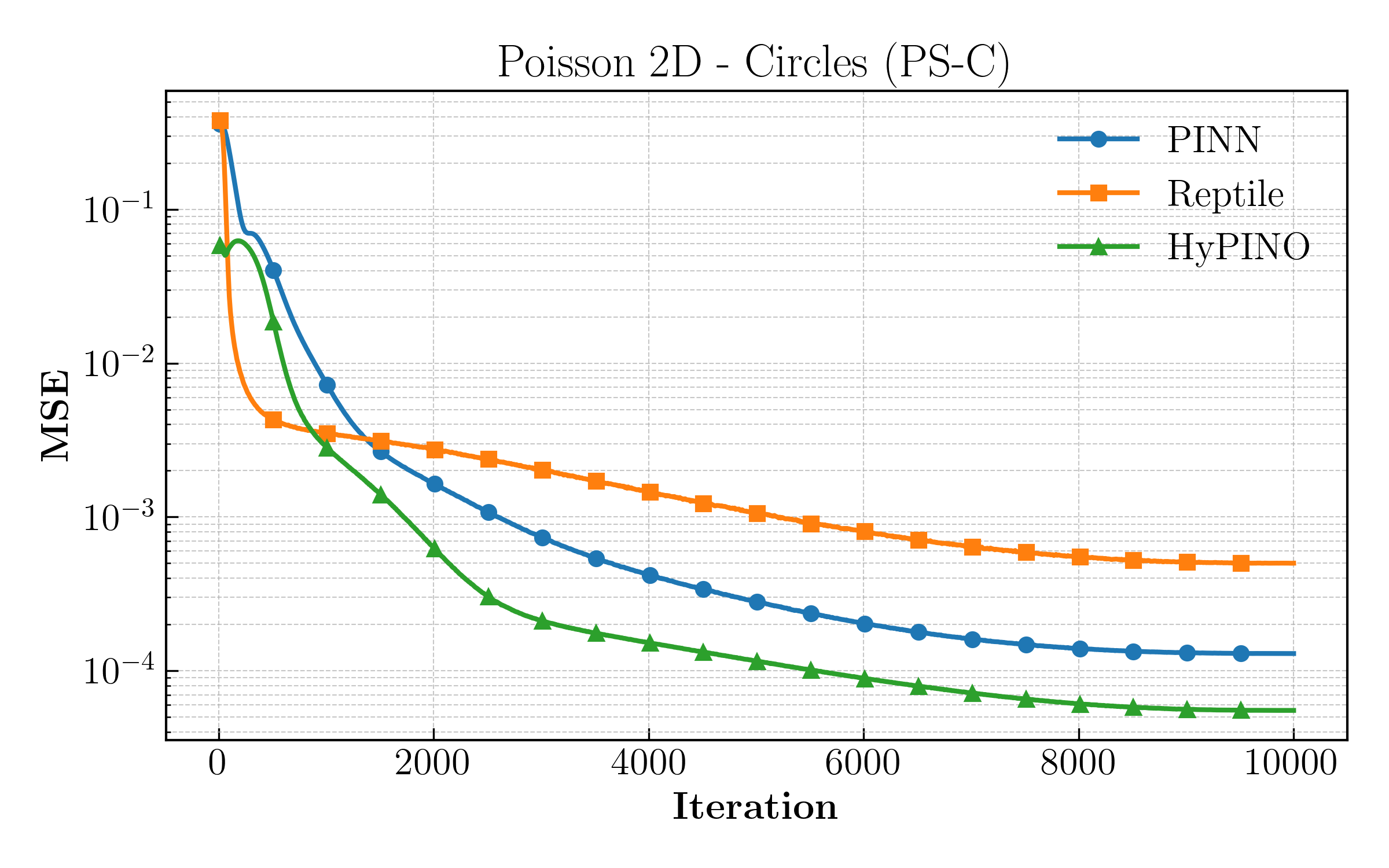}
        \caption{}
    \end{subfigure}
    \begin{subfigure}[b]{0.32\linewidth}
        \includegraphics[width=\linewidth]{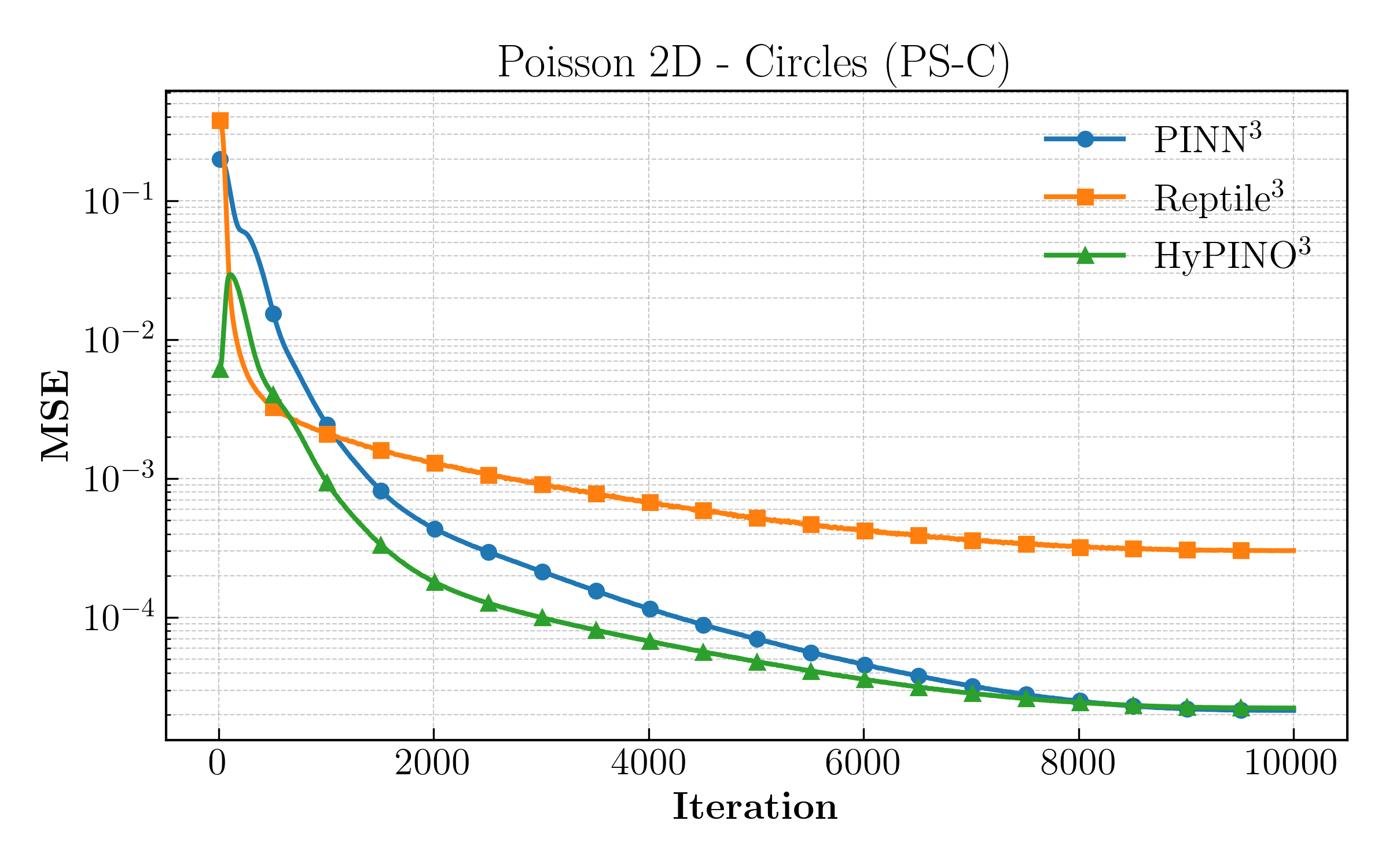}
        \caption{}
    \end{subfigure}
    \begin{subfigure}[b]{0.32\linewidth}
        \includegraphics[width=\linewidth]{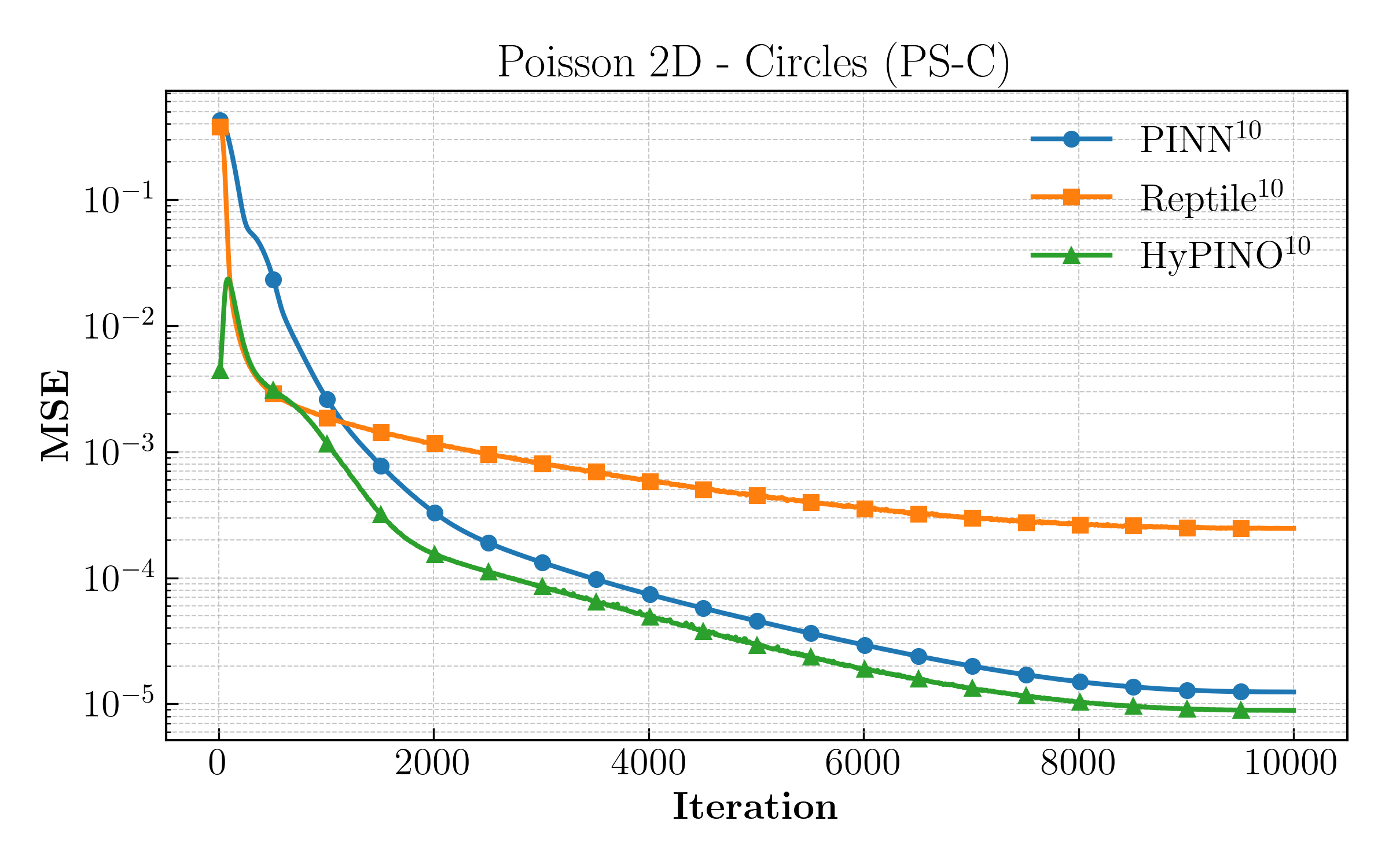}
        \caption{}
    \end{subfigure}

    \caption{Convergence of PINNs when fine-tuned on each of the benchmark PDE problems. We compare the convergence of different ensemble sizes: (a) single PINN, (b) ensemble of size 4 (c) ensemble of size 11, where an ensemble of size $i$ is an ensemble of $i$ randomly initialized PINNs (blue), $i$ PINNs initialized via Reptile (orange), or one PINN initialized via HyPINO followed by $i-1$ refinement rounds (green).}
    \label{fig:finetuning_convergence}
\end{figure}

\begin{figure}[ht]
    \centering

    \begin{subfigure}[b]{0.32\linewidth}
        \includegraphics[width=\linewidth]{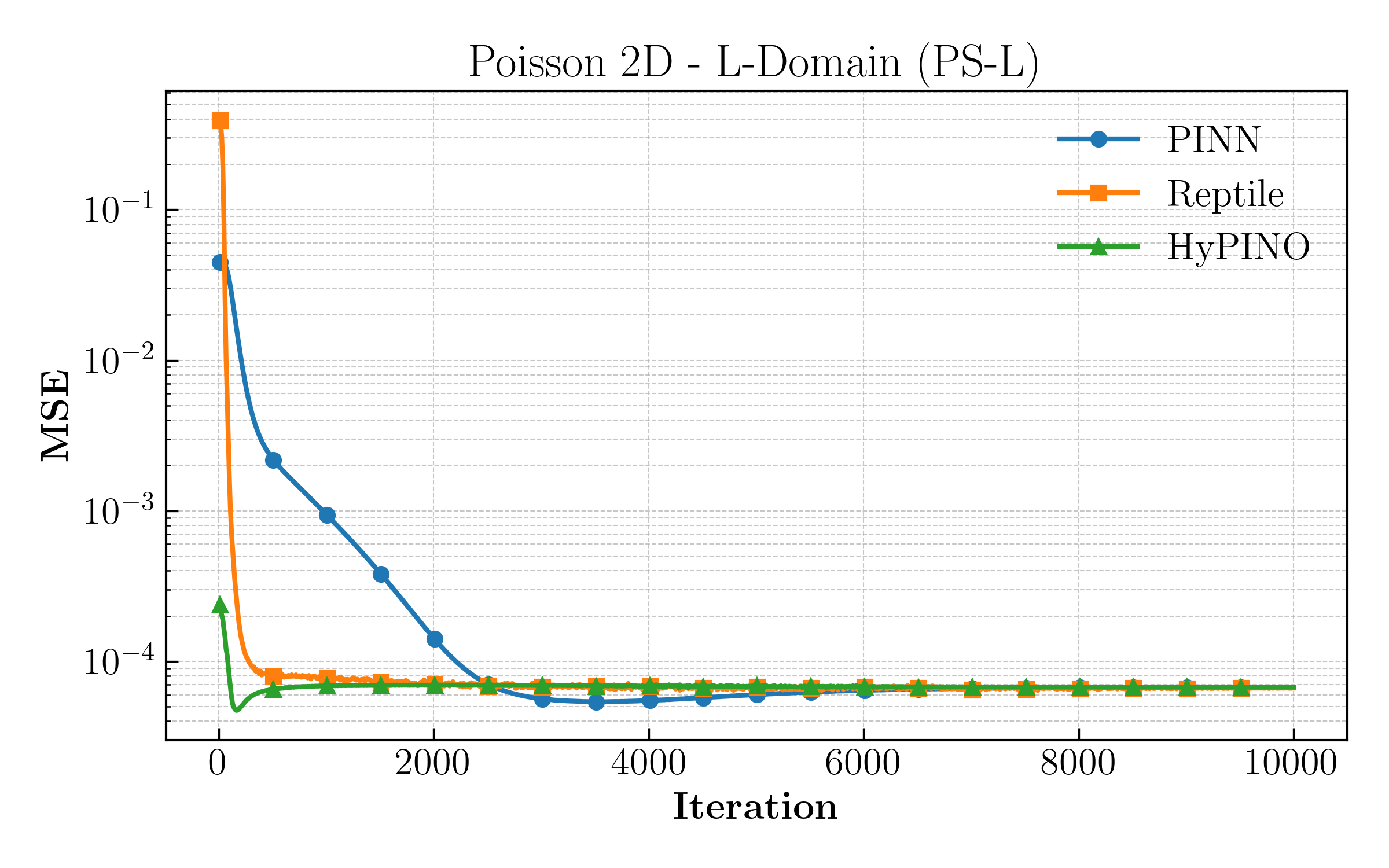}
    \end{subfigure}
    \begin{subfigure}[b]{0.32\linewidth}
        \includegraphics[width=\linewidth]{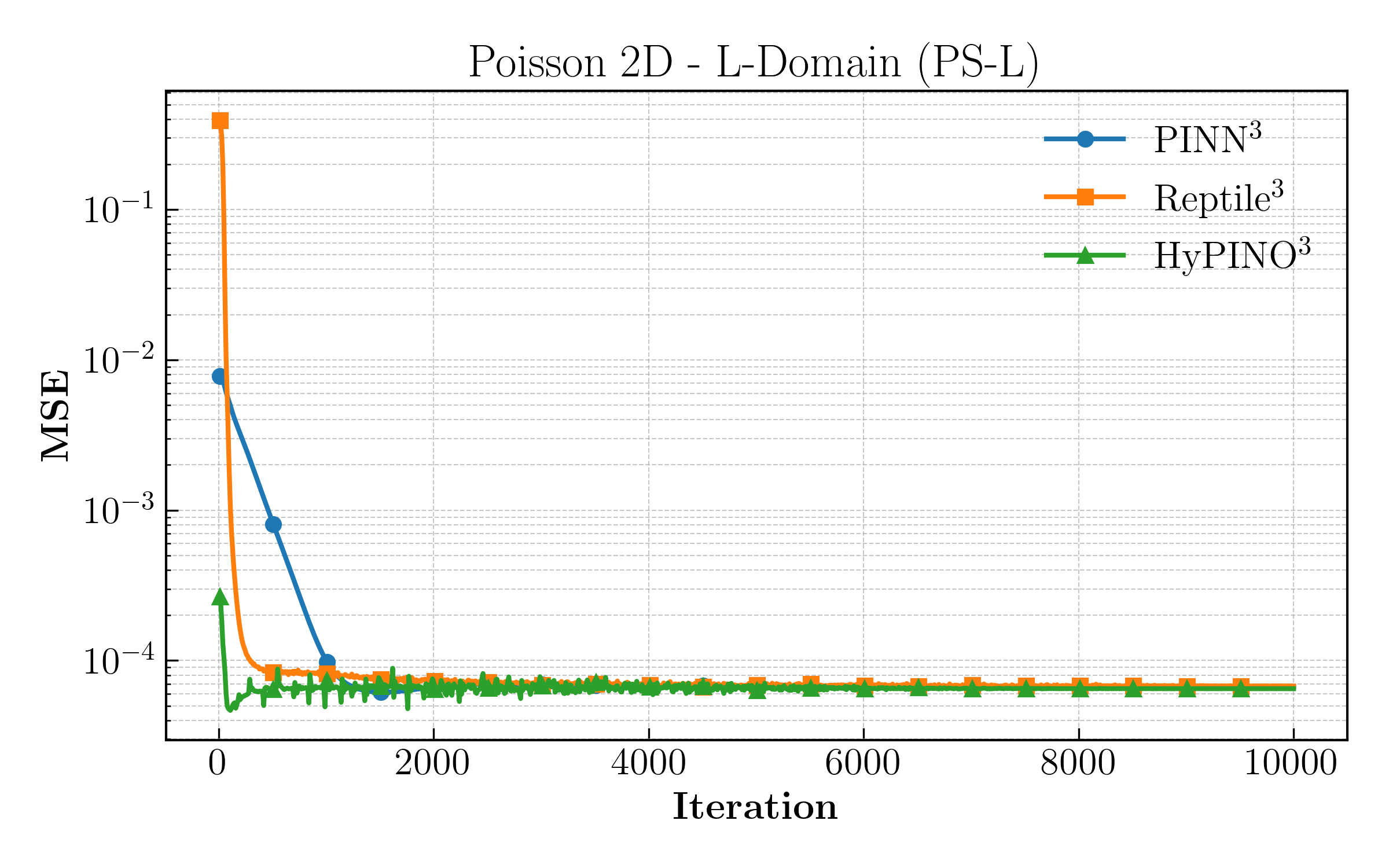}
    \end{subfigure}
    \begin{subfigure}[b]{0.32\linewidth}
        \includegraphics[width=\linewidth]{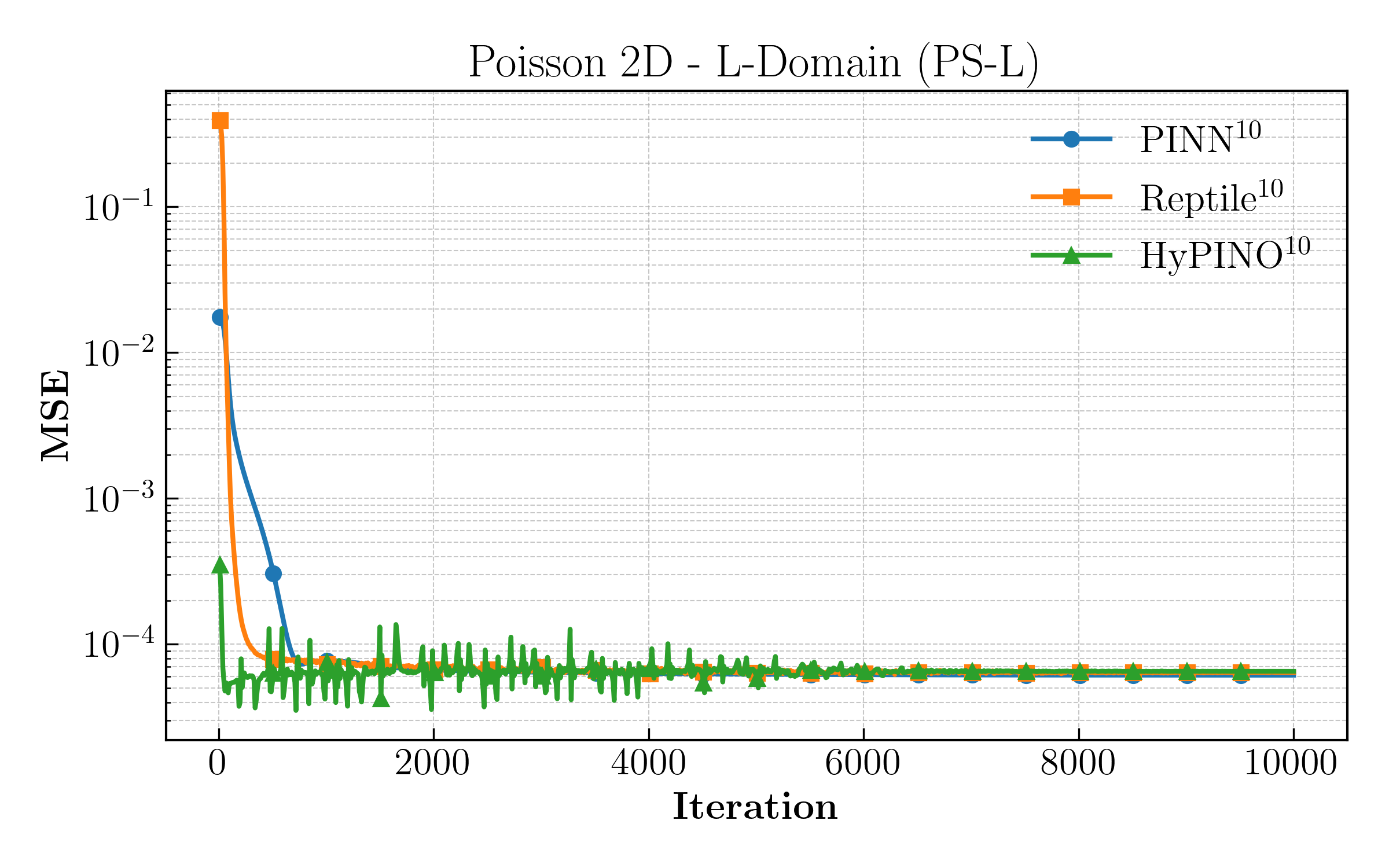}
    \end{subfigure}
    
    \vspace{1em}

    \begin{subfigure}[b]{0.32\linewidth}
        \includegraphics[width=\linewidth]{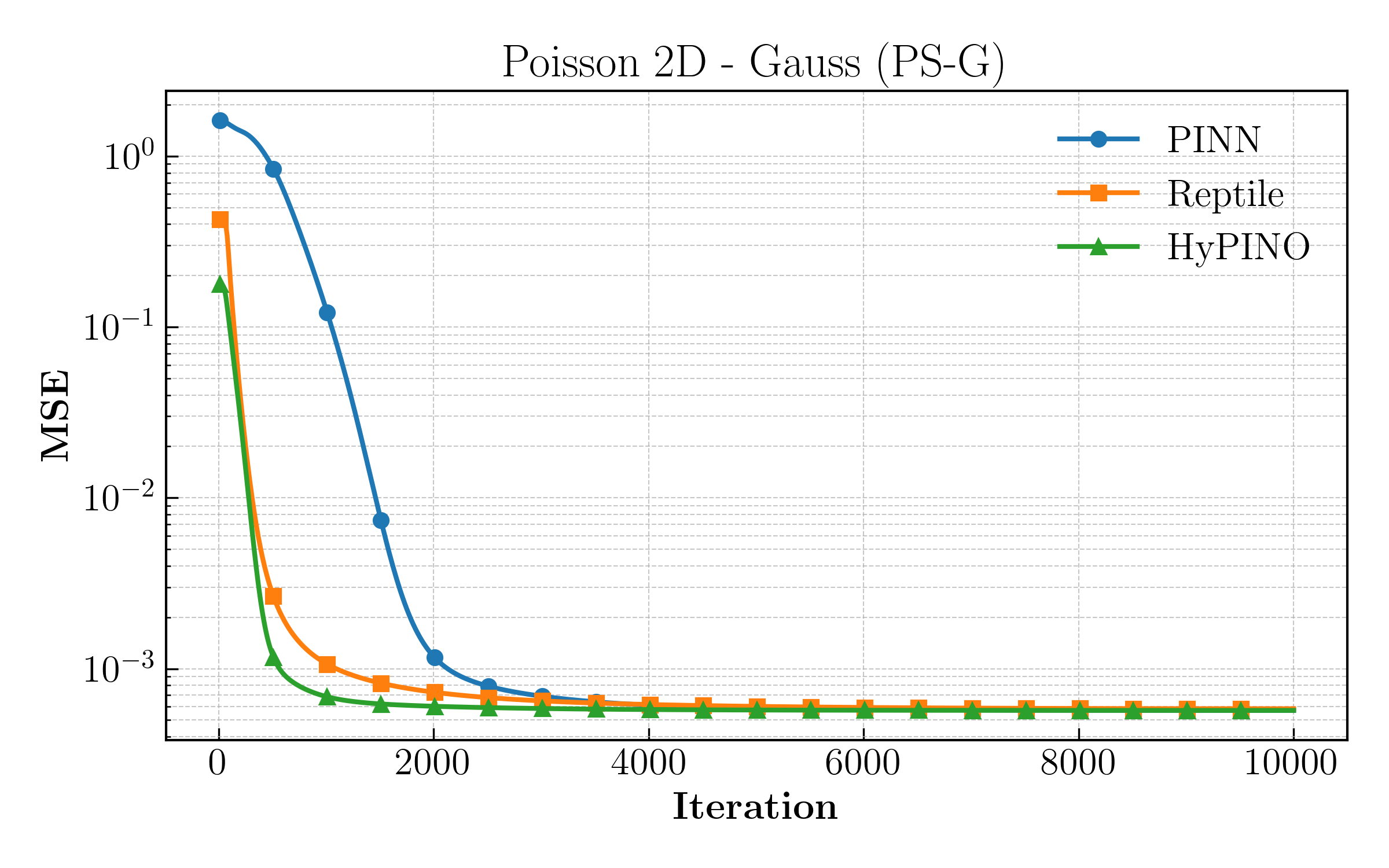}
    \end{subfigure}
    \begin{subfigure}[b]{0.32\linewidth}
        \includegraphics[width=\linewidth]{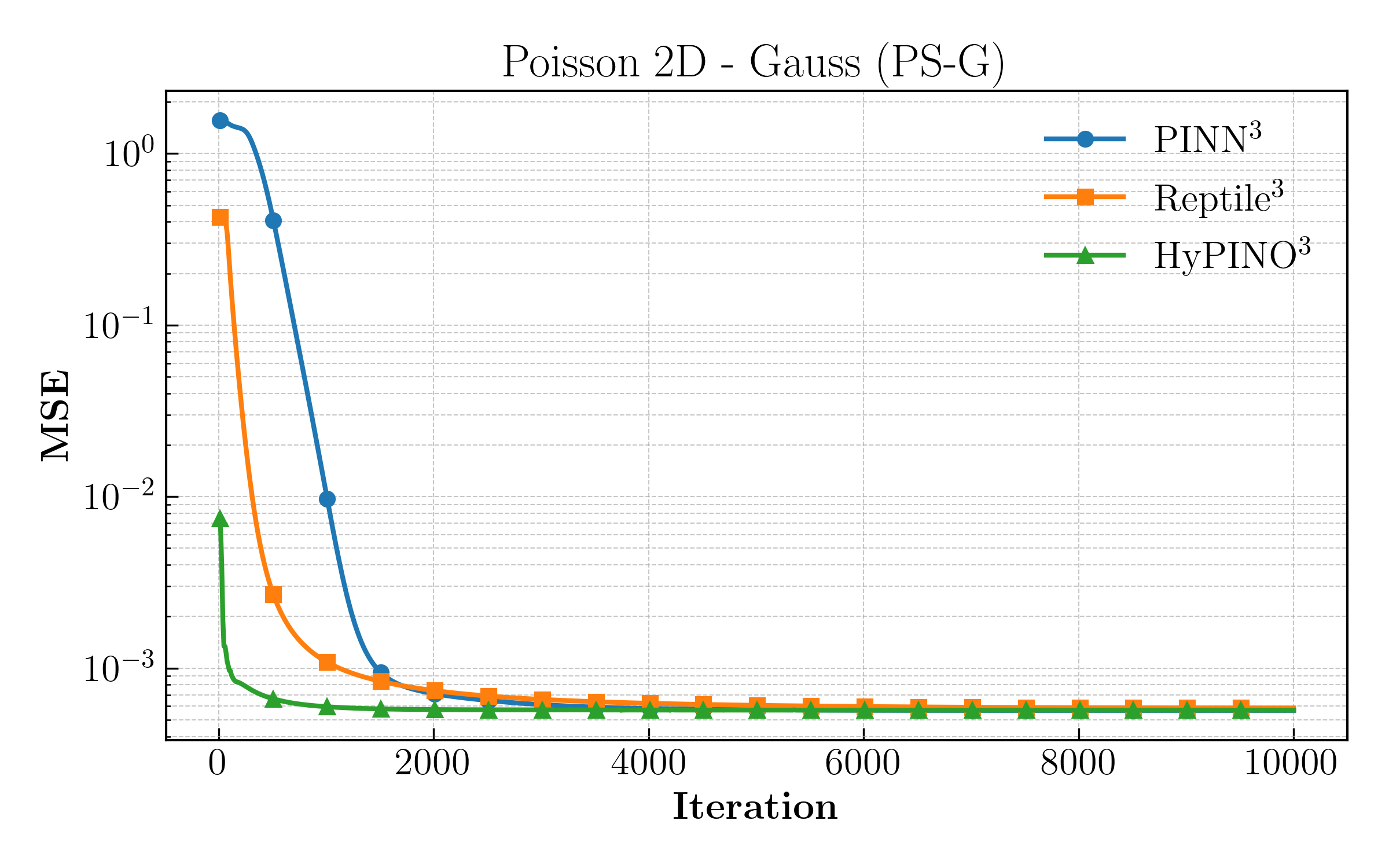}
    \end{subfigure}
    \begin{subfigure}[b]{0.32\linewidth}
        \includegraphics[width=\linewidth]{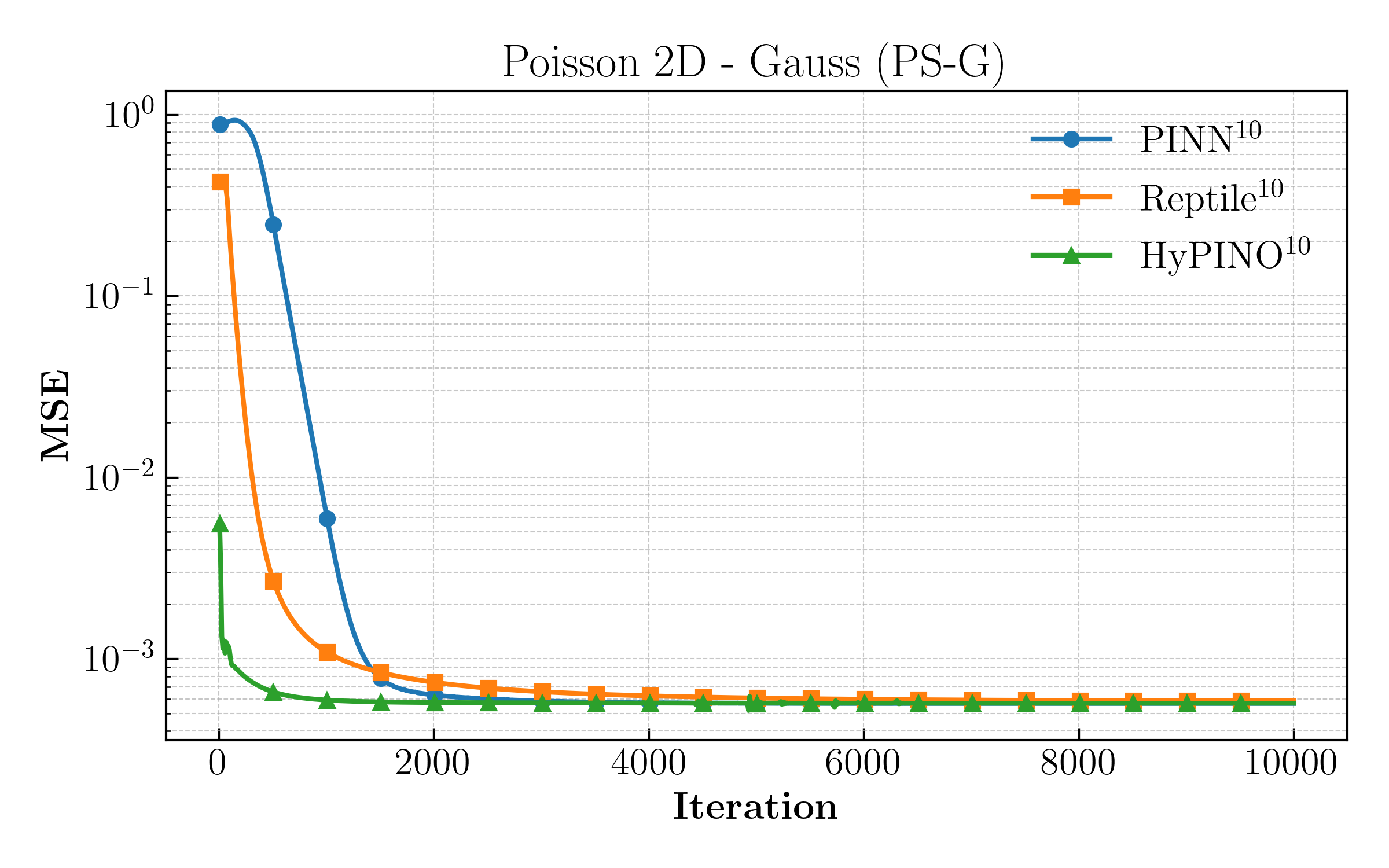}
    \end{subfigure}
    
    \vspace{1em}

    \begin{subfigure}[b]{0.32\linewidth}
        \includegraphics[width=\linewidth]{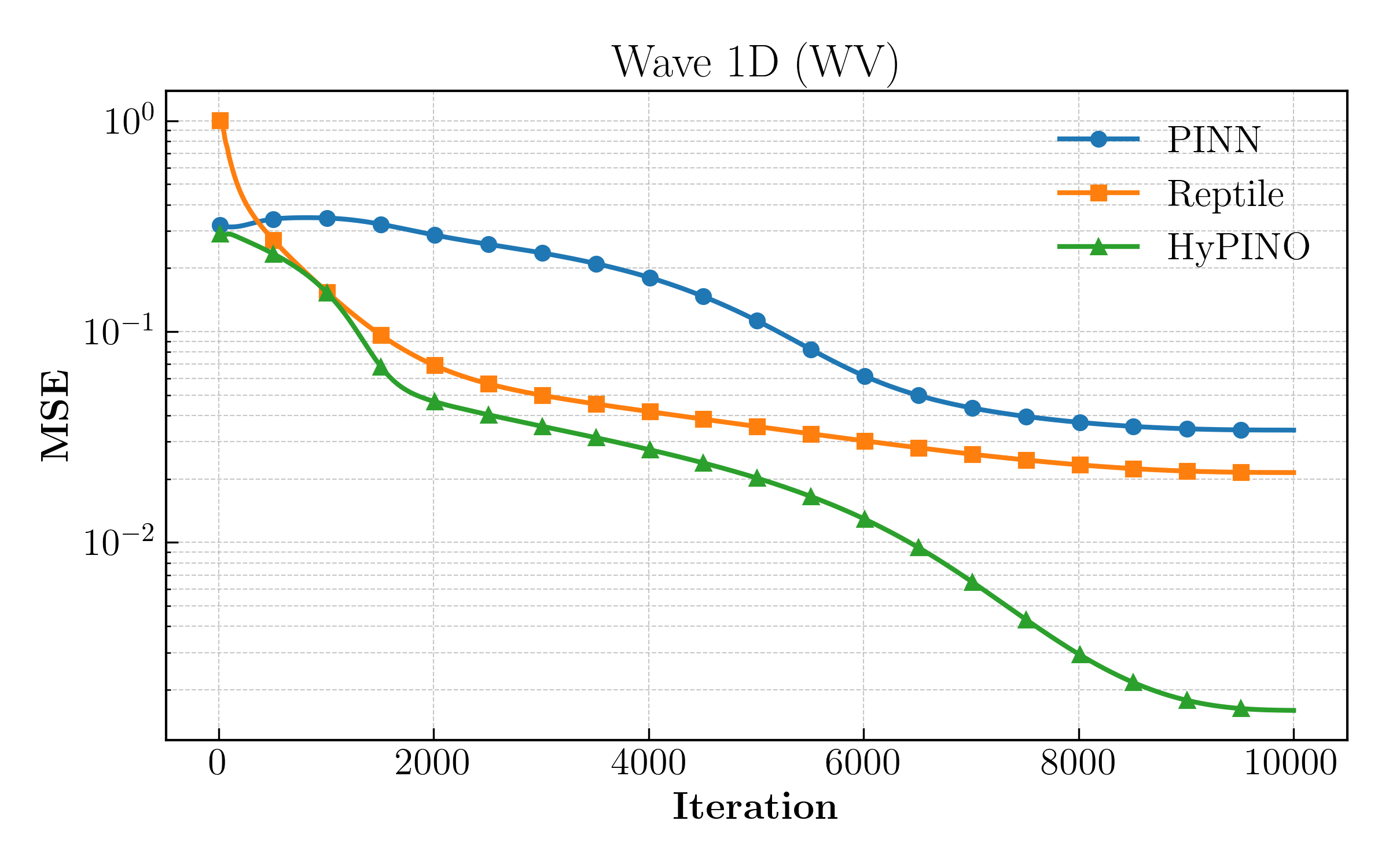}
        \caption{}
    \end{subfigure}
    \begin{subfigure}[b]{0.32\linewidth}
        \includegraphics[width=\linewidth]{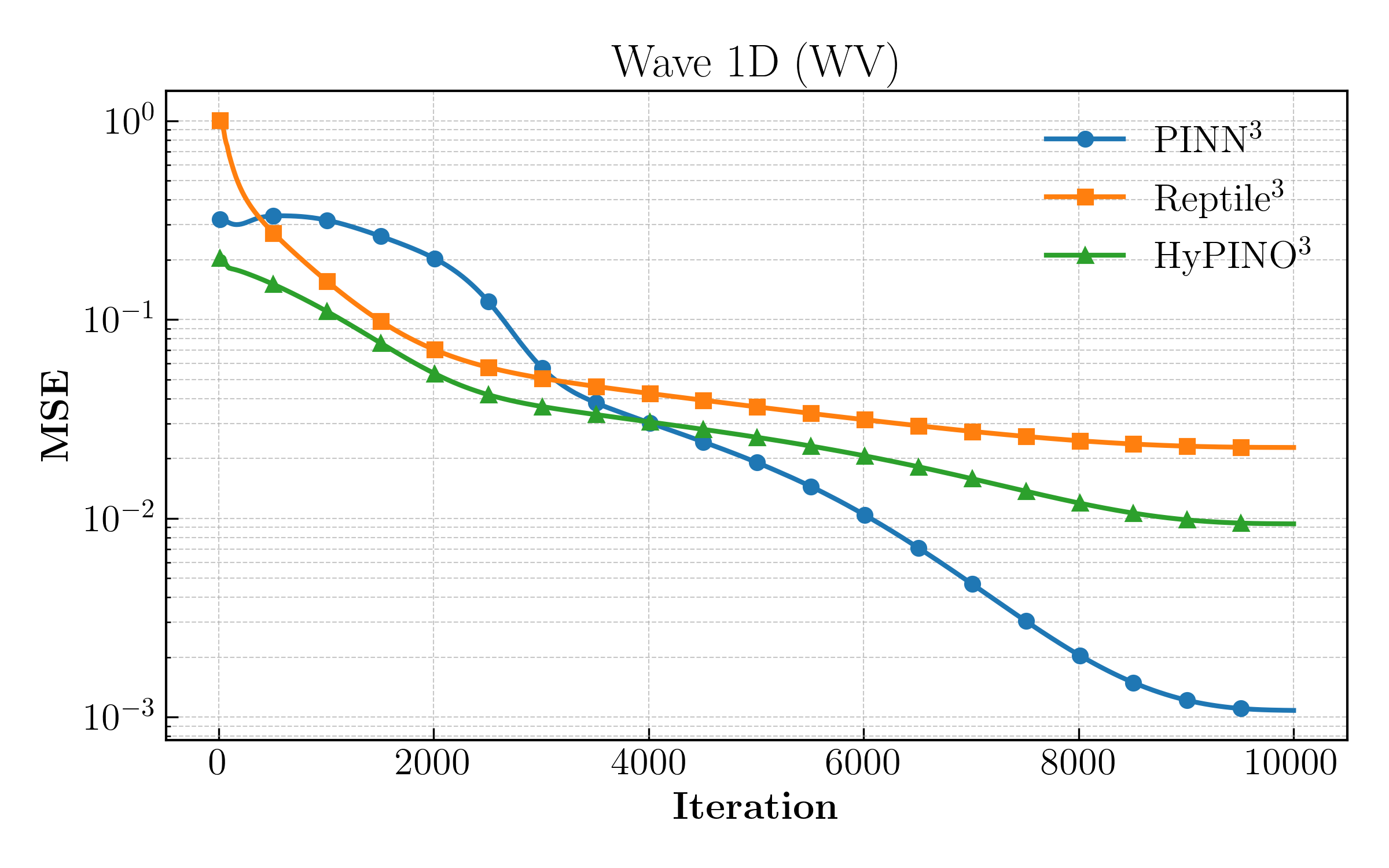}
        \caption{}
    \end{subfigure}
    \begin{subfigure}[b]{0.32\linewidth}
        \includegraphics[width=\linewidth]{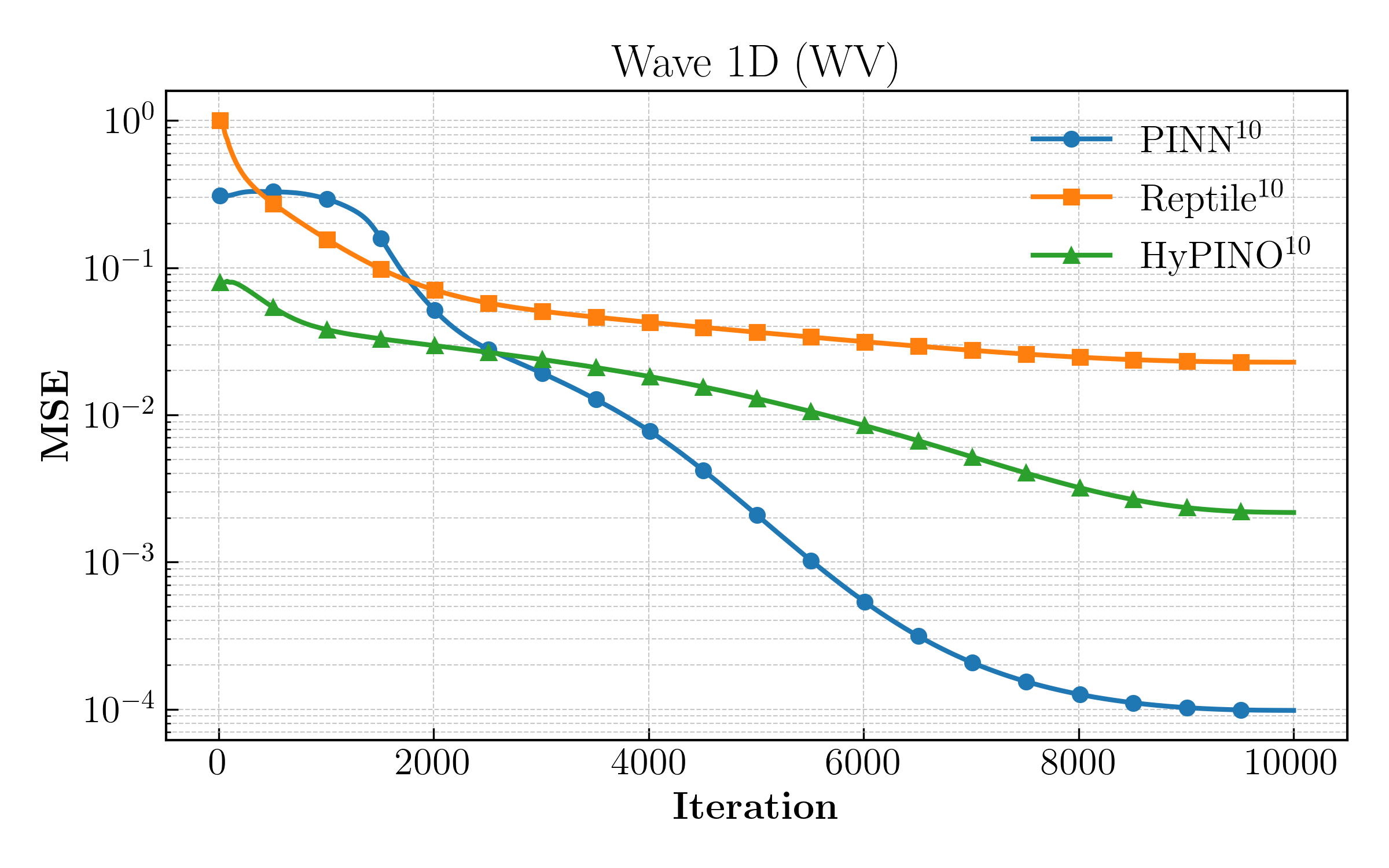}
        \caption{}
    \end{subfigure}

    \caption{Convergence of PINNs when fine-tuned on each of the benchmark PDE problems. We compare the convergence of different ensemble sizes: (a) single PINN, (b) ensemble of size 4 (c) ensemble of size 11, where an ensemble of size $i$ is an ensemble of $i$ randomly initialized PINNs (blue), $i$ PINNs initialized via Reptile (orange), or one PINN initialized via HyPINO followed by $i-1$ refinement rounds (green).}
    \label{fig:finetuning_convergence_2}
\end{figure}

Convergence behavior across all PDE classes is reported in Figures~\ref{fig:finetuning_convergence} and~\ref{fig:finetuning_convergence_2}.

\newpage

\subsection*{Resolution Invariance Ablation}

Discretization-invariance is an important property for neural operators. While the output of HyPINO is a continuous PINN that can be evaluated at arbitrary spatial coordinates, the input PDE parameterization (source function and boundary masks/values) is discretized on a fixed-size grid ($224 \times 224$) to match the Swin Transformer’s input resolution. Following prior work~\cite{pde_foundation_model_2024_Herde}, this limitation can be mitigated by demonstrating test-time resolution invariance when varying the input grid resolution and resizing it to ($224 \times 224$). 

We performed this ablation on the Helmholtz benchmark (HZ) by changing the source function resolution between $28$ and $448$. The results are shown in Table~\ref{tab:resolution_invariance}.

\begin{table}[h]
  \centering
  \caption{Resolution invariance ablation on the Helmholtz benchmark (HZ). 
  Each cell reports SMAPE across different input grid sizes, resized to $224 \times 224$.}
  \vspace{5pt}
  \label{tab:resolution_invariance}
  \begingroup
  \setlength{\tabcolsep}{5pt}
  \small
  \renewcommand{\arraystretch}{1.1}
  \begin{tabular}{l *{6}{c}}
    \toprule
     & \textbf{28} & \textbf{56} & \textbf{96} & \textbf{112} & \textbf{140} & \textbf{168} \\
    \midrule
    \textbf{SMAPE} & 38.04 & 35.78 & 35.91 & 36.00 & 36.05 & 36.05 \\
    \midrule
     & \textbf{196} & \textbf{224} & \textbf{280} & \textbf{336} & \textbf{392} & \textbf{448} \\
    \midrule
    \textbf{SMAPE} & 36.05 & 36.04 & 36.05 & 36.03 & 36.04 & 36.04 \\
    \bottomrule
  \end{tabular}
  \endgroup
\end{table}

Between resolutions of $56$ and $448$, SMAPE varies by less than $0.3$, which indicates approximate invariance. Only at very coarse resolutions ($28 \times 28$) does the performance begin to deteriorate.

\subsection*{L-BFGS Fine-Tuning with Different Initializations}

Our initial choice to evaluate fine-tuning performance using the Adam optimizer was motivated by its wide adoption in the PINN literature. To test whether HyPINO initializations also benefit second-order optimization, we conducted additional fine-tuning experiments using L-BFGS, chosen for its broad adoption and ease of use within PyTorch. All runs used standard L-BFGS hyperparameters without tuning.

\begin{table}[h]
  \centering
  \caption{Iterations required to match the initial MSE of a HyPINO-initialized PINN.}
  \vspace{5pt}
  \label{tab:lbfgs_iterations}
  \begingroup
  \setlength{\tabcolsep}{6pt}
  \small
  \renewcommand{\arraystretch}{1.1}
  \begin{tabular}{l *{7}{c}}
    \toprule
     & \textbf{HT} & \textbf{HZ} & \textbf{HZ-G} & \textbf{PS-C} & \textbf{PS-L} & \textbf{PS-G} & \textbf{WV} \\
    \midrule
    \textbf{Random Init}  & 4 & 20 & N/A & 36 & 34 & 11 & 35 \\
    \textbf{Reptile Init} & 4 & 22 & 211 & 22 & 65 & 9  & 27 \\
    \bottomrule
  \end{tabular}
  \endgroup
\end{table}

On PS-C and PS-L, Reptile requires 22 and 65 L-BFGS steps, respectively, to match HyPINO’s starting error, while random initialization needs 36 and 34. On HZ-G, Reptile requires 211 steps, while random never reaches HyPINO’s initial accuracy.

\begin{table}[h]
  \centering
  \caption{Final MSE after L-BFGS fine-tuning.}
  \vspace{5pt}
  \label{tab:lbfgs_final}
  \begingroup
  \setlength{\tabcolsep}{6pt}
  \small
  \renewcommand{\arraystretch}{1.1}
  \begin{tabular}{l *{7}{c}}
    \toprule
     & \textbf{HT} & \textbf{HZ} & \textbf{HZ-G} & \textbf{PS-C} & \textbf{PS-L} & \textbf{PS-G} & \textbf{WV} \\
    \midrule
    \textbf{Random Init}  & 2.93e-9 & \textbf{1.15e-7} & 2.89e-1 & 3.18e-4 & 7.05e-5 & 5.69e-4 & 2.68e-2 \\
    \textbf{Reptile Init} & 2.69e-9 & 2.18e-7 & 3.55e-2 & 9.34e-4 & 8.66e-5 & \textbf{5.68e-4} & \textbf{3.80e-4} \\
    \textbf{HyPINO Init}  & \textbf{1.62e-9} & 1.52e-7 & \textbf{1.74e-2} & \textbf{8.19e-5} & \textbf{6.87e-5} & 5.69e-4 & 1.94e-2 \\
    \bottomrule
  \end{tabular}
  \endgroup
\end{table}

The results show that HyPINO initializations remain effective with L-BFGS. HyPINO achieves the lowest final MSE on four benchmarks (HT, PS-C, PS-L, HZ-G) and is competitive on PS-G. Only on WV does Reptile achieve the best result, while on HZ, random slightly outperforms HyPINO. These differences are especially meaningful given the high cost of L-BFGS iterations.